\documentclass[acmsmall]{acmart}
\usepackage{graphicx}
\usepackage{todonotes}
\usepackage{xcolor}
\usepackage{subcaption}
\usepackage{wrapfig}
\usepackage{footmisc}
\usepackage{tabu}
\usepackage{algorithm}
\usepackage{algpseudocode}
\usepackage{multirow}
\usepackage{adjustbox}
\usepackage{amsthm}
\usepackage{amsmath,amsfonts}
\usepackage{hhline}
\PassOptionsToPackage{normalem}{ulem}
\usepackage{ulem}
\algnewcommand\algorithmicforeach{\textbf{for each}}
\algnewcommand{\LineComment}[1]{\(\triangleright\) #1}
\algdef{S}[FOR]{ForEach}[1]{\algorithmicforeach\ #1\ \algorithmicdo}

\definecolor{darkgreen}{RGB}{21,176,26}
\definecolor{orange}{RGB}{242,148,7}

\newtheorem{definition}{Definition}[section]

\providecolor{newadded}{rgb}{0,1,0}
\providecolor{omidadded}{rgb}{0,0,1}
\providecolor{deleted}{rgb}{1,0,0}

\newcommand{\re}[1]{{\color{black}{}#1}}

\newcommand{\reject}[1]{{\color{red}{}#1}}

\newcommand{\rsm}{RSM}

\graphicspath{{figures/}}

\begin{document}


\title{Dealing with Drift of Adaptation Spaces in Learning-based Self-Adaptive Systems using Lifelong Self-Adaptation}

\author{Omid Gheibi}
\affiliation{
	\institution{Katholieke Universiteit Leuven}
	\city{Celestijnenlaan 200A, 3001 Leuven}
	\country{Belgium}
}
\email{omid.gheibi@gmail.com}
 
\author{Danny Weyns}
\affiliation{
	\institution{Linnaeus University, Sweden}
	\city{Universitetsplatsen 1, 352 52 Växjö}
	\country{Katholieke Universiteit Leuven, Belgium}
}
\email{danny.weyns@gmail.com}

\begin{abstract}
    
Recently, machine learning (ML) has become a popular approach to support self-adaptation. ML has been used to deal with several problems in self-adaptation, such as maintaining an up-to-date runtime model under uncertainty and scalable decision-making. Yet, exploiting ML comes with inherent challenges. In this paper, we focus on a particularly important challenge for learning-based self-adaptive systems: drift in adaptation spaces. With adaptation space, we refer to the set of adaptation options a self-adaptive system can select from to adapt at a given time based on the estimated quality properties of the adaptation options. A drift of adaptation spaces originates from uncertainties, affecting the quality properties of the adaptation options. Such drift may imply that the quality of the system may deteriorate, eventually, no adaptation option may satisfy the initial set of adaptation goals, or adaptation options may emerge that allow enhancing the adaptation goals. In ML, such a shift corresponds to a novel class appearance, a type of concept drift in target data that common ML techniques have problems dealing with. To tackle this problem, we present a novel approach to self-adaptation that enhances learning-based self-adaptive systems with a lifelong ML layer. We refer to this approach as \textit{lifelong self-adaptation}. The lifelong ML layer tracks the system and its environment, associates this knowledge with the current learning tasks, identifies new tasks based on differences, and updates the learning models of the self-adaptive system accordingly. A human stakeholder may be involved to support the learning process and adjust the learning and goal models. We present a general architecture for lifelong self-adaptation and apply it to the case of drift of adaptation spaces that affects the decision-making in self-adaptation. We validate the approach for a series of scenarios with a drift of adaptation spaces using the DeltaIoT exemplar. 

\end{abstract}



\maketitle

\section{Introduction}\label{sec:introduction}

Self-adaptation equips a software system with a feedback loop that maintains a set of runtime models, including models of the system, its environment, and the adaptation goals. The feedback loop uses these up-to-date models to reason about changing conditions, analyze the options to adapt the system if needed, and if so, select the best option to adapt the system realizing the adaptation goals~\cite{Roadmap2009,weyns2020introduction}. \re{The key drivers for applying self-adaptation are automating tasks that otherwise need to be realized by operators (operators may be involved, e.g., to provide high-level goals to the system)~\cite{Kephart,garlan2004rainbow}, and mitigating uncertainties that the system may face during its lifetime that are hard or even impossible to be resolved before the system is in operation~\cite{esfahani2013usa,3487921}.} 

In the past years, we have observed an increasing trend in the use of machine learning (ML in short) to support self-adaptation~\cite{gheibi2021}. ML has been used to deal with a variety of tasks, such as learning and improving scaling rules of a cloud infrastructure~\cite{7515437}, efficient decision-making by reducing a large number of adaption options~\cite{quin2019efficient}, detecting abnormalities in the flow of activities in the  environment of the system~\cite{krupitzer2017adding}, and 
learning changes of the system utility dynamically~\cite{8498142}.
We use the common term \textit{learning-based self-adaptive systems} to refer to such systems. 

While ML techniques have already demonstrated their usefulness, these techniques are subject to several engineering challenges, such as reliable and efficient testing, handling unexpected events, and obtaining adequate quality assurances for ML applications~\cite{Kumeno2019,amershi2019software}. In this paper, we focus on one such challenge, namely novel class appearance, a particular type of concept drift in target data that common ML
techniques have problems dealing  with~\cite{masud2010classification, webb2016characterizing}. Target data refers to the data about which the learner wants to gain knowledge. Target data in learning-based self-adaptive systems typically correspond to predictions of quality attributes for the different adaptation options.

Concept drift in the form of novel class appearance is particularly important for learning-based self-adaptive systems in the form of drift in adaptation spaces. With adaptation space, we mean the set of adaptation options from which the feedback loop can select to adapt the system at a given point in time based on the estimated quality properties of the adaptation options and the adaptation goals. Due to the uncertainties the self-adaptive system is subjected to, the quality properties of the adaptation options typically fluctuate, which may cause concept drift\,\cite{8787137,metzger2020triggering,vieira2021}, in particular drift of the adaptation space over time. Eventually, this drift may have two effects. On the one hand, it may result in a situation where none of the adaptation options can satisfy the set of adaptation goals initially defined by the stakeholders. This may destroy the utility of the system. As a fallback, the self-adaptive system may need to switch to a fail-safe strategy, which may be sub-optimal. On the other hand, due to the drift, the quality properties of adaptation options may have changed such that adaptation options could be selected that would enhance the adaptation goals, i.e., new regions emerge in the adaptation space with adaptation options that have superior predicted quality properties. This offers an opportunity for the system to increase its utility. The key problem with drift of adaptation spaces, or novel class appearance, 
is that new classes of data emerge over time that are not known before they appear~\cite{masud2010classification, webb2016characterizing}, so training of a learner cannot anticipate such changes. This may deteriorate the precision of the learning model, which may jeopardize the reliability of the system. Hence, the research problem that we tackle in this paper is: 

\begin{quote}
    \textit{How to enable learning-based self-adaptive systems to deal with a drift of adaptation spaces during operation, i.e., concept drift in the form of novel class appearance?}
\end{quote}

To tackle this research problem, we propose \emph{lifelong self-adaptation}: a novel approach to self-adaptation that enhances learning-based self-adaptive systems with a lifelong ML layer. The lifelong ML layer: (i) tracks the running system and its environment, (ii) associates the collected knowledge with the current classes of target data, i.e., regions of adaptation spaces determined by the quality attributes associated with the adaptation goals, (iii) identifies new classes based on differentiations, i.e., new emerging regions of adaptation spaces, (iv) visualizes the new classes providing feedback to the stakeholder who can then rank all classes (new and previously detected classes), and (v) finally updates the learning models of the self-adaptive system accordingly. 

Lifelong self-adaptation leverages the principles of lifelong machine learning~\cite{thrun1998lifelong,chen2018lifelong}, which offers an architectural approach for continual learning of a machine learning system.
Lifelong machine learning adds a layer on top of a machine learning system that selectively transfers the knowledge from previously learned tasks to facilitate the learning of new tasks within an existing or new domain~\cite{chen2018lifelong}. 
Lifelong machine learning has been successfully combined with a wide variety of learning techniques~\cite{chen2018lifelong}, including supervised~\cite{silver2015consolidation}, interactive~\cite{ammar2015autonomous}, and unsupervised learning~\cite{shu2016lifelong}.

Our focus in this paper is on self-adaptive systems that rely on architecture-based adaptation~\cite{KramerMagee,FORMS,Roadmap2009,Lemos2013}, where a self-adaptive system consists of a managed system that operates in the environment to deal with the domain goals and a managing system that manages the managed system to deal with the adaptation goals. We focus on managing systems that comply with the MAPE-K reference model, short for Monitor-Analyse-Plan-Execute-Knowledge~\cite{Kephart,empiricalMAPE}. Our focus is on managing systems that use an ML technique to support any of the MAPE-K functions. We make the assumption that dealing with a drift of adaptation spaces, i.e., novel class appearance, does not require any runtime evolution of the software of the managed and managing system.

\re{The concrete contribution of this paper is two-fold: 
\begin{enumerate}
    \item A general architecture for lifelong self-adaptation with a concrete instance to deal with a drift of adaptation spaces, i.e., novel class appearance;    
    \item A validation of the instance of the architecture using the DeltaIoT artifact~\cite{iftikhar2017deltaiot}.
\end{enumerate}

We evaluate the instantiated architecture in terms of effectiveness in dealing with a drift of adaptation spaces, the robustness of the approach to changes in the appearance order of classes, and the effectiveness of feedback of an operator in dealing with a drift of adaptation spaces.} 

In~\cite{10.1145/3524844.3528052}, we introduced an initial version of lifelong self-adaptation and we applied it to two types of concept drift: sudden covariate drift and incremental covariate drift. Covariate drift refers to drift in the \textit{input features} of the learning model of a learner under the assumption that the labeling functions of the source and target domains are identical for a classification task~\cite{adel2015probabilistic}. In contrast, novel class appearance concerns drift in the \textit{target} of a learner, i.e., the prediction space of the learning model. Handling this type of concept drift often requires interaction with stakeholders. 

The remainder of this paper is structured as follows. In Section~\ref{sec:background}, we provide background on novel class appearance and lifelong machine learning.
Section~\ref{sec:problemContext} introduces DeltaIoT and elaborates on the problem of drift of adaptation spaces. Section~\ref{sec:methodology} then presents lifelong self-adaptation.   
We instantiate the architecture of a lifelong self-adaptive system for the case of drift in adaptation spaces using DeltaIoT. 
In Section~\ref{sec:evaluation}, we evaluate lifelong self-adaptation for drift of adaptation spaces using different scenarios in DeltaIoT and we discuss threats to validity.   
Section~\ref{sec:relatedWork} presents related work. Finally, we wrap up and outline opportunities for future research in  Section~\ref{sec:conclusions}.

\section{Background}\label{sec:background}

We start this section with a brief introduction to novel class appearance. Then we provide a short introduction to lifelong machine learning, the basic framework underlying lifelong self-adaptation.

\subsection{Concept Drift and Novel Class Appearance}
Mitchell et. al~\cite{mitchell1997machine} defined machine learning as follows: ``A computer program is said to learn from experience $E$ concerning some class of tasks $T$ and performance measure $P$, if its performance at tasks in $T$, as measured by $P$, improves with experience $E$''. 
Consider a self-adaptive sensor network that should keep packet loss and energy consumption below specific criteria.
The analysis of adaptation options could serve as the training experience $E$ from which the system learns.
The task $T$ could be classifying the adaptation options to predict which of them comply with the goals (need to be analyzed) and which do not (not necessary to be analyzed).
To perform this task, the performance measure $P$ could be the comparison of the classification results for adaptation options with their actual classes as ground truth. 
Here, learning (classification) assists the analysis step of the feedback loop by lowering a large number of adaptation options to improve analysis efficiency.

Static supervised machine learning models, used for prediction tasks, i.e., regression and classification, are trained based on historical data. These models face significant issues in dynamic worlds. In particular, the learning performance of these models may deteriorate as the world changes. In the context of non-stationary distributions~\cite{webb2016characterizing}, world changes are commonly called \emph{concept drift}.
Different types of concept drift can be characterized based on: (i) how the distribution of data shifts over time, and (ii) where this shift takes place in data for prediction learning tasks. 

\begin{figure}[!ht]
	\centering
	\includegraphics[scale=0.5]{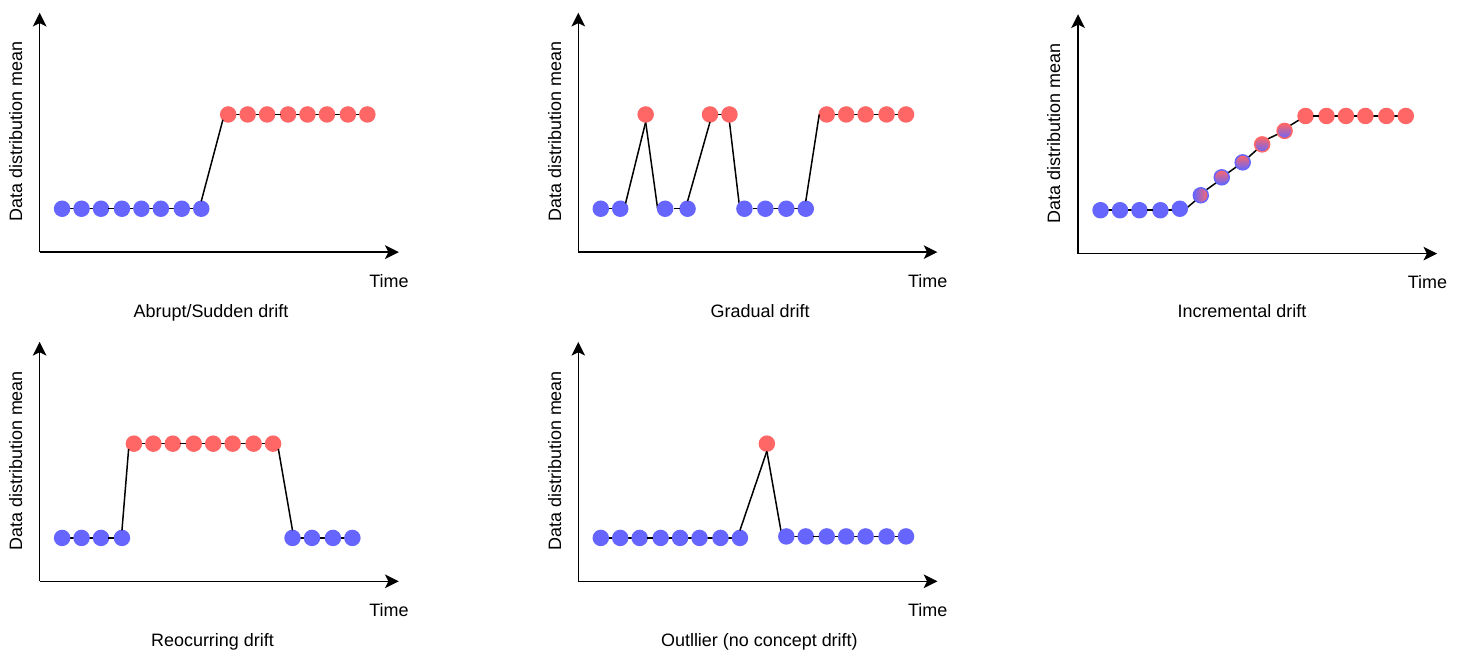}
	\caption{Different patterns of occurring concept drift in comparison with the outlier pattern.}
	\label{fig:concept drift patterns}
\end{figure}

Regarding data shifts over time, Figure~\ref{fig:concept drift patterns} shows four main patterns of concept drift and their difference with an outlier~\cite{gama2014survey}. An outlier is different compared to the patterns as the distribution of the data will not significantly shift for a significant duration of time. Note that concept drift in practice can be a combination of some of these patterns, e.g., incremental recurring drift. 

Regarding where the shift takes place, we can distinguish input features to the corresponding learning model, and targets of the prediction space of the model. Based on this, we can distinguish three important types of concept drift. 

First, \textit{covariate drift} arises when the distribution of attributes changes over time, and the 
conditional distribution of the target with respect to the attributes
remains invariant~\cite{gama2014survey,webb2016characterizing}. 
For instance, take business systems that use socio-economic factors (attributes) for user classification (the class of users is the target here). Shifting the demography of the customer base (as part of attributes) during the time changes the probability of demographic factors~\cite{webb2016characterizing}. This type of drift can occur both in classification and regression tasks. 

Second, \textit{target drift} (also known as concept drift or class drift) arises when the conditional distribution of the target with respect to attributes changes over time, while the distribution of attributes remains unaffected~\cite{gama2014survey,webb2016characterizing}.
An example of this type of drift may occur in network intrusion detection~\cite{vzliobaite2016overview}. 
An abrupt usage pattern (usage metrics are attributes here) in the network can be an adversary behavior or a weekend activity (class labels are targets here) by normal users. In this case, we confront the same set of data attributes for the decision but with different classes (target) of meanings behind these scenarios that depend on the activity time (attributes). This type of drift can also occur in both types of prediction tasks, regression and classification.

Third, \textit{novel class appearance}  is a type of concept drift about emerging new classes in the target over time~\cite{masud2010classification, webb2016characterizing}.
Hence, for a new class, the probability of having any data with the new class in the target is zero before it appears.\footnote{The novel class appearance is a type of drift in the distribution of the target ($\mathbb{P}(Y)$). In contrast, the target drift focuses on the drift in the posterior distribution ($\mathbb{P}(Y|X)$) and the target distribution ($\mathbb{P}(Y)$) may be unaffected.} 
Over time the data with the new class emerges, and there is a positive probability of observing such data. 
Examples of this type of drift~\cite{mustafa2017unsupervised} are novel intrusions or attacks (as new targets) appearing in the security domain~\cite{araujo2016engineering, araujo2014patches} and new physical activities (as new targets) in the monitoring stage of wearable devices~\cite{reiss2012introducing}. In contrast to covariate and target drifts, this type of drift can only occur in classification tasks.

\subsection{Lifelong Machine Learning}

Lifelong machine learning enables a machine-learning system to learn new tasks that were not predefined when the system was designed~\cite{THRUN199525}. It mimics the learning processes of humans and animals that accumulate knowledge learned from earlier tasks and use it to learn new tasks and solve new problems. 
Technically, lifelong machine learning is a continuous learning process of a learner~\cite{chen2018lifelong}. Assume that at some point in time, the learner has performed a sequence of $n$ learning tasks, $\mathcal{T}_1, \mathcal{T}_2,\cdots, \mathcal{T}_n$, called the \textit{previous tasks}, that have their corresponding data sets $\mathcal{D}_1, \mathcal{D}_2,\cdots, \mathcal{D}_n$. Tasks can be of different \textit{types} and from different \textit{domains}. When faced with 
task $\mathcal{T}_{n+1}$ (called the \textit{new} or \textit{current task}) with its data set  $\mathcal{D}_{n+1}$, the learner can leverage \textit{past knowledge} maintained in a \textit{knowledge-base} to help learn task  $\mathcal{T}_{n+1}$. The new task may be given by a stakeholder or it may be discovered automatically by the system. 
Lifelong machine learning aims to optimize the performance of the learner for the new task (or for an existing task by treating the rest of the tasks as previous tasks). After completing the learning task $\mathcal{T}_{n+1}$, the knowledge base is updated with the newly gained knowledge, e.g., using intermediate and ﬁnal results obtained via learning. Updating the knowledge can involve checking consistency, reasoning, and mining meta-knowledge. 

For example, consider a lifelong machine learning system for the never-ending language learner~\cite{mitchell2018never} (NELL in short). NELL aims to answer questions posed by users in natural language. To that end,  it sifts the Web 24/7 extracting facts, e.g., ``Paris is a city." 
The system is equipped with a set of classifiers and deep learners to categorize nouns and phrases (e.g., ``apple'' can be classified as ``Food'' and ``Company'' falls under an ontology), and detecting relations (e.g., ``served-with'' in ``tea is served with biscuits'').
NELL can infer new beliefs from this extracted knowledge, and based on the recently collected web documents, NELL can expand relations between existing noun phrases or ontology. This expansion can be a change within existing ontological domains, e.g., politics or sociology, or be a new domain like internet-of-things. Hence, the expansion causes an emerging task like classifying new noun phrases for the expanded part of the ontology.

Lifelong machine learning works together with different types of learners. In lifelong supervised learning, every learning task aims at recognizing a particular class or concept. For instance, in cumulative learning, identifying a new class or concept is used to build a new multi-class classifier for all the existing and new classes using the old classifier~\cite{2939672.2939835}. Lifelong unsupervised learning focuses on topic modeling and lifelong information extraction, e.g., by mining knowledge from topics resulting from previous tasks to help generate better topics for new tasks~\cite{2872427.2883086}. In lifelong semi-supervised learning, the learner enhances the number of relationships in its knowledge base by learning new facts, for instance, in the NELL system~\cite{mitchell2018never}. Finally, in lifelong reinforcement learning each environment is treated as a task~\cite{Tanaka1998AnAT},
or a continual-learning agent solves complex tasks by learning easy tasks first~\cite{Ring1997}. Recently, lifelong learning has gained increasing attention, in particular for autonomous learning agents and robots based on neural networks~\cite{abs-1802-07569}. 

One of the challenges for lifelong machine learning is dealing with catastrophic forgetting, i.e., the loss of what was previously learned while learning new information, which may eventually lead to system failures~\cite{abs-1908-01091}.  Another more common challenge is under-specification, i.e., a significant decrease of the performance of a learning model from training to deployment (or testing)~\cite{d2020underspecification}.
Promising approaches have been proposed, e.g.,~\cite{abs-1802-07569} for catastrophic forgetting and~\cite{ribeiro-etal-2020-beyond} for under-specification. Yet, more research is needed to transfer these techniques to real-world systems.  

\section{Problem of Drift of Adaptation Spaces}\label{sec:problemContext}

In this section, we introduce the setting of DeltaIoT that we use in this paper. Then we illustrate and elaborate on the problem of drift of adaptation spaces using a scenario of DeltaIoT. 

\subsection{DeltaIoT}\label{sec:delta-iot}

\begin{figure}
	\centering
	\includegraphics[width=\textwidth]{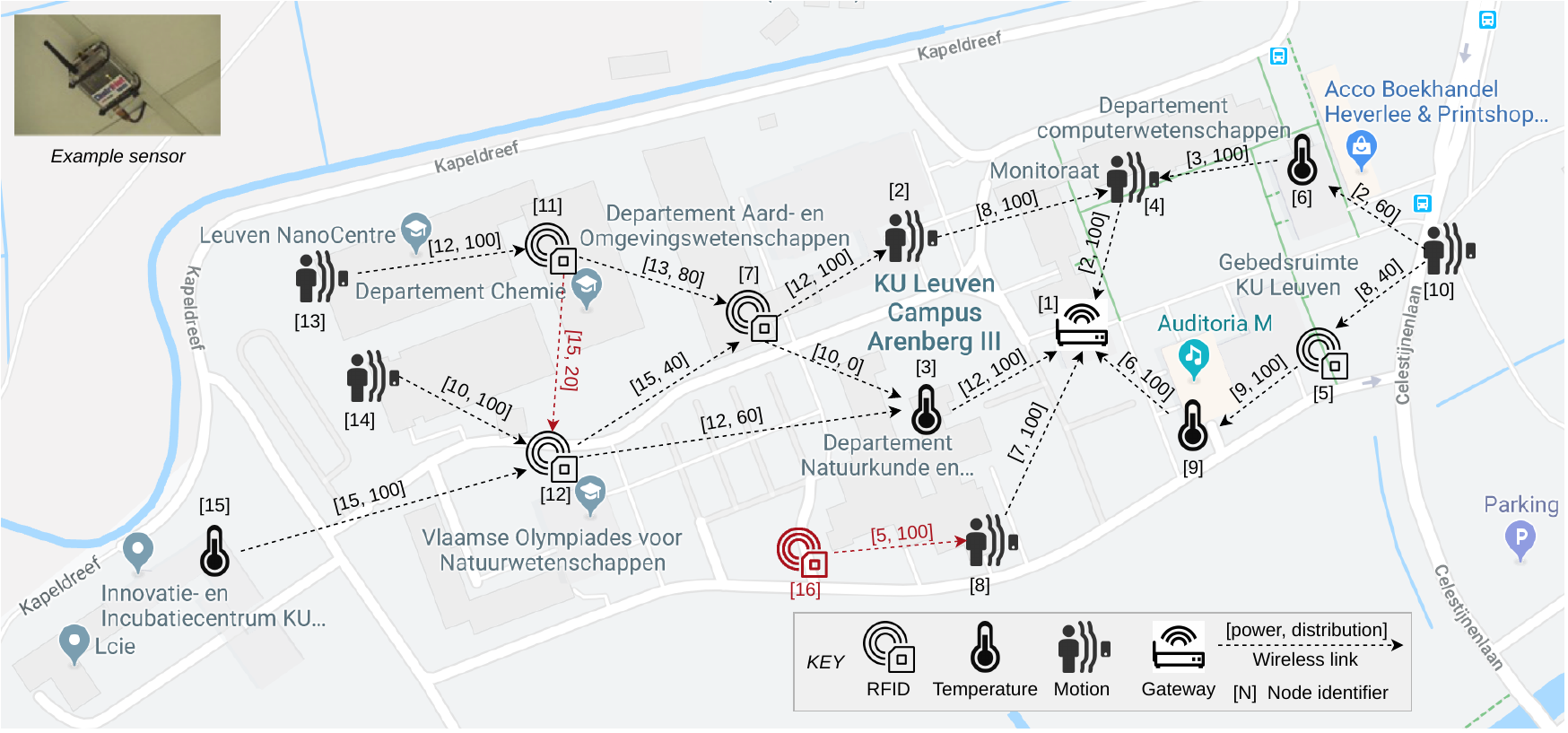}
	\caption{DeltaIoTv1.1 setup.}
	\label{fig:deltaiotv1.1}
\end{figure}

DeltaIoT~\cite{iftikhar2017deltaiot} is an examplar in the domain of the Internet of Things that supports research in engineering self-adaptive systems, see e.g.,\,\cite{10.1007/978-3-030-00761-4_4,10.1145/3522585,10.1145/3530192,10.1145/3194133.3194142}. Figure~\ref{fig:deltaiotv1.1} shows the setup of the IoT network that we used for the research presented in this paper. The network consists of 16 battery-powered sensor motes\footnote{The IoT system is deployed at the campus of the Computer Science department of KU Leuven and a simulated version is available for experimentation. Inspired by~\cite{quin2022reducing},  we used an extension of version v1.1 of the DeltaIoT network~\cite{iftikhar2017deltaiot}, called DeltaIoTv1.1. This version adds an extra mote (marked with number [16]) to the network. With this extension, the adaptation space increases by a factor of four compared to version v1.1 as we explain further in the paper.} that measure parameters in the environment and send the data via a wireless multi-hop network to a central gateway that connects with users at an application server that can use the data. Motes are equipped with various sensors, in particular RFID, temperature sensors, and motion sensors. The data collected by the motes can be used to monitor the campus area and take action when needed. The communication in the network is time-synchronized~\cite{mills2017computer}, i.e., the communication is organized in cycles (a number of minutes) where neighboring motes are allocated slots that they can use to send and receive messages over a wireless link as shown in the figure. 

Two quality properties of interest in DeltaIoT are packet loss and energy consumption. In general, stakeholders want to keep both the packet loss and the energy consumption low. Yet, these quality properties are conflicting as using more energy will increase the signal strength and hence reduce packet loss. Furthermore, uncertainties have an impact on the qualities. We consider two types of uncertainties: the load of messages produced by motes, which varies depending on several aspects, including the number of humans sensed in the surroundings, and network interference caused by environmental circumstances, such as other networks and weather changes. Network interference affects the Signal-to-Noise Ratio (SNR)~\cite{haenggi2009stochastic}, which then influences the packet loss. 

\subsubsection{Self-adaptation}

To mitigate the uncertainties and satisfy the goals of the stakeholders, we add a feedback loop at the gateway (i.e., a managing system) that monitors the IoT network (i.e., the managed system) and its environment and can adapt the network settings in each cycle. 

The managing system can adapt two settings for each mote: (1) the power setting (a value in the range of 0 to 15), which will affect the SNR and hence the packet loss, and (2) the distribution of the messages along the outgoing network links (for motes with two links, a selection among the following options is possible: 0/100, 20/80, 40/60, 60/40, 80/20, 100/0). Because the power setting of each mote can be determined by the values of the sensed SNRs of it links, these values are determined in each cycle and used for all adaptation options. The adaptation options are then determined by the distribution of messaging for each mote with two links. Hence, the total number of possible adaptation options is equal to the possible configurations (0/100, 20/80, 40/60, 60/40, 80/20, 100/0) for 4 motes with two parent links (motes with the index of 7, 10, 11, and 12 in Figure~\ref{fig:deltaiotv1.1}). This creates in total $6^4 = 1296$ different configurations from which the managing system can select an option to adapt the system. 

\begin{figure}
	\centering
        \begin{subfigure}[b]{0.49\textwidth}
		\includegraphics[width=\textwidth]{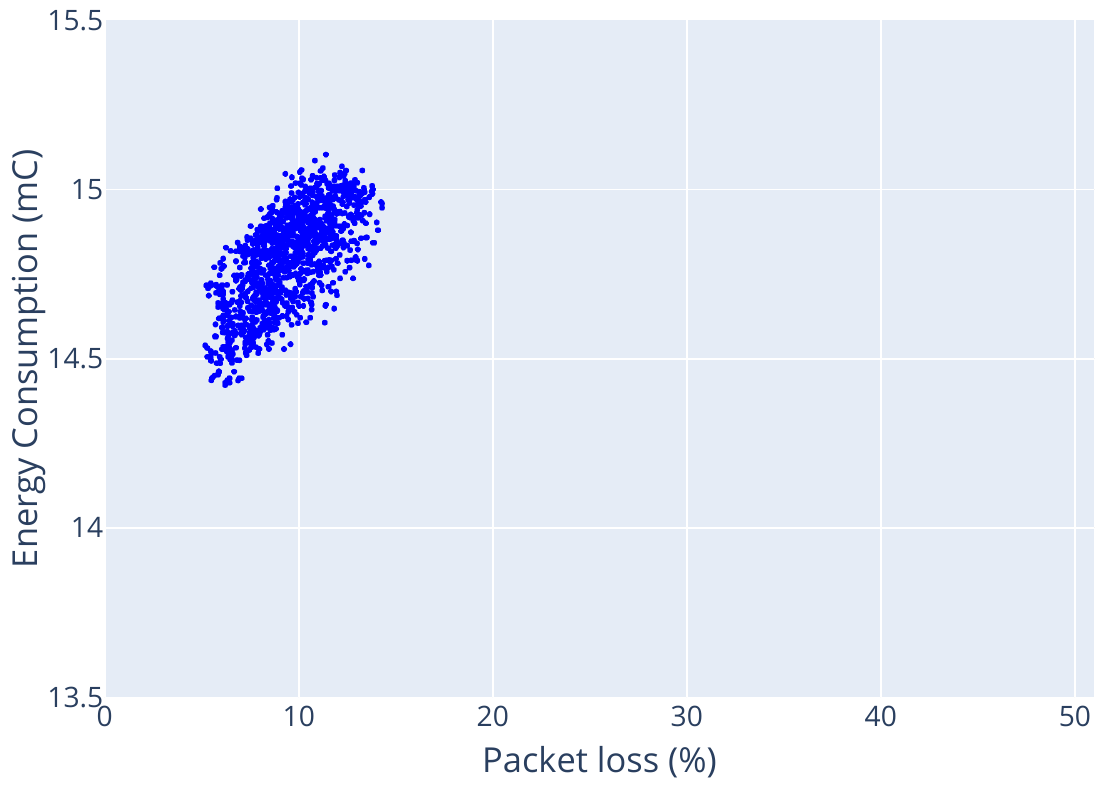}
		\caption{Adaptation space in one cycle}
		\label{fig: sample of quality attributes distribution at one point}
	\end{subfigure}
	\begin{subfigure}[b]{0.49\textwidth}
		\includegraphics[width=\textwidth]{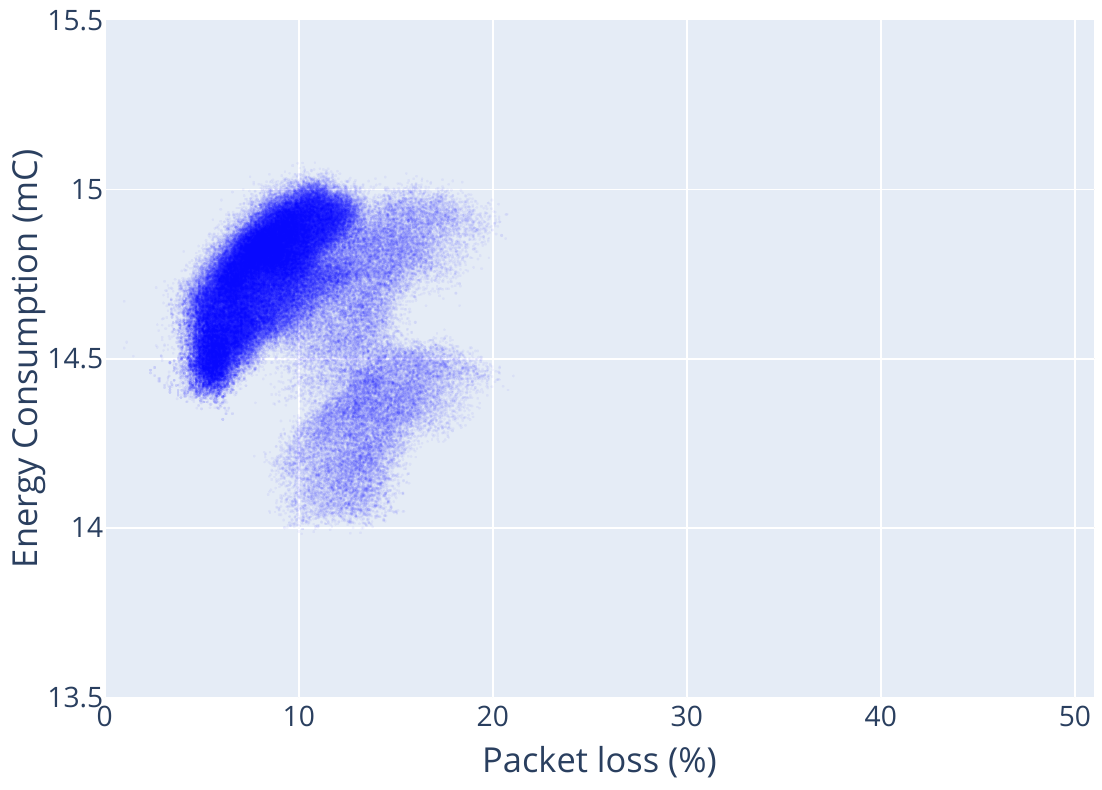}
		\caption{Adaptation spaces over 180 cycles}
		\label{fig: sample of quality attributes distribution in a period}
	\end{subfigure}
	
	\caption{Left: the estimated quality properties of all the adaptation options of DeltaIoTv1.1 at one point in time. Right: the distribution of the quality properties for all adaptation options over 180 adaptation cycles.} 
	\label{fig:sample of quality attributes distribtion}
\end{figure}

The left part of Figure~\ref{fig:sample of quality attributes distribtion} shows the estimated quality properties of the adaptation options in DeltaIoTv1.1 made by a verifier at runtime in one particular cycle, i.e., the adaptation space at that time. Each point shows the values of the estimated quality properties for one adaptation option. The right part of Figure~\ref{fig:sample of quality attributes distribtion} shows the distribution of the quality properties for all adaptation options over 180 adaptation cycles, i.e., a plot of all the adaptation spaces over 180 cycles.

\subsubsection{Adaptation Goals}

\begin{figure}
	\centering
	\includegraphics[scale=0.5]{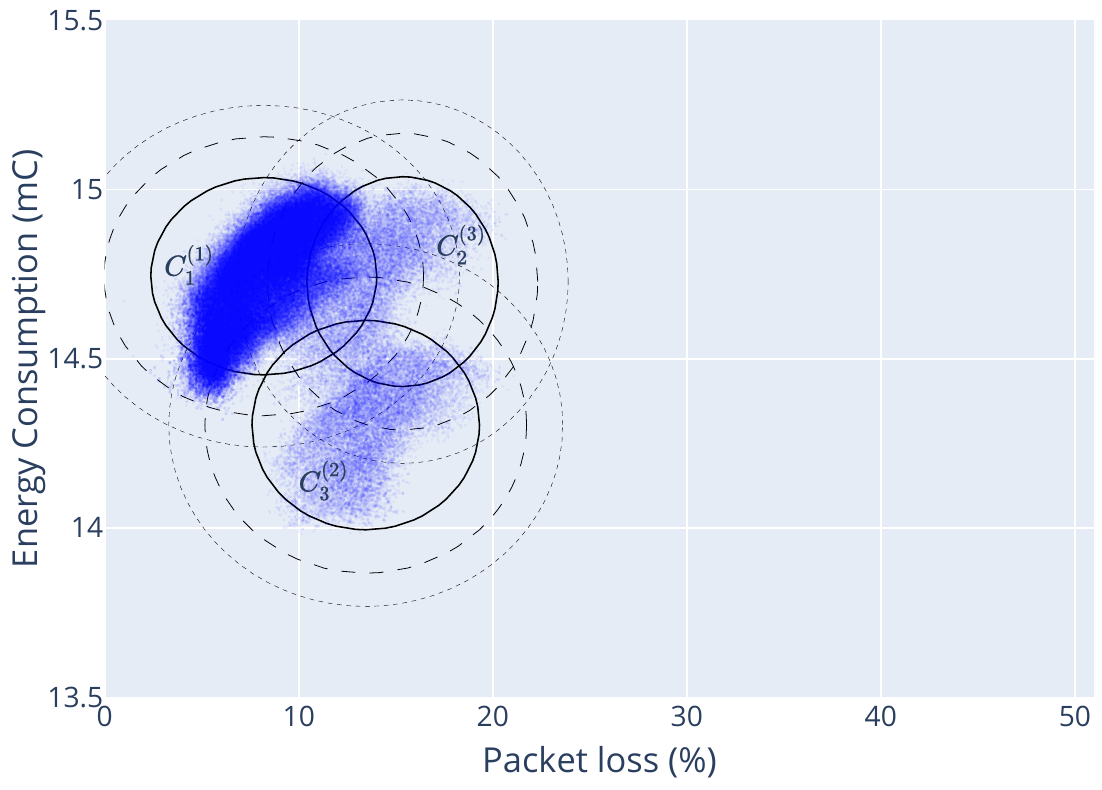}
	\caption{Stakeholder-defined classes (elliptic curves) created by fitting Gaussian distributions over selected points. The preference order of the stakeholders over these classes is determined by $m$ in $C_i^{(m)}$ for class $i$.
		\label{fig:sample of quality attributes distribtion with GMM}
	}
\end{figure}

Figure~\ref{fig:sample of quality attributes distribtion with GMM} shows how the adaptation goals of the system are defined. 
Specifically, the adaptation goals are defined by the stakeholders as a ranking of regions in the plane of the quality properties of the system. This ranking is based on the \emph{preference order} of the stakeholders. As an example, stakeholders may have a preference order $\langle$\emph{``less packet loss'', ``less energy consumption''}$\rangle$, which means stakeholders prefer less packet loss over less energy consumption. 
Technically, the overall goal of the system is defined based on a ranking over a set of classes characterized by a mixture of Gaussian distributions.\footnote{A Gaussian Mixture Model (GMM) is a statistical method to represent the distribution of data by a mixture of Gaussian distributions~\cite{bishop2006pattern}.} 
This ranking then expresses the preference order of the stakeholders in terms of configurations of the system with particular quality properties. 
The stakeholder-defined classes (represented by contour elliptic curves) are created by fitting\footnote{To fit a distribution model over specific data, we use a common algorithm in statistics called Expectation-Maximization. This algorithm allows performing maximum likelihood estimation~\cite{543975,bishop2006pattern}.} of Gaussian distributions over selected points (pairs of quality attributes) pertinent to each class. Each Gaussian distribution 
comprises three ellipsoids that show sigma, 2-sigma, and 3-sigma boundaries around the mean, from inside to outside, respectively.
The preference order of the stakeholders over classes is determined by $m$ in $C_i^{(m)}$ for each class $i$. 
As each Gaussian distribution is defined over the infinity range (from $-\infty$ to $+\infty$), each point is assigned to every class (here, three classes) with a probability. 
Thus, the class corresponding to the highest probability is assigned to the data point, i.e., the adaptation option with its estimated qualities at that point in time. Note that the approach for lifelong self-adaptation we propose in this paper allows an operator to interact with the system to determine or adjust the ordering of the classes on behalf of the stakeholders. 

\subsubsection{Learning-based Decision-making}

\begin{figure}
	\centering
	\includegraphics[width=\textwidth]{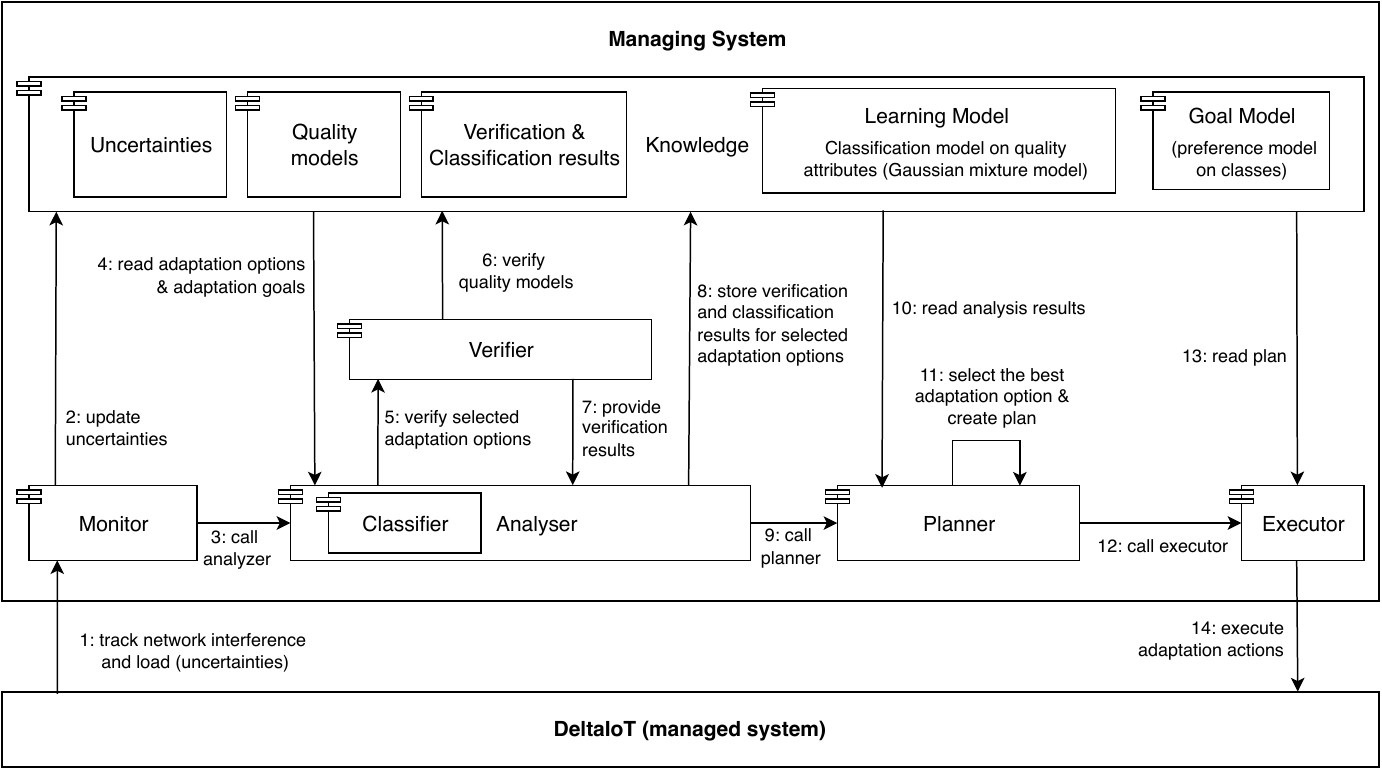}
	\caption{Learning-based self-adaptive system for DeltaIoT}
	\label{fig: self-adaptive system instance }
\end{figure}

The internal structure of the managing system shown in Figure~\ref{fig: self-adaptive system instance } follows the Monitor-Analyse-Plan-Execute-Knowledge reference model (MAPE-K)~\cite{Kephart, empiricalMAPE}.
An operator initiates the learning model and the goal model \re{(or interchangeably preference model)} based on offline analysis of historical data before deploying the system. The learning model consists of a classification model of quality attributes. The goal model expresses the ordering of the classes in the classification model based on the preference order of the stakeholders (e.g., see Figure~\ref{fig:sample of quality attributes distribtion with GMM}). 

The monitor component tracks the uncertainties (network interference of the wireless links and load of messages generated by the motes) and stores them in the knowledge component (steps 1 and 2). Then the analyzer component is triggered (step 3). 

Algorithm~\ref{algo: DM best option selection} describes the analysis process (steps 4 to 8). The analyzer starts with reading the necessary data and models from the knowledge component (step 4, lines~\ref{line: load uncertainties} and \ref{line: random shuffle}). Then, the analyzer uses the classifier to find an adaptation option with quality attributes that is classified as a member of a class with the highest possible ranking in the preference model. The selection of adaptation options happens randomly (by random shuffling of adaptation options in line~\ref{line: random shuffle}). This option is then verified using the verifier (step 5, line~\ref{line: verification}). To that end, the verifier configures the parameterized quality models, one for each quality property, and initiates the verification (step 6). The parameters that are set are the actual uncertainties and the configuration of the adaptation option. We use in our work statistical model checking for the verification of quality models that are specified as parameterized stochastic timed automata, using Uppaal-SMC~\cite{david2015uppaal,10.1145/2593929.2593944} (line~\ref{line: verification}). For details, we refer to~\cite{10.1145/3522585,10.1145/2593929.2593944}.   
When the analyzer receives the verification results, i.e., the estimated qualities for the adaptation option (step 7), it uses the classifier to classify the data point based on the Gaussian Mixture Model (GMM) (line~\ref{line: classification}).\footnote{Technically, this means that the probability of observing the data point based on the distribution pertinent to the assigned class is higher than other distributions in the GMM.}
This loop of verification and classification (steps 5 to 7) is repeated until the analyzer finds an adaptation option with a probability of being a member of the best class that is higher than a specific threshold\footnote{The probability threshold (PROB\_THR) for detecting outliers, i.e., options for which the estimated verification results are not a member of any of the classes provided by the stakeholders, is determined by 3-sigma rule~\cite{3SigmaCzitrom1997statistical}, here $0.001$.} (line~\ref{line: probability threshold}).
If no option of the best class is found\footnote{Note that due to drift caused by the uncertainties, there may not always be one member of each class among the available adaptation options in each adaptation cycle.} (line~\ref{line: is class member}), the loop is repeated for the next best-ranked class in the preference model (line~\ref{line: first loop}), etc., until an option is found. 
Alternatively, the analysis loop ends when the number of detected outliers exceeds a predefined threshold\footnote{The threshold for the number of outliers (COUNTER\_THR) is determined based on the ratio of outliers among pairs of quality attributes for adaptation options. Assume that $40$ percent of data can be an outlier. The likelihood of hitting $10$ consecutive outliers while the analyzer randomly iterates through adaptation options (line~\ref{line: random shuffle}) is less than $(0.4)^{10} \approx 0.0001$.} (line~\ref{line: probability threshold}). 
To ensure that each candidate adaptation option is verified only once (line~\ref{line: check duplicate key} to \ref{line: end loading cache}), all verification and classification results are stored (we empirically determined data of the 1000 most recent cycles) in a dictionary of the knowledge (lines~\ref{line: store verification} and \ref{line: store classification}). Finally, the analyzer stores the verification and classification results in the knowledge component (step 8). 


 \begin{algorithm}[htbp]
     \caption{Analysis stage of the managing system}
     \label{algo: DM best option selection}
     \begin{algorithmic}[1]
     	 \State $\mathrm{PROB\_THR}\leftarrow 0.001$
      	 \State $\mathrm{COUNTER\_THR}\leftarrow 10$
         \State $\mathrm{uncs} \gets \mathrm{K.tracked\_uncertainties[current\_cycle\_num]}$ 
         \label{line: load uncertainties}
         \Comment{SNRs + load of messages}
         \State $\mathrm{adpt\_opts} \gets \mathrm{random\_shuffle}(\mathrm{K.adaptation\_options})$ 
         \label{line: random shuffle}
		 \State $\mathrm{outlier\_counter}\leftarrow 0$
		 \ForEach{$\mathrm{p\_class\_index}$ $\mathrm{\textbf{in}}$ $\mathrm{preference\_model}.\mathrm{class\_rankings}$}
   \label{line: first loop}
		 \ForEach{$\mathrm{option}$ $\mathrm{\textbf{in}}$ $\mathrm{adpt\_opts}$}
   \label{line: second loop}
		 \State $\mathrm{key} \leftarrow (\mathrm{current\_cycle\_num}, \mathrm{uncs}, \mathrm{option})$
		 \If{$\mathrm{key}$ $\mathrm{\textbf{exists}}$ $\mathrm{\textbf{in}}$ $\mathrm{K.results}$}
		 \label{line: check duplicate key}
		 	\State $\mathrm{packet\_loss}, \mathrm{energy\_consumption} \leftarrow  \mathrm{K.results}[\mathrm{key}].\mathrm{verification}$
		 	\State $\mathrm{class\_index}, \mathrm{class\_prob} \leftarrow \mathrm{K.results}[\mathrm{key}].\mathrm{classification}$
    \label{line: end loading cache}
		 \Else
		 	\State $\mathrm{packet\_loss}, \mathrm{energy\_consumption} \leftarrow$ $\mathrm{Verifier}.\mathrm{verify}(\mathrm{uncs}, \mathrm{option})$
		 	\label{line: verification}
		 	\State $\mathrm{K.results}[\mathrm{key}].\mathrm{verification}\leftarrow (\mathrm{packet\_loss}, \mathrm{energy\_consumption}) $
		 	\label{line: store verification}
		 	\State $\mathrm{class\_index}, \mathrm{class\_prob}\leftarrow\mathrm{Classifier}.\mathrm{classify}(\mathrm{packet\_loss},\mathrm{energy\_consumption})$
		 	\label{line: classification}
		 	\State $\mathrm{K.results}[\mathrm{key}].\mathrm{classification} \leftarrow (\mathrm{class\_index}, \mathrm{class\_prob})$
		 	\label{line: store classification}
		 \EndIf

		 \If{$\mathrm{class\_prob} > \mathrm{PROB\_THR}$ \textbf{or} 
		 	$\mathrm{outlier\_counter} > \mathrm{COUNTER\_THR}$} 
		 \label{line: probability threshold}
		 
		 \Comment{3-sigma rule for detecting possible outliers}
 		 \If{$\mathrm{class\_index} == \mathrm{p\_class\_index}$}
	 	 \label{line: is class member}
		 \State	\Return
		 \EndIf
		 \Else
		 \State $\mathrm{outlier\_counter}\leftarrow \mathrm{outlier\_counter} + 1$ 
		 \EndIf
		 \EndFor
		 \EndFor

     \end{algorithmic}
 \end{algorithm}

After the analysis stage, the planner is triggered (step 9). The planner uses the analysis results (step 10) to select the best adaptation option, i.e., the option found by the analyzer in the highest possible ranked class according to the preference model of the stakeholders and generates a plan to adapt the managed system (step 11 in Figure~\ref{fig: self-adaptive system instance }). Finally, the executor is called (step 12).

The executor then reads the adaptation plan (step 13) and enacts the adaptation actions of this plan via the gateway to the motes of the IoT network (step 14).

\subsection{Problem of Drift in Adaptation Spaces in DeltaIoT}
\label{sec: problem}
The problem with a classification approach as exemplified by the approach we described for DeltaIoT is that the quality attributes of the adaptation options may change over time due to uncertainties. 
\re{
For example, during construction works, equipment or physical obstacles may increase the noise in the environment on a set of specific links. Consequently, the packet loss and energy consumption of the configurations that route packets via these links will be affected.
}
As a result, the classes of adaptation options with particular values for quality attributes may disappear or new classes may appear over time that are not part of the initial classification model. We refer to this problem as a drift of adaptation spaces.  This phenomenon may deteriorate the precision of the classification model and, hence, the reliability and quality of the system.

\begin{figure}[!ht]
	\centering
	\includegraphics[scale=0.5]{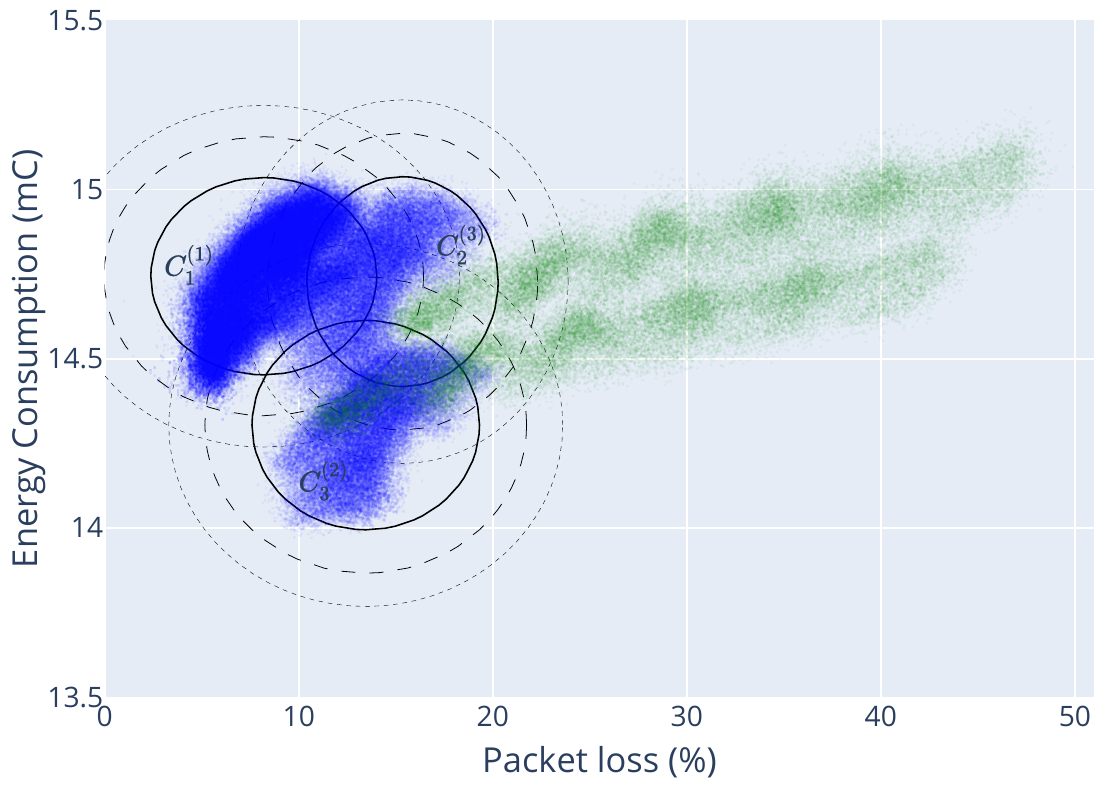}
	\caption{An example of novel class appearance in DeltaIot (cf. Figure~\ref{fig:sample of quality attributes distribtion with GMM}). Blue points refer to the quality attributes for adaptation options over the initial 250 adaptation cycles. The green points refer to cycles 250 to 300.}
	\label{fig: class drift example}
\end{figure}

Figure~\ref{fig: class drift example} shows an instance of novel class appearance (cf. Figure~\ref{fig:sample of quality attributes distribtion with GMM}) for the setup shown in Figure\,\ref{fig:deltaiotv1.1}. This plot represents the distribution of the quality attributes of adaptation options over 300 cycles. The blue points are related to cycles 1 to 250, and the green points to cycles 250 to 300. Although the position of some of the adaptation options (with their quality attributes) after cycle 250 (green points) were derived from initially defined classes (distributions), clearly most of the adaptation options after cycle 250 are not part of these initial distributions and form a new class or classes.

\begin{figure}[!ht]
	\centering
	\includegraphics[scale=0.5]{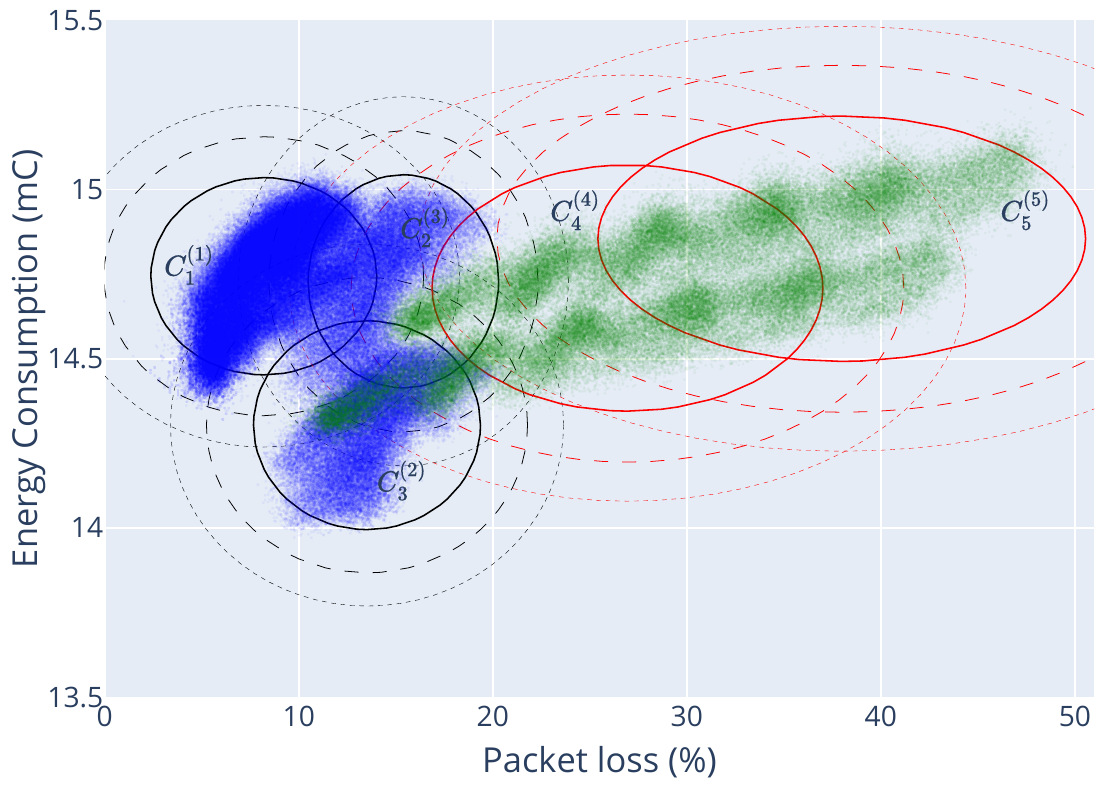}
	\caption{Ideal classifier with classes based on verification results of all adaptation options in all 350 adaptation cycles. The red elliptic curves (marked with $C^{4}_4$ and $C^{5}_5$) represent distributions of new data points that are not a member of any initially introduced classes (black elliptic curves) defined by the stakeholder.}
	\label{fig: ideal classification}
\end{figure}

To demonstrate the impact of this drift on the quality attributes of the IoT system, we compare the packet loss and energy consumption of the self-adaptive systems with a predefined classifier and with an ideal classifier (baseline) over 350 cycles. The ideal classifier used the verification results of all adaptation options in all cycles (determined offline as it takes three days to compute the ideal classifier). The stakeholders could then mark the different classes as shown in 
Figure~\ref{fig: ideal classification}. 
\re{
Note that the distribution of the population of each class obtained with the ideal classifier is close to its corresponding perfect Gaussian distribution, measured using sliced-Wasserstein distance\footnote{\re{The Wasserstein distance is a common metric to measure the similarity between two distributions. This distance has a possible range of values from zero to infinity, where a value of zero implies that the two distributions are identical. However, as it is computationally intractable, a simpler method like the sliced-Wasserstein is commonly used to approximate the distance. The python library~\cite{flamary2021pot} supports measuring the sliced-Wasserstein distance.}}~\cite{bonneel2015sliced} 
($0.004\pm 1\mathrm{e}{-4}$, $0.003\pm 1\mathrm{e}{-4}$, $0.002\pm 2\mathrm{e}{-4}$, $0.008\pm 4\mathrm{e}{-4}$, and $0.009\pm 6\mathrm{e}{-4}$ --- normalized to $[0\ldots1]$ 
by the maximum distance between any two points of the data --- for classes 1 to 5, respectively\footnote{\re{As sliced-Wassertein is using the Monte Carlo method for approximating the Wasserstein metric, the measurement has some errors. Here, we reported the expected value and the standard deviation of the measurements for each class.}}). }

\begin{figure}[htbp]
	\centering
	\begin{subfigure}[b]{0.49\textwidth}
		\includegraphics[width=\textwidth]{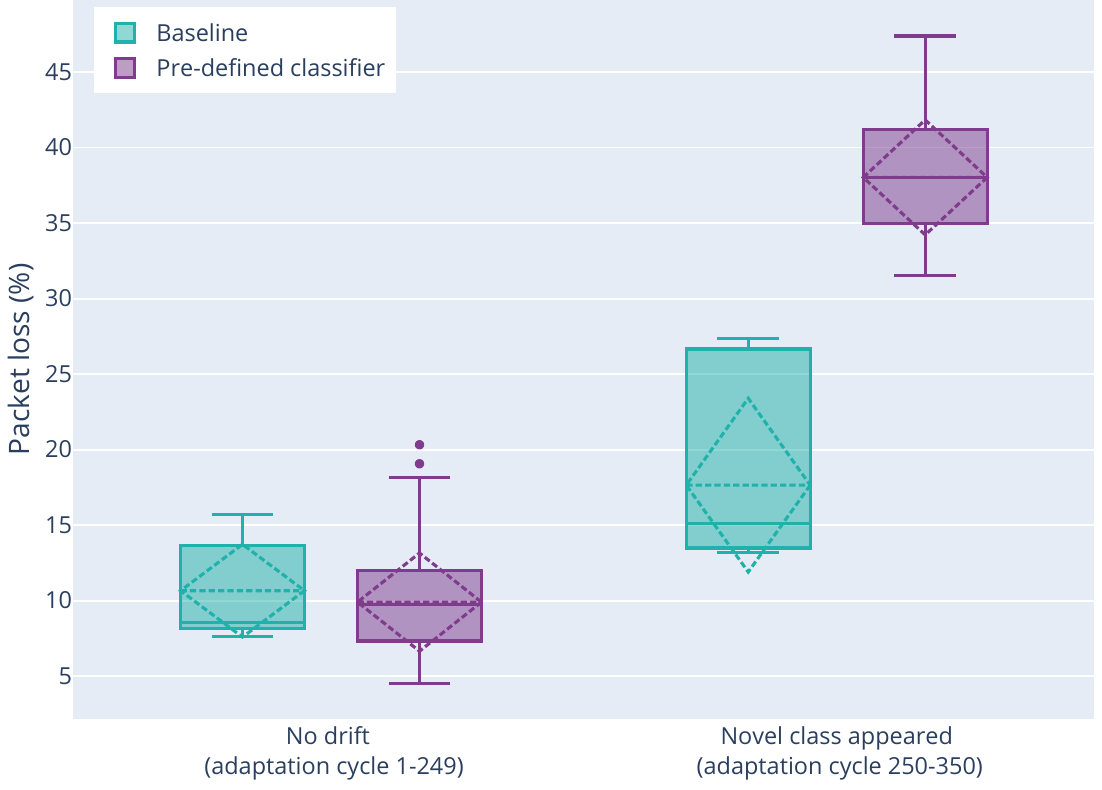}
		\caption{Impact on packet loss.}
		\label{fig: impact on pl to show the issue}
	\end{subfigure}
	\begin{subfigure}[b]{0.49\textwidth}
		\includegraphics[width=\textwidth]{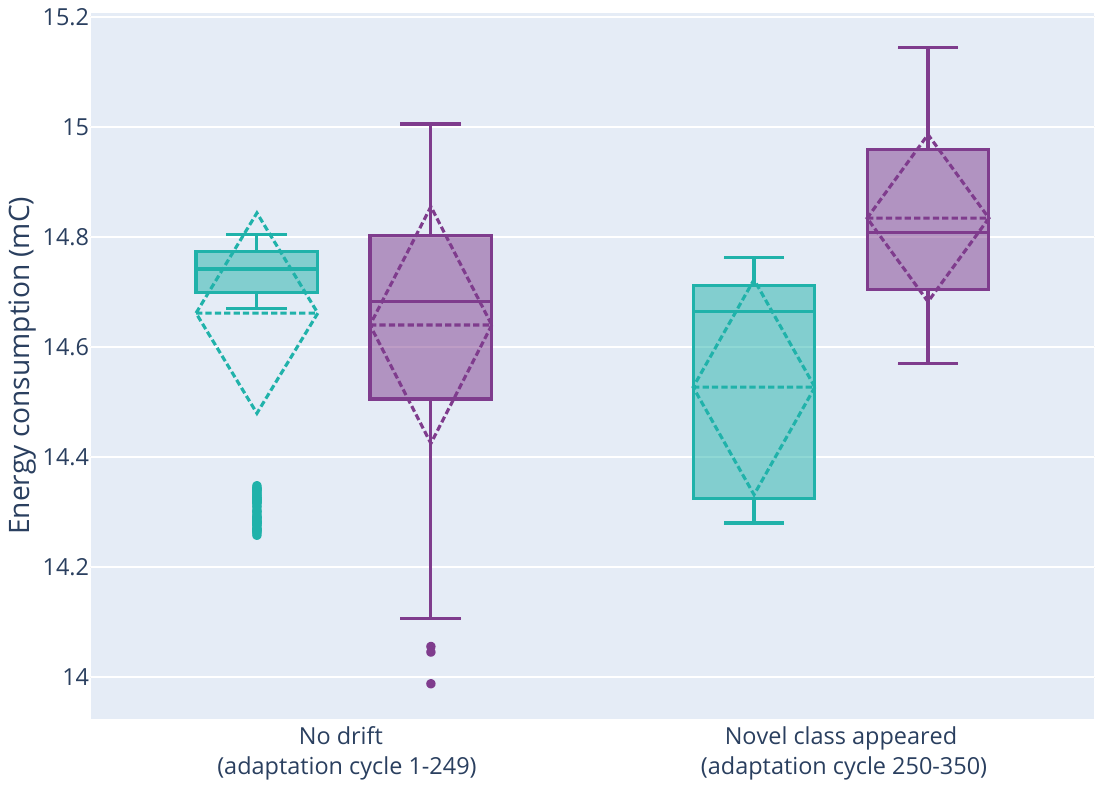}
		\caption{Impact on energy consumption.}
		\label{fig: impact on ec to show the issue}
	\end{subfigure}

	\caption{Impact of a shift of adaptation spaces (novel class appearance) on the quality attributes of the system for a predefined classifier compared to an ideal classifier (that serves as a baseline).}\vspace{-10pt}
	\label{fig: imapct on quality attribures to show the issue}
\end{figure}

We compare the two versions of the classifier with and without drift over 350 cycles for DeltaIoTv1.1 shown in Figure\,\ref{fig:deltaiotv1.1}. Figure~\ref{fig: imapct on quality attribures to show the issue} shows the impact of the drift on the quality attributes of the system. For packet loss (\ref{fig: impact on pl to show the issue}), we observe an increase of the difference of the mean from $-0.75$\,\% (mean $9.91$\,\% for pre-defined classifier and $10.66$\,\% for the baseline) to $20.38$\,\% ($38.03$\,\% for pre-defined classifier and $17.65$\,\% for the baseline). For energy consumption (\ref{fig: impact on ec to show the issue}), we also observe an increase of the difference of the mean from a small difference of $-0.02$\,mC ($14.64$\,mC for pre-defined classifier and $14.66$\,mC for the baseline) to $0.30$\,mC ($14.83$\,mC for pre-defined classifier and $14.53$\,mC for the baseline).\footnote{Note that the classifiers randomly select adaptation options of the best class. This explains why the pre-defined classifier achieves a marginally better packet loss in the period without drift.}
The results make clear that the impact on packet loss is dramatic (over 20\% extra packet loss for the pre-defined classifier), while the effect on energy consumption is relatively small (0.30\,mC on 14.83\,mC is only 2$\%$ of the consumed energy).

\begin{figure}[htbp]
	\centering
	\begin{subfigure}[b]{0.55\textwidth}
		\includegraphics[width=\textwidth]{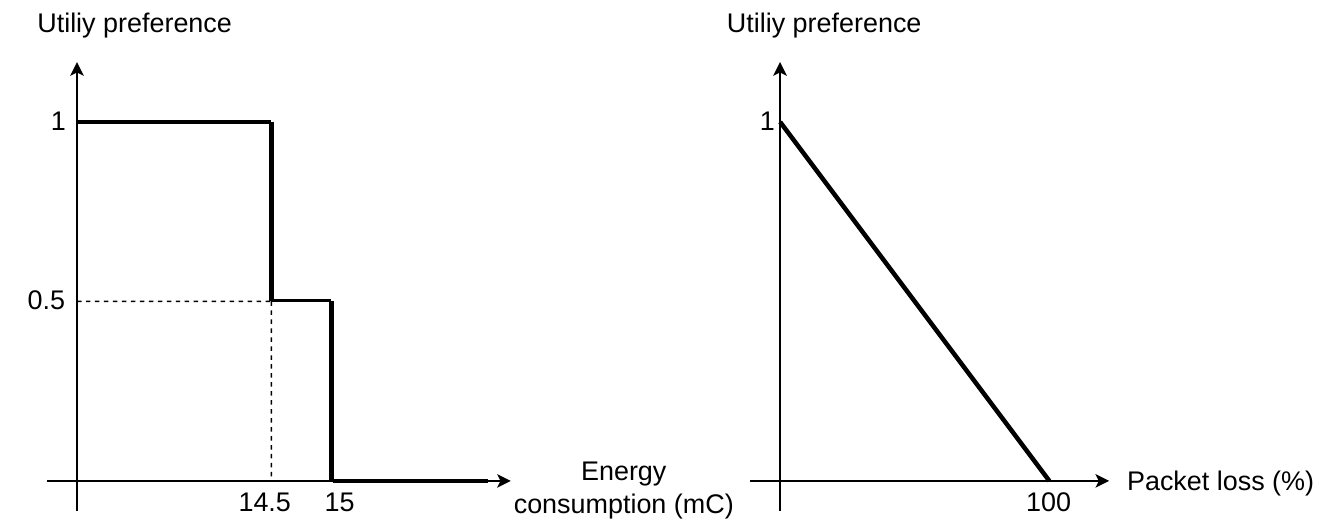}
		\caption{The utility response curves.}
		\label{fig: utility functions}
	\end{subfigure}
	\begin{subfigure}[b]{0.44\textwidth}
		\includegraphics[width=\textwidth]{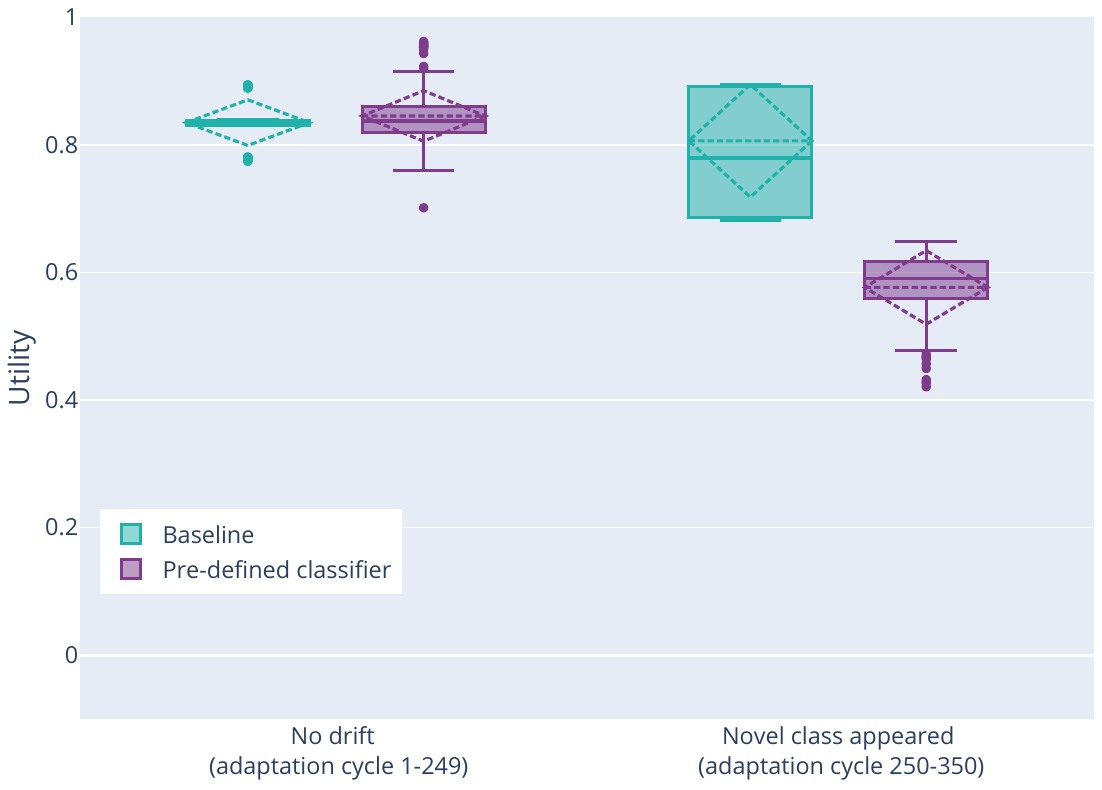}
		\caption{Impact on the utility.}
		\label{fig: impact on utility to show the issue}
	\end{subfigure}

	\caption{Left: utility response curves. Right: the impact of drift of adaptation spaces on the utility. }\vspace{-10pt}
	\label{fig: impact on the utility to show the issue}
\end{figure}

Since the impact on individual quality properties does not express the overall impact of the drift of adaptation spaces on the system, we also computed the utility of the system.  We used the definition of utility as used in\,\cite{10.1145/3524844.3528052} where the utility is defined as follows: 

\vspace{-2pt}
\begin{equation}
    U_c = 0.2 \cdot p_{ec} + 0.8 \cdot p_{pl}
\end{equation}

with $p_{ec}$ and $p_{pl}$ the utility for energy consumption and packet loss respectively, and $0.2$ and $0.8$ the weights associated with the quality properties. The values of the utilities are determined based on the utility response curves shown in\,\ref{fig: utility functions}. These functions show that stakeholders give maximum preference to an energy consumption below 14.5~mC, medium preference to an energy consumption of 14.5~mC to 15~mC, and zero preference to energy consumption above 15~mC. On the other hand, the utility for packet loss decreases linearly from one to zero for packet losses from 0\% to 100\%. 
Figure\,\ref{fig: impact on utility to show the issue} shows the results for the utility of the pre-defined classifier and the baseline split for the period before drift (cycles 1-249) and the period when drift of adaptation spaces occurs (cycles 250-350). The results show that the utility with the pre-defined classifier is close to the baseline for the period of no drift (mean 0.85 versus 0.83 respectively). Yet, for the period with drift, the utility for the pre-defined classifier drastically drops (mean 0.58 versus 0.81 for the baseline).

Comparing the individual quality attributes obtained with the two classifiers or using the utilities (derived from the quality attributes) provides useful insights into the effect of the drift of adaptation spaces on the system. Yet, these metrics offer either only a partial view of the effect of the drift of adaptation spaces (for individual quality attributes) or the interpretation is dependent on specific utility response curves and the weights of the utility function used to compute the utilities. Therefore, we introduced a new metric to measure 
of the drift of adaptation spaces that is based on the sum of the differences between the class ranking of the selected adaptation option in each adaptation cycle over a number of cycles for a pre-defined classifier and an ideal classifier (the baseline), respectively.\footnote{The ideal classifier classifies the options such that based on the stakeholder-defined ranking of the classes all classified options are in the correct class, while the classification of a practical classifier may make predictions that are not correct.}  
Then, this sum is normalized by the sum of the maximum value of this difference in each adaptation cycle over the number of cycles.  
Note that this number of cycles is domain-specific and needs to be determined empirically.\footnote{This number refers to the number of adaptation cycles within a cycle of the lifelong learning loop as we will explain later in the paper.}
We call this metric \textit{Ranking Satisfaction Mean} (\rsm).
The value for the \rsm\ falls within the range of $[0, 1]$. 
An \rsm\ of zero indicates that the 
performance (the accuracy of the predictions to classify adaptation options correctly according to the class ranking defined by the stakeholders) obtained with a given classifier is equal to the performance of an ideal classifier. 
An \rsm\ of $1$ represents the worst performance of a given classifier compared to an ideal classifier, i.e., the classifier classifies the options such that based on the ranking of the classes none of the classified options are in the correct class and the assigned class is the furthest class to the correct class based on the ranking. 
Formally, \rsm\ is defined as follows: 

\begin{definition}[Ranking Satisfaction Mean (RSM)]
Take a given classifier $R$ and an ideal classifier $R^*$ (here GMM classifiers) and a set of ranked classes  $C_1^{*(i_1)}, C_2^{*(i_2)}, \ldots, C_m^{*(i_m)}$ ($\langle i_1, i_2, \ldots, i_m\rangle$ is a permutation of $\langle1,2, \ldots, m\rangle$, ranking over classes $C_1^*$ to $C_m^*$). Also, suppose that ranks 
of selected adaptation options based on each of these classifiers, $R$ and $R^*$, employed by a managing system 
for $n$ adaptation cycles are respectively denoted by $r = \langle r_1, r_2, \ldots, r_n\rangle$ and $r^*=\langle r^*_1, r^*_2, \ldots, r^*_n\rangle$ ($r_i, r^*_i \in \{1,2,\ldots, m\}$ for all $i$ from $1$ to $m$). The \rsm\ of $r$ compared to $r^*$ is then defined as follows:

\begin{equation}
\mathrm{\textit{RSM}}_{r^*}(r) = \frac{1}{n\times(m-1)} \sum_{i=1}^n \left(r_i - r^*_i\right)
\end{equation}
 
\end{definition}

\begin{figure}[!ht]
	\centering
	\includegraphics[scale=0.4]{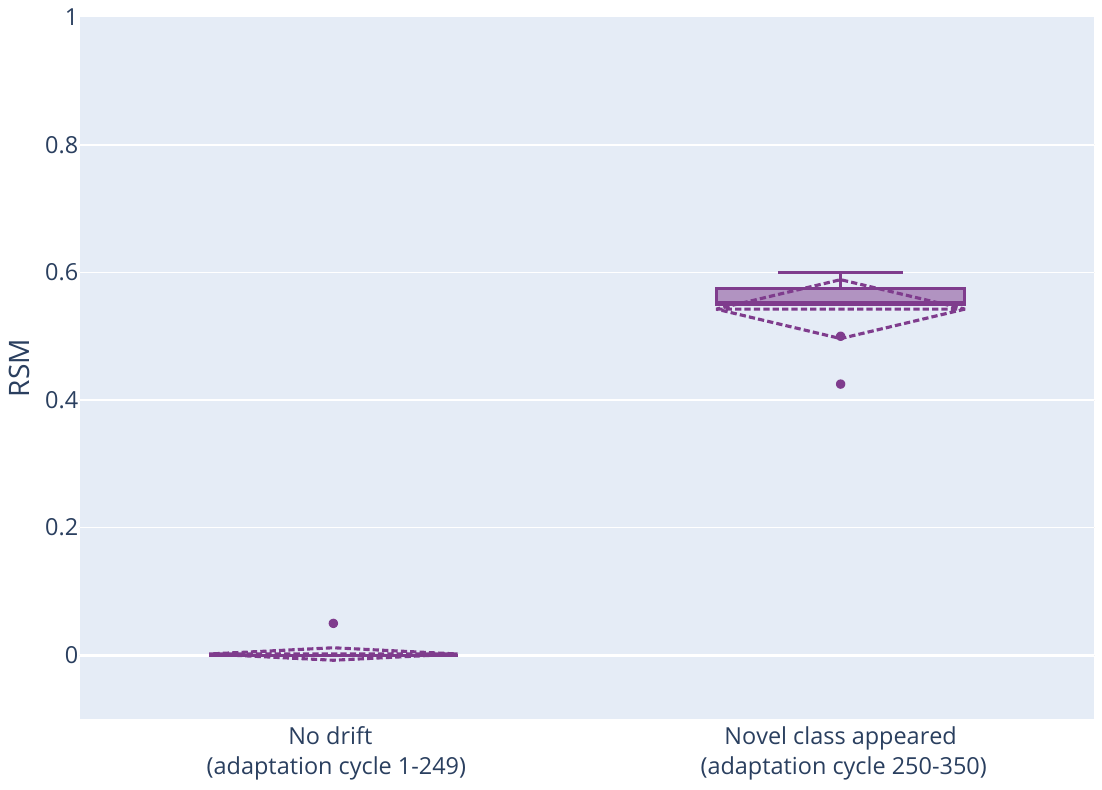}
	\caption{The impact of drift of adaptation spaces on \rsm.}
	\label{fig: rsm issue}\vspace{-10pt}
\end{figure}

Figure~\ref{fig: rsm issue} shows the impact of the drift of adaptation spaces (novel class appearance) on the \rsm, for every 10 adaptation cycles (i.e., $n = 10$) (the drift is illustrated in Figure~\ref{fig: class drift example}).
The mean of \rsm\ value remarkably increases from $0.002$ without drift to $0.543$ with drift. The 
results show that the performance of the pre-defined classifier before the drift occurs is quite close to that of the ideal classifier (baseline). However, once the drift occurs and novel classes appear, the performance of the pre-defined classifier drops significantly as demonstrated by the increased \rsm.

Based on Figures~\ref{fig: imapct on quality attribures to show the issue},~\ref{fig: impact on the utility to show the issue} and~\ref{fig: rsm issue}, we can conclude that drift of adaptation spaces (novel class appearance) in learning-based self-adaptive systems can hinder the system from reaching its adaptation goals and may drastically reduce the level of satisfaction of the stakeholders. 

\section{Lifelong Self-Adaptation to Deal with Shift of Adaptation Spaces}\label{sec:methodology}

We now introduce the novel approach of lifelong self-adaptation. We start with a general approach of lifelong self-adaptation that enables learning-based self-adaptive systems to deal with new learning tasks during operation. Then we instantiate the general architecture for the problem of learning-based self-adaptive systems that need to deal with shifts in adaptation spaces. 

\subsection{General \re{Architecture} of Lifelong Self-Adaptation}

We start with assumptions and requirements for lifelong self-adaptation. Then we present the architecture of a lifelong self-adaptive system and we explain how it deals with new tasks. 

\subsubsection{Assumptions for Lifelong Self-Adaptation}
 
\begin{sloppypar}
The assumptions that underlie lifelong self-adaptation are: 
\end{sloppypar}

\begin{itemize}
    \item The self-adaptive system comprises a managed system that realizes the domain goals for users in the environment and a managing system that interacts with the managed system to realize the adaptation goals; 
    \item The managing system is equipped with a learner that supports the realization of self-adaptation; 
    \item The self-adaptive system provides the probes to collect the data that is required for realizing lifelong self-adaptation; this includes an interface to the managing system and the environment, and an operator to support the lifelong self-adaptation process if needed;
    \item The managing system provides the necessary interface to adapt the learning models.  
\end{itemize}

In addition, \re{we assume that the (target) data relevant to new class appearances comprises a mixture of Gaussian distributions. Lastly,} we only consider new learning tasks that require an evolution of the learning models; runtime evolution of the software of the managed or managing system is out of the scope of the research presented in this paper.

\subsubsection{Requirements for Lifelong Self-Adaptation}

A lifelong self-adaptive system should:  

\begin{enumerate}
    \item[R1] Provide the means to collect and manage the data that is required to deal with new tasks; 
    \item[R2] Be able to discover new tasks based on the collected data; 
    \item[R3] Be able to determine the required evolution of the learning models to deal with the new tasks; 
    \item[R4] Evolve the learning models such that they can deal with the new tasks.   
\end{enumerate}

\subsubsection{Architecture of Lifelong Self-Adaptive Systems}

\begin{figure*}[t!]
    \centering
     \includegraphics[width=\linewidth]{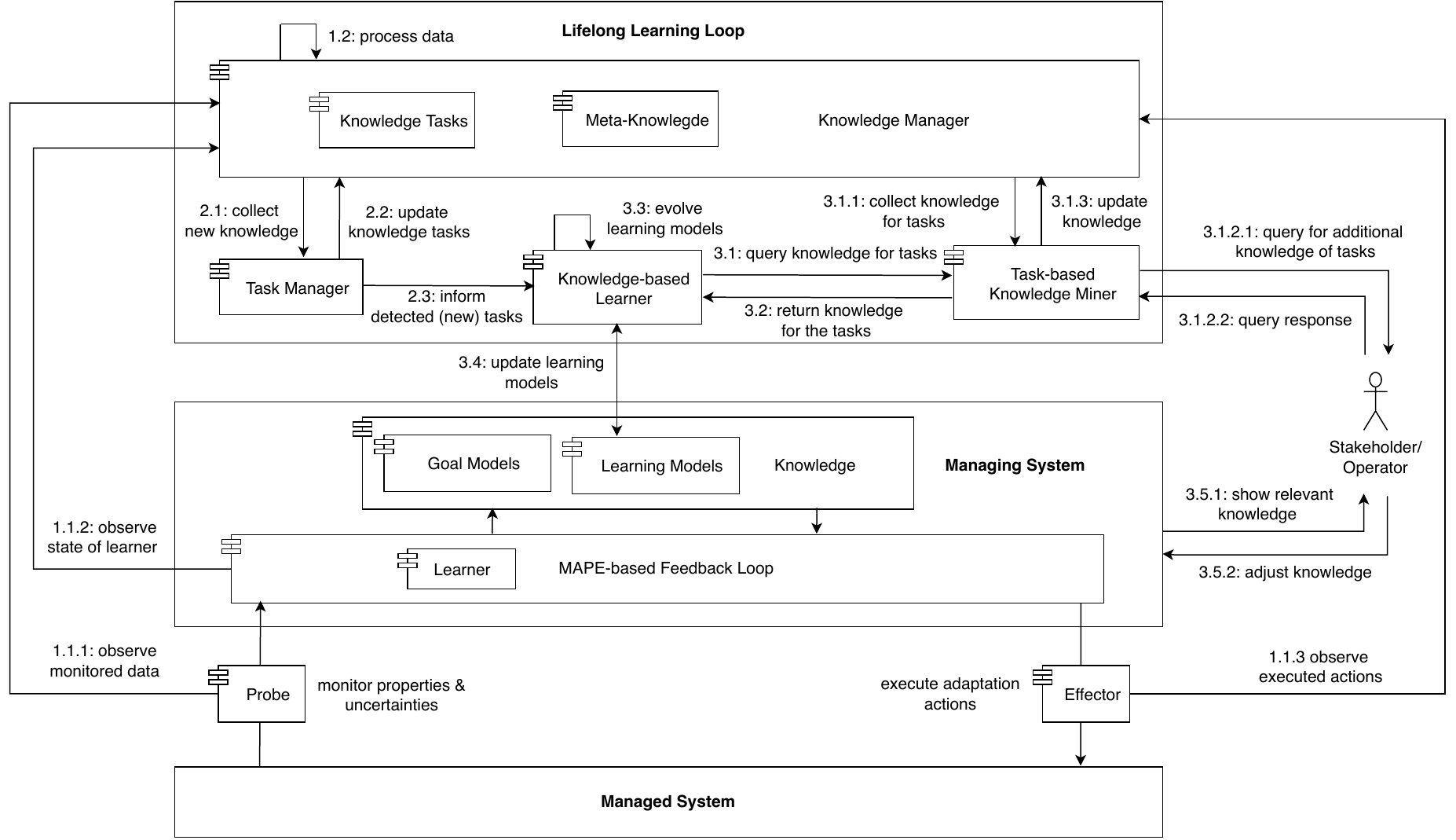}
    \caption{The architecture of a lifelong self-adaptive system
    \label{fig:lsa}\vspace{-5pt}}
\end{figure*}

Figure~\ref{fig:lsa} shows the architecture of a lifelong self-adaptive system. We explain the role of each component and the flow of activities among them. 

\paragraph{Managed System} Interacts with the users to realize the domain goals of the system. 

\paragraph{Managing System} Monitors the managed system and environment and executes adaptation actions on the managed system to realize the adaptation goals. The managing system comprises the MAPE components that share knowledge.
\re{The MAPE functions are supported by a learner, which primary aim is to solve a learning problem~\cite{gheibi2021}. Such problems can range from keeping runtime models up-to-date to reducing large adaptation spaces and updating adaptation rules or policies. The managing system may interact with an operator for input (explained below).
}

\paragraph{Lifelong Learning Loop} Adds a meta-layer on top of the managing system, leveraging the principles of lifelong machine learning. This layer tracks the layers beneath and when it detects a new learning task, it will evolve the learning model(s) of the learner \re{of the managing system} accordingly. We elaborate now on the components of the lifelong learning loop and their interactions.  

\paragraph{Knowledge Manager} Collects and stores all knowledge that is relevant to the learning tasks of the learner of the managing system (realizing requirement R1). In each adaptation cycle, the knowledge manager collects a knowledge triplet: $k_i$\,=\,$\langle$input$_i$,\,state$_i$,\,output$_i$$\rangle$. Input is the properties and uncertainties of the system and its environment (activity 1.1.1). State refers to data of the managing system relevant to the learning tasks, e.g., settings of the learner (1.1.2). Output refers to the actions applied by the managing system to the managed system (1.1.3). Sets of knowledge triplets are labeled with tasks $t_i$, i.e., $\langle$t$_i$, $\{$k$_u$,\,k$_v$,\,k$_w$$\}$$\rangle$, a responsibility of the task manager. The labeled triplets are stored in the repository with knowledge tasks. Depending on the type of learning tasks at hand, some parts of the knowledge triples may not be required by the lifelong learning loop.  

Depending on the problem, the knowledge manager may reason about new knowledge, mine the knowledge, and extract (or update) meta-knowledge, such as a cache or an ontology (1.2). The meta-knowledge can be used by the other components of the lifelong learning loop to enhance their performance. The knowledge manager may synthesize parts of the knowledge to manage the amount of stored knowledge (e.g., outdated or redundant tuples may be marked or removed). 

\paragraph{Task manager} Is responsible for detecting new learning tasks (realizing R2). The task manager periodically retrieves new knowledge triplets from the knowledge manager (activity 2.1). The duration of a period is problem-specific and can be one or more adaptation cycles of the managing system. The task manager then identifies task labels for the retrieved knowledge triplets. A triplet can be assigned the label of an existing task or a new task. Each new task label represents a (statistically) significant change in the data of the knowledge triplets, e.g., a significant change in the distribution of the data observed from the environment and managed system.
Hence, a knowledge triplet can be associated with multiple task labels, depending on the overlap of their corresponding data (distributions).  
The task manager then returns the knowledge triplets with the assigned task labels to the knowledge manager, which updates the knowledge accordingly (2.2). Finally, the task manager informs the knowledge-based learner about the new tasks (2.3).

\paragraph{Knowledge-based learner} Decides how to evolve the learning models of the learner of the managing system based on the collected knowledge and associated tasks (realizing R3), and then enacts the evolution of the learning models (realizing R4). To collect the knowledge it needs for the detected learning tasks, the knowledge-based learner queries the task-based knowledge miner (3.1) that returns task-specific data (3.2); the working of the task-based knowledge miner is explained below.
The knowledge-based learner then uses the collected data to evolve the learning models of the managing system (3.3). This evolution is problem-specific and depends on the type of learner at hand, e.g., tuning or retraining the learning models for existing tasks, or generating and training new learning models for newly detected tasks. The knowledge-based learner then updates the learning models (3.4). Optionally, the managing system may show these updates to the operator who may provide feedback (3.5.1 and 3.5.2), for instance ordering a set of existing and newly detected tasks  (e.g., the case where learning tasks correspond to classes that are used by a classifier).  

\paragraph{Task-based knowledge miner} Is responsible for collecting the data that is required for evolving the learning models for the given learning task by the knowledge-based learner (supports realizing R3). As a basis, the task-based knowledge miner retrieves the knowledge triplets associated with the given task from the knowledge tasks repository, possibly exploiting meta-knowledge, such as a cache (3.1.1). Additionally, the task-based knowledge miner can mine the knowledge repositories of the knowledge manager, e.g., to retrieve knowledge of learning tasks that are related to the task requested by the knowledge-based learner. Optionally, the task-based knowledge miner may collect knowledge from stakeholders, for instance, to confirm or modify the labels of knowledge triplets or newly detected tasks (activities 3.1.2.1 and 3.1.2.2). Finally, the miner uses new knowledge to update the knowledge maintained in the repositories by the knowledge manager (3.1.3), e.g., it may update meta-knowledge about related tasks or add data to the knowledge provided by stakeholders. 

\paragraph{Interplay between the knowledge-based learner and the learner of the managing system}

\begin{figure}
	\centering
	\includegraphics[scale=0.5]{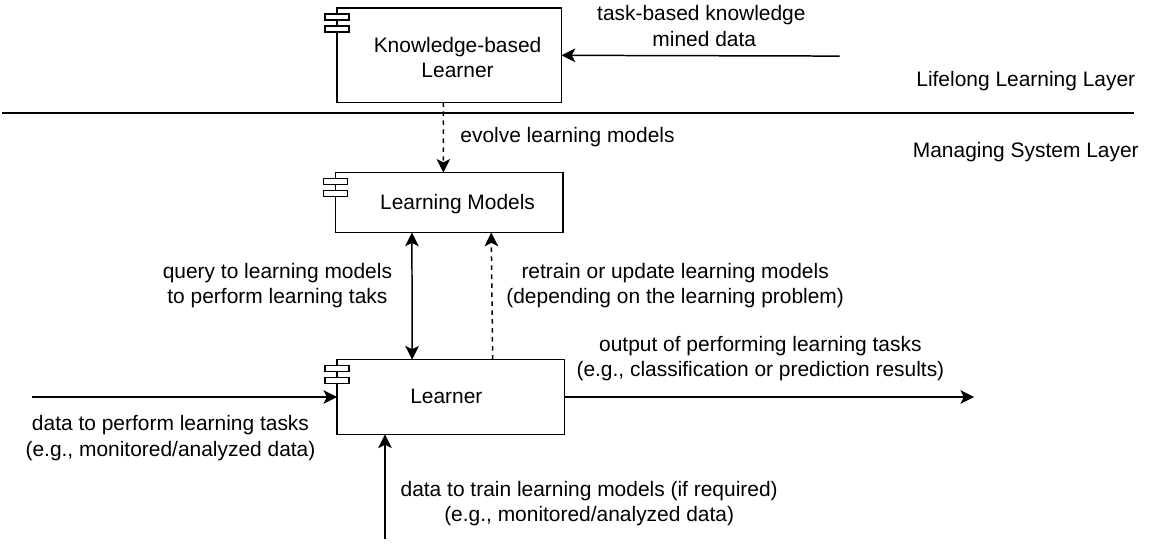}
	\caption{\re{Interplay between the knowledge-based learner and the learner of the managing system
 }}
	\label{fig: simplified lsa to show learners}
\end{figure}

Since both the knowledge-based learner and the learner of the managing system act upon the learning model as shown in Figure~\ref{fig: simplified lsa to show learners}, it is important to ensure that this interaction occurs in a consistent manner. 

Changes applied to learning models can be categorized into two main types: parametric changes and structural changes. Parametric changes pertain to adjustments made to the parameters of the current learning models. This may involve incrementally modifying the parameters of the learning model, like adjusting the vectors of a support vector machine based on newly observed training data or retraining the model with a new set of training data. On the other hand, structural changes involve alterations to the structure of learning models, such as adjusting the number of neurons or layers in a neural network, or even replacing an existing learning model with a new type of model, such as substituting a support vector machine with a Hoeffding adaptive tree.

Lifelong self-adaptation supports two approaches for changing the learning models: (1) the learner in the managing system uses the learning models to perform the learning tasks; the lifelong learning loop can apply structural changes to the learning models, and (2) the learner in the managing system performs the learning task and can apply parametric changes to the learning models (dotted lines in Figure~\ref{fig: simplified lsa to show learners}); the lifelong learning loop can apply structural changes to the learning models. The instance of architecture for lifelong self-adaptation used in the evaluation case in this paper applies approach (1). The instances of the architecture used in the cases of~\cite{10.1145/3524844.3528052} (that are summarised in Section~\ref{subsec:sidrift} below) apply approach (2). 

By properly allocating the concerns, i.e., applying the learning task and parametric changes of the learning models allocated to the managing system, and structural changes of the learning models allocated to the lifelong learner, we can optimally deal with potential conflicts when changing the learning models. 
Since there are only structural changes in the learning models for approach (1), no conflicts can occur (under the condition that the learning models are evolved atomically, i.e., without the interference of performing learning tasks by the learner of the managing system). For approach (2), a conflict may occur if the training data used by the managing system to update the learning models contradicts the data mined by the task-based knowledge miner to evolve the learning models. This may degrade the performance of the evolved learning models due to drift in the training data. To avoid this issue, the learner of the managing system should only update the evolved learning models based on new data that is observed between two consecutive cycles of the lifelong learning loop (to be certain that this data pertains to the current task of the system). Hereby, it is important to note that the cycle time of the lifelong learner is often multiple times longer than the cycle time of the learner of the managing system.\footnote{\re{Note that the cycle time may be adjusted based on changes in the frequency that drift occurs using methods as ADWIN~\cite{bifet2007learning}. However, considering the dynamic cycle time of the knowledge-based learner is out of the scope of this paper.}}

\re{
\subsubsection{Lifelong Self-Adaptation to Deal with Sudden and Incremental Covariate Drift}\label{subsec:sidrift}

In~\cite{10.1145/3524844.3528052}, we instantiated the general architecture for lifelong self-adaptive systems (shown in Figure~\ref{fig:lsa}) for two types of drift: recurrent sudden and incremental covariate drift. We briefly summarise these two instances of the general architecture here, for further details we refer the interested reader to~\cite{10.1145/3524844.3528052}.

For the case of recurrent sudden covariate drift, we instantiated the architecture of lifelong self-adaptive systems for DeltaIoT. 
Recurrent sudden covariate drift occurs when the distributions of input data that are used  by the managing system suddenly change and at some point may return to the same state, for instance, due to machinery that periodically produces noise patterns affecting the
signal-to-noise ratio of the network links in the IoT network. 
In this instance, the underlying managing system predicts the quality attributes of the adaptation options either using a stochastic gradient descent (SGD) regressor~\cite{saad1998online} or by switching between an SGD and a Hoeffding Adaptive Tree~\cite{HAT}.
The task manager of the lifelong learning loop in this instance uses auto-encoders~\cite{jaworski2020concept,yang2021cade, andresini2021insomnia} to detect new tasks (i.e., new distribution of features in the input data). 
The knowledge-based learner optimizes the hyper-parameters of each learning model using a Bayesian optimizer and then trains the best model based on the collected training data. 
The task-based knowledge miner simply fetches newly collected knowledge from the knowledge manager.
Hence, the lifelong learning loop operates without human intervention. 
When the learning models are trained, they are updated in the knowledge repository of the 
managing system.

For the case of incremental covariate drift, we instantiated the architecture of lifelong self-adaptive systems for a gas delivery station~\cite{vergara2012chemical}. The gas station composes substances to produce different types of gas that are routed to specific users. When composing the gas, there is uncertainty about the type of gas that is produced. To mitigate this uncertainty, a feedback loop collects data from sensors at the gas tank and uses a classifier, a multi-class support vector machine (SVC)~\cite{angulo2003k}, to predict the gas type. %
The valves for gas delivery are then set such that the gas is routed to the right users.  Similar to the first instance, the task manager of the lifelong learning loop uses auto-encoders to detect new tasks that emerge from drift in the measurements of gas sensors over time. 
The knowledge-based learner also uses a Bayesian optimizer to tune the hyper-parameters of the SVC learning model and train the most optimal model using the acquired training data. 
In contrast to the previous instance, the task-based knowledge miner requires feedback from the stakeholder on the labeling of some newly collected data  (because data labeling here needs some chemical tests by the stakeholder). The aim is to minimize the interaction with the stakeholder by minimizing the number of required data labeling using active learning. After completing the training, the knowledge repository of the managing system is updated with the latest model. 

In these two instances concept drift occurs in \textit{input features} of the learning model of the learner. In novel class appearance, on the other hand, the type of concept drift we study in this paper, drift occurs in the \textit{target} of the learner, i.e., the prediction space of the learning model. 
}

\subsection{Lifelong Self-Adaptation to Deal with Shift of Adaptation Spaces}

\begin{figure}
	\centering
	\includegraphics[width=\textwidth]{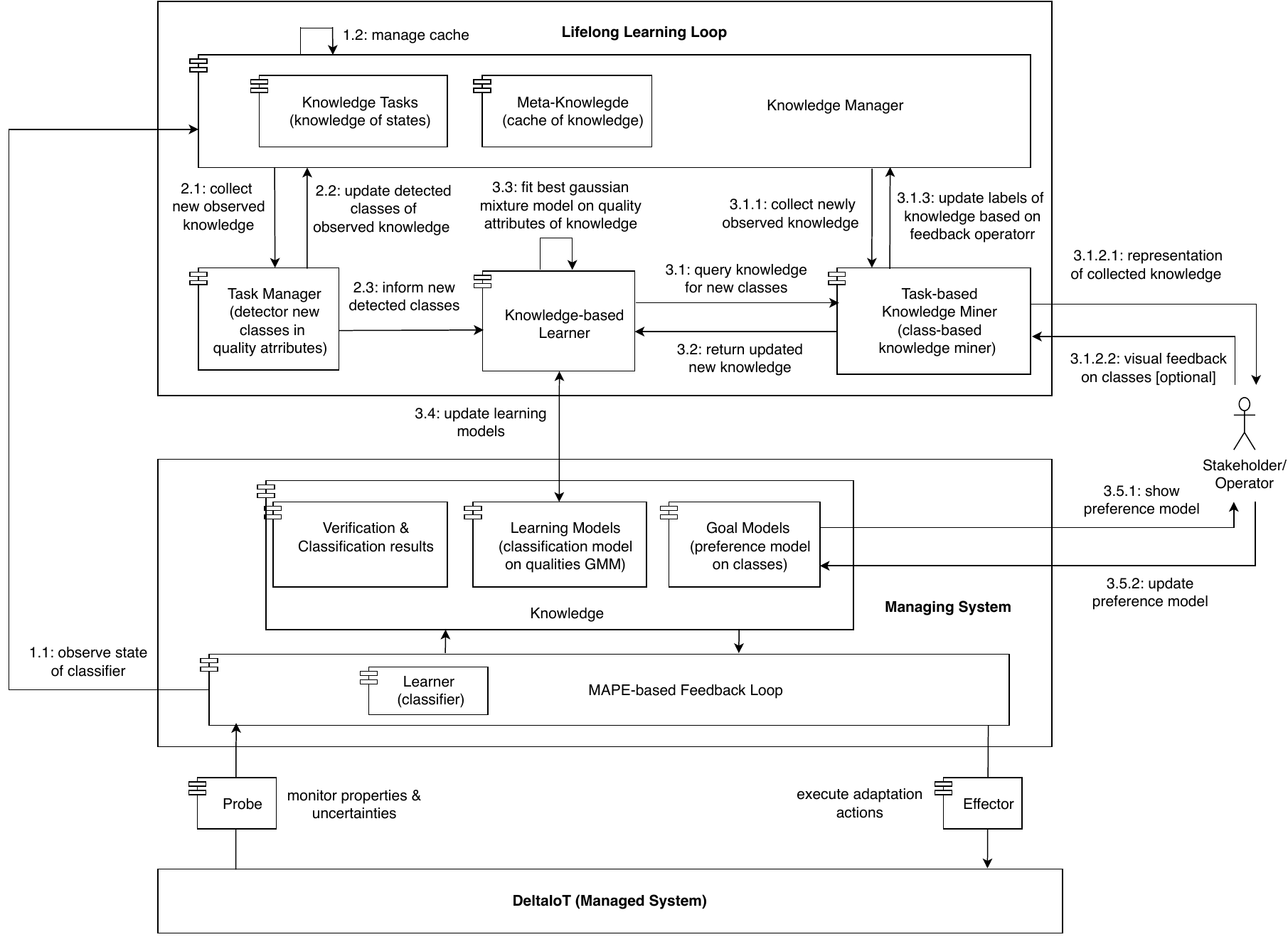}
	\caption{Architecture of lifelong self-adaptation to deal with drift of adaptation spaces illustrated for DeltaIoT.}
	\label{fig: solution to tackle the problem}
\end{figure}

We now instantiate the general architecture for lifelong self-adaptation to tackle the problem of shift in adaptation spaces in learning-based self-adaptive systems. Figure~\ref{fig: solution to tackle the problem} shows the instantiated architecture illustrated for DeltaIoT (the managed system). We elaborate on the high-level components and their interactions that together solve the problem of shift of adaptation spaces. 

\paragraph{Knowledge Manager (KM)}
The knowledge manager starts the lifelong learning loop in the $i$-th adaptation cycle by collecting the state of the classifier from the managing system,  denoted by $\mathrm{state}_i$ (link 1.1). This state includes the verification and classification results, the classification model of the learner on the quality attributes, and the preference model on classes (goal model). The knowledge manager stores and maintains a history of knowledge over a given window in a cache (link 1.2 in Figure\,\ref{fig: solution to tackle the problem}). Note that this instantiation of the architecture of lifelong self-adaptation does not use the input and the output of the knowledge triplet, see the explanation of the general architecture (links 1.1.1 and 1.1.3 in Figure\,\ref{fig: solution to tackle the problem} of the general architecture are not required in the instantated architecture, and link 1.1.2 the general architecture is represented as link 1.1).

\paragraph{Task Manager}
A ``task'' in the general architecture corresponds to a ``class'' in the instantiated architecture,  expressed by mixed Gaussian distributions on quality attributes. Hence, the task manager is responsible for detecting (new) classes based on the Gaussian Mixture Model (GMM) previously defined by the stakeholders (stored in the knowledge of states collected by the knowledge manager). Algorithm~\ref{algo: task manager} describes the detection algorithm in detail. 

\begin{figure}
	\centering
	\includegraphics[scale=0.5]{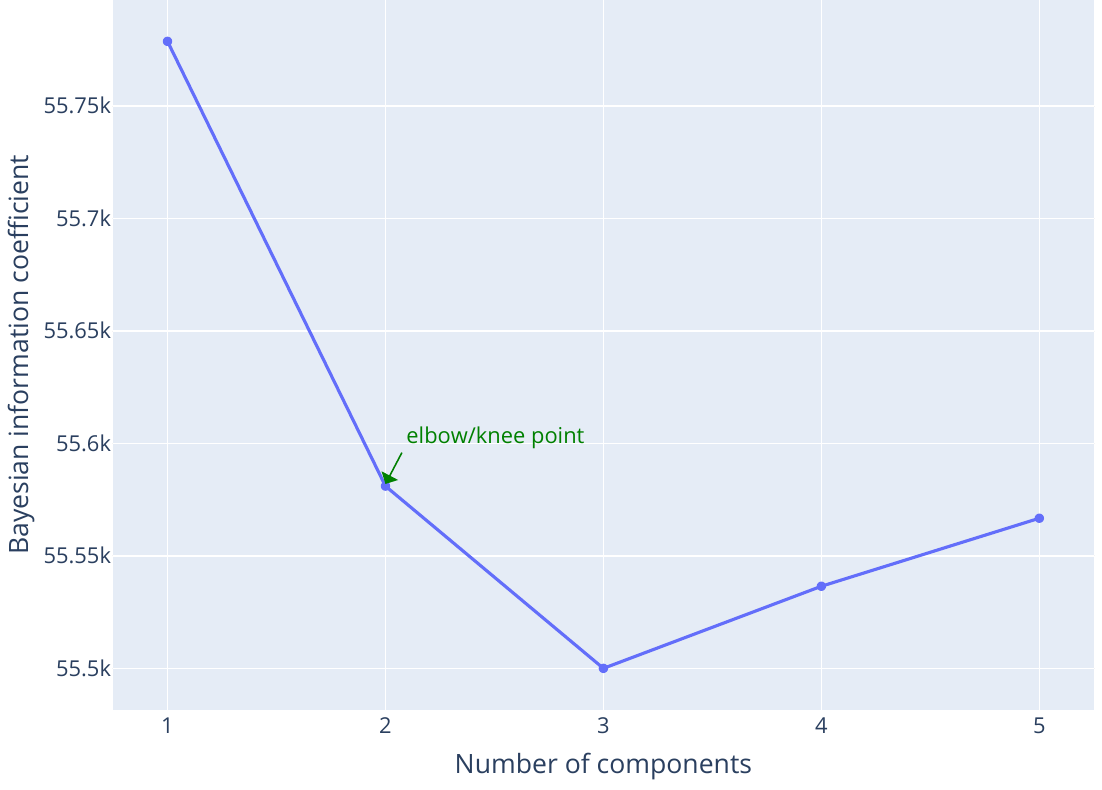}
	\caption{A BIC curve (based on the different number of components) for the quality attribute pairs corresponding to all adaptation options in adaptation cycles 151 to 160.}
	\label{fig: elbow example}
\end{figure}

The task manager starts with collecting the required knowledge of states (link 2.1 in Figure\,\ref{fig: solution to tackle the problem} and lines~\ref{line: start collecting} and \ref{line: end collecting} in the algorithm), i.e., recently observed states of the classifier\footnote{Each cycle of lifelong learning loop corresponds to multiple adaptation cycles. Here, we assume that the detection algorithm operates every $10$ adaptation cycles to detect new emerging classes. This factor $10$ that was determined empirically is a balance between minimising unnecessary computational overhead on the system and  being timely to mitigate destructive effects of the emerge of possible new classes in the system. \re{More specifically, as an adaptation cycle takes 10 minutes, experiments have shown that a 100-minute time frame, i.e., 10 cycles, is suitable to verify whether any alterations have occurred in the data, given the uncertainties in the environment. This duration appears to be neither too brief to overuse the lifelong learning loop nor excessively long to overlook a shift in this timeframe.}}, including the last goal model.
Then, using the classification model and the 3-sigma method~\cite{3SigmaCzitrom1997statistical}, the algorithm finds out which pairs of verified quality attributes in the collected states are characterized as data not belonging to existing classes, i.e., out-of-classes-attributes (lines~\ref{line: start out_of_class_detection} to \ref{line: end out_of_class_detection}).
The algorithm then computes the percentage out-of-classes-attributes of the total number of considered quality attributes in the collected states (line 13) and compares this with a threshold\footnote{This threshold for detecting newly emerged class(es) is determined based on domain knowledge \re{(e.g., the number of adaptation options and the rate of occurring drift that affects on the maximum possible number of classes that can appear)} and possibly empirically checked; $20$\,\% is a plausible threshold in the DeltaIoT domain. } (line~\ref{line: check out-of-class percentage}). If the percentage out-of-classes-attributes does not cross the threshold, the algorithm will terminate (do  nothing, line~\ref{line: do nothing}). However, if the threshold is crossed, some new classes have emerged, and the algorithm fits a GMM over the data (lines~\ref{line: find component num} and \ref{line: fit new classifier}).
The first step to fitting a GMM is determining the number of classes (or components) (line~\ref{line: find component num}). To that end, the algorithm uses the Bayesian Information Criterion curve~\cite{schwarz1978estimating, konishi2008information} (BIC curve) for the different number of classes.\footnote{The number of classes (components) in a BIC curve changes from 1 to 5, with the assumption that not more than 5 classes appear in the domain in each lifelong learning loop.}
Afterward, the algorithm employs a common method, called Kneedle algorithm~\cite{satopaa2011finding}, to find an elbow (or a knee) in this curve to specify a Pareto-optimal number of classes. 
For example, Figure~\ref{fig: elbow example} represents a BIC curve with an indicated elbow/knee point that occurs at 2 components, i.e., the distribution of the quality attribute pairs (during the specified adaptation cycles) can be reasonably expressed as the sum of two Gaussian distributions, meaning two classes.
After specifying the number of classes, the algorithm uses the expectation-maximization algorithm to fit a GMM over the out-of-classes-attributes (line~\ref{line: fit new classifier}). 
Then a new classification model is constructed by integrating the last classification model of the system and the newly detected one (line~\ref{line: integrating classifiers}). 
Finally, the task labels of all collected states are updated based on classification results obtained by applying the new classification model to the quality attributes related to the state (link 2.2 in Figure\,\ref{fig: solution to tackle the problem} and lines~\ref{line: start update labels} to \ref{line: end update labels} in the algorithm). This concludes the detection algorithm of the task manager. When the task manager detects some new class(es) it triggers the knowledge-based learner (link 2.3 in Figure\,\ref{fig: solution to tackle the problem}). 

 \begin{algorithm}[htbp]
	\caption{Detection of (new) classes in quality attributes (in task manager)}
	\label{algo: task manager}
	\begin{algorithmic}[1]
		\State  $\mathrm{OUT\_OF\_CLASS\_PERCENT\_THR}\leftarrow 20$
		\State $\mathrm{ADAPT\_CYCLE} \leftarrow 10$
		\State $\mathrm{out\_of\_class\_attrs}\leftarrow []$
		\State $\mathrm{states} \leftarrow \mathrm{KM}.\mathrm{Knowledge}[-\mathrm{ADAPT\_CYCLE}:]$ 
		\label{line: start collecting}
		\Comment{collect recent states of adaptation cycles}
		\State $\mathrm{classification\_model} \leftarrow \mathrm{states[end]}.\mathrm{Classfication\_Model}$
		\label{line: end collecting}

		\ForEach{$\mathrm{state} \in \mathrm{states}$}
		\label{line: start out_of_class_detection}
		\ForEach{$\mathrm{attr} \in \mathrm{state}.\mathrm{quality\_attributes}$ }
		\If{$\mathrm{classification\_model}.\mathrm{is\_out\_of\_class}(\mathrm{attr})$}
		\Comment{using 3-sigma method}
		\State $\mathrm{out\_of\_class\_attrs}.\mathrm{append}(attr)$
		\EndIf
		\EndFor
		\EndFor
		\label{line: end out_of_class_detection}
		\State $\mathrm{out\_of\_class\_percent} \leftarrow {100 \times|\mathrm{out\_of\_class\_attrs}|}/\left( {\sum_{\mathrm{state}\in\mathrm{states}}|\mathrm{state}.\mathrm{quality\_attributes}|}\right)$
		\If{$\mathrm{out\_of\_class\_percent} < \mathrm{OUT\_OF\_CLASS\_PERCENT\_THR}$}
		\label{line: check out-of-class percentage}
		\State \Comment{do nothing}\label{line: do nothing}
		\Else
		\Comment{new class(es) detected}
		\State $\mathrm{component\_num} \leftarrow \mathrm{find\_component\_num}(\mathrm{out\_of\_class\_attrs})$
		\label{line: find component num}
		\State $\mathrm{new\_model} \leftarrow \mathrm{fit\_GMM\_model}(\mathrm{out\_of\_class\_attrs}, \mathrm{component\_num})$
		\Comment{using EM}
		\label{line: fit new classifier}
		\State$\mathrm{classification\_model} \leftarrow \mathrm{classification\_model} + \mathrm{new\_model}$
		\label{line: integrating classifiers}
		\Comment{GMM model}
		\ForEach{$\mathrm{state}\in \mathrm{KM}.\mathrm{Knowledge}[-\mathrm{ADAPT\_CYCLE}:]$}
		\label{line: start update labels}
		\Comment{update task labels}
		\State $\mathrm{task\_labels}\leftarrow\mathrm{classification\_model}.\mathrm{classify}(\mathrm{state.Quality\_Attributes})$
		\State $\mathrm{KM}.\mathrm{Task\_Label}[\mathrm{state}]\leftarrow \mathrm{task\_labels}$
		\EndFor
		\label{line: end update labels}
		\EndIf
	\end{algorithmic}
\end{algorithm}

\paragraph{Knowledge-Based Learner}
When the knowledge-based learner is triggered by the task manager, it queries the task-based knowledge miner (link 3.1 in Figure\,\ref{fig: solution to tackle the problem}) to collect knowledge connected to the newly detected class(es) (link 3.2).
The knowledge-based learner then fits a GMM on the gathered data (verification results) and integrates with the last state of the GMM classification model of the system (link 3.3). Finally, the knowledge-based learner updates the goal classification model in the managing system with the created GMM (link 3.4). We elaborate on the interaction with the operator below (i.e., links 3.5.1 and 3.5.1). 

\begin{figure}[htbp]
	\centering
	\begin{subfigure}[b]{0.49\textwidth}
		\includegraphics[width=\textwidth]{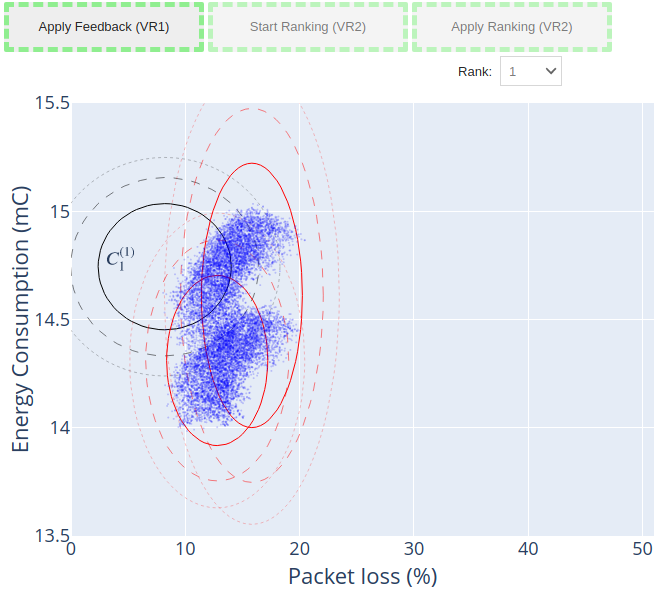}
		\caption{Initial classification shown to the operator.\vspace{10pt}}
		\label{fig: visualization example}
	\end{subfigure}
	\begin{subfigure}[b]{0.49\textwidth}
		\includegraphics[width=\textwidth]{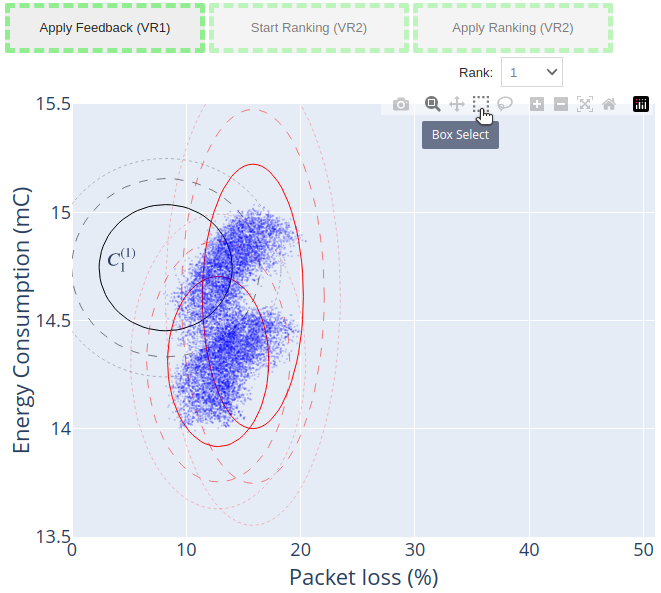}
		\caption{The operator select the ``Box Selection'' option.\vspace{10pt}}
		\label{fig: choosing box slection}
	\end{subfigure}
 \bigskip
	\begin{subfigure}[b]{0.49\textwidth}
		\includegraphics[width=\textwidth]{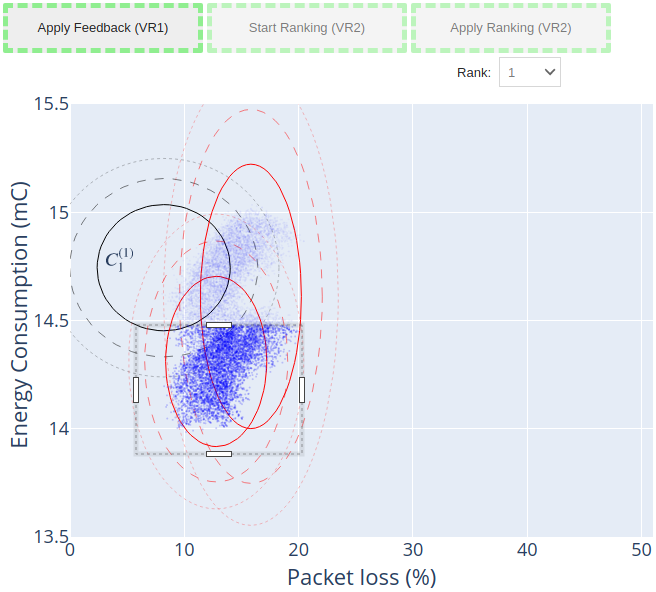}
		\caption{The operator marks a reasonable area of a new class and confirms by clicking ``Apply Feedback.''}
		\label{fig: box selecting}
	\end{subfigure}
        \begin{subfigure}[b]{0.49\textwidth}
		\includegraphics[width=\textwidth]{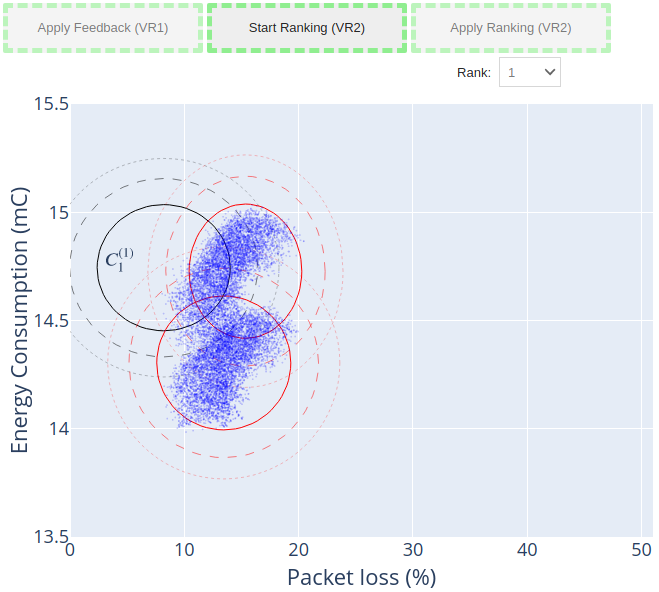}
		\caption{The operator starts ranking the new classification by clicking ``Start Ranking'' using the ``Rank'' option.  
        }
		\label{fig: apply feedback}
        \end{subfigure}
	\caption{Illustration of the 
  interaction of the operator with the task-based knowledge miner via a GUI.	\label{fig: feecdback VR1 example}}
\end{figure}

\paragraph{Task-Based Knowledge Miner}
Based on the query from the knowledge-based learner  (link 3.1), the task-based knowledge miner collects newly observed and labeled knowledge of states from the knowledge manager (link 3.1.1). Then, similar to Algorithm~\ref{algo: task manager}, the task-based knowledge miner initiates a GMM classification model on the gathered data and combines it with the last classification model of the system (from the last observed state). 
The task-based knowledge miner then provides the operator with a visual representation of this classification model (link 3.1.2.1).  
The operator can then give feedback on the proposed classification model (link 3.1.2.2).
Figure\,\ref{fig: feecdback VR1 example} illustrates the different steps of the interaction of the operator with the task-based knowledge miner via a GUI. 


Figure~\ref{fig: visualization example} shows a visualization of the classification model using the verified quality attributes in the specified adaptation cycles (indicated by blue points). 
The black elliptic curve shows a previously detected class. The red elliptic curves show distributions of newly detected classes based on recently observed quality attributes of adaptation options. 
In this example, there is a visual distinction between two new classes (groups of quality attributes mapping to adaptation options) based on energy consumption. However, the GMM has not represented this distinction well. As the stakeholders may desire less energy consumption, the operator uses a box selector to separate the two groups by enclosing one of the groups (the other group is outside of it) (Figure~\ref{fig: choosing box slection} and Figure~\ref{fig: box selecting}). By clicking the ``Apply Feedback'' button the operator will provide the feedback to the task-based knowledge miner (link 3.1.2.2 in Figure\,\ref{fig: solution to tackle the problem}). The task-based knowledge miner then applies the feedback (fitting a new GMM on the newly collected data based on the feedback) and shows the new classification to the stakeholder (Figure~\ref{fig: apply feedback}). 
Finally, the task-based knowledge miner updates the labels of the collected data using the feedback from the operator (link 3.1.3) and returns the updated knowledge to the knowledge-based learner responding to the query (link 3.2).

\begin{figure}[htbp]
	\centering
	\begin{subfigure}[b]{0.49\textwidth}
		\includegraphics[width=\textwidth]{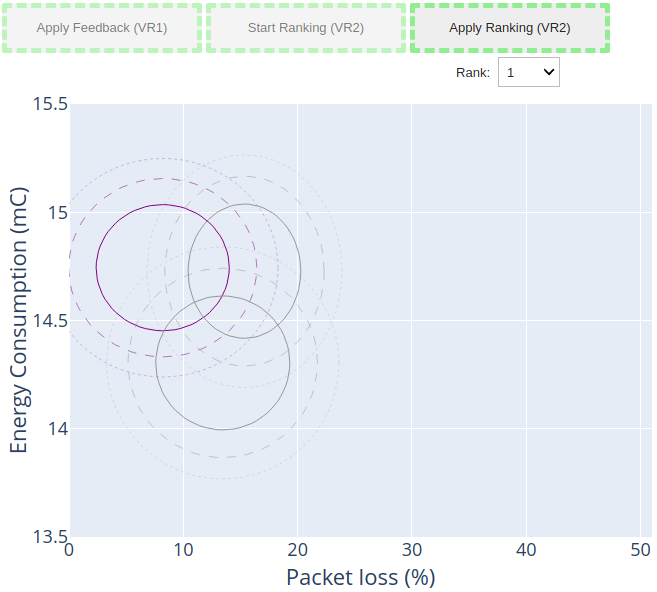}
		\caption{The operator can rank the (purple) marked class.\newline}
		\label{fig: ranking step 1}
	\end{subfigure}
	\begin{subfigure}[b]{0.49\textwidth}
		\includegraphics[width=\textwidth]{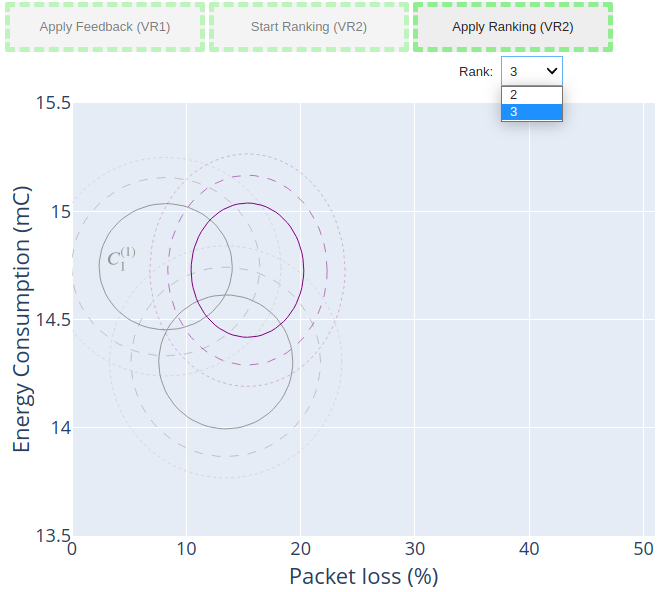}
		\caption{The operator can rank the next marked class.\newline.}
		\label{fig: ranking step 2}
	\end{subfigure}
 \bigskip
	\begin{subfigure}[b]{0.49\textwidth}
		\includegraphics[width=\textwidth]{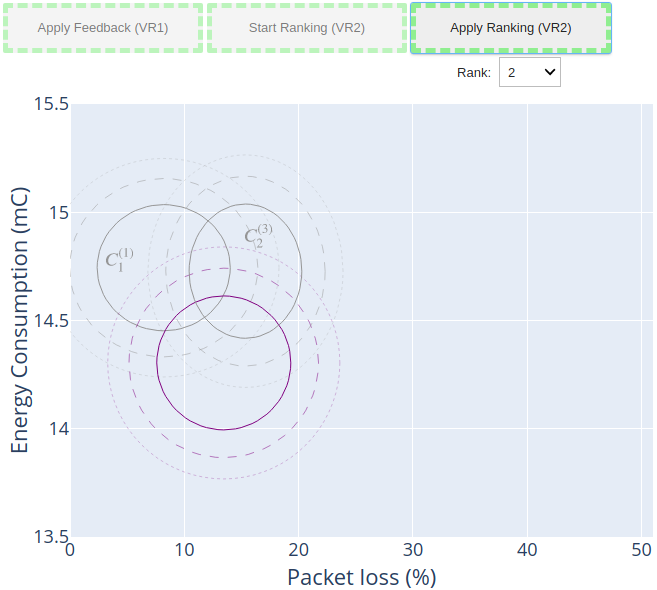}
		\caption{The operator can rank the last marked class.\newline}
		\label{fig: ranking step 3}
	\end{subfigure}
        \begin{subfigure}[b]{0.49\textwidth}
		\includegraphics[width=\textwidth]{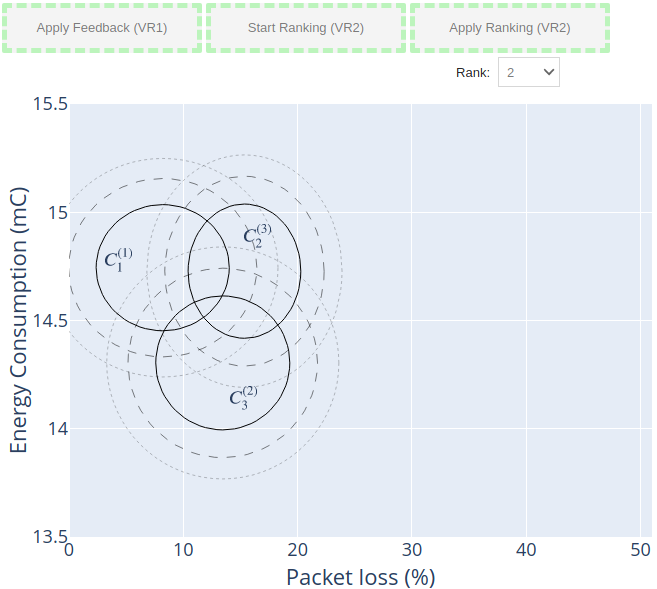}
		\caption{Total ordering of classes that the task-based knowledge can use to update the preference model.}
		\label{fig: final ranking}
        \end{subfigure}\vspace{-10pt}
	\caption{Illustration of the interaction of the operator to update the preference model.
	\label{fig: feecdback VR2 example}}
\end{figure}

We now come back on the interaction of the operator with the managing system (links 3.5.1 and 3.5.2). 
When new classes are detected, the operator should update the ranking of the classes in the preference model (including previously and newly detected classes). This is illustrated in Figure\,\ref{fig: feecdback VR2 example}. The operator starts the ranking process by clicking the ``Start Ranking'' button in the GUI (Figure~\ref{fig: ranking step 1}). The managing system then shows the preference model to the operator (link 3.5.1). During the ranking process (Figure\ref{fig: ranking step 1} to Figure~\ref{fig: ranking step 3}), one class is highlighted (with purple color) in each step that needs to be ranked. To that end, the operator selects the desirable rank for the class from the menu (e.g., the operator selects class 3 for the highlighted class in Figure~\ref{fig: ranking step 2}) and assigns the rank by clicking the ``Apply Ranking'' button. After ordering all classes, the final total ranking of the classification model is shown to the stakeholder (Figure~\ref{fig: final ranking}) and this ranking is applied to the preference model of the managed system (link 3.1.2.3 in Figure\,\ref{fig: solution to tackle the problem}). 

\paragraph{Tasks of the learners}

Recall that the adaptation goals are defined by the stakeholders as a ranking of regions in the
plane of the quality properties of the system. 
In our solution, we  assume that the borders of previously identified classes remain static over time. The knowledge-based learner then fits the optimal mixture of Gaussian distributions on the newly detected data and incrementally incorporates them into the existing mixture of Gaussian models in the managing system (i.e., a structural change). Hence, the learner in the managing system solely performs queries to classify data points and does not perform any parametric change on the learning model over time.  
Additionally, as stated in the assumptions of the approach, the knowledge-based learner uses the same interface definition for the GMM as the learner of the managing system.

\section{Evaluation}\label{sec:evaluation}

We evaluate now lifelong self-adaptation to deal with a drift of adaptation spaces. To that end, we use the DeltaIoT simulator with a setup of 16 motes as explained in Section~\ref{sec:delta-iot}. We applied multiple scenarios with novel class appearance over 350 adaptation cycles that represent three days of wall clock time. The evaluation was done on a computer with an i7-3770 @ 3.40GHz processor and 16GB RAM. A replication package is available on the project website~\cite{project-website}. The remainder of this section starts with a description of the evaluation goals. Then we explain the evaluation scenarios we use. Next, we present the evaluation results. Finally, we discuss threats to validity. 

\subsection{Evaluation Goals}
\label{sec: evalutation goals}
To validate the approach of lifelong self-adaptation for drift of adaptation spaces in answer to the research question, we study to following evaluation questions: 

\begin{itemize}
	
	\item [EQ1] How effective is lifelong self-adaptation in dealing with drift of adaptation spaces?
	
	\item [EQ2] How robust is the approach to changing the appearance order of classes and different preference orders of the stakeholders?
	
	\item [EQ3] How effective is the feedback of the operator in dealing with drift of adaptation spaces? 
	
\end{itemize}

To answer EQ1, we measured the values of the quality attributes over all adaptation cycles and based on these results we computed the utility and the \rsm\ (as defined in Section\,\ref{sec: problem}) before and after novel class(es) appear. We performed the measurements and computations for four approaches: (i) the managing system equipped with an ideal classifier that uses a correct ranking of classes, we refer to this approach as the baseline; (ii) the managing system equipped with a pre-defined classifier with ML2ASR~\cite{quin2022reducing}; this is a representative state-of-the-art approach of learning-based self-adaptation that applies adaptation space reduction, (iii) a pre-defined classifier with lifelong self-adaptation (no operator feedback)\footnote{Note that, this approach is equivalent to the pre-defined classifier explained in Section~\ref{sec:problemContext}. Because in case of no operator feedback, newly detected classes by the lifelong learning loop will not be ranked, and the goal model (the preference model of the stakeholders) in the managing system will not evolve. }, and (iv) an evolving classifier with lifelong self-adaptation with operator feedback. All classifiers rely on mixed Gaussian distributions on quality attributes (GMM). 
For approach (ii), we implemented a \re{self-adaptation} approach that leverages Machine Learning to Adaptation Space Reduction (ML2ASR)~\cite{quin2022reducing}.
This approach uses a regressor to predict quality attributes in the analysis stage and then uses the prediction result to rank and select a subset of the adaptation options for verification, i.e., the approach verifies those adaptation options that are classified as a higher-ranking class as in Algorithm~\ref{algo: DM best option selection}. 
Note that to the best of our knowledge, there are no competing approaches for dealing with the novel class appearance in the context of self-adaptive systems that interact with an operator on behalf of stakeholders to order classes.   
Hence, we compared the proposed approach with a perfect baseline and a related state-of-the-art approach that uses learning to predict quality attributes of the adaptation options at runtime.

To answer the evaluation questions EQ2 and EQ3 we applied the evaluations of EQ1 for multiple scenarios that combined different preference orders of stakeholders and different orders of emerging classes with and without feedback from an operator. For questions EQ2 and EQ3 we focused on the period of new appearance of classes within each scenario. 

\re{Before explaining the evaluation scenarios, we acknowledge that we evaluated the instantiated architecture of lifelong self-adaptation for dealing with a novel class appearance in only one domain. Finding and evaluating solutions beyond one domain goes beyond the scope of the research presented in this paper and offers opportunities for future research. We anticipate that the instance of the architecture presented in this paper may lay a foundation for such future studies.} 

\subsection{Evaluation Scenarios}

Table~\ref{tab: evalutation scenarios} shows the different scenarios that we use for the evaluation comprising three factors.

\small
\begin{table}[h!]
	\caption{Evaluation scenarios\label{tab: evalutation scenarios}}
	\begin{tabular}{|c|c|c|}
		\hline
		\begin{tabular}[c]{@{}c@{}}Preference order of\\ stakeholders\end{tabular}                                                                                                                & \begin{tabular}[c]{@{}c@{}}Classes appearance order
  \end{tabular}                                                                                                                                                                                                     & Operator feedback \\ \hline
		\begin{tabular}[c]{@{}c@{}}$\langle$``less packet loss'', \\ ``less energy consumption''$\rangle$,\\ $\langle$``less energy consumption'',\\ ``less packet loss'' $\rangle$\end{tabular} & \begin{tabular}[c]{@{}c@{}}
            $\langle$(B), R, G$\rangle$,\\ 
            $\langle$(B), G, R$\rangle$,\\ 
            $\langle$(R), B, G$\rangle$,\\ 
            $\langle$(R), G, B$\rangle$,\\ 
		$\langle$(B, R), G$\rangle$,\\
		$\langle$(B, G), R$\rangle$,\\
  \end{tabular} & $\langle$active$\rangle$, $\langle$inactive$\rangle$     \\ \hline
	\end{tabular}
\end{table}
\normalsize

The first factor (left column of Table\,\ref{tab: evalutation scenarios}) shows the two options for the preference order of the stakeholders.\footnote{The simulator used these options to automate the ranking of the classes as the stakeholders' feedback in the experiments.} This factor allows us to evaluate the robustness against different preference orders of the stakeholder (EQ2) of the lifelong self-adaptation approach.  

\begin{figure}
	\centering
	\includegraphics[scale=0.4]{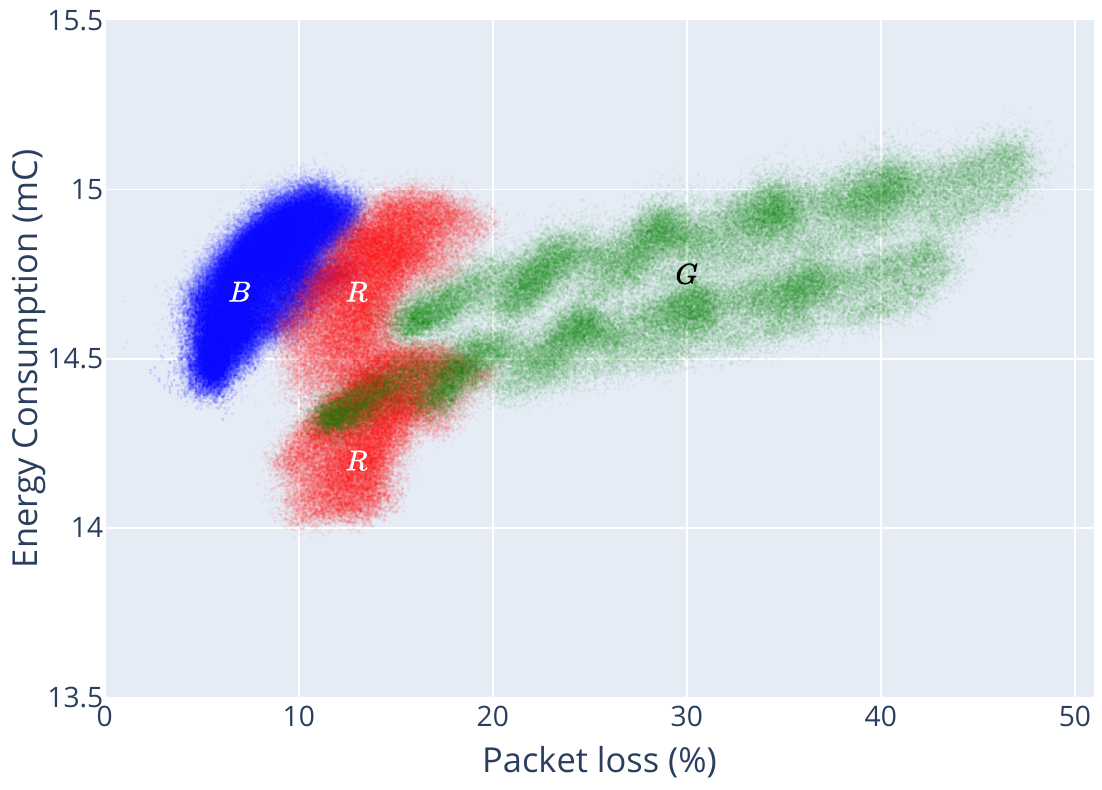}
	\caption{Groups of classes in the quality attribute panel where adaptation options appear.}
	\label{fig:zones}
\end{figure}

\begin{figure}[htbp]
	\centering
	\begin{subfigure}[b]{0.325\textwidth}
		\includegraphics[width=\textwidth]{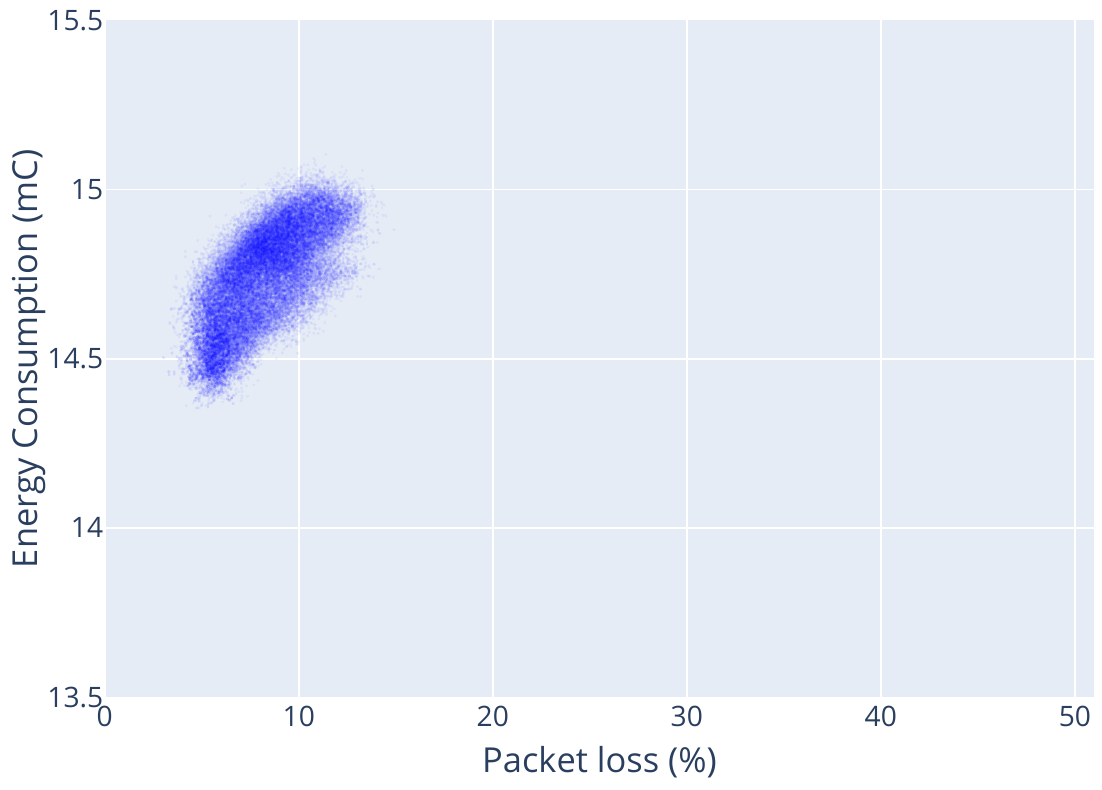}
		\caption{Adaptation cycle 1-40}
		\label{fig:adapt-cycle-1-40}
	\end{subfigure}
	\begin{subfigure}[b]{0.325\textwidth}
		\includegraphics[width=\textwidth]{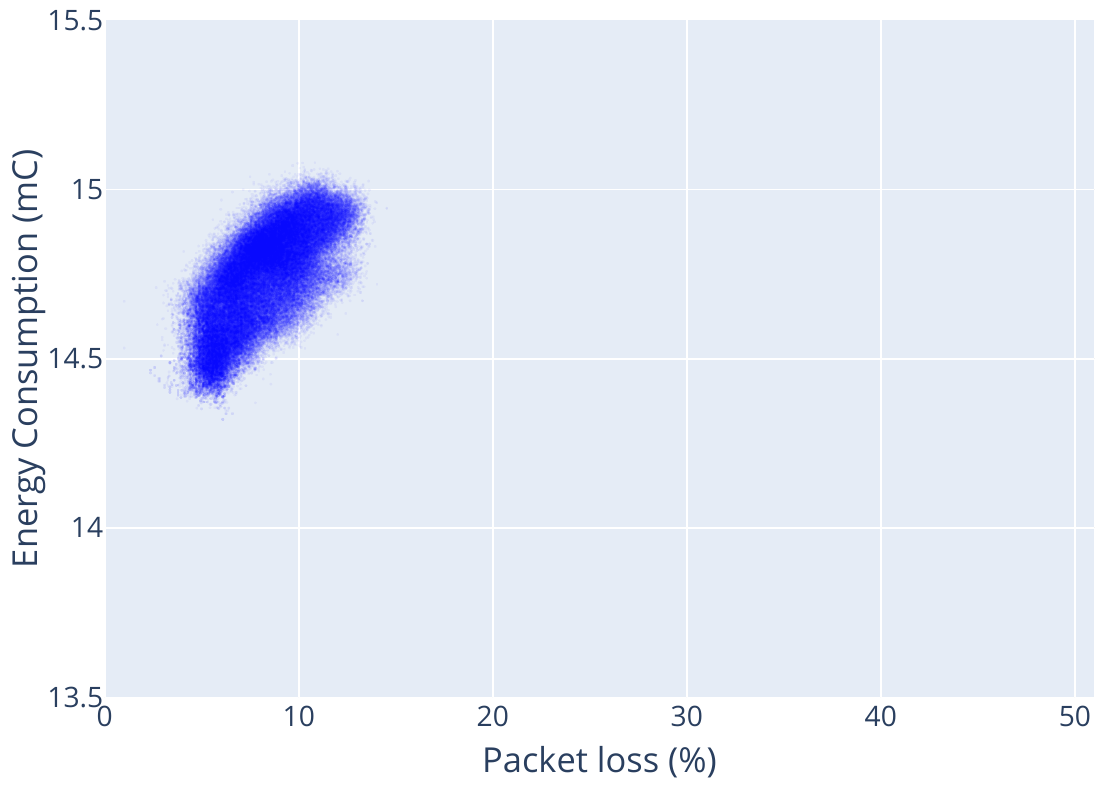}
		\caption{Adaptation cycle 41-140}
		\label{fig:adapt-cycle-41-150}
	\end{subfigure}
	\begin{subfigure}[b]{0.325\textwidth}
		\includegraphics[width=\textwidth]{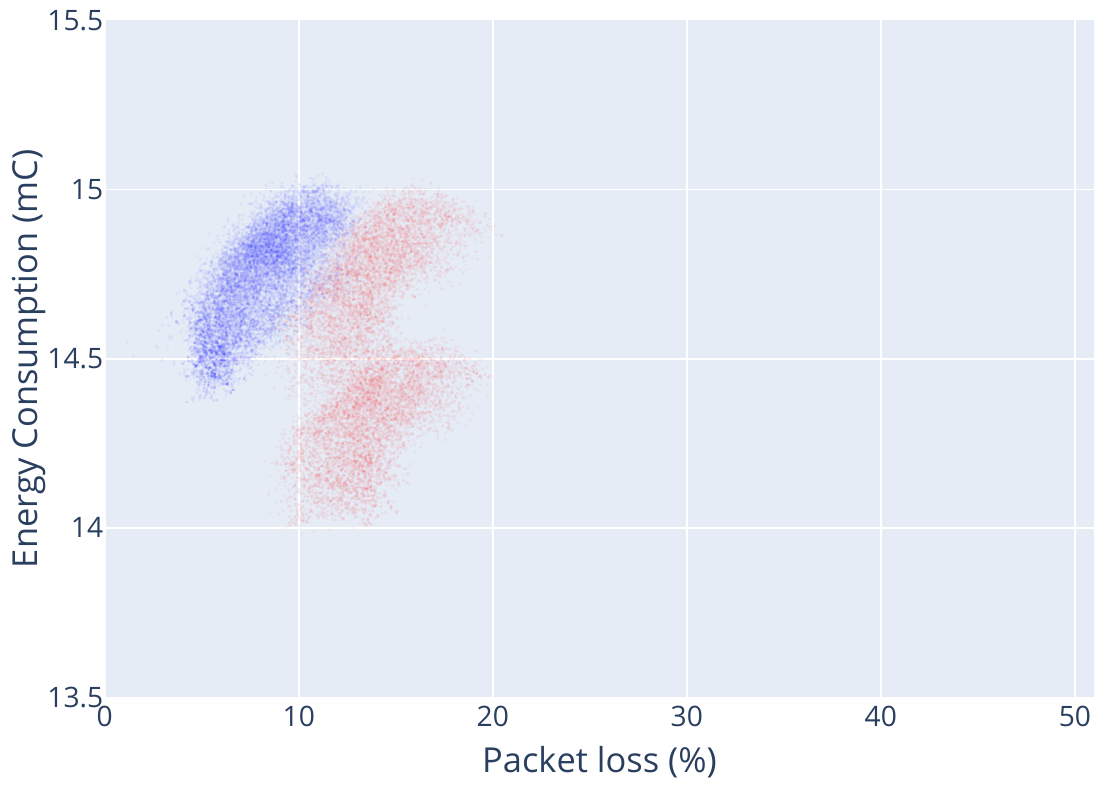}
		\caption{Adaptation cycle 141-160}
		\label{fig:adapt-cycle-146-180}
	\end{subfigure}
	
	\bigskip
	\begin{subfigure}[b]{0.325\textwidth}
		\includegraphics[width=\textwidth]{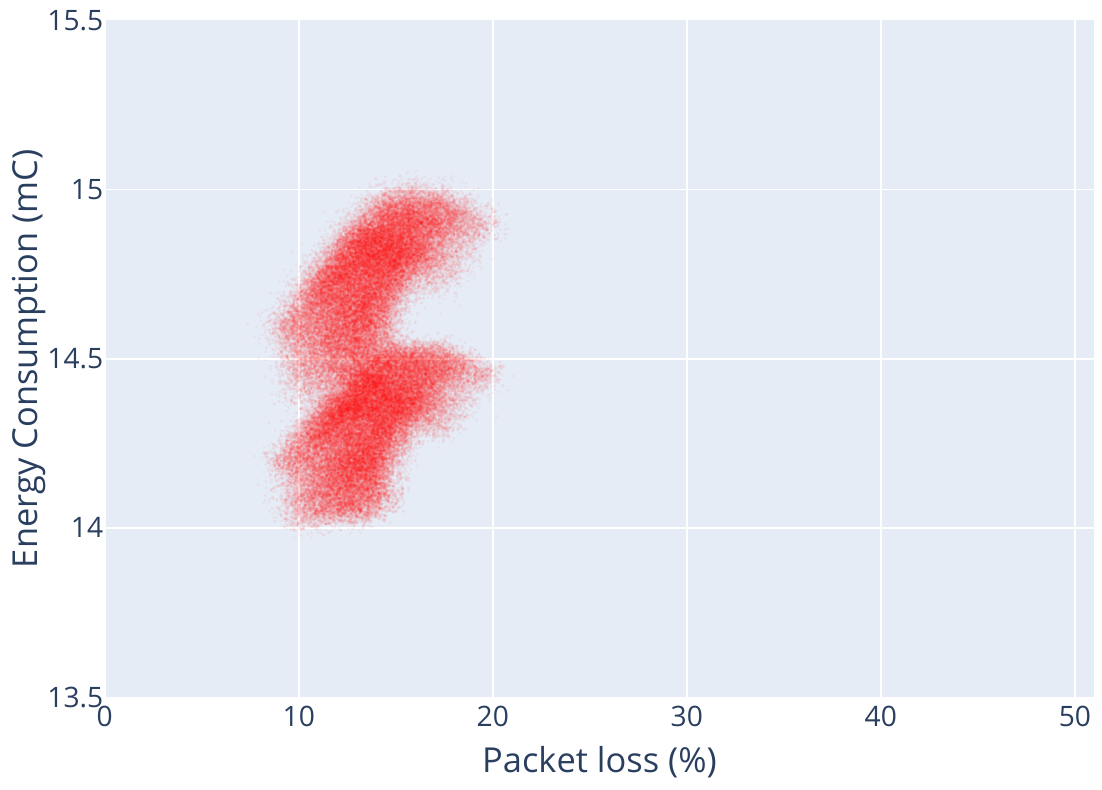}
		\caption{Adaptation cycle 161-240}
		\label{fig:adapt-cycle-181-245}
	\end{subfigure}
	\begin{subfigure}[b]{0.325\textwidth}
		\includegraphics[width=\textwidth]{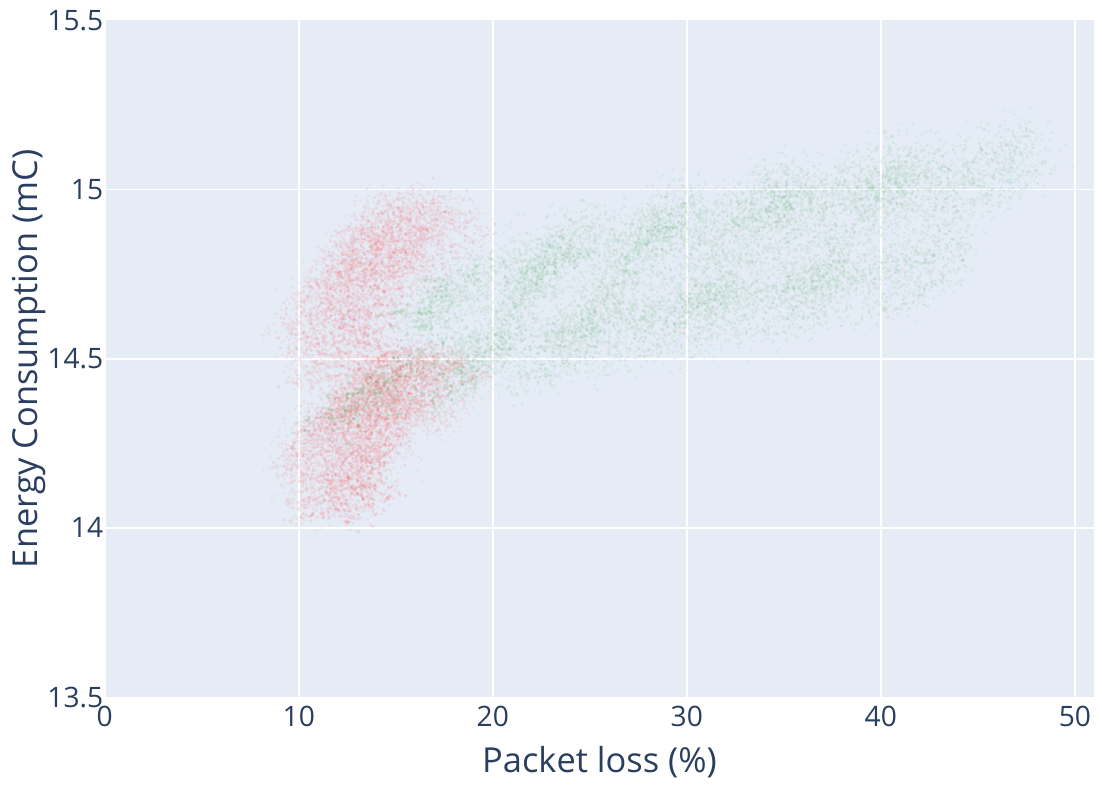}
		\caption{Adaptation cycle 241-260}
		\label{fig:adapt-cycle-246-275}
	\end{subfigure}
	\begin{subfigure}[b]{0.325\textwidth}
		\includegraphics[width=\textwidth]{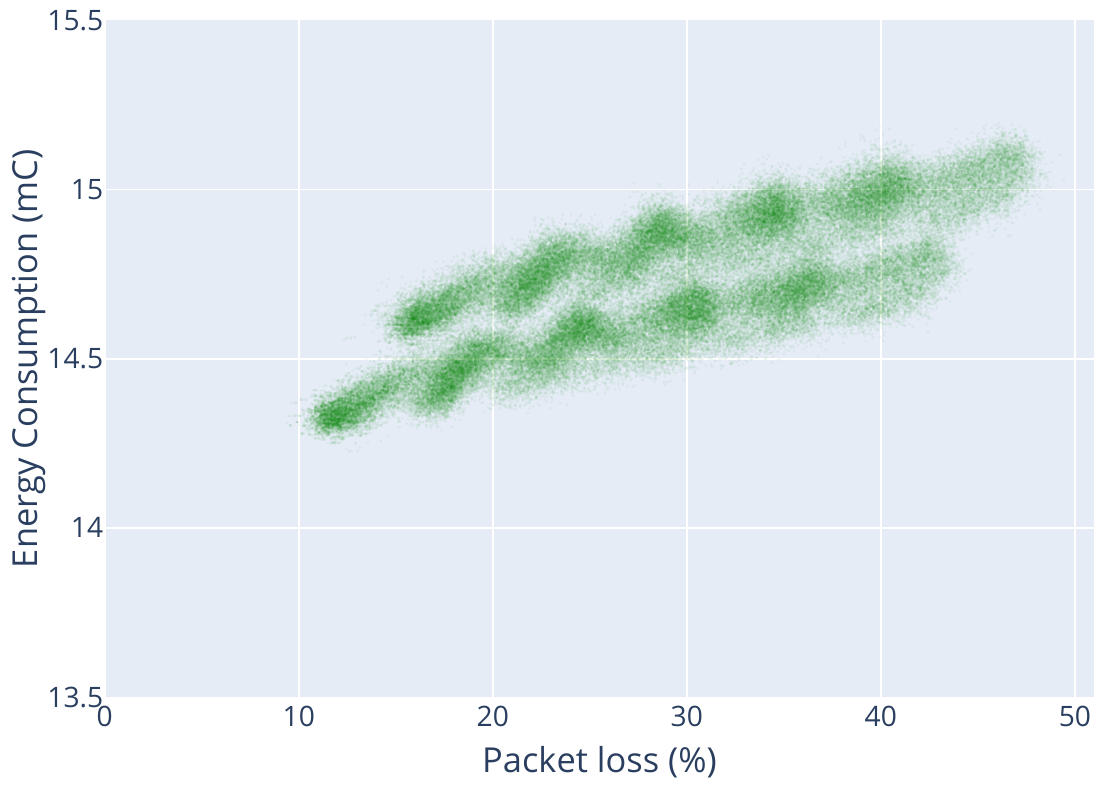}
		\caption{Adaptation cycle 261-350}
		\label{fig:adapt-cycle-276-350}
	\end{subfigure}
	\caption{Adaptation spaces for the base scenario illustrating novel class appearance over time.}\vspace{-10pt}
	\label{fig:3:emerging-classes}
\end{figure}

The second factor (middle column of Table\,\ref{tab: evalutation scenarios}) shows six options for the appearance order of classes over time. Each character, $B$, $R$, and $G$, refers to a group of classes in the quality attribute plane, as illustrated in Figure~\ref{fig:zones}. The order expresses the appearances of groups over time. For instance, $\langle$(B), R, G$\rangle$ means that first, the group of classes marked with $B$ appears, then the group $R$ appears, and finally group $G$ appears. Figure~\ref{fig:3:emerging-classes} illustrates this scenario. The groups of classes marked between round brackets are known before deployment (order-invariant) and can be used for training the learners. 
The classifiers were trained for a number of cycles (between 40 and 180 cycles) depending on the appearance of new classes in each scenario. 
Since the order of appearance of classes may affect the effectiveness of lifelong self-adaptation, we analyzed the different scenarios to validate the robustness to changing the appearance order of classes (EQ2) of the proposed approach. 

Finally, the third factor (right column of Table\,\ref{tab: evalutation scenarios}) expresses whether the operator is actively involved in the self-adaptation process (active)  or not (inactive). This involvement refers to the activities related to links  3.1.2.1, 3.1.2.2, 3.5.1, and 3.5.2 in Figure~\ref{fig: solution to tackle the problem}. 
This factor allows us to evaluate the effectiveness of feedback from the operator in dealing with a drift of adaptation spaces (EQ3). 

By combining the three factors (preference order of stakeholders, appearance order of classes, and operator feedback), we obtain a total of 24 scenarios (i.e., $2 \times 6 \times 2$) for evaluation. We refer to the scenario with preference order of stakeholders $\langle$``less packet loss'',  ``less energy consumption''$\rangle$, class appearance  $\langle$(B), R, G$\rangle$ and both settings of operator feedback as the base scenario.

\subsection{Evaluation Results}

We start with answering EQ1 using the base scenario. Then we answer EQ2 and EQ3 using all 24 scenarios derived from Table\,\ref{tab: evalutation scenarios}.  For EQ1 we collect the data of both the period before and after new classes appear. For EQ2 and EQ3 we focus only on data from the period when new classes appear. All basic results of the evaluation (median, mean, sd) are available in Appendix~\ref{sec: appendix statistical test results}. 
We provide p-values of statistical tests for the relevant evaluation results, that is, results that are important to the evaluation questions and cannot obviously be answered without a test.\footnote{For the statistical analyses we used the Scipy library\,\cite{2020SciPy-NMeth}.}

\subsubsection{Effectiveness of Lifelong Self-Adaptation in Dealing with Drift of Adaptation Spaces.}

\begin{figure}[htbp]
  \centering
  	\begin{subfigure}[b]{0.45\textwidth}
    \includegraphics[width=\textwidth]{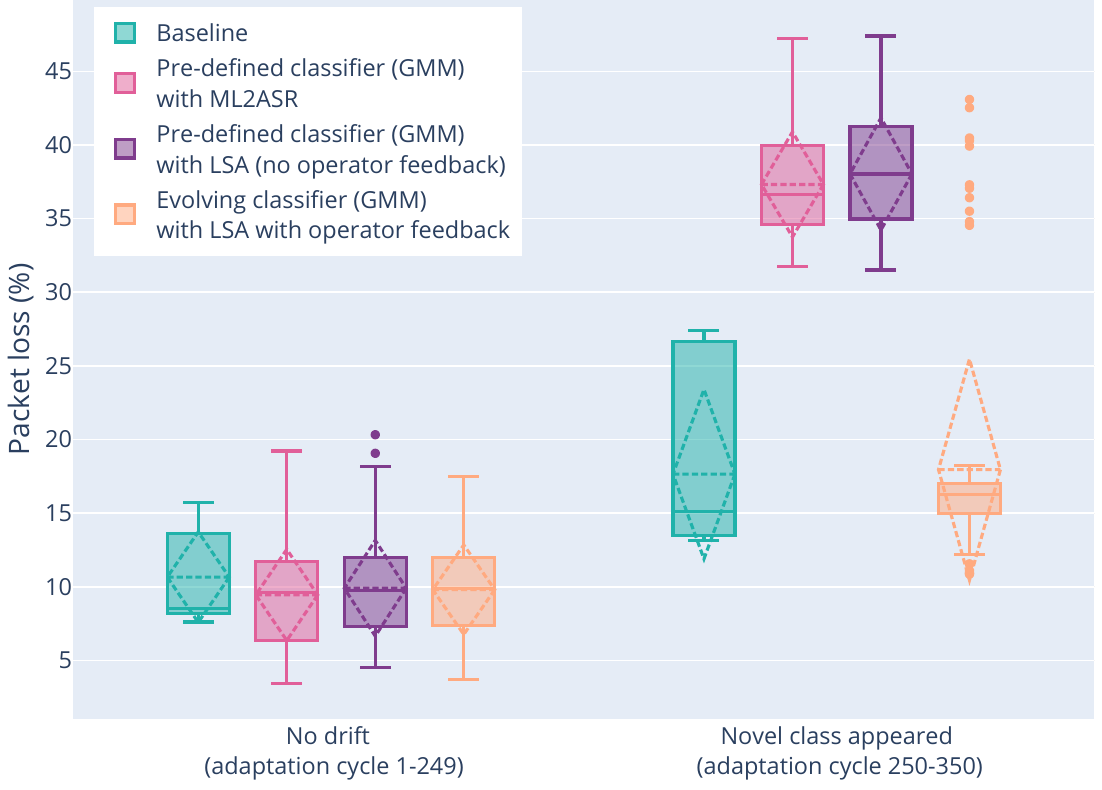}
    \caption{Packet loss}
    \label{fig:packet loss distribution over adaptation cycles}
      \end{subfigure}
   \begin{subfigure}[b]{0.45\textwidth}
  	\includegraphics[width=\textwidth]{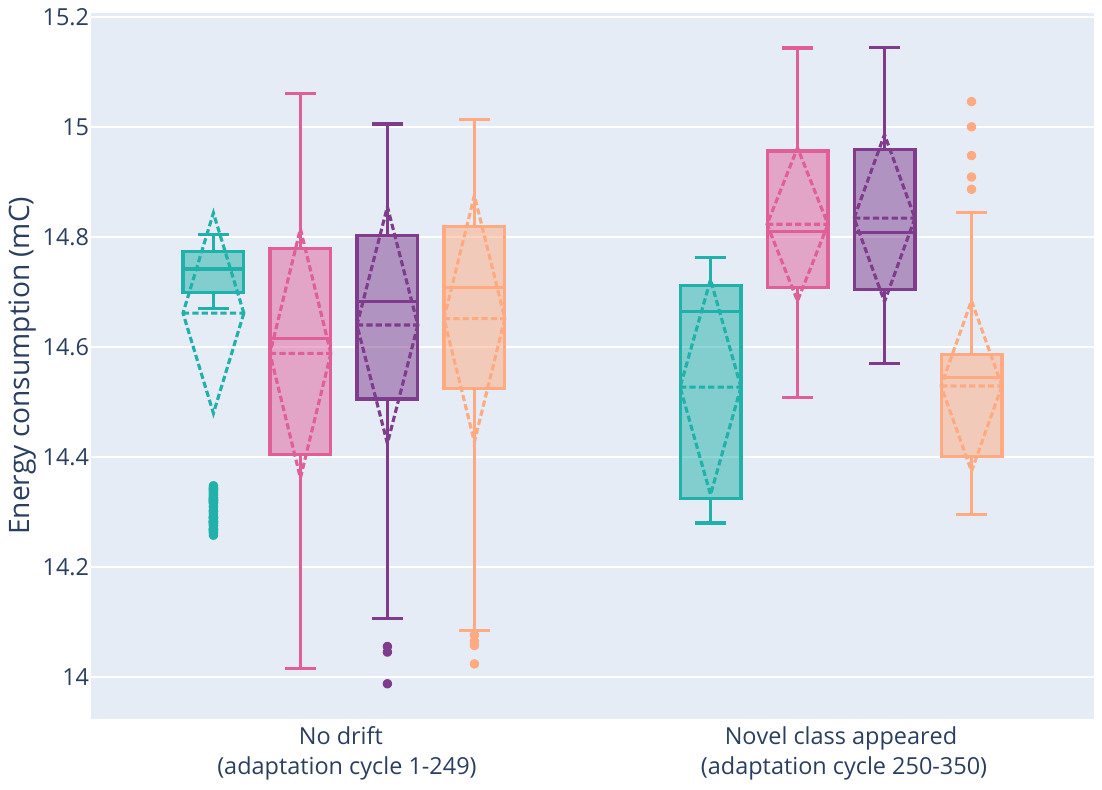}
  	\caption{Energy consumption}
  	\label{fig:energy consumption distribution over adaptation cycles}
  \end{subfigure}
  \caption{Quality properties of lifelong self-adaptation compared to the other approaches for the base scenario.}
  \label{fig:compare performance on quality attributes}
\end{figure}

To answer EQ1, we use the base scenario. 
Figure~\ref{fig:compare performance on quality attributes} shows the distributions of quality attributes of the selected adaptation options. The results are split into two periods: the period before the drift occurs (adaptation cycles 1-249) and the period with shift of adaptation spaces when novel classes appear, i.e., the emerging green dots in Figure~\ref{fig:3:emerging-classes}(e) and the green dots in  Figure~\ref{fig:3:emerging-classes}(f) (cycles 250-350).  

Figure\,\ref{fig:compare performance on quality attributes} shows that the four approaches perform similarly for both quality attributes without drift of adaptation spaces (mean values between 9.90 and 10.67 for packet loss and 14.59 and 14.66 for energy consumption). 
Yet, once the drift appears, the pre-defined classifier with ML2ASR and the pre-defined classifier with LSA (no operator feedback) degrade substantially (mean 37.32\,\% for packet loss and 14.82~mC for energy consumption for the pre-defined classifier with ML2ASR, 38.04\,\% and 14.83~mC for the pre-defined classifier with LSA (no operator feedback), compared to 17.65\,\% and 14.63~mC for the baseline). \re{On the other hand, the evolving classifier with LSA with operator feedback maintains its performance 
(17.96\,\% for packet loss compared to 17.65\,\% for the baseline, a difference of 0.31\,\% of packet loss on a total of 17.96\,\% is negligible in practice; and    
14.53~mC for energy consumption compared to 14.52~mC for the baseline.} Note that we do not use statistical tests to compare individual quality properties as the approaches optimize for utility. 

\begin{figure}[!ht]
	\centering
	\includegraphics[scale=0.4]{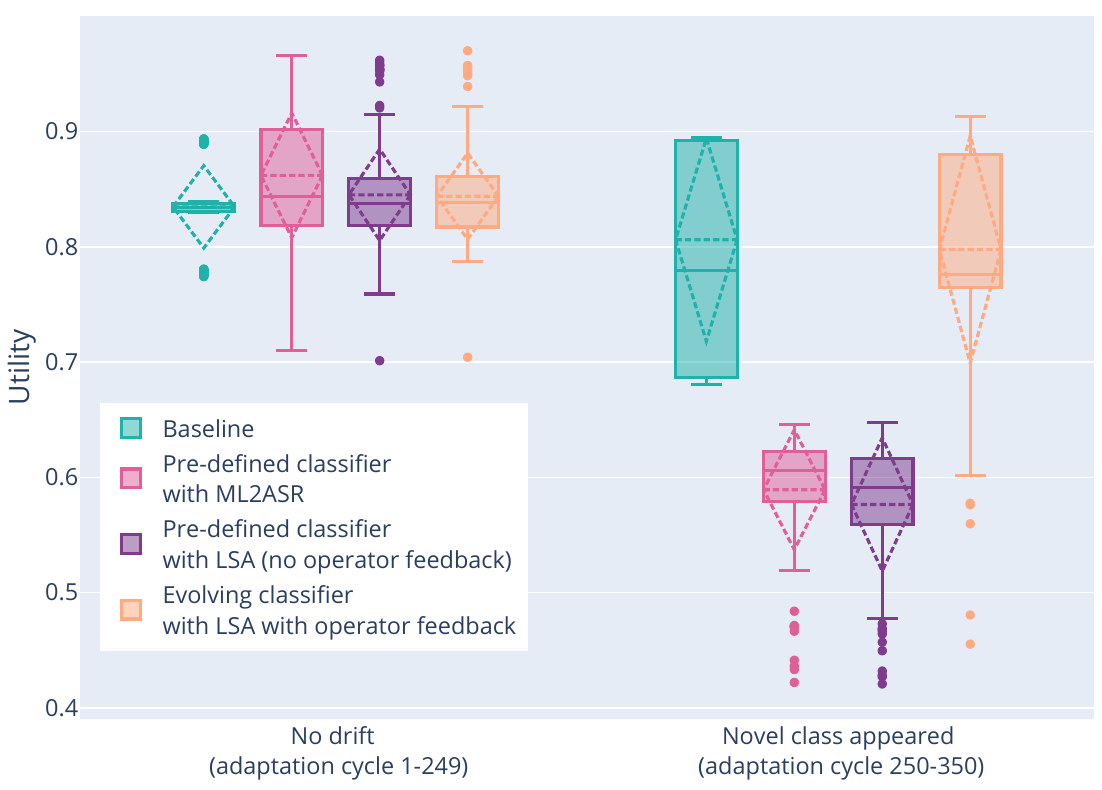}
	\caption{Impact of drift of adaptation spaces on the utility of the system. }
	\label{fig:resuls-utilities}
\end{figure}

Figure\,\ref{fig:resuls-utilities} shows the results for the impact on the utilities. \re{Under drift, for a pre-defined classifier with ML2ASR the mean utility is 0.59, for the pre-defined classifier with LSA (no operator feedback) it is 0.58, compared to 0.81 for the baseline approach. On the other hand, the mean utility for the evolving classifier with LSA with operator feedback is 0.80. With a significance level of 0.05, the results of a Mann-Witney U test do not support the hypothesis that the utility of the baseline is higher than the utility of the evolving classifier with LSA with operator feedback; $p=0.114$}. 

\begin{figure}[!ht]
	\centering
	\includegraphics[scale=0.4]{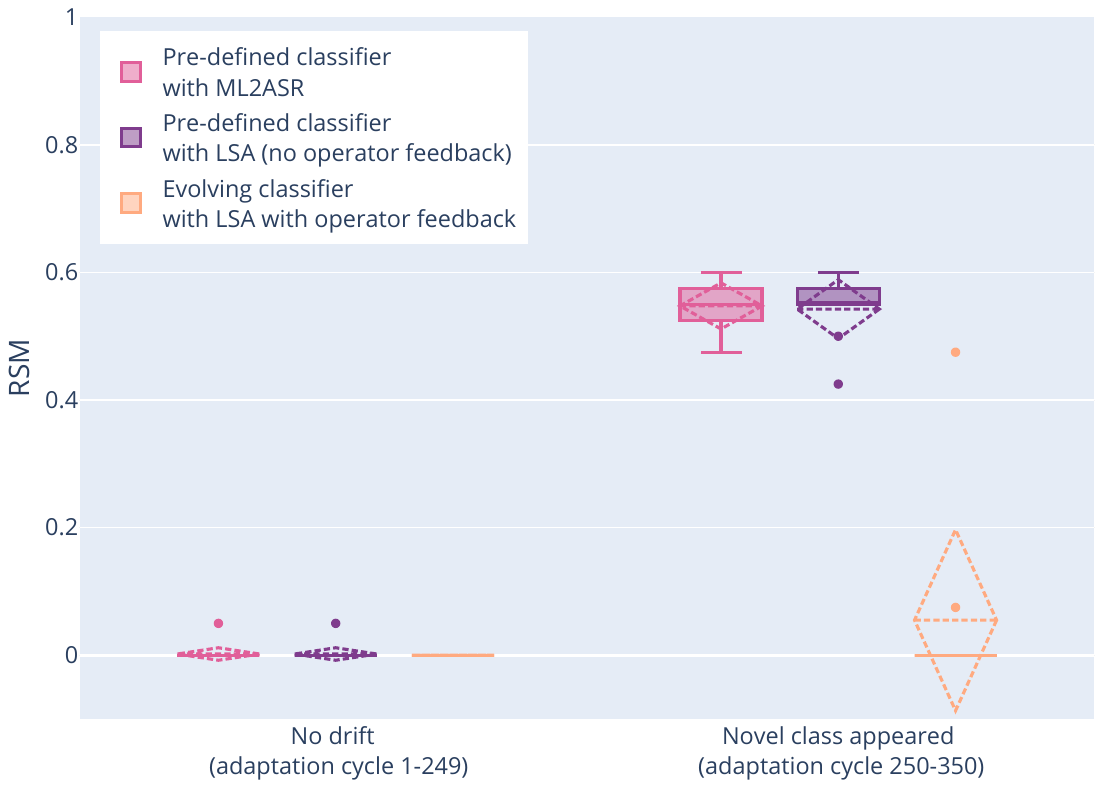}
	\caption{Impact of drift of adaptation spaces on the \rsm\ for different approaches. 
 }
	\label{fig: rsm improvement}
\end{figure}

In terms of \rsm\ we observe similar results, see Figure~\ref{fig: rsm improvement}. Without drift the three approaches perform close to the baseline (\rsm\ of 0.002 for the pre-defined classifier with ML2ASR, 0.002 for the pre-defined classifier with LSA (no operator feedback), and 0.000 for the pre-defined classifier with LSA with operator feedback.
On the other hand, with a drift of adaptation spaces, the \rsm\ for the pre-defined classifier with ML2ASR, and the pre-defined classifier with LSA (no operator feedback) increase dramatically (to 0.55 and 0.54 respectively).
\re{On the contrary, the predictions of the pre-defined classifier with LSA with operator remain accurate with an \rsm\ of 0.06.}

\paragraph{Conclusion.}
In answer to EQ1, we can conclude that an evolving classifier with LSA with operator feedback is particularly effective in dealing with a drift of adaptation spaces, with a performance close to an ideal classifier with a perfect classification. 

\subsubsection{Robustness of Lifelong Self-Adaptation in Dealing with Drift of Adaptation Spaces.}

To answer EQ2, we evaluated the 24 scenarios based on Table\,\ref{tab: evalutation scenarios}.  
We measured the quality attributes, the utilities, and the \rsm\ values for all scenarios. The detailed results are available in Appendix~\ref{sec: appendix validation scenarios} (including all validation scenarios, see Figures~\ref{fig:all compare performance on packet loss}, 
\ref{fig:all compare performance on energy consumption}, 
\ref{fig:all compare performance on utility}, 
\ref{fig:all compare performance on rsm}, 
\ref{fig:all compare performance on packet loss 2}, 
\ref{fig:all compare performance on energy consumption 2}, 
\ref{fig:all compare performance on utility 2}, 
\ref{fig:all compare performance on rsm 2}).

Here we summarize the results for the utilities and \rsm\ over all scenarios (during the period that new classes emerge). We start by looking at the robustness with respect to the appearance order of classes. Then we look at robustness with respect to the preference order of stakeholders. 

\paragraph{Robustness with respect to the appearance order of classes.}

\begin{figure}[!ht]
	\centering
	\begin{subfigure}[b]{0.49\textwidth}
		\includegraphics[width=\textwidth]{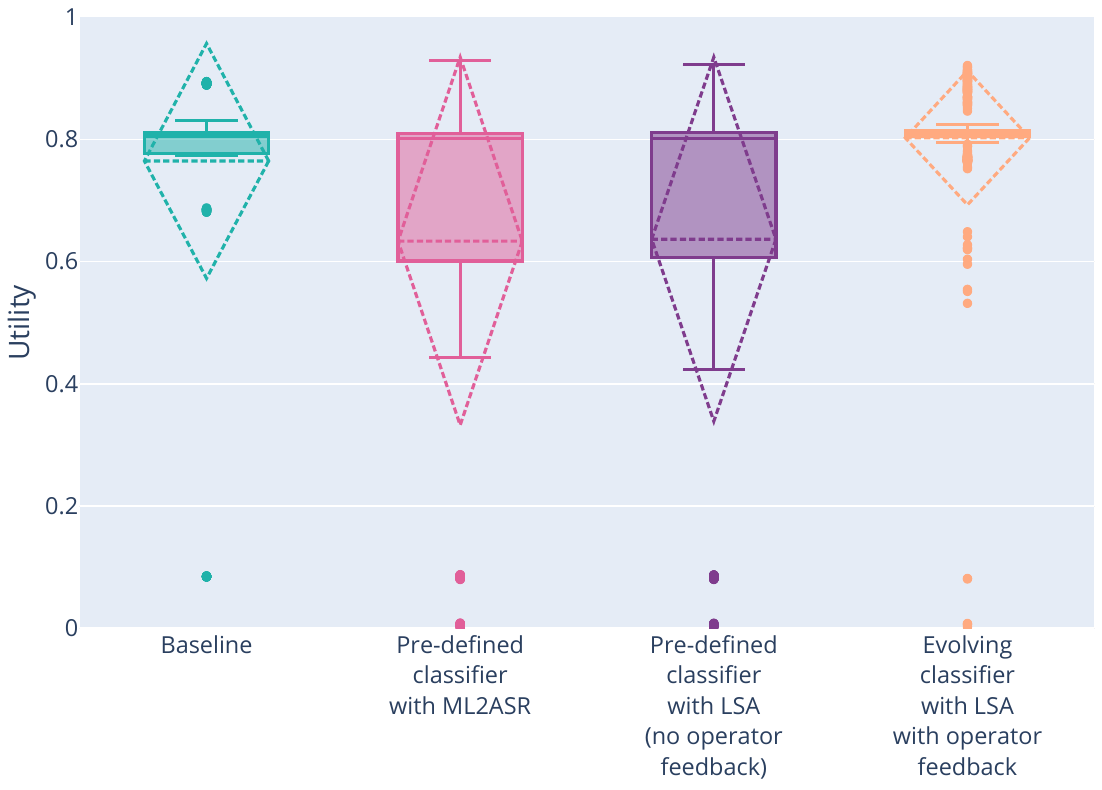}
		\caption{$\langle$(B),R, G$\rangle$}
		\label{fig: util total (b)rg}
	\end{subfigure}
	\begin{subfigure}[b]{0.49\textwidth}
		\includegraphics[width=\textwidth]{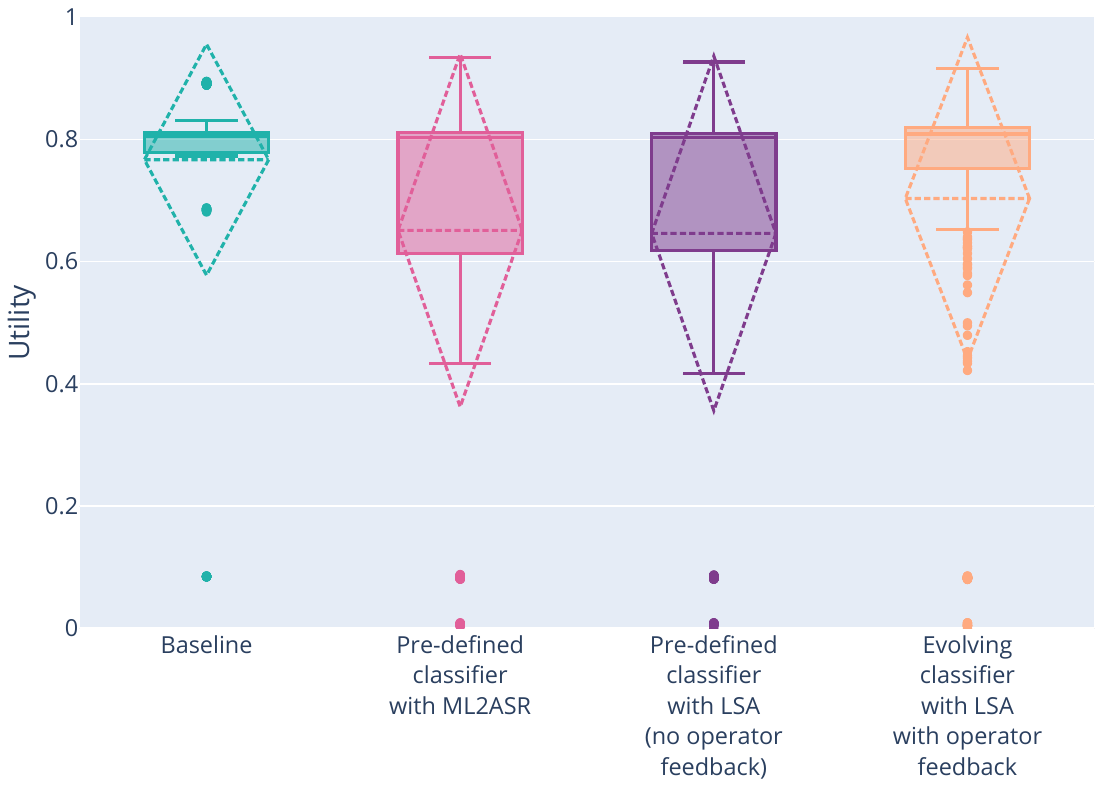}
		\caption{$\langle$(B),G,R$\rangle$}
		\label{fig: util total (b)gr}
	\end{subfigure}
 \bigskip
    \begin{subfigure}[b]{0.49\textwidth}
		\includegraphics[width=\textwidth]{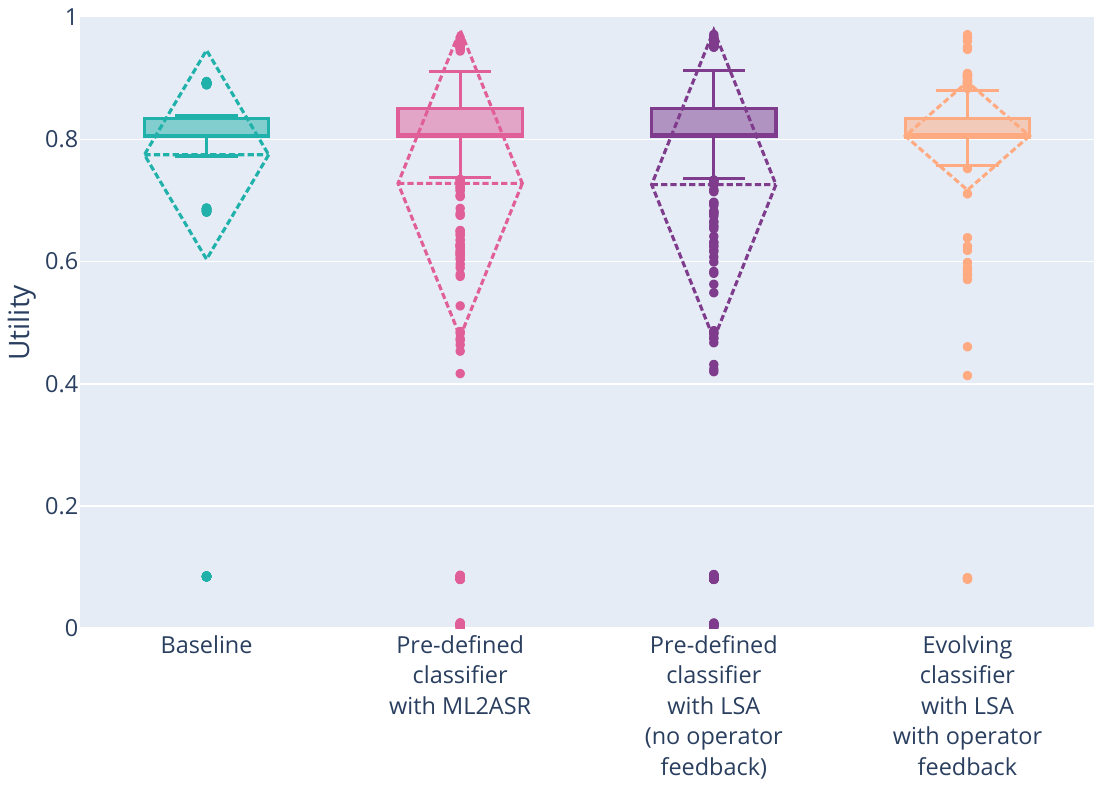}
		\caption{$\langle$(R),B,G$\rangle$}
		\label{fig: util total (r)bg}
	\end{subfigure}
 \begin{subfigure}[b]{0.49\textwidth}
		\includegraphics[width=\textwidth]{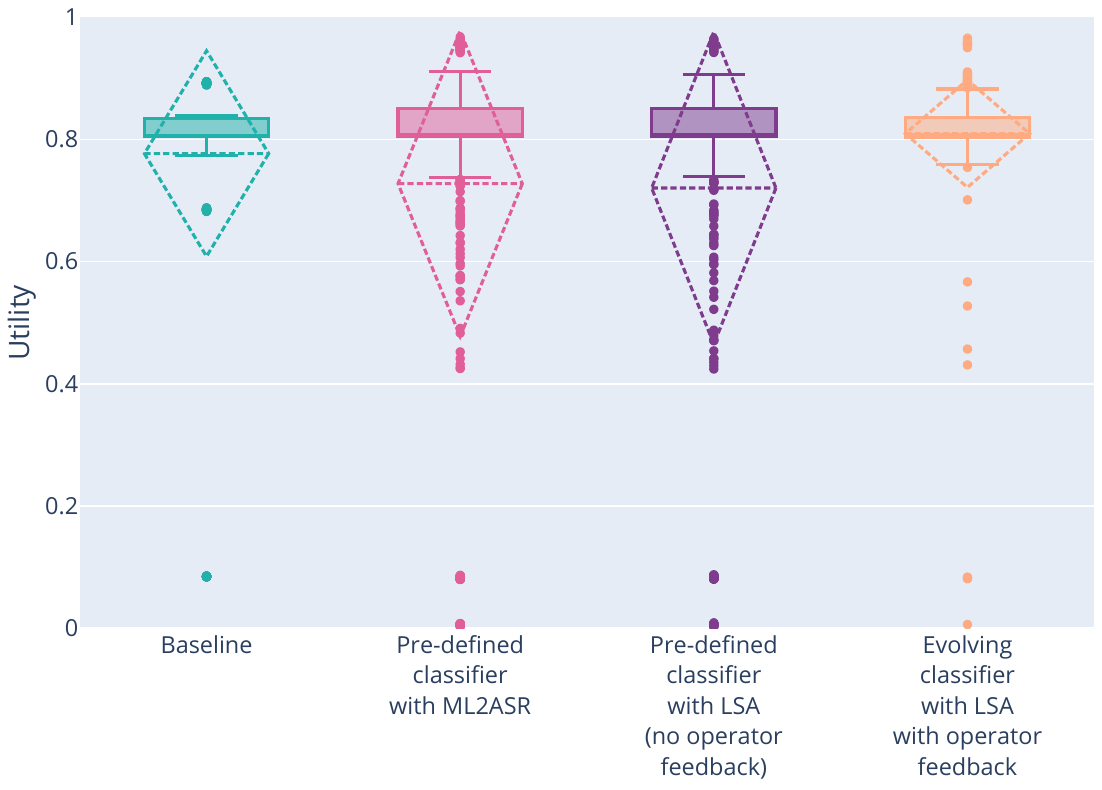}
		\caption{$\langle$(R),G, B$\rangle$}
		\label{fig: util total (r)gb}
	\end{subfigure}
 \bigskip
	\begin{subfigure}[b]{0.49\textwidth}
		\includegraphics[width=\textwidth]{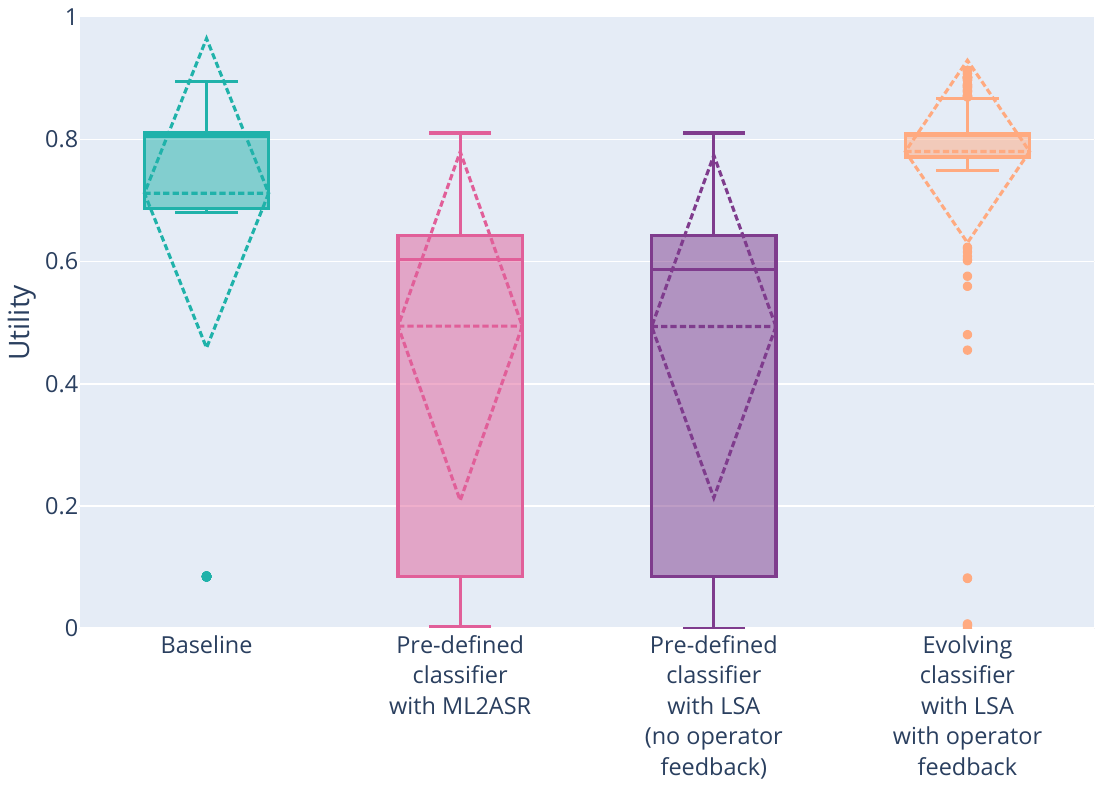}
		\caption{$\langle$(B, R),G$\rangle$}
		\label{fig: util total (b,r),g}
	\end{subfigure}
    \begin{subfigure}[b]{0.49\textwidth}
		\includegraphics[width=\textwidth]{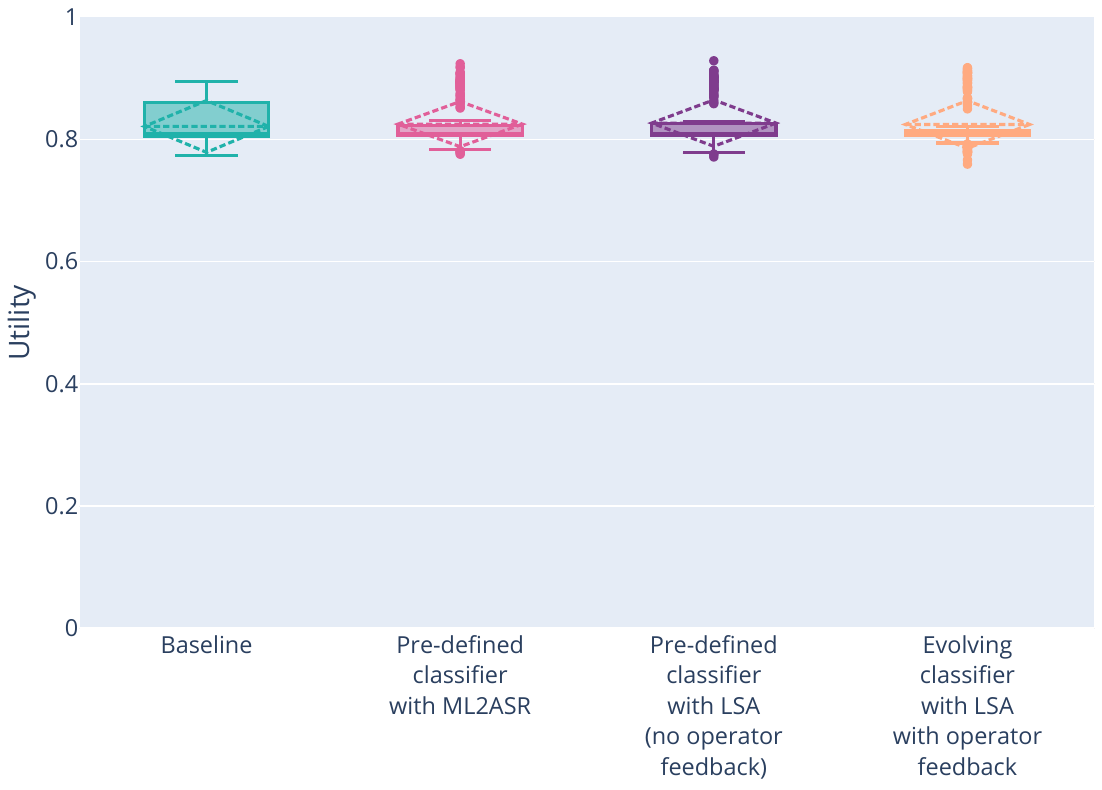}
	\caption{$\langle$(B,G),R$\rangle$}
		\label{fig: util total (b,g),r}
	\end{subfigure}
	\caption{Utility for all preference orders of stakeholders split for classes appearance orders. \label{fig: utility total for class appearances}}
	
\end{figure}

Figure~\ref{fig: utility total for class appearances} shows the results of the utilities for the six scenarios of the appearance order of classes. The results indicate that the evolving classifier with LSA and operator feedback outperforms the pre-defined classifier with ML2ASR and the pre-defined classifier with LSA (no operator feedback) for scenarios (a), (b), and (e). For scenario (a) the mean utility is 0.80 for the evolving classifier with LSA and operator feedback versus 0.63 and 0.64 for the pre-defined classifier with ML2ASR and the pre-defined classifier with LSA (no operator feedback) respectively; for scenario (b) the mean utility was 0.70 versus 0.65 for both other approaches, and for scenario (e) the results are 0.78 versus 0.50 and 0.49 respectively. With a significance level of 0.05, the results of Mann-Withney U tests support the hypotheses that the utility of the evolving classifier with LSA and operator feedback is higher than the utility of the pre-defined classifier with ML2ASR and the utility of the pre-defined classifier with LSA in these scenarios (p = 0.000 for the three scenarios). For the other scenarios, the test results do not support the hypotheses. These scenarios do not seem to present a real challenge as all classifiers were able to achieve high mean utilities. On the other hand, the evolving classifier with LSA and operator feedback performs similarly to the baseline for all scenarios (difference in mean utilities between 0.004 and 0.068). With a significance level of 0.05, the results of Mann-Whitney U tests do not support the hypothesis that the utility of the baseline would be higher than the utility of the evolving classifier with LSA and operator feedback (p-values between 0.638 and 0.997 for the six scenarios).  

\begin{figure}[!ht]
	\centering
	\begin{subfigure}[b]{0.49\textwidth}
		\includegraphics[width=\textwidth]{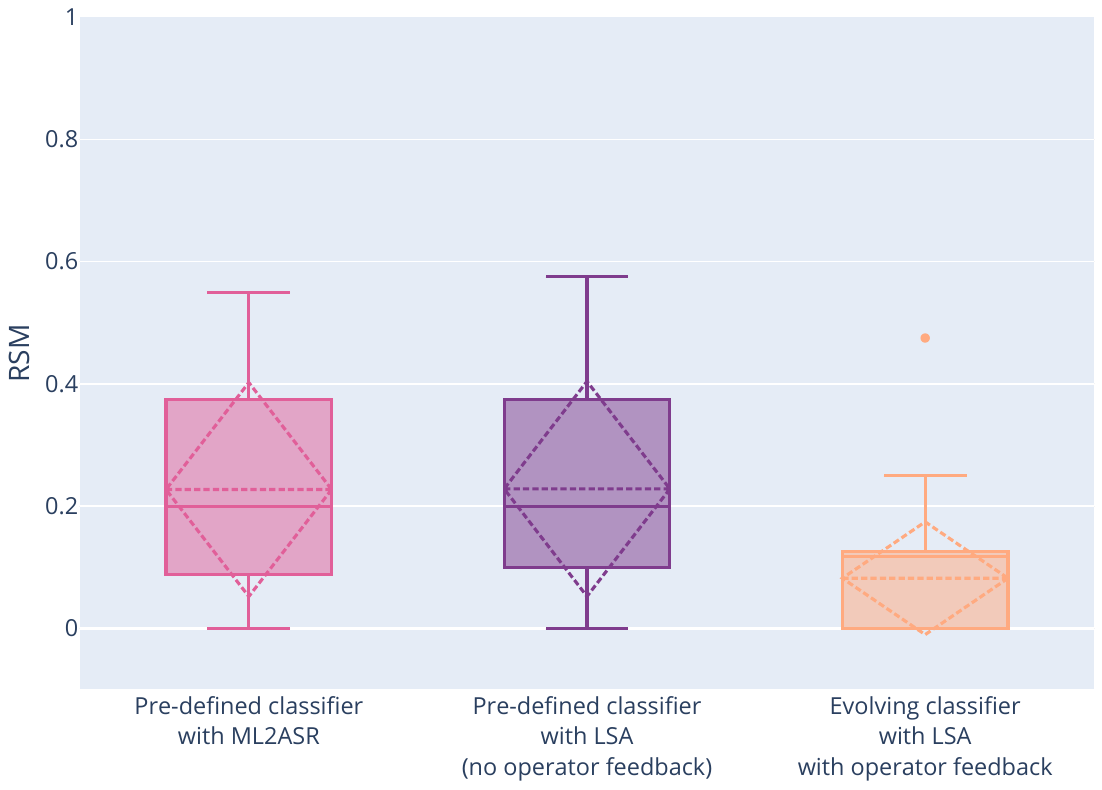}
		\caption{$\langle$(B),R, G$\rangle$}
		\label{fig: rsm total (b)rg}
	\end{subfigure}
	\begin{subfigure}[b]{0.49\textwidth}
		\includegraphics[width=\textwidth]{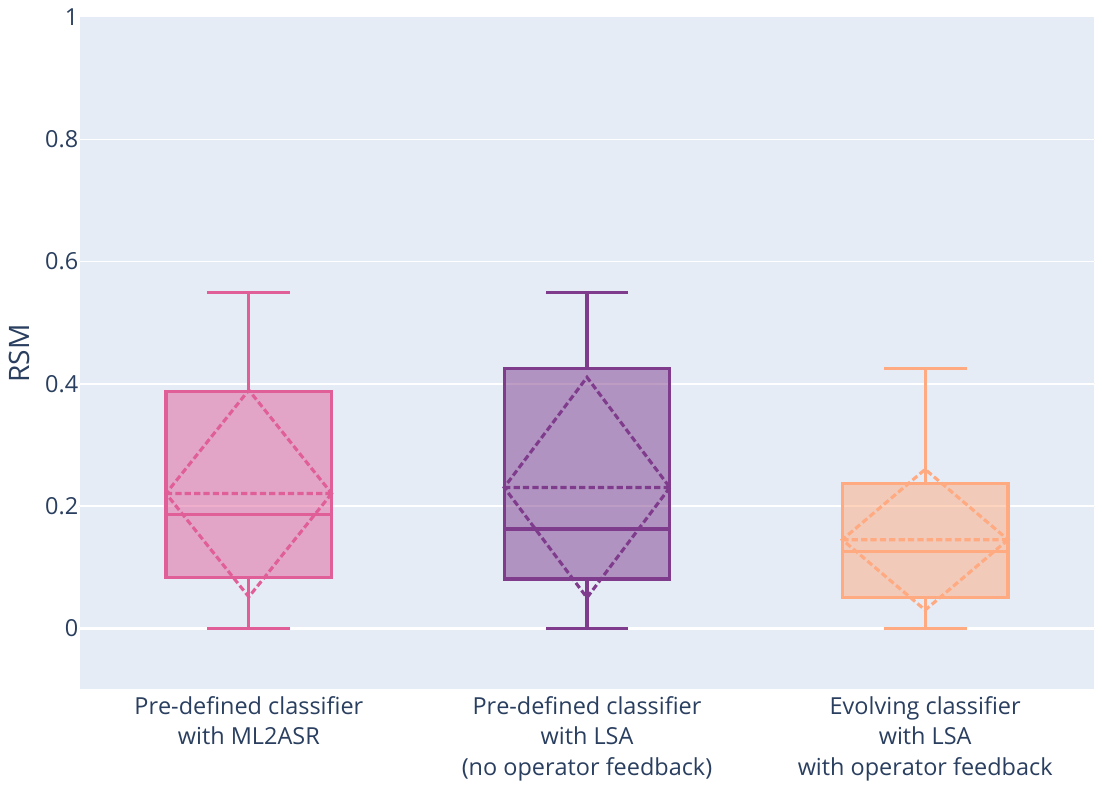}
		\caption{$\langle$(B),G,R$\rangle$}
		\label{fig: rsm total (b)gr}
	\end{subfigure}
 \bigskip
    \begin{subfigure}[b]{0.49\textwidth}
		\includegraphics[width=\textwidth]{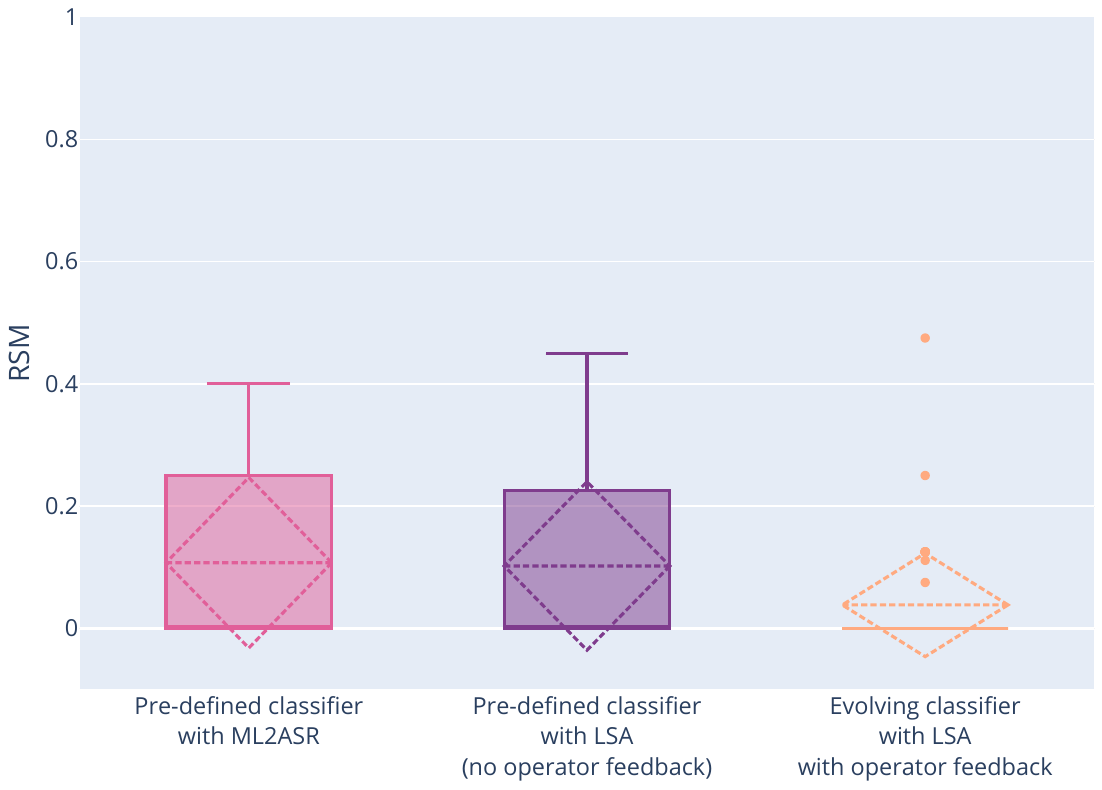}
		\caption{$\langle$(R),B,G$\rangle$}
		\label{fig: rsm total (r)bg}
	\end{subfigure}
 \begin{subfigure}[b]{0.49\textwidth}
		\includegraphics[width=\textwidth]{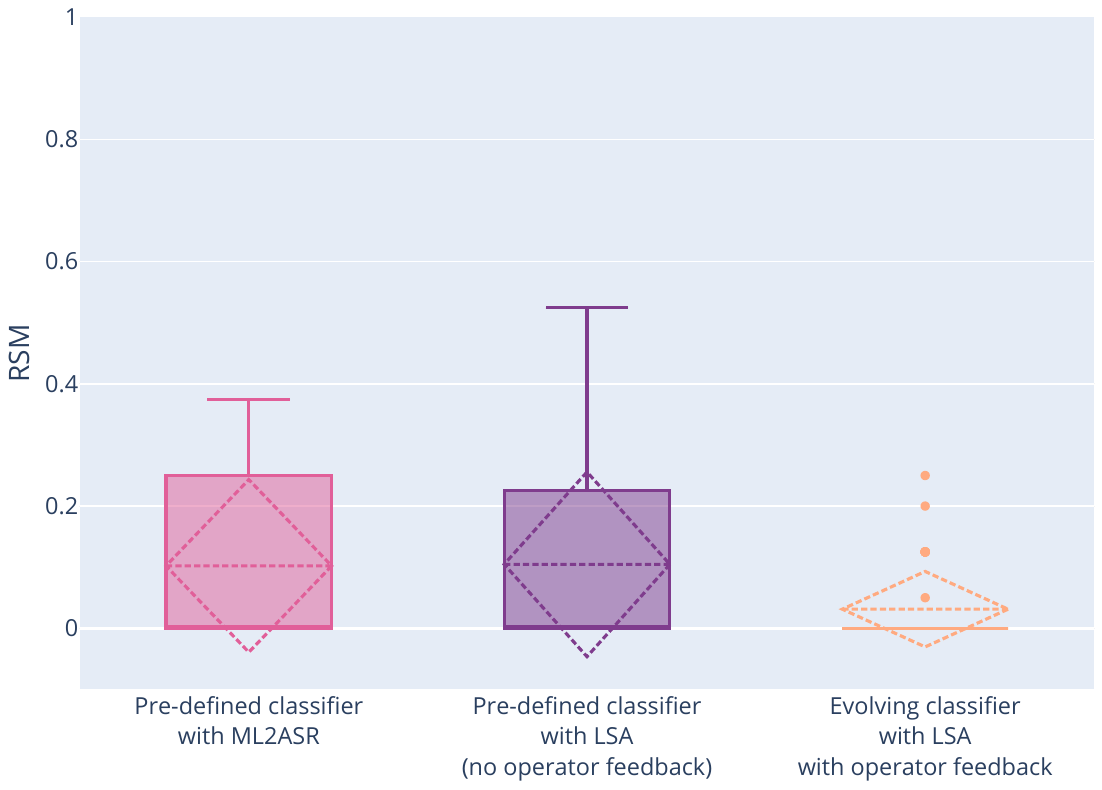}
		\caption{$\langle$(R),G, B$\rangle$}
		\label{fig: rsm total (r)gb}
	\end{subfigure}
 \bigskip
	\begin{subfigure}[b]{0.49\textwidth}
		\includegraphics[width=\textwidth]{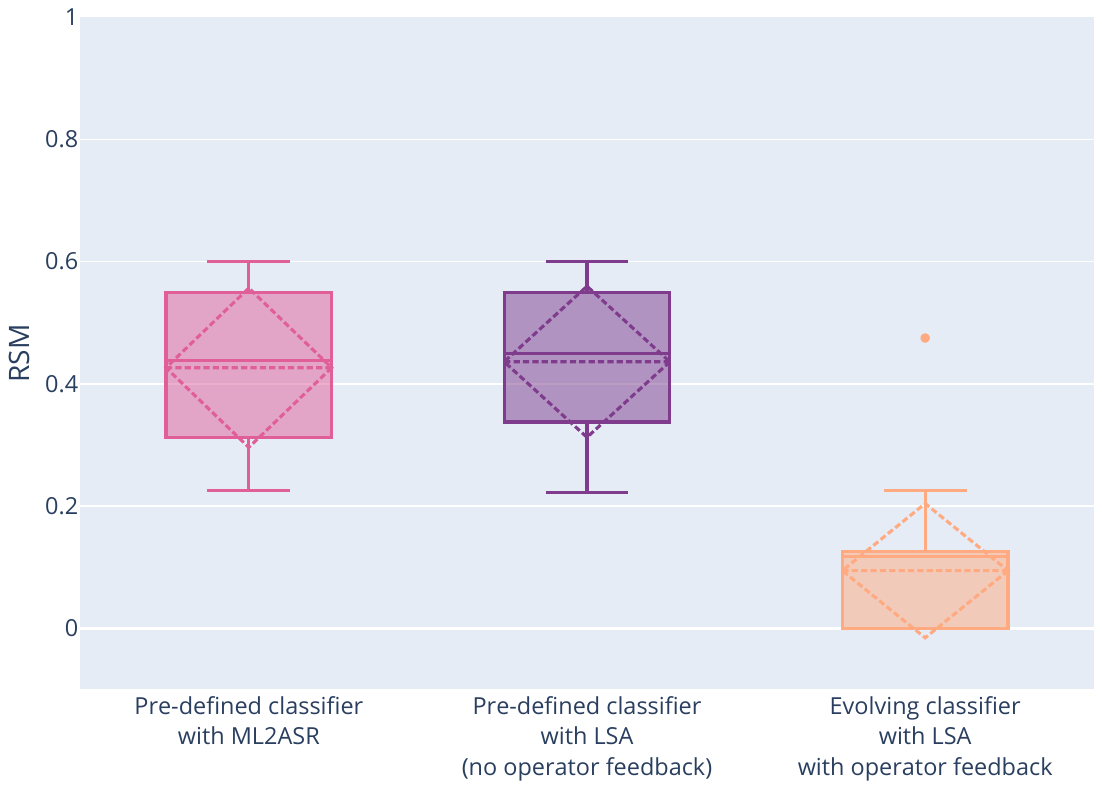}
		\caption{$\langle$(B, R),G$\rangle$}
		\label{fig: rsm total (b,r),g}
	\end{subfigure}
    \begin{subfigure}[b]{0.49\textwidth}
		\includegraphics[width=\textwidth]{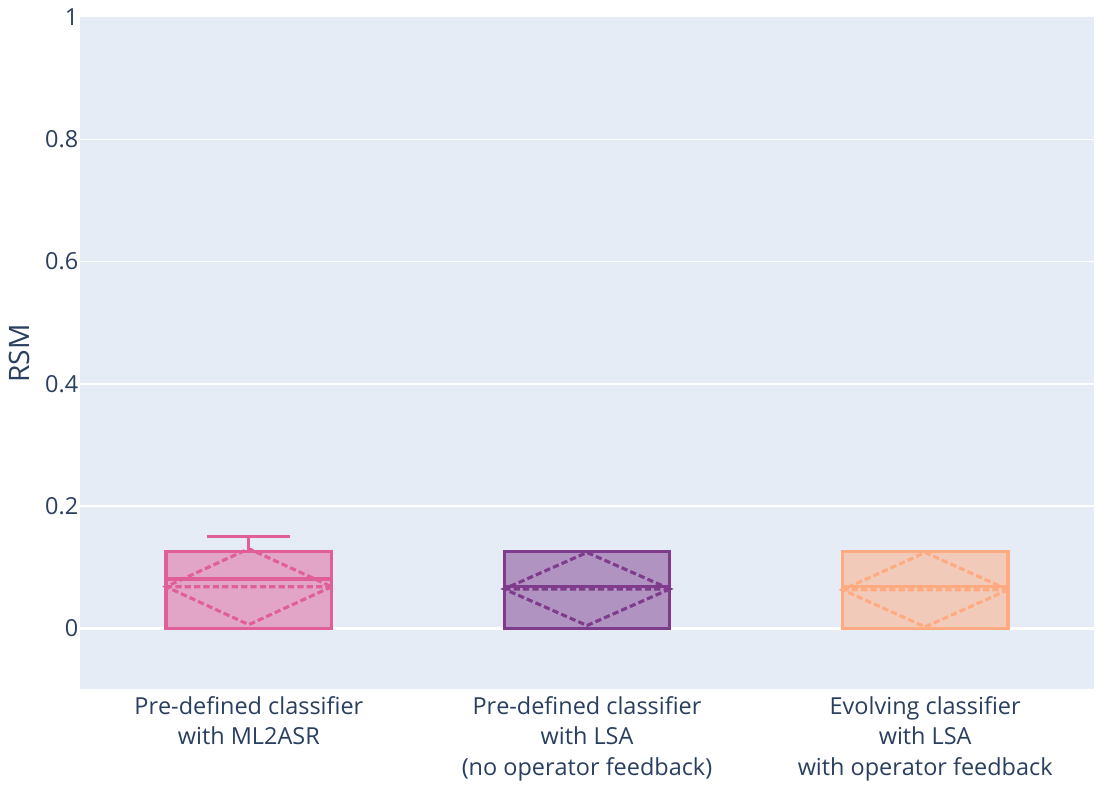}
	\caption{$\langle$(B,G),R$\rangle$}
		\label{fig: rsm total (b,g),r}
	\end{subfigure}
	\caption{\rsm\ values for all preference orders of stakeholders split for classes appearance order. \label{fig: rsm total for class appearances} }
	
\end{figure}

The results for \rsm\ shown in Figure~\ref{fig: rsm total for class appearances} confirm the superiority of the evolving classifier with LSA and operator feedback compared to the other approaches. For scenarios (a) to (e), the difference between the mean \rsm\ of the evolving classifier with LSA and operator feedback and the pre-defined classifier with ML2ASR is between 0.069 and 0.401, while the difference with the pre-defined classifier with LSA (no operator feedback) is between 0.065 and 0.342. With a significance level of 0.05, the results of Mann-Whitney U tests support the hypotheses that the  \rsm\ of the evolving classifier with LSA and operator feedback is lower than the other approaches (p-values between 0.000 and 0.028). For scenario (f), the difference between the mean \rsm\ of the evolving classifier with LSA and operator feedback and the pre-defined classifier with ML2ASR is 0.005 and the difference with the pre-defined classifier with LSA (no operator feedback) is 0.001. With a significance level of 0.05, for this scenario, the results of Mann-Whitney U tests do not support the hypotheses that the  \rsm\ of the evolving classifier with LSA and operator feedback is lower than the two other approaches (p-values 0.341 and 0.447 respectively). The absolute results of the mean values for \rsm\ for the evolving classifier with LSA and operator feedback (between 0.031 and 0.145 for the six scenarios) indicate a very good performance of the approach, comparable with an ideal classifier. 

\paragraph{Robustness with respect to the preference order of stakeholders.}

\begin{figure}[!ht]
	\centering
	\begin{subfigure}[b]{0.49\textwidth}
		\includegraphics[width=\textwidth]{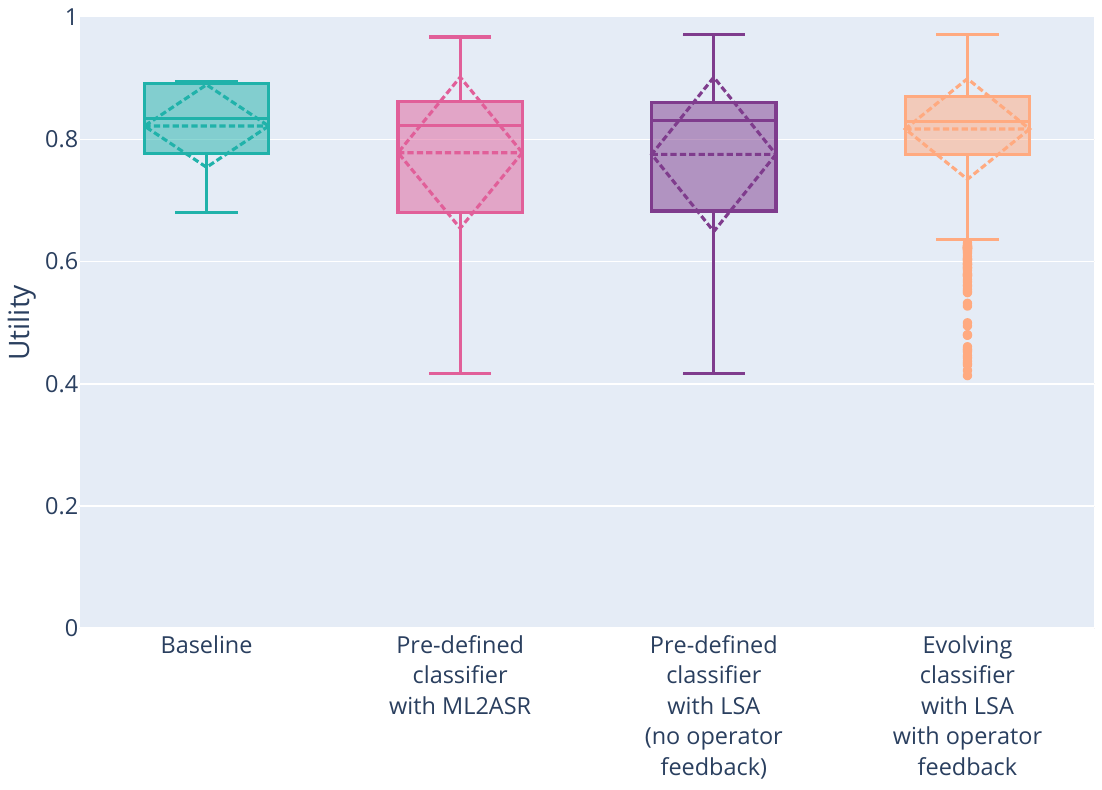}
		\caption{$\langle$``less packet loss'', ``less energy consumption''$\rangle$}
		\label{fig: utility total 1}
	\end{subfigure}
	\begin{subfigure}[b]{0.49\textwidth}
		\includegraphics[width=\textwidth]{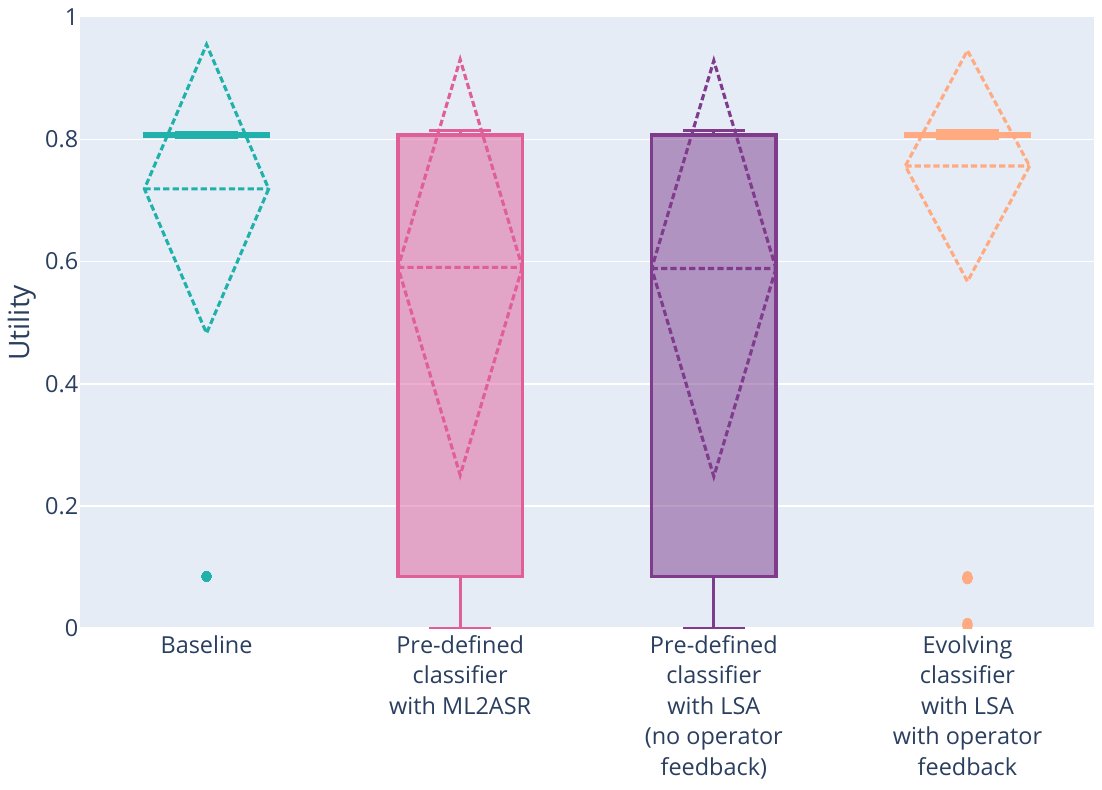}
		\caption{$\langle$``less energy consumption'', ``less packet loss''$\rangle$}
		\label{fig: utility total 2}
	\end{subfigure}
	\caption{Utility for all class appearance order scenarios split for preference order of stakeholders. }
	\label{fig: utility total}
\end{figure}

For the preference order of stakeholders, we look at two scenarios
(a) $\langle$''less packet loss,'' ''less energy consumption''$\rangle$, and (b) $\langle$''less energy consumption,'' ''less packet loss''$\rangle$. 
Figure\,\ref{fig: utility total 2} shows the utilities for scenario (b). The pre-defined classifier with ML2ASR and the pre-defined classifier with LSA (no operator feedback) perform similarly with the same mean utility of 0.59. The evolving classifier with LSA and operator feedback with a mean utility of 0.76 outperforms the pre-defined classifier with ML2ASR  and the pre-defined classifier with LSA (no operator feedback), i.e., a difference in mean utility of 0.16 and 0.17 respectively. When we compare the mean utility of the evolving classifier with LSA and operator feedback with the baseline we observe a small difference in utility of 0.04. 
Figure\,\ref{fig: utility total 1}) shows the utilities for scenario (a). Here too, the pre-defined classifier with ML2ASR and the pre-defined classifier with LSA (no operator feedback) perform similarly with the same mean utility of 0.78. Yet, the difference in the mean utility for the evolving classifier with LSA and operator feedback with the other approaches is smaller, namely 0.03 and 0.04 compared to the pre-defined classifier with ML2ASR and the pre-defined classifier with LSA (no operator feedback). For the mean utility of the evolving classifier with LSA and operator feedback with the baseline, we observe again a small difference in utility of 0.05. 
With a significance level of 0.05, the test results of Mann-Withney U tests for both scenarios and additionally Wilcoxon signed-rank tests for scenario (a) support the following hypotheses in both scenarios: (i) the utility of evolving classifier with LSA and operator feedback is higher than the utility of the pre-defined classifier with ML2ASR and (ii) the utility of evolving classifier with LSA and operator feedback is higher than the utility of the pre-defined classifier with LSA. Furthermore, for both scenarios, the test results do not provide support that the utility of the baseline is higher than the utility of the evolving classifier with LSA and operator feedback.  From a practical point of view, the difference in the mean utility of the evolving classifier with LSA and operator feedback with the baseline is negligible; so we can conclude that the evolving classifier with LSA and operator feedback performs close to the ideal classifier.

\begin{figure}[!ht]
	\centering
	\begin{subfigure}[b]{0.49\textwidth}
		\includegraphics[width=\textwidth]{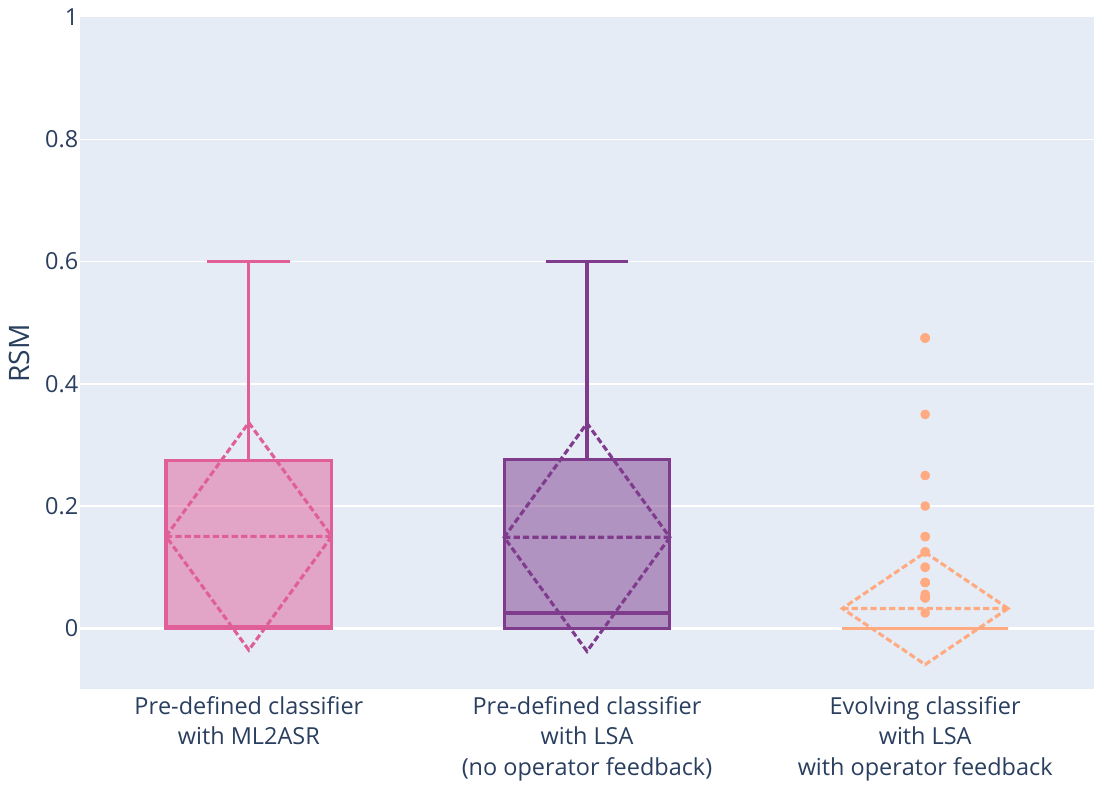}
		\caption{$\langle$``less packet loss'', ``less energy consumption''$\rangle$}
		\label{fig: rsm total 1}
	\end{subfigure}
	\begin{subfigure}[b]{0.49\textwidth}
		\includegraphics[width=\textwidth]{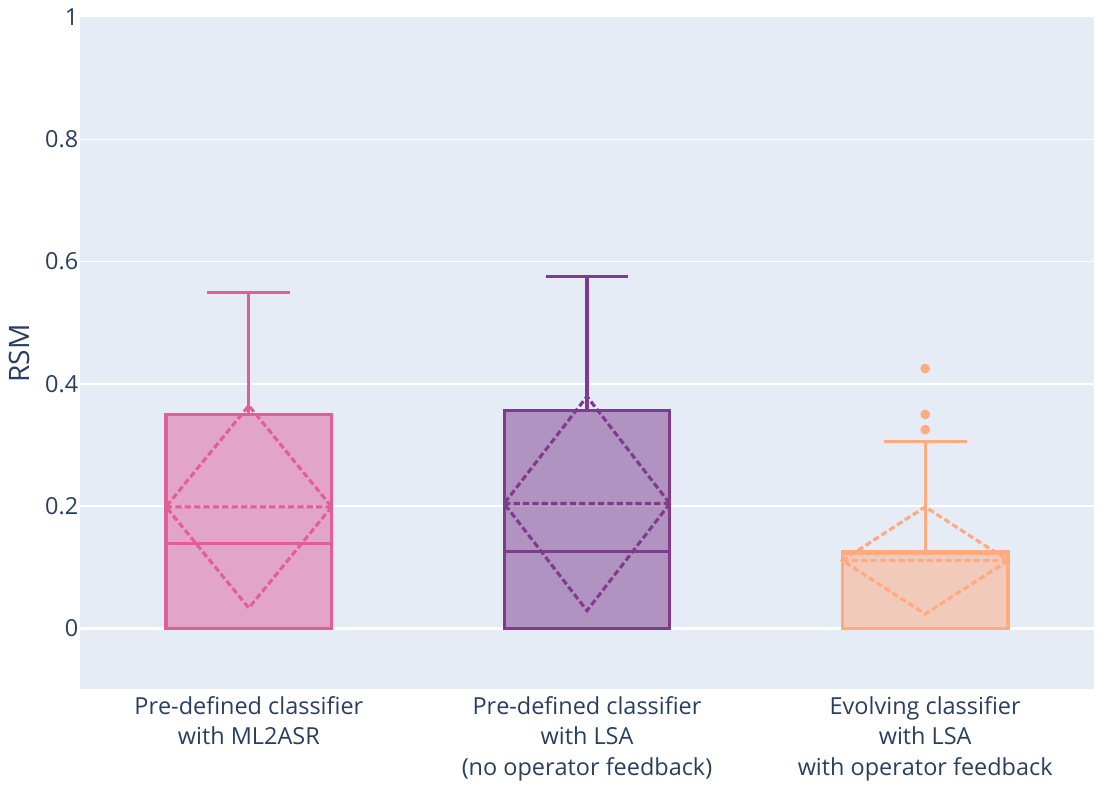}
		\caption{$\langle$``less energy consumption'', ``less packet loss''$\rangle$}
		\label{fig: rsm total 2}
	\end{subfigure}
	\caption{\rsm\ values for all class appearance order scenarios split for  preference order of stakeholders. }
	\label{fig: rsm total}
\end{figure}

The results for the mean \rsm\ confirm these analyses. Figure\,\ref{fig: rsm total 2} shows the results for scenario (b). The pre-defined classifier with ML2ASR and the pre-defined classifier with LSA (no operator feedback) perform similarly, with the same mean \rsm\ of 0.20. The mean \rsm\ is 0.09 lower for the evolving classifier with LSA and operator feedback compared to both other approaches (0.11 versus 0.20 respectively). Figure\,\ref{fig: rsm total 1} shows the results for scenario (a). Here too, the pre-defined classifier with ML2ASR and the pre-defined classifier with LSA (no operator feedback) perform similarly, with the same mean \rsm\ of 0.15. The mean \rsm\ of the evolving classifier with LSA and operator feedback is 0.12 lower compared to the pre-defined classifier with ML2ASR and the pre-defined classifier with LSA (no operator feedback) (0.03 versus 0.15 respectively). 
With a significance level of 0.05, the results of Mann-Withney U tests for both scenarios support the hypotheses: (i) the \rsm\ of evolving classifier with LSA and operator feedback is less than the \rsm\ of the pre-defined classifier with ML2ASR, and (ii) the \rsm\ of evolving classifier with LSA and operator feedback is less than the \rsm\ of the pre-defined classifier with LSA. The \rsm\ for the evolving classifier with LSA and operator feedback are in both scenarios close to zero (0.11 and 0.03 for scenarios (b) and (a) respectively) indicating that the performance of the classifier is close to the ideal classifier. 

\paragraph{Conclusion.}
In answer to EQ2, we can conclude that an evolving classifier with LSA with operator feedback is robust both to different appearance order of classes and to different preference orders of the stakeholder, and this is for all evaluated scenarios (while the results indicate that the competing approaches are not robust in half of the scenarios; the other half seem non-challenging scenarios).

\subsubsection{Effectiveness of Operator Feedback in Lifelong Self-Adaptation for Dealing with Drift of Adaptation Spaces.}

To answer EQ3, we leverage the results presented above for the 24 scenarios. 

\begin{figure}[!ht]
	\centering
	\begin{subfigure}[b]{0.49\textwidth}
		\includegraphics[width=\textwidth]{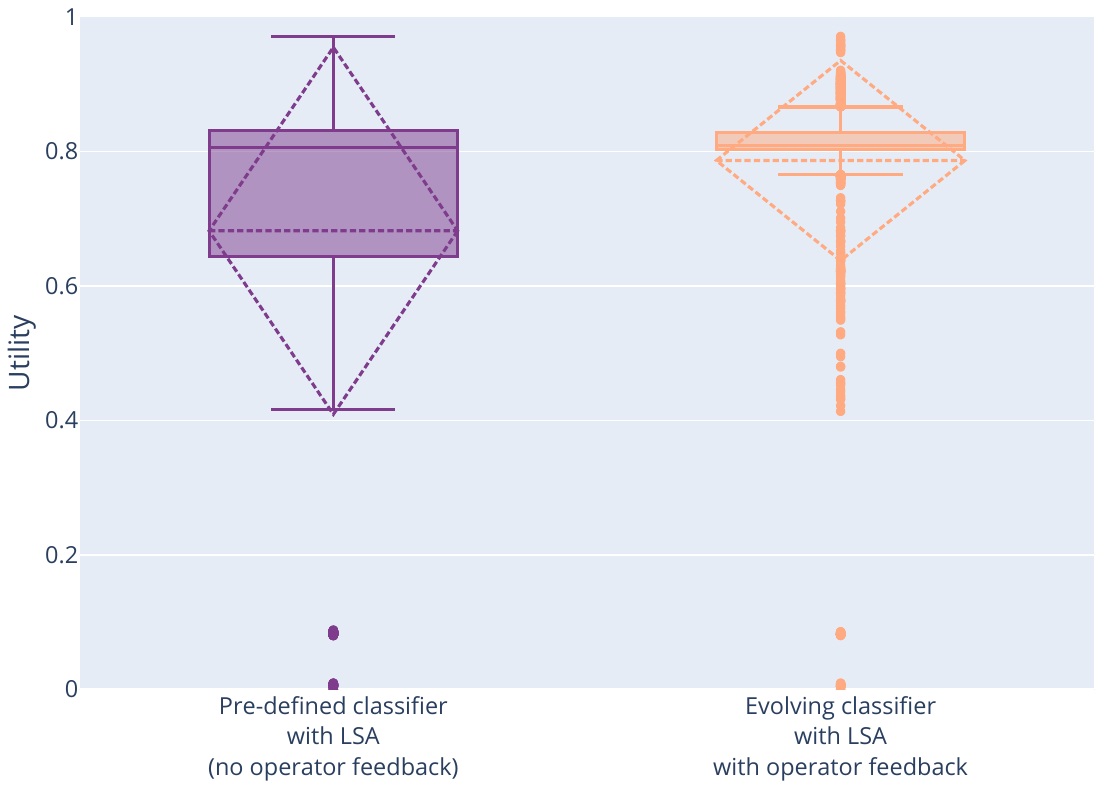}
		\caption{Impact on utilities.}
		\label{fig: utility total total}
	\end{subfigure}
	\begin{subfigure}[b]{0.49\textwidth}
		\includegraphics[width=\textwidth]{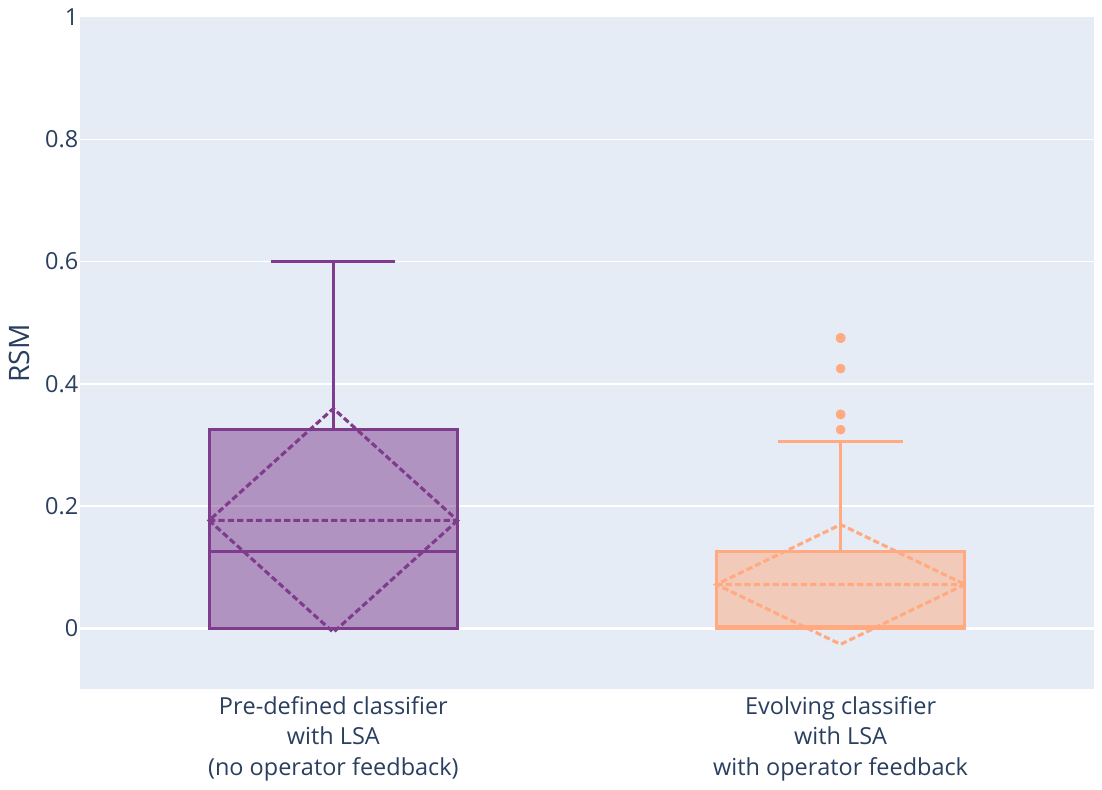}
		\caption{Impact on \rsm.}
		\label{fig: rsm total total}
	\end{subfigure}
	\caption{Impact of the operator feedback on utilities and \rsm\ in all scenarios.}
	\label{fig: utility rsm total total}
\end{figure}

Figure\,\ref{fig: utility total total} summarises the effect on the utility of the system over all scenarios for the pre-defined classifier with LSA with and without operator feedback. The difference of 0.11 in the mean utility (mean 0.79 with operator feedback and 0.68 without) shows that operator feedback contributes substantially to the utility of a self-adaptive system that faces drift of adaptation options.  With a significance level of 0.05, the result of a Mann-Whitney U test supports the hypothesis that the utility of the evolving classifier with LSA and operator feedback is higher than the utility of the evolving classifier with LSA without operator feedback (p-value 0.000). 

Figure\,\ref{fig: rsm total total} summarises the effect of operator involvement on the \rsm. The results for \rsm\ confirm the important role of the operator in dealing with a drift of adaptation spaces in self-adaptive systems. The difference in the mean of \rsm\ is 0.11 lower with operator feedback (mean 0.07 without operator feedback and 0.18 with feedback). With a significance level of 0.05, the result of a Mann-Withney U test supports the hypothesis that the \rsm\ of the evolving classifier with LSA and operator feedback is less than the \rsm\ of the evolving classifier with LSA without operator feedback (p-value 0.000). 

The boxplots also show that the interquartile ranges (i.e., the range between the 25 and the 75 percentile) for both metrics are substantially smaller for the classifier with operator feedback ([0.80, 0.82], i.e., 0.02 versus [0.64, 0.83], i.e., 0.19 for utility and [0.00, 0.13] versus [0.00, 0.33] for \rsm). This shows that the classifier with operator feedback provides high-quality decisions in most of the adaptation cycles, which is not the case for the classifier without operator feedback.

\paragraph{Conclusion.}
In answer to EQ3, we can conclude that operator feedback is particularly effective in dealing with drift of adaptation spaces; in fact operator feedback is essential.

\subsection{Threats to Validity}

The evaluation of lifelong \re{self-adaptation} to deal with drift of adaptation spaces is subject to a number of validity threats. To that end, we follow the guidelines provided in\,\cite{Wohlin:2012,wieringa2014design}. 

\subsubsection{Construct Validity}

Construct validity is about whether we have obtained the right measures to evaluate the proposed approach. Since there is (to the best of our knowledge) no competing approach that deals with shift of adaptation spaces in learning-based self-adaptive systems, we compared the proposed approach with a baseline (with an ideal classifier) and a state-of-the-art approach that uses learning in the analysis stage of self-adaptation. We used the impact on the quality properties as a primary metric to measure the usefulness of the approach in comparison with the other approaches. To measure the satisfaction level of the stakeholders of the system, we then computed the utility (using the data of the quality attributes) and the ranking satisfaction mean based on the classification of the selected adaptation options to compare the different approaches. 

\subsubsection{Internal Validity}

Internal validity concerns drawing a causal conclusion based on the study. We have evaluated the instances of the architecture for the DeltaIoT case using particular settings. The type and number of new emerging classes generated in these settings may have an effect on the difficulty of the problems. We mitigated this threat by instantiating the architecture for in total of 24 different scenarios. These scenarios consider different patterns in new emerging classes. However, additional evaluation for other instances and in other domains is required to increase the validity of the results for the type of concept drift we studied.

\subsubsection{External Validity}

External validity concerns the generalization of the study results. We evaluated the approach only for one type of learner used in self-adaptive systems, so we cannot generalize the findings for other types of learners that require dealing with a new shift in adaptation spaces. Additional research is required to study the usefulness of the approach and the architecture for other use cases of learning that may be affected by a drift of adaptation spaces.  Additionally, we validated the architecture with a single application. Evaluation in different domains is required to increase the validity of the results for the type of concept drift considered in this paper. 

\subsubsection{Reliability}

For practical reasons, we used the simulator of the DeltaIoT network. For the evaluation we used data that contains uncertainty. Hence, the results may not necessarily be the same if the study would be repeated. We minimized this threat by considering stochastic data that was based on observations from real settings of the network, and we evaluated the different scenarios over long periods of time. We also provide a replication package for the study~\cite{project-website}.

\subsubsection{Statistical Conclusion}

Statistical conclusion validity concerns the degree to which conclusions drawn from statistical data analyses are accurate and appropriate~\cite{iftikhar2017deltaiot}. To ensure that we have used proper statistical tests, we checked the distribution of the data and applied appropriate statistical tests based on the characteristics of the data. On the other hand, we acknowledge that the p-values that demonstrate statistical relevance are based on a limited set of experiments. To enhance the statistical significance of these results, repeated experiments or studies are needed that can confirm the results obtained in this paper. We leave this as an option for future work. 

\section{Related Work}\label{sec:relatedWork}

We look at a selection of work at the crossing of machine learning and self-adaptation, focusing on approaches for (i) \re{dealing with concept drift in machine learning,} (ii) dealing with concept drift in self-adaptive systems, (iii) improving the performance of machine learning in self-adaptive systems, and (iv) dealing with unknowns in self-adaptive systems. 

\re{
\subsection{Dealing with Concept Drift in Machine Learning}

Lu et al.~\cite{lu2018learning} studied concept drift in the machine learning literature and identified two main research areas: drift detection and drift adaptation.
Based on the methods described in the literature they proposed a general architecture that
comprises four stages of concept drift detection in data stream analysis. Stage 1 retrieves chunks of data from data streams and organizes them to form a meaningful pattern. Stage 2 (optional) 
abstracts the data and extracts key features. Stage 3 calculates dissimilarity between data sets and was considered as the most challenging aspect of concept drift detection. Stage 4 uses a specific hypothesis test to evaluate the statistical significance of the change observed in Stage 3 to accurately determine the detection of drift. Compared with our general architecture, Stage 1 is part of the process of data collection performed by the knowledge manager. Stage 2 to Stage 4 can be part of task identification in the task manager. Therefore, existing detection methods seem to fit appropriately into our proposed general architecture.

When drift is detected, it needs to be managed. The method required to manage drift depends on the type of detected drift. Three main groups of methods were proposed by Lu et al.~\cite{lu2018learning}: training a new model, ensemble training, and model adjusting. Training a new model uses the latest data to replace the obsolete model, which maps to a structural change. Ensemble methods reused old models for recurring drifts, while model adjusting develops a model that adaptively learns from changing data; hence, decision tree algorithms were commonly used for this approach. Both ensemble methods and model-adjusting approaches could be considered instances of parametric changes. These three adaptation methods could be applied at any point in the data stream. 

The general architecture for lifelong self-adaptation includes all the necessary elements for drift adaptation. The task manager is responsible for drift detection and understanding, such as when, how, and where the drift occurs (e.g., using previously detected tasks to handle recurrent drift). The task-based knowledge miner is responsible for mining useful data for adapting the learning model. The knowledge-based learner is responsible for adapting the learning model based on the task-based knowledge-mined data. Meanwhile, the knowledge manager collects all required data for the operation of other components in the lifelong learning layer. Therefore, all proposed drift adaptation methods can be integrated into our proposed general architecture. 

}

\subsection{Dealing with Concept Drift in Self-Adaptation} 
T. Chen~\cite{chen2019} studied the impact of concept drift on machine learning models caused by uncertainties; focusing on models used by a self-adaptive system to evaluate and predict performance to make proper adaptation decisions. Two methods were studied: retrained modeling that always discarded the old model and retrained a new one using all available data, and incremental modeling that retained the existing model and tuned it using one newly arrived data sample. Usually, the choice for one of them was based on general beliefs. In contrast, the author reported an empirical study that examined both modeling methods for distinct domains of adaptable software and identified evidence-based factors that could be used to make well-informed decisions for choosing a method. 

Bierzynski et al.~\cite{Bierzyn2019ski} presented the architecture of a self-learning lighting system that equips a MAPE-K loop with a learner that learns activities and user preferences. The learner is equipped with a dedicated component that realizes another feedback loop on top of the learner. This inner feedback loop determines when the predictions of a model start to drift and then adapts the learning model accordingly. Previously recognized drift patterns are exploited to enhance the performance of new drift and minimize the need for human intervention. 
The authors proposed a concrete implementation based on \re{the micro-service pattern and evaluated the result with different other architectural patterns, e.g., Mikrokernel and Monolith.} 

Vieira et al.~\cite{vieira2021} proposed Driftage, a multi-agent systems framework to simplify implementing concept drift detectors. The framework divides concept drift detection responsibilities between agents. The approach is realized as a MAPE-K loop, where monitor and analyzer agents capture and predict concept drifts in data, and planner and executor agents determine whether the detected concept drift should be alerted. The authors illustrated their approach in the domain of health monitoring of muscle cells, combining different types of learners who vote for detecting drift. 

Casimiro et al.~\cite{Casimiro0GMKK21} discussed the implications of unexpected changes, e.g., drifts in the input data on using machine learning-based systems. The authors proposed a framework for self-adaptive systems that relies on machine-learned components. The paper outlined (i) a set of causes of misbehavior of machine-learned component\re{s} and a set of adaptation tactics inspired by the literature, (ii) the required changes to the MAPE-K loop for dealing with the unexpected changes, and (iii) the challenges associated with developing this framework.

\begin{sloppypar}
\re{
In contrast to these approaches, we provide a general domain-independent architecture to deal with all types of concept drift in learning modules with variant learning tasks (i.e., regression and classification) used by self-adaptive systems. 
}
The work of~\cite{chen2019} and~\cite{Casimiro0GMKK21} offer valuable solutions that can be used when instantiating the architecture for lifelong self-adaptation. 

\end{sloppypar}

\subsection{Improving the Performance of Machine Learning for Self-Adaptation}
Jamshidi et al.~\cite{jamshidi2018} proposed an efficient approach for transferring knowledge across highly configurable environments to simplify the configuration, e.g., hardware, workload, and software release. The approach, called L2S (Learning to Sample), selects better samples in the target environment based on information from the source environment. L2S progressively shrinks and adaptively concentrates on interesting regions of the configuration space. The authors demonstrated that L2S outperformed state-of-the-art learning and transfer-learning approaches in terms of measurement effort and learning accuracy.

T. Chen and Bahsoon~\cite{Chen2017} presented a self-adaptive modeling approach that leverages information theory and machine learning algorithms to create a quality model for predicting a quality property over time by using data on environmental conditions and the control settings as inputs. Concretely, the authors used self-adaptive hybrid dual-learners that partition the input space in two sub-spaces, each applying a different symmetric uncertainty-based selection technique; the results of sub-spaces were then combined. The authors then used adaptive multi-learners for building the model of the QoS function, supporting selecting the best model for prediction on the fly. The approach was evaluated in a cloud environment. X.~Chen et al.~\cite{chen19} applied a similar approach to deal with the problem of resource allocation for cloud-based software services.

T.~Chen et al.~\cite{chen2010experience} proposed the notion of ``experience transfer’’ to utilize knowledge learned from one system to another similar system. The transfer process discovers and represents transferable experiences, extracts the experiences during the modeling process in the original system, and embeds learned experiences into the management of the new system. The authors demonstrated the process and benefits of experience transfer for system configuration tuning using a Bayesian network. In this case, the dependencies between configuration parameters are valuable experiences. 

These related approaches targeted the efficiency of machine learning methods in the context of self-adaptation. Our work complements these approaches focusing on enhancing learning to handle new learning tasks as required for concept drift and drift of adaptation spaces in particular. 

\subsection{Dealing with Unknowns in Self-Adaptive Systems}

Kinneer et al.\,\cite{10.1145/3194133.3194145} focused on the change of the adaptive logic in response to changes, such as the addition or removal of adaptation tactics. The authors argue that such changes in a self-adaptive system often require a human planner to redo expensive planning. To address this problem the authors proposed a planner based on genetic programming that reuses existing plans. The authors demonstrated that na\"ively reusing existing plans for planning in self-adaptive systems results in a loss of utility. This work fits in a line of research on automatic  (re-)generation of adaptive logic to deal with circumstances that are hard to anticipate before the deployment of the system. 

Palm et al.~\cite{palm2020online} integrated a policy-based reinforcement learner with the MAPE-K architecture to deal with 
environment changes that are difficult or impossible to anticipate before deployment of the system. Different from traditional online reinforcement learning approaches for self-adaptive systems that require manual fine-tuning of the exploration rate and quantizing environment states, the proposed approach automates these manual activities. To demonstrate the feasibility and applicability, the authors validated the proposed approach in two domains, namely to balance workloads in a web application subject to multiple types of drifts in the distribution of the workload, and to predict the behavior of a process in a business process management system when there is a shift in process behavior. The experiments show that there is room for improvement in the convergence of the reinforcement learning method and the ability to handle large adaptation spaces. 

Krupitzer et al.~\cite{7573157} coin the term self-improvement within self-adaptive systems as an adaptation of the adaptation logic that helps shift the integration tasks from the static design time to the runtime. The authors survey approaches for self-improvement, compare these approaches, and categorize them. The categorization highlights that the approaches focus either on structural or parameter adaptation but seldom combine both. From this insight, the authors outline a set of challenges that need to be addressed by future approaches for self-improvement.  

Recently, Alberts and Gerostathopoulos\,\cite{10.1007/978-3-031-19759-8-15} focused on context shifts in learning-based self-adaptive systems. The authors proposed a new metric, convergence inertia, to assess the robustness of reinforcement learning policies against context shifts. This metric is then used to assess the robustness of different policies within a family of multi-armed bandits against context shifts. The authors argue through an experiment with a self-adaptation web server that inertia and the accompanying interpretation of the unknown-unknowns problem is a viable way to inform the selection of online learning policies for self-adaptive systems. 

These related approaches exploited learning approaches to solve problems with unknown or unforeseen conditions in self-adaptive systems. Lifelong self-adaptation contributes to this line of research with an approach that enables a learning-based self-adaptive system to deal with new learning tasks with a focus on a drift of adaptation spaces. 
\section{Conclusions and Future Work}\label{sec:conclusions}

This paper started from the research problem ''how to enable learning-based self-adaptive systems to deal with drift of adaptation spaces
during operation, i.e., concept drift in the form of novel class appearance?'' We illustrated the potentially severe effect of a shift of adaptation in terms of achieving the quality attributes of the adaptation goals, the utility of self-adaptation, and the ranking satisfaction mean, a novel metric to measure the satisfaction level of stakeholders in terms of class ranking of the selected adaptation option of a pre-defined classifier versus an ideal classifier. 

To tackle the research problem we presented \re{a general architecture for lifelong self-adaptation that supports learning-based self-adaptive systems to deal with new learning tasks during operation. We instantiated the general architecture to deal with a drift of adaptation spaces using the DeltaIoT exemplar.} Empirical results of 24 different scenarios demonstrate that lifelong self-adaptation is effective and robust to drift in adaptation spaces. The operator takes a key role in ranking classes, including new classes, in the lifelong learning loop that underlies lifelong self-adaptation. 

As future work, the proposed approach of lifelong self-adaptation could be applied to the problem of drift of adaptation spaces \re{for different scenarios and different application domains, adding to the external validity of the proposed approach. The knowledge extracted from such studies 
may yield solutions that may be reusable across domains.} 
\re{Another interesting line of future work could be dealing with dynamically changing goals. Yet another option for future work would be to study the challenges associated with catastrophic forgetting when applying lifelong self-adaptation.} 
Beyond that, it would be interesting to study other problems that a learning-based self-adaptive system may face leveraging the generic approach of lifelong self-adaptation. One example could be
the automatic generation of adaptation strategies (e.g., to reconfigure) under new encountered conditions. An inspiring example based on automatic synthesis is presented in\,\cite{NAHABEDIAN2022101850}. 
Another example is employing lifelong self-adaptation for realizing situation awareness in self-adaptive systems, inspired by~\cite{lesch2022self}. The task-manager component can play a central role in this way. 
In the long term, an interesting topic for research could be to investigate how lifelong self-adaptation can be enhanced to build systems that can truly evolve themselves. Inspiration in that direction can be found in research on self-improvement\,\cite{7573157} and self-evolution\,\cite{self-evolve}.




\newpage
\appendix

\section{Complete Validation Scenarios}
\label{sec: appendix validation scenarios}

\begin{figure}[htbp]
	\centering
	\begin{subfigure}[b]{0.32\textwidth}
		\includegraphics[width=\textwidth]{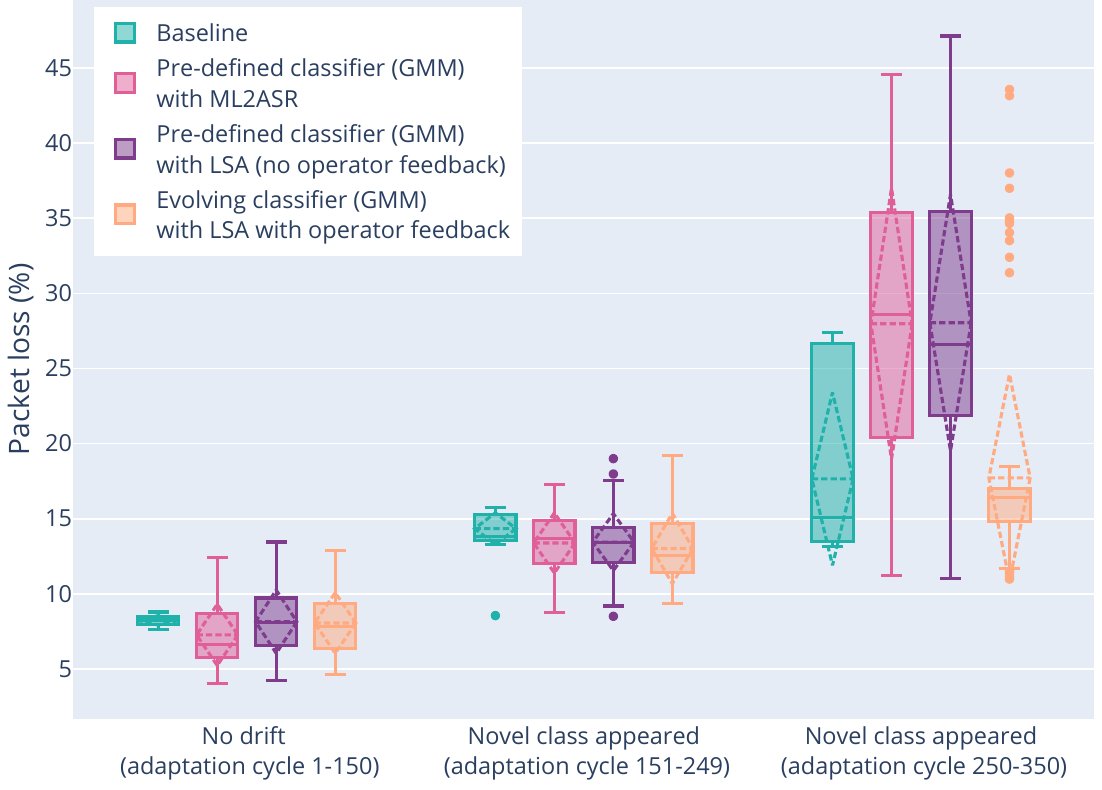}
		\caption{$\langle$(B), R, G$\rangle$}
		\label{fig: pl blue red green}
	\end{subfigure}
	\begin{subfigure}[b]{0.32\textwidth}
		\includegraphics[width=\textwidth]{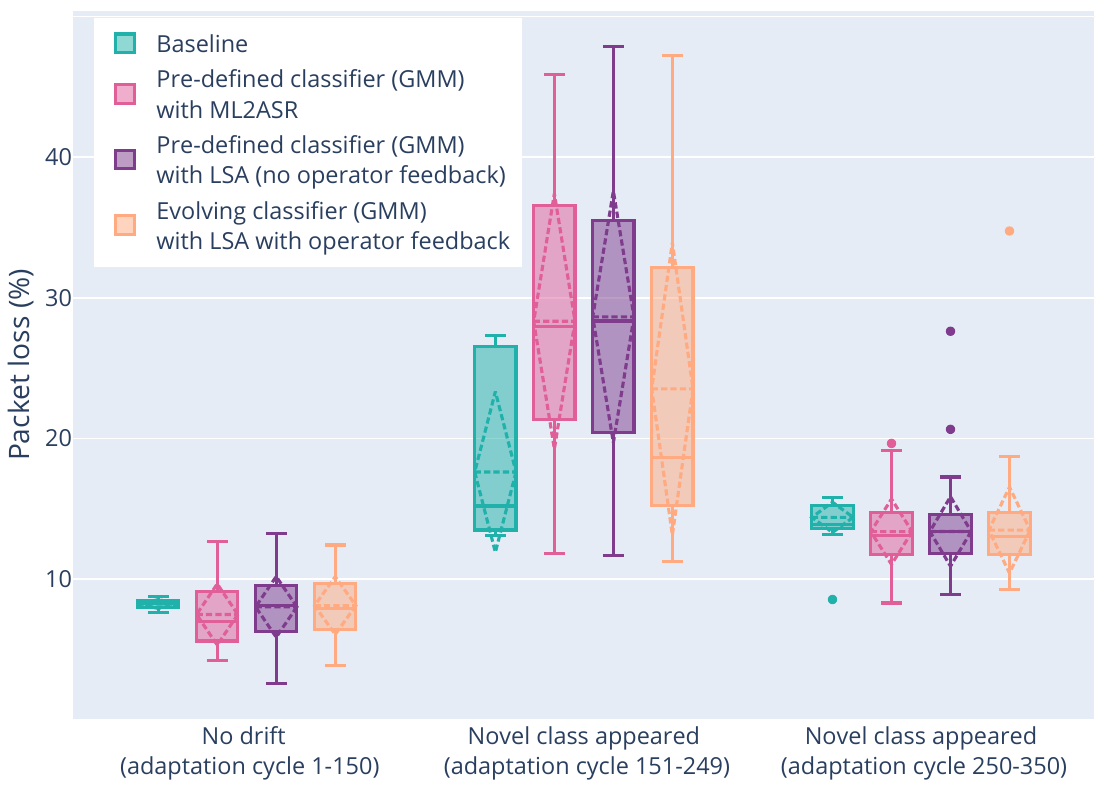}
		\caption{$\langle$(B), G, R$\rangle$}
		\label{fig: pl blue green red}
	\end{subfigure}
	\begin{subfigure}[b]{0.32\textwidth}
	\includegraphics[width=\textwidth]{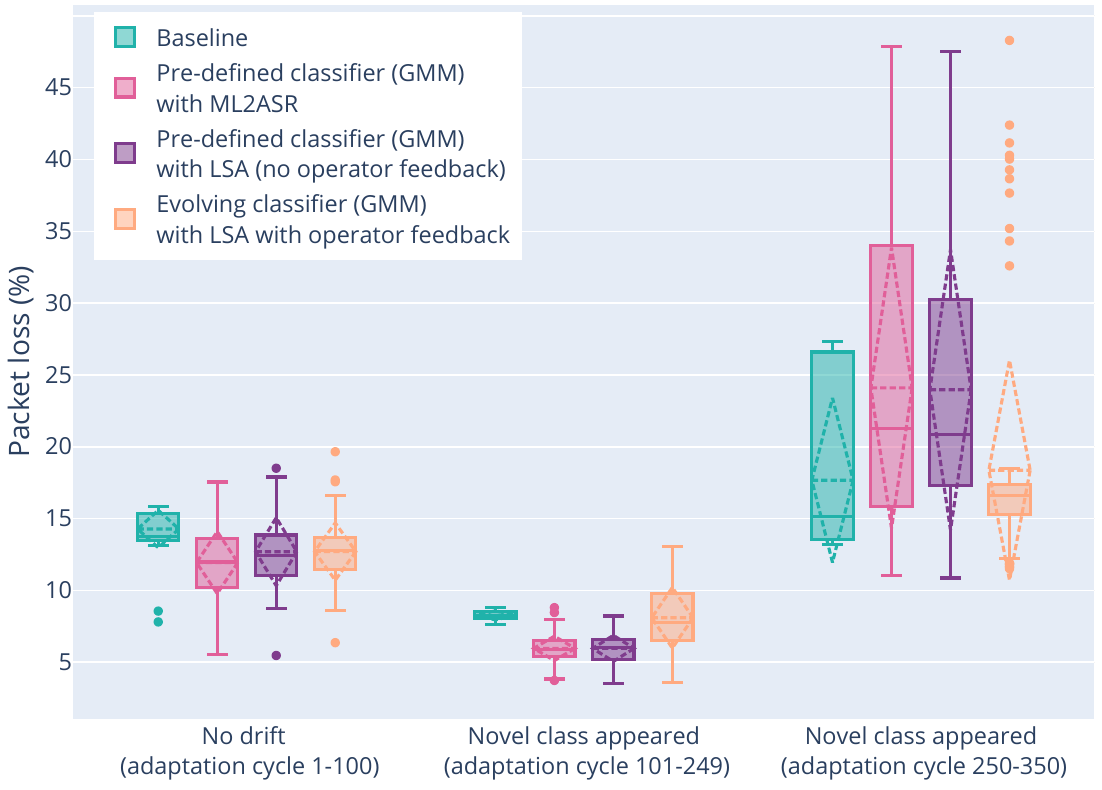}
	\caption{$\langle$(R), B, G$\rangle$}
	\label{fig: pl red blue green}
	\end{subfigure}
\bigskip
	\begin{subfigure}[b]{0.32\textwidth}
	\includegraphics[width=\textwidth]{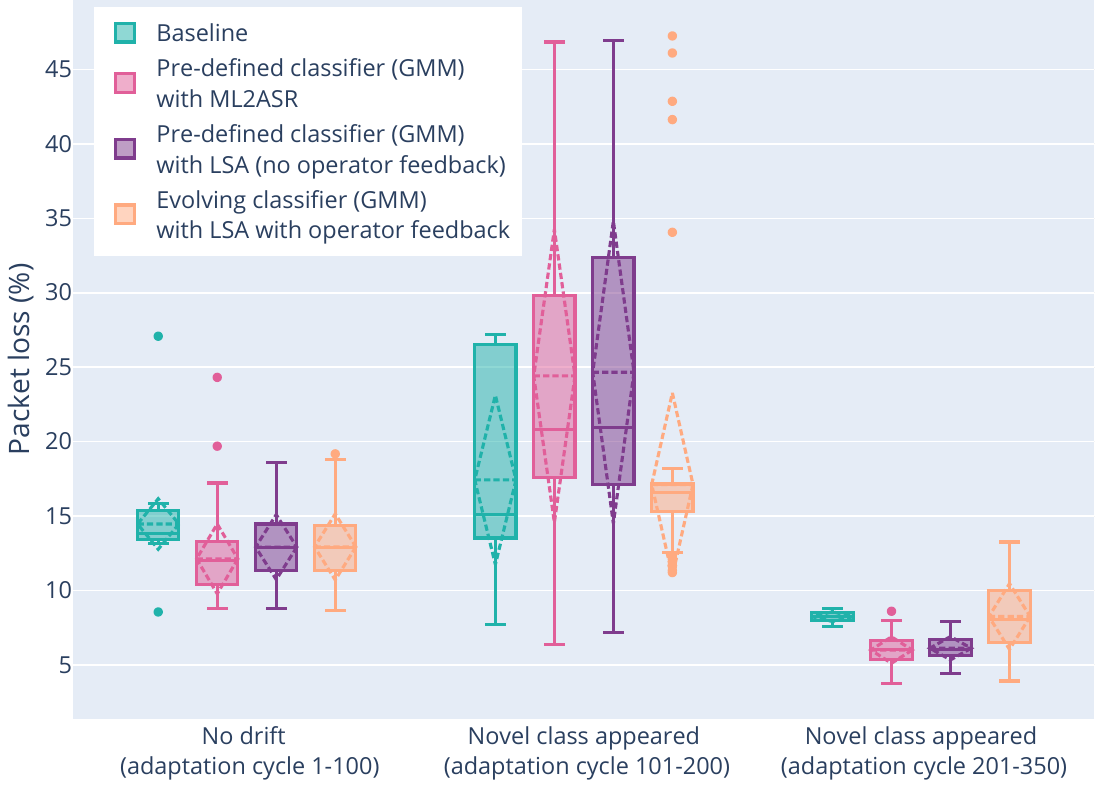}
	\caption{$\langle$(R), G, B$\rangle$}
	\label{fig: pl red green blue}
    \end{subfigure}
	\begin{subfigure}[b]{0.32\textwidth}
	\includegraphics[width=\textwidth]{figures/scenarios_1/pl/pl_problem_quality_attributes_dist_7.pdf}
	\caption{$\langle$(B, R), G$\rangle$}
	\label{fig: pl (blue red) green}
\end{subfigure}
\begin{subfigure}[b]{0.32\textwidth}
	\includegraphics[width=\textwidth]{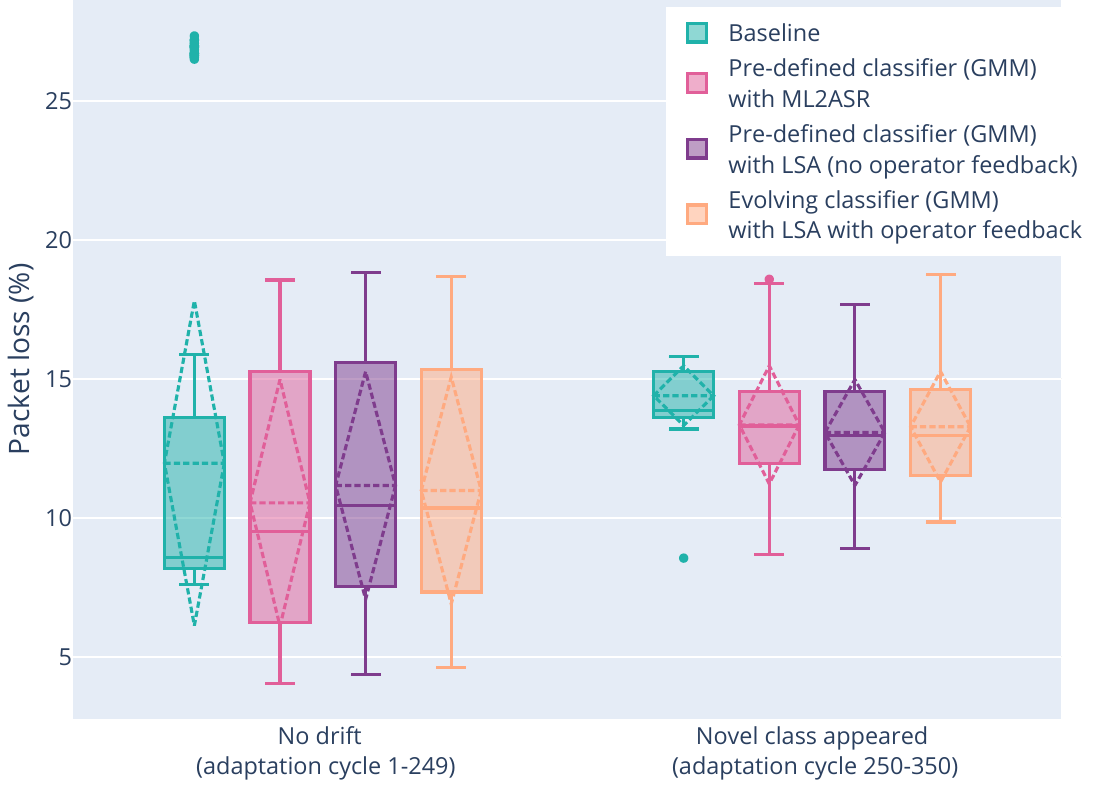}
	\caption{$\langle$(B, G), R$\rangle$}
	\label{fig: pl (blue green) red}
\end{subfigure}\vspace{-20pt}
	\caption{In terms of packet loss, evaluation of the lifelong self-adaptation  with and without the feedback of the stakeholder by comparing with the state-of-the-art (the pre-defined classifier supported by ML2ASR), and the baseline. The preference of the stakeholder is here $\langle$``less packet loss'', ``less energy consumption''$\rangle$. The appearance order of classes corresponding to each plot is mentioned in its caption.}\vspace{-20pt}
	\label{fig:all compare performance on packet loss}
\end{figure}

\begin{figure}[htbp]
	\centering
	\begin{subfigure}[b]{0.32\textwidth}
		\includegraphics[width=\textwidth]{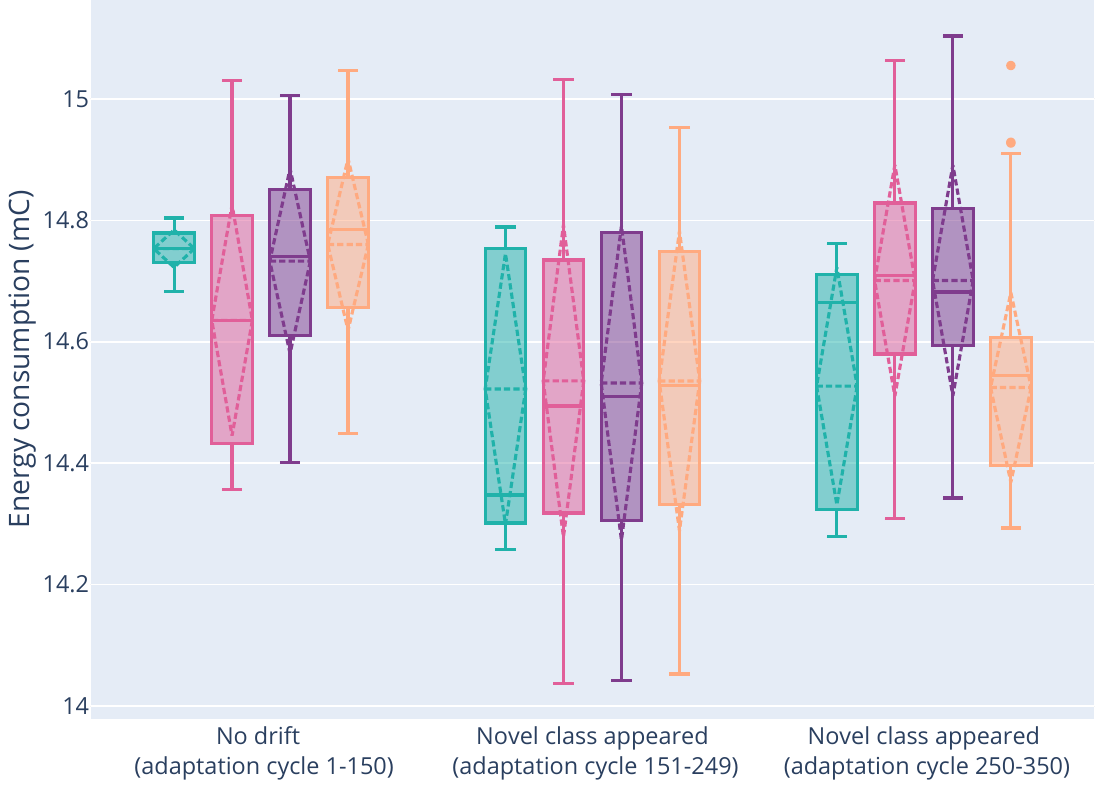}
		\caption{$\langle$(B), R, G$\rangle$}
		\label{fig: ec blue red green}
	\end{subfigure}
	\begin{subfigure}[b]{0.32\textwidth}
		\includegraphics[width=\textwidth]{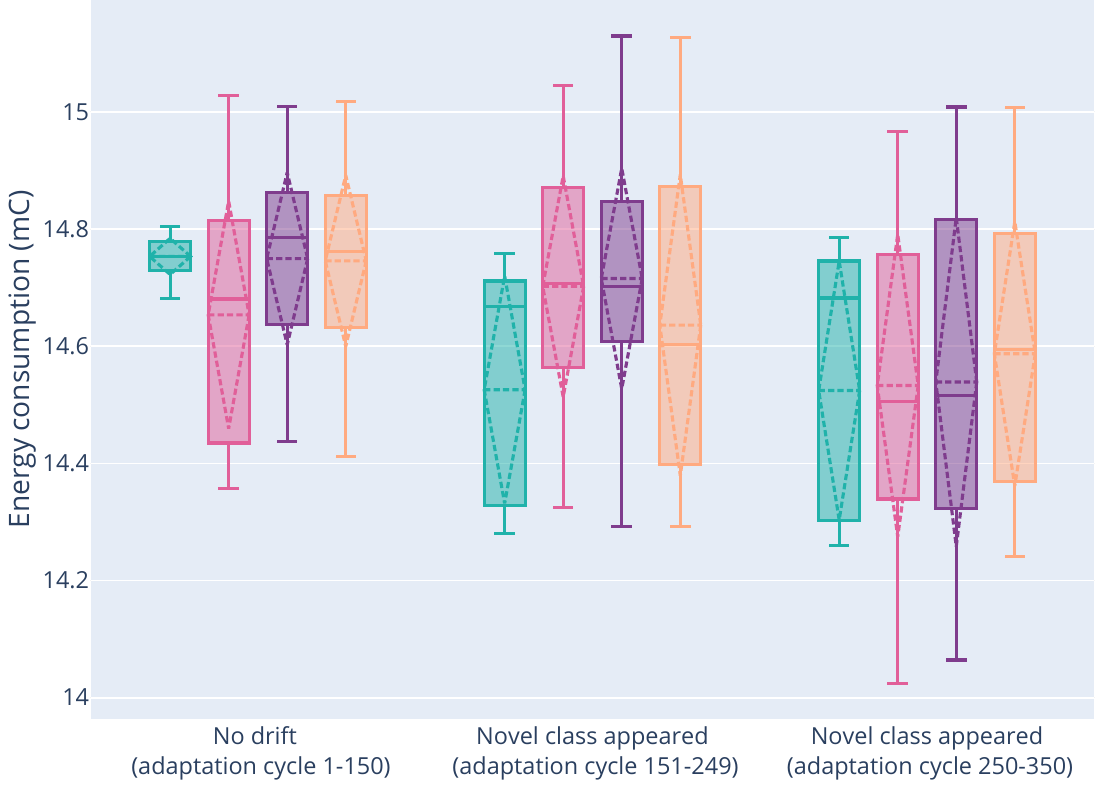}
		\caption{$\langle$(B), G, R$\rangle$}
		\label{fig: ec blue green red}
	\end{subfigure}
	\begin{subfigure}[b]{0.32\textwidth}
		\includegraphics[width=\textwidth]{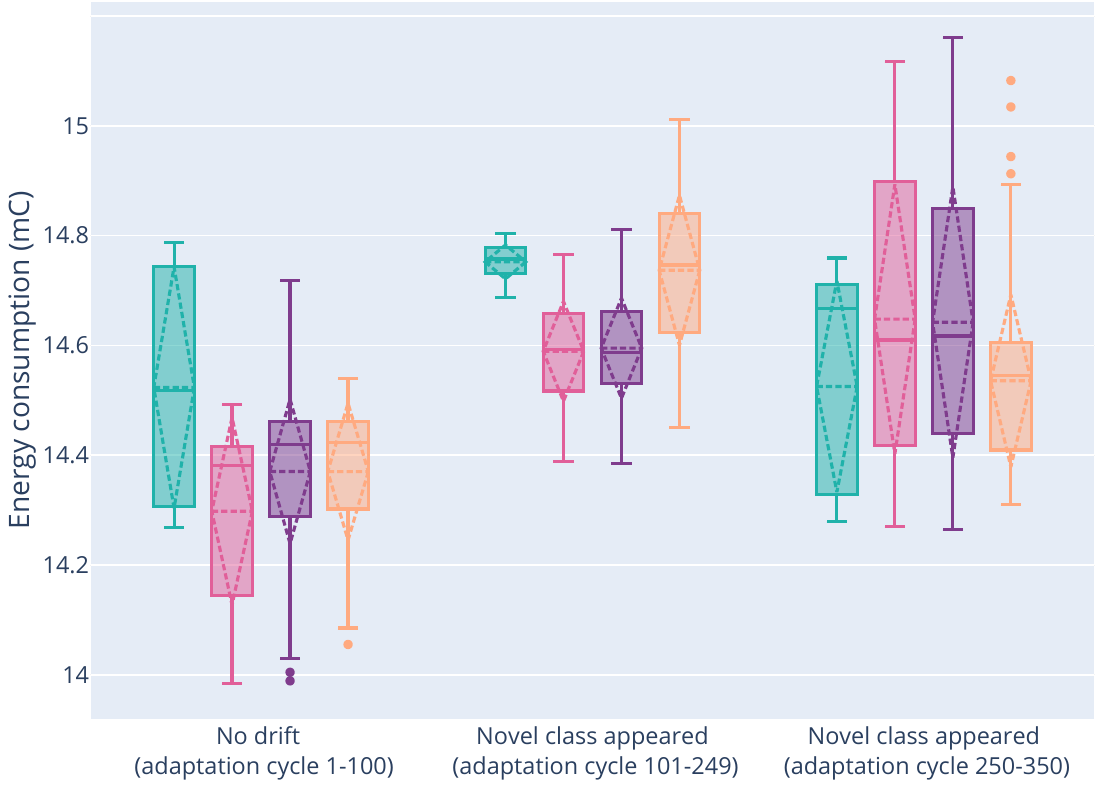}
		\caption{$\langle$(R), B, G$\rangle$}
		\label{fig: ec red blue green}
	\end{subfigure}
	\bigskip
	
	\begin{subfigure}[b]{0.32\textwidth}
		\includegraphics[width=\textwidth]{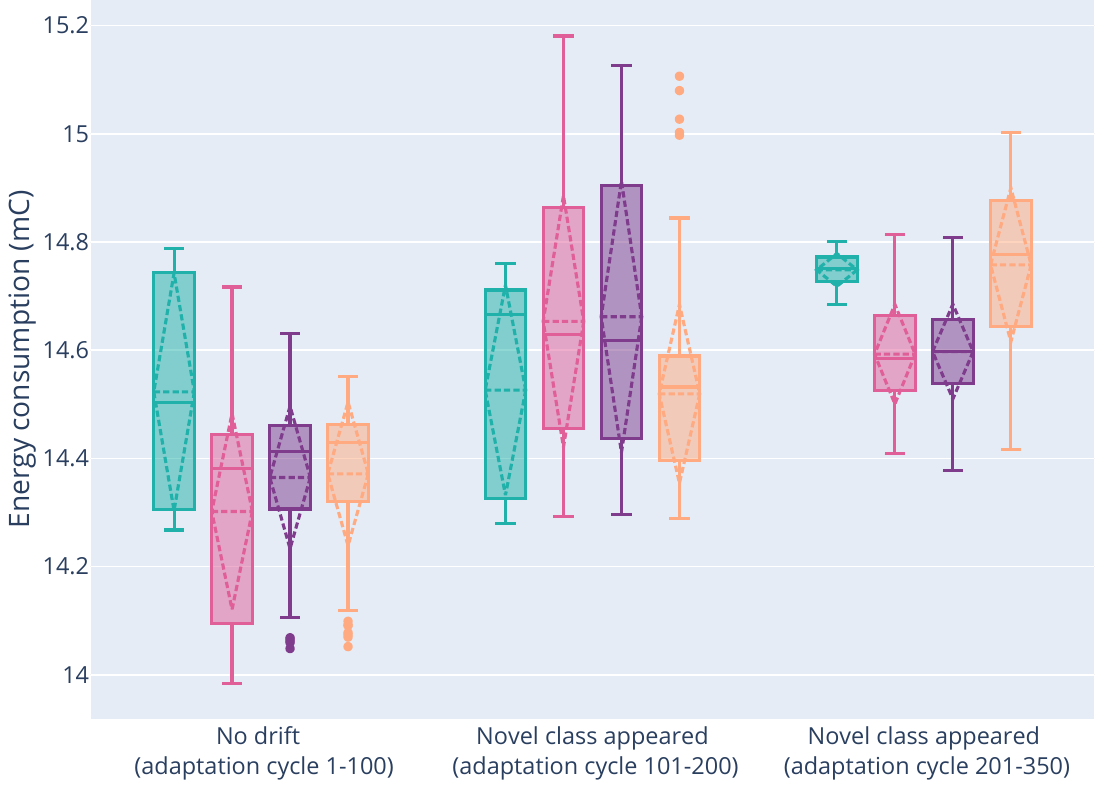}
		\caption{$\langle$(R), G, B$\rangle$}
		\label{fig: ec red green blue}
	\end{subfigure}
	\begin{subfigure}[b]{0.32\textwidth}
		\includegraphics[width=\textwidth]{figures/scenarios_1/ec/ec_problem_quality_attributes_dist_7.pdf}
		\caption{$\langle$(B, R), G$\rangle$}
		\label{fig: ec (blue red) green}
	\end{subfigure}
	\begin{subfigure}[b]{0.32\textwidth}
		\includegraphics[width=\textwidth]{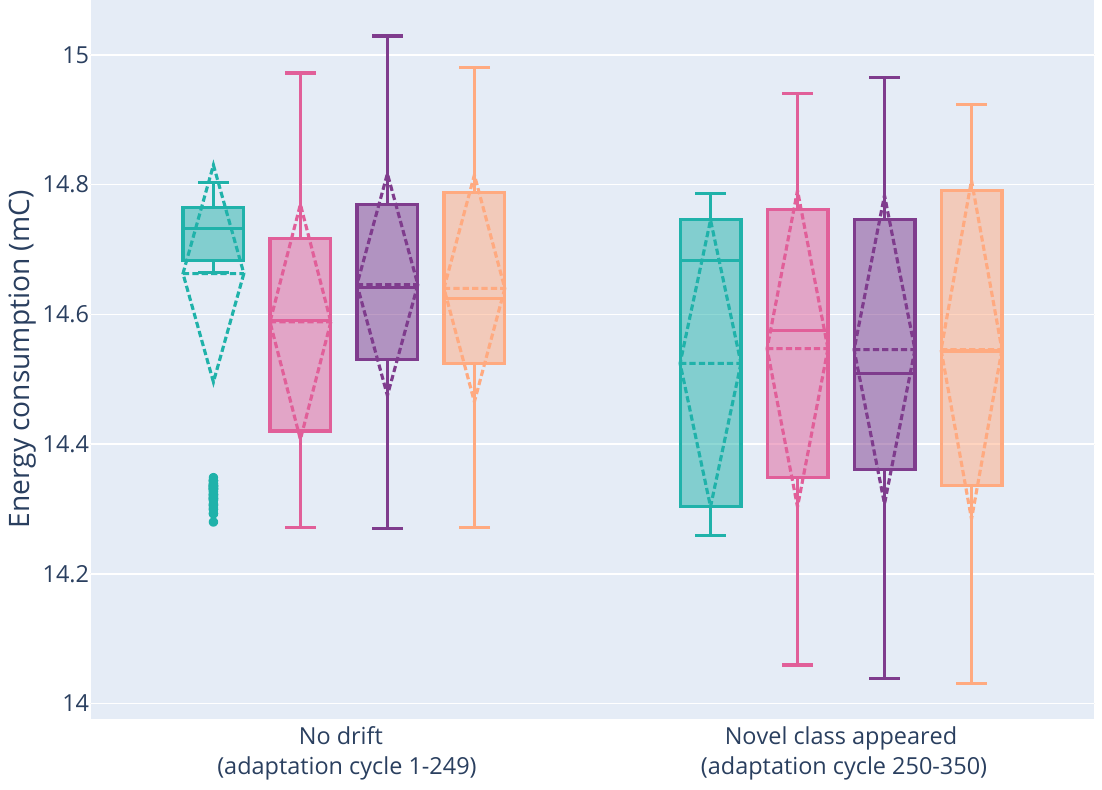}
		\caption{$\langle$(B, G), R$\rangle$}
		\label{fig: ec (blue green) red}
	\end{subfigure}
	\caption{In terms of energy consumption, evaluation of the lifelong self-adaptation  with and without the feedback of the stakeholder by comparing with the state-of-the-art (the pre-defined classifier supported by ML2ASR), and the baseline. The preference of the stakeholder is here $\langle$``less packet loss'', ``less energy consumption''$\rangle$.  The appearance order of classes corresponding to each plot is mentioned in its caption.}\vspace{-10pt}
	\label{fig:all compare performance on energy consumption}
\end{figure}

\begin{figure}[htbp]
	\centering
	\begin{subfigure}[b]{0.32\textwidth}
		\includegraphics[width=\textwidth]{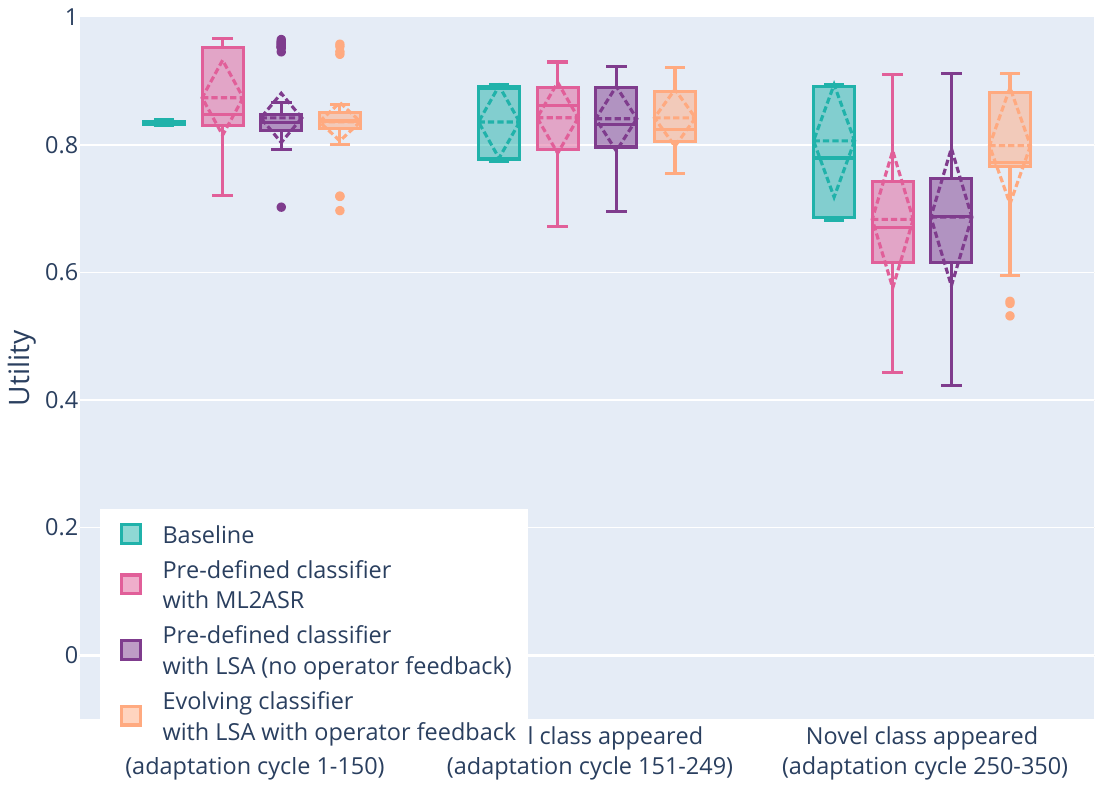}
		\caption{$\langle$(B), R, G$\rangle$}
		\label{fig: utility blue red green}
	\end{subfigure}
	\begin{subfigure}[b]{0.32\textwidth}
		\includegraphics[width=\textwidth]{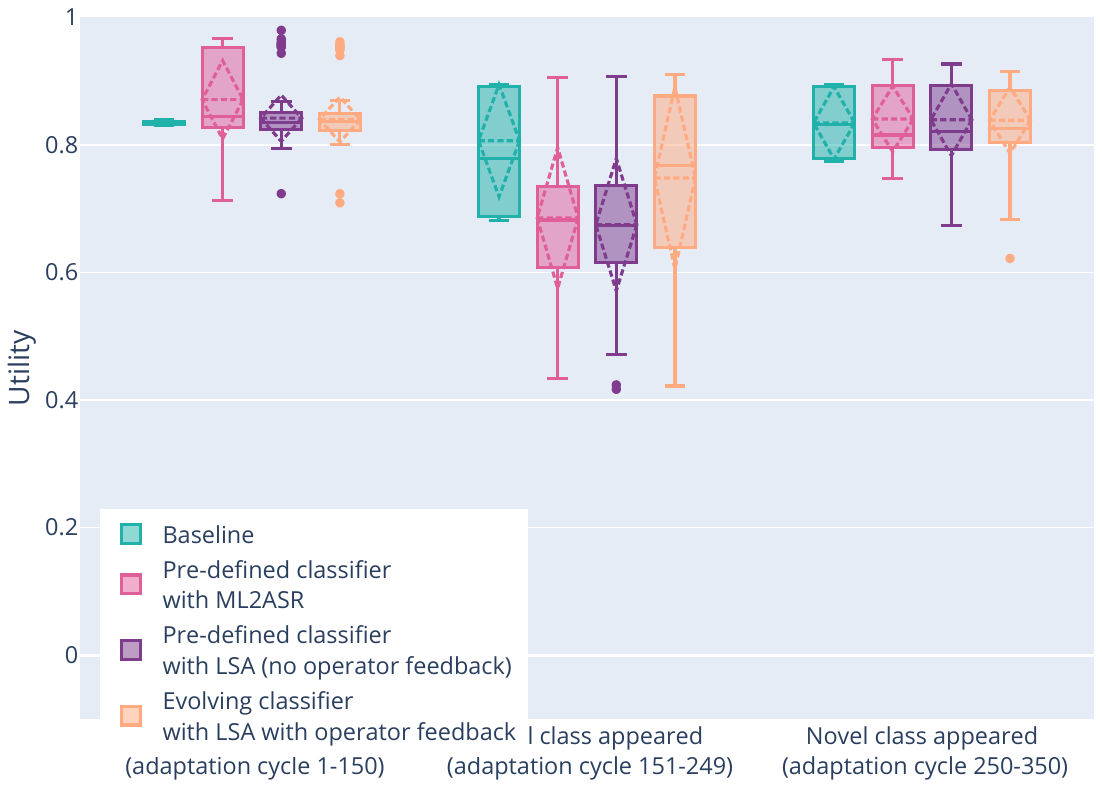}
		\caption{$\langle$(B), G, R$\rangle$}
		\label{fig: utility blue green red}
	\end{subfigure}
	\begin{subfigure}[b]{0.32\textwidth}
		\includegraphics[width=\textwidth]{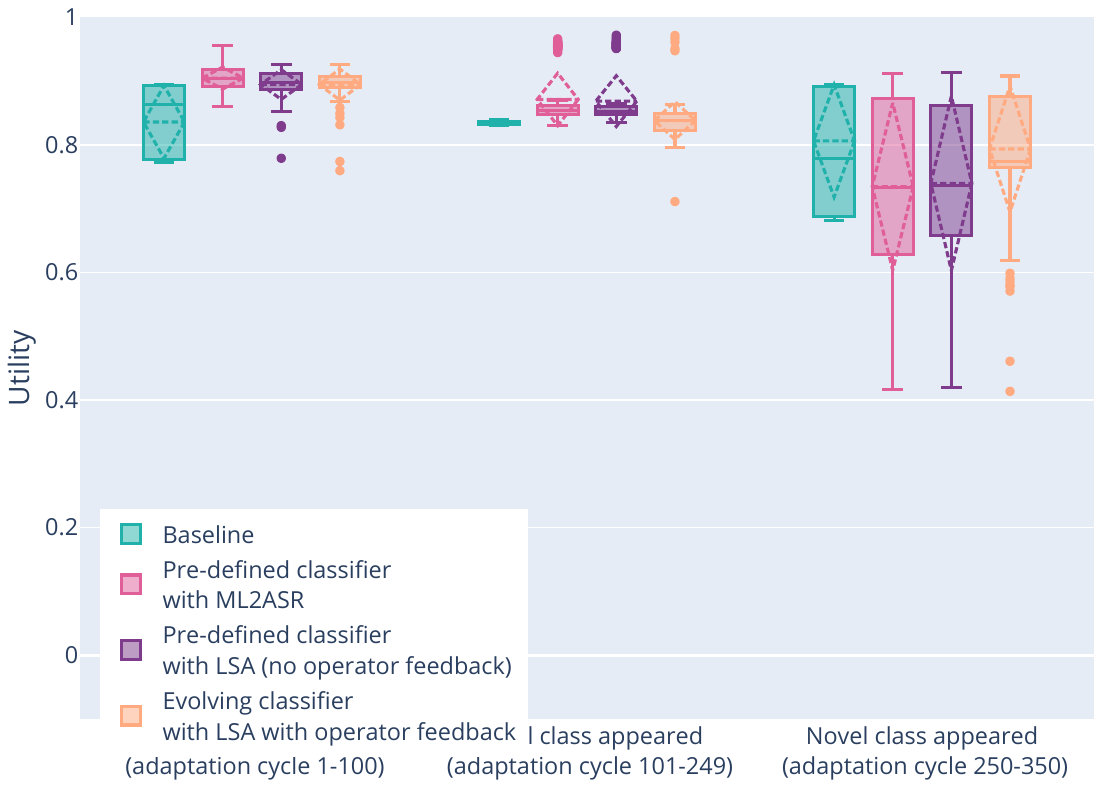}
		\caption{$\langle$(R), B, G$\rangle$}
		\label{fig: utility red blue green}
	\end{subfigure}
	\bigskip
	
	\begin{subfigure}[b]{0.32\textwidth}
		\includegraphics[width=\textwidth]{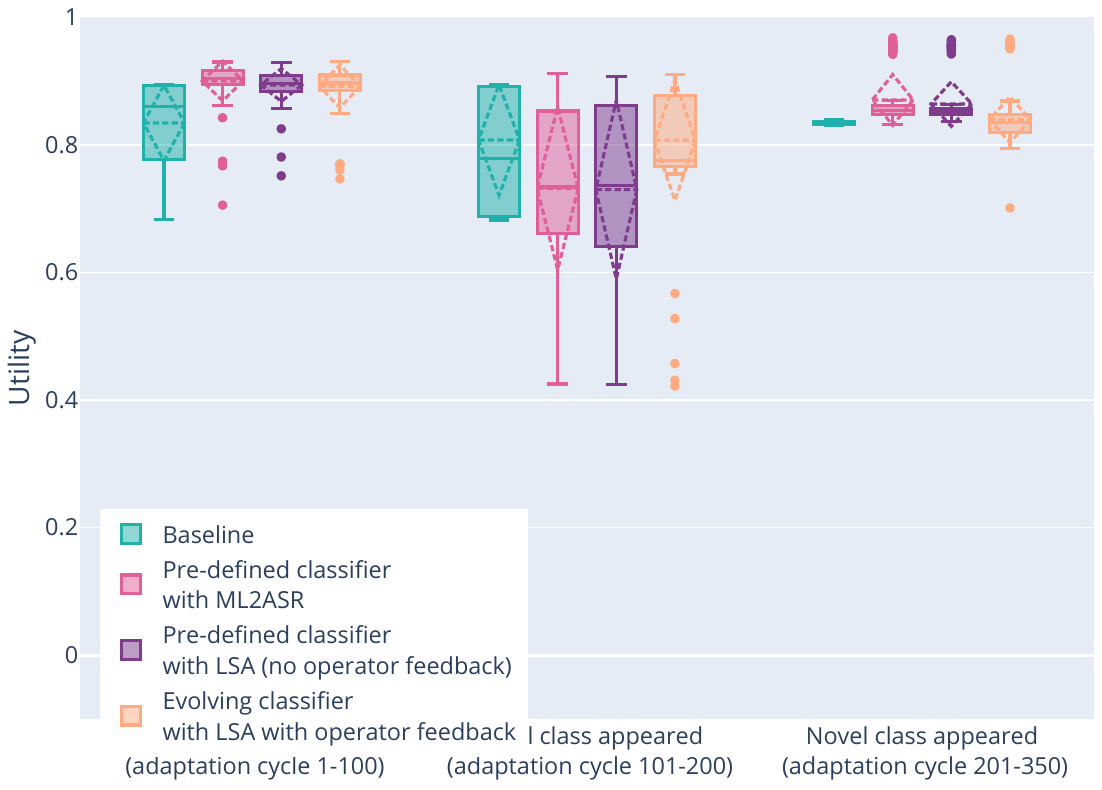}
		\caption{$\langle$(R), G, B$\rangle$}
		\label{fig: utility red green blue}
	\end{subfigure}
	\begin{subfigure}[b]{0.32\textwidth}
		\includegraphics[width=\textwidth]{figures/scenarios_1/utility/utility_problem_quality_attributes_dist_7.pdf}
		\caption{$\langle$(B, R), G$\rangle$}
		\label{fig: utility (blue red) green}
	\end{subfigure}
	\begin{subfigure}[b]{0.32\textwidth}
		\includegraphics[width=\textwidth]{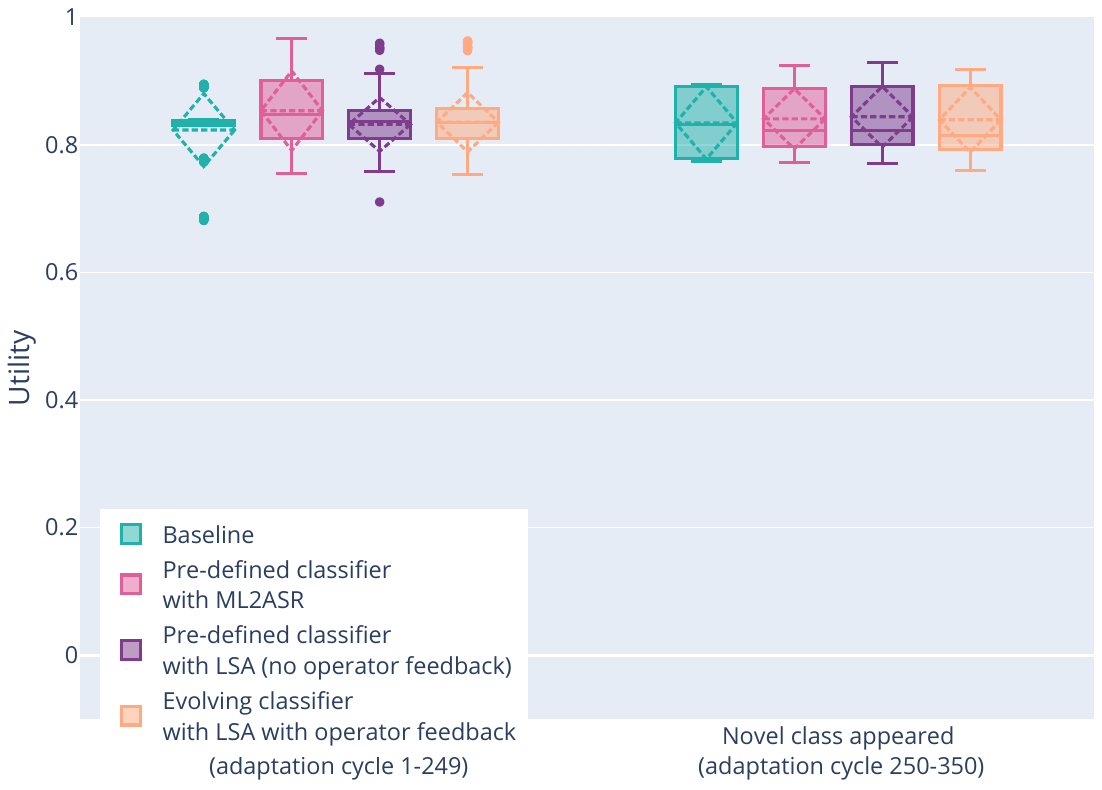}
		\caption{$\langle$(B, G), R$\rangle$}
		\label{fig: utility (blue green) red}
	\end{subfigure}
	\caption{In terms of utility, evaluation of the lifelong self-adaptation  with and without the feedback of the stakeholder by comparing with the state-of-the-art (the pre-defined classifier supported by ML2ASR), and the baseline. The preference of the stakeholder is here $\langle$``less packet loss'', ``less energy consumption''$\rangle$ (by weight of 0.8 and 0.2).  The appearance order of classes corresponding to each plot is mentioned in its caption.}\vspace{-10pt}
	\label{fig:all compare performance on utility}
\end{figure}

\begin{figure}[htbp]
	\centering
	\begin{subfigure}[b]{0.32\textwidth}
		\includegraphics[width=\textwidth]{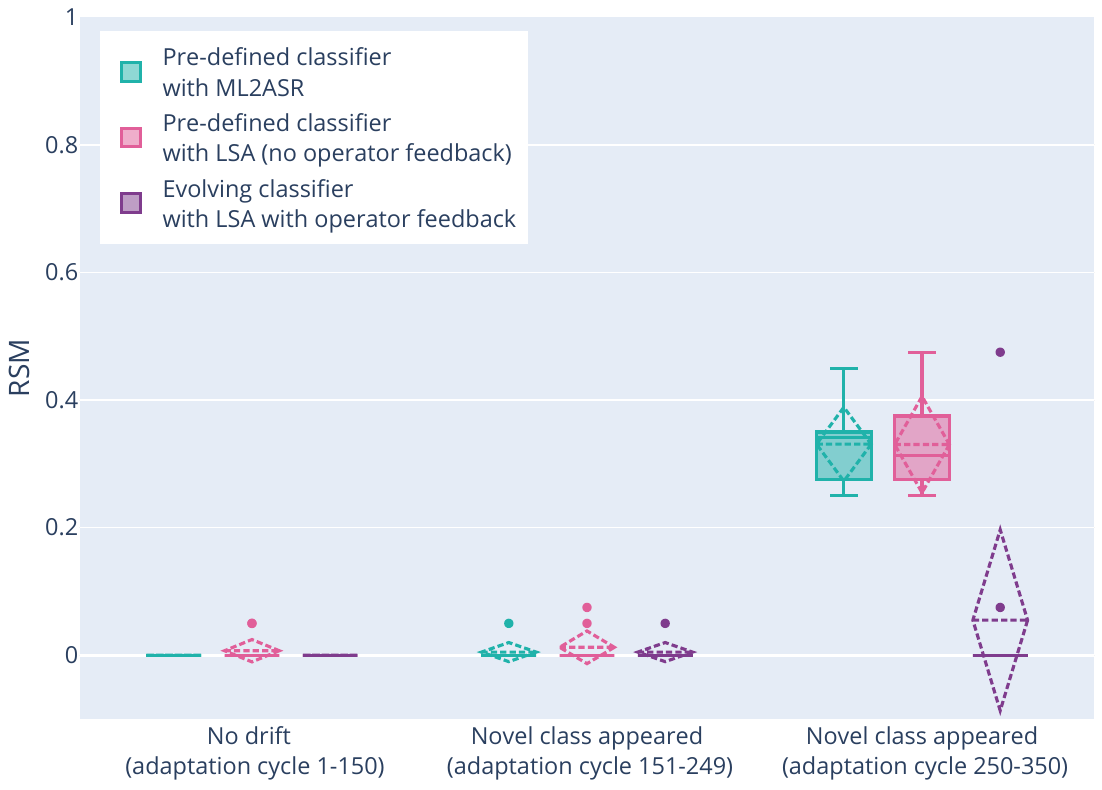}
		\caption{$\langle$(B), R, G$\rangle$}
		\label{fig: rsm blue red green}
	\end{subfigure}
	\begin{subfigure}[b]{0.32\textwidth}
		\includegraphics[width=\textwidth]{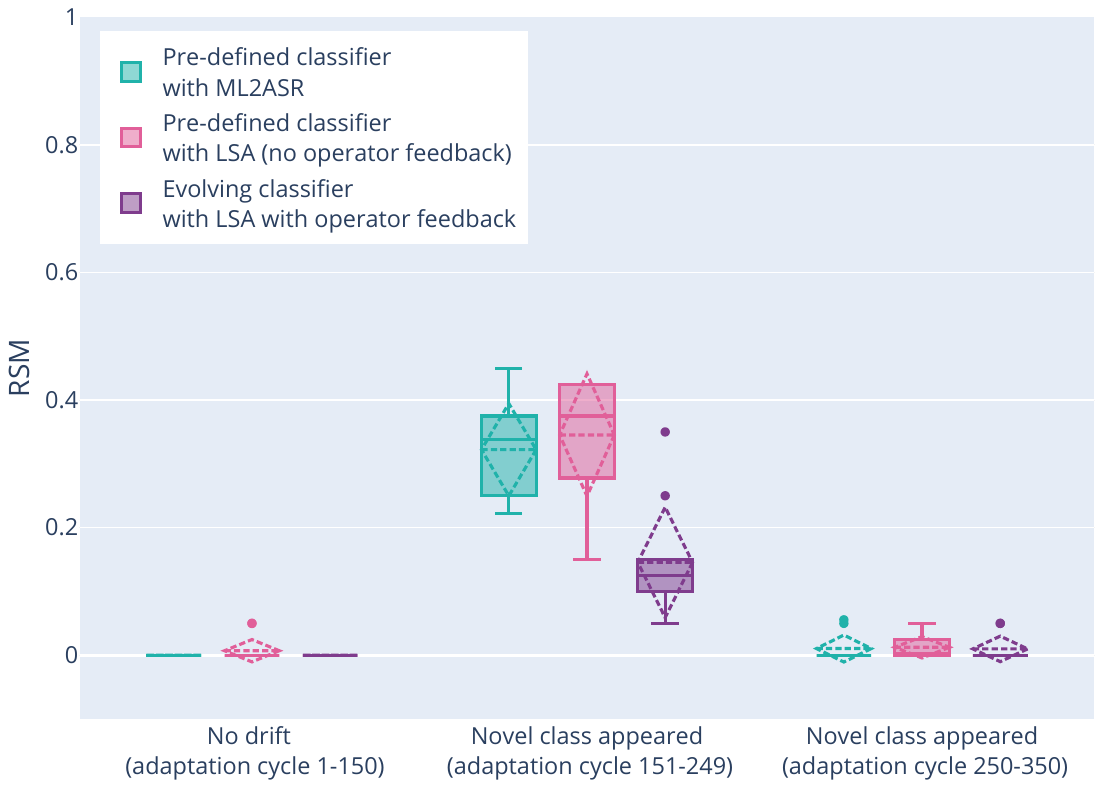}
		\caption{$\langle$(B), G, R$\rangle$}
		\label{fig: rsm blue green red}
	\end{subfigure}
	\begin{subfigure}[b]{0.32\textwidth}
		\includegraphics[width=\textwidth]{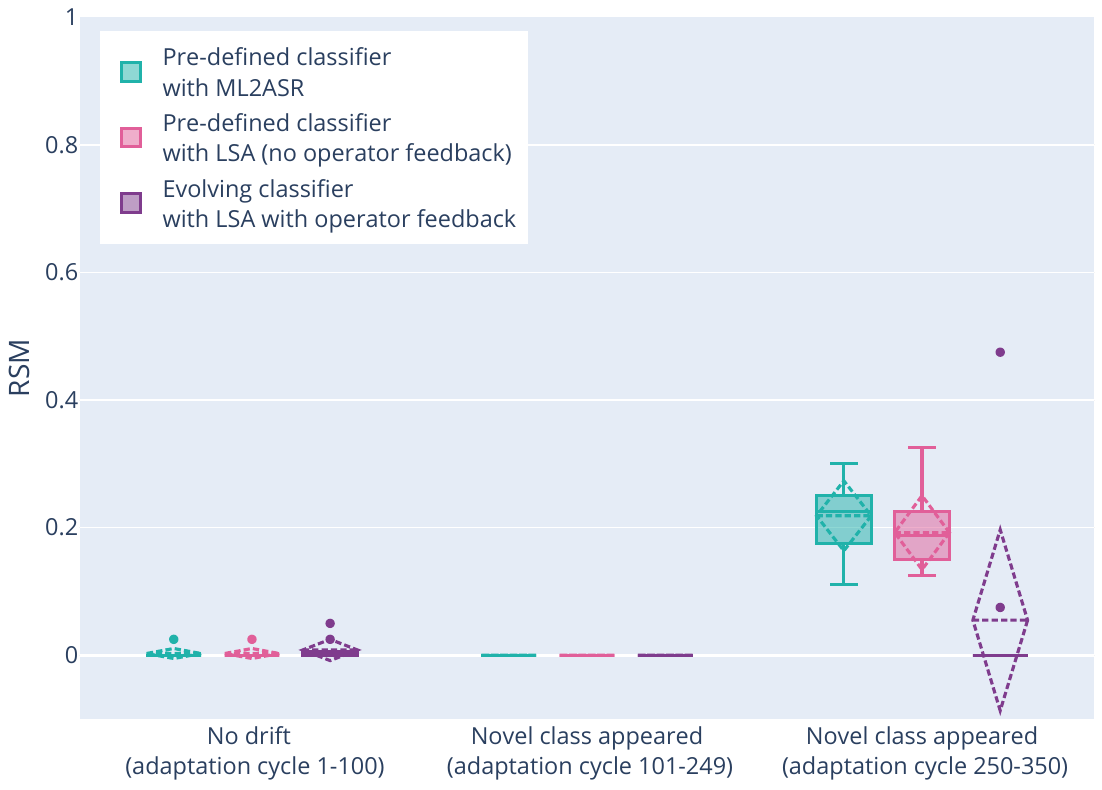}
		\caption{$\langle$(R), B, G$\rangle$}
		\label{fig: rsm red blue green}
	\end{subfigure}
	\bigskip
	
	\begin{subfigure}[b]{0.32\textwidth}
		\includegraphics[width=\textwidth]{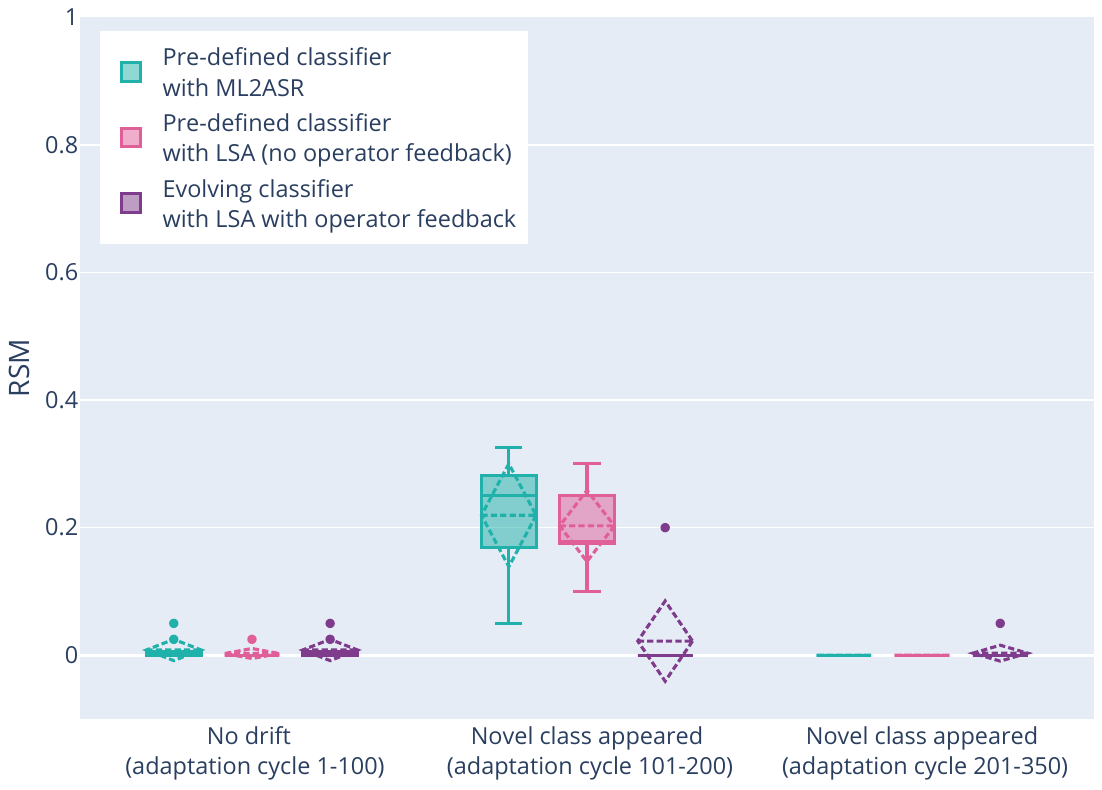}
		\caption{$\langle$(R), G, B$\rangle$}
		\label{fig: rsm red green blue}
	\end{subfigure}
	\begin{subfigure}[b]{0.32\textwidth}
		\includegraphics[width=\textwidth]{figures/scenarios_1/rsm/rsm_7.pdf}
		\caption{$\langle$(B, R), G$\rangle$}
		\label{fig: rsm (blue red) green}
	\end{subfigure}
	\begin{subfigure}[b]{0.32\textwidth}
		\includegraphics[width=\textwidth]{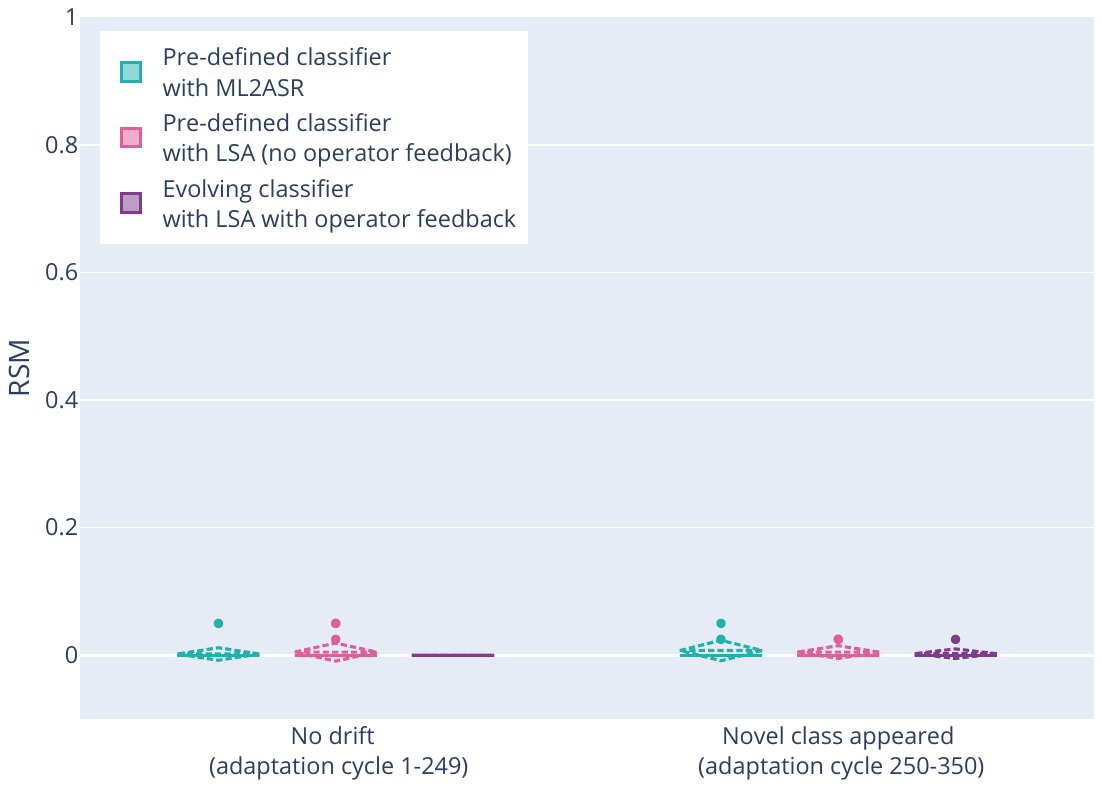}
		\caption{$\langle$(B, G), R$\rangle$}
		\label{fig: rsm (blue green) red}
	\end{subfigure}
	\caption{The impact of drift on the \rsm\ value. In each group, bars indicate the value of \rsm\ for: (i) the predefined classifier, (ii) the state-of-the-art (the pre-defined classifier supported by ML2ASR), (iii) lifelong self-adaptation without and (iv) with the feedback of the stakeholder, respectively from left to right. Also, the related class appearance order is determined in the caption of each figure. The preference of the stakeholder is here $\langle$``less packet loss'', ``less energy consumption''$\rangle$. }\vspace{-10pt}
	\label{fig:all compare performance on rsm}
\end{figure}

\begin{figure}[htbp]
	\centering
	\begin{subfigure}[b]{0.32\textwidth}
		\includegraphics[width=\textwidth]{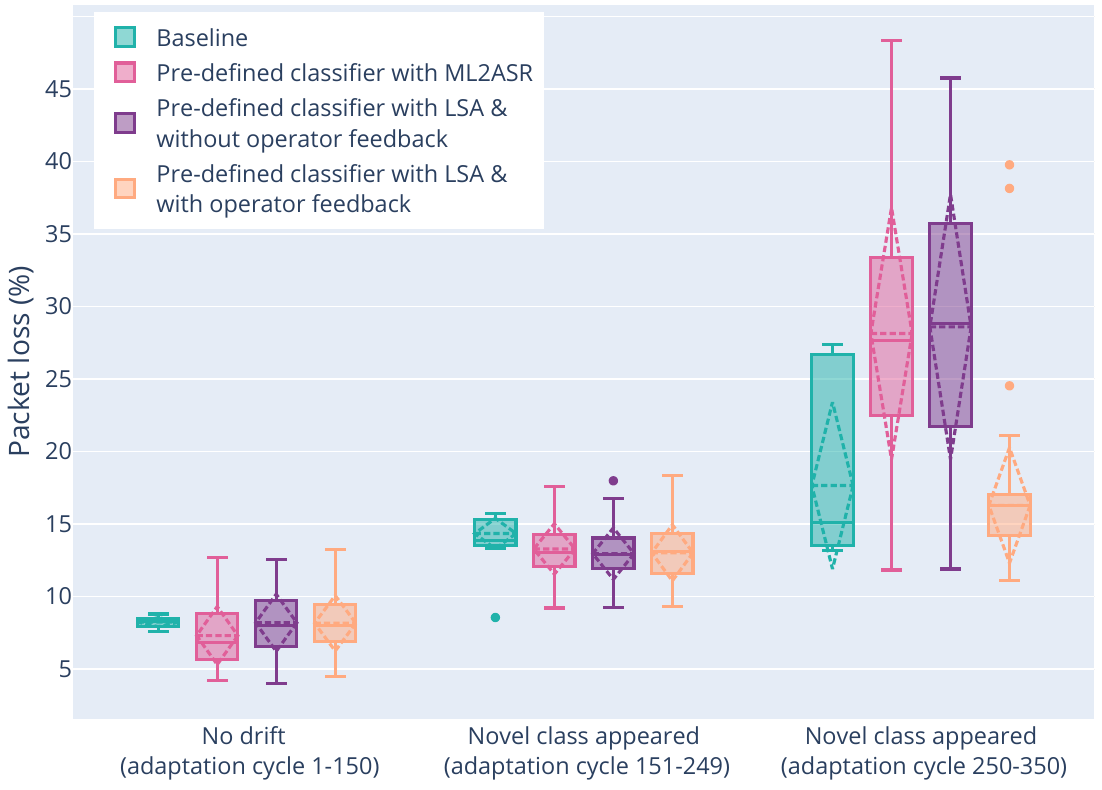}
		\caption{$\langle$(B), R, G$\rangle$}
		\label{fig: pl blue red green 2}
	\end{subfigure}
	\begin{subfigure}[b]{0.32\textwidth}
		\includegraphics[width=\textwidth]{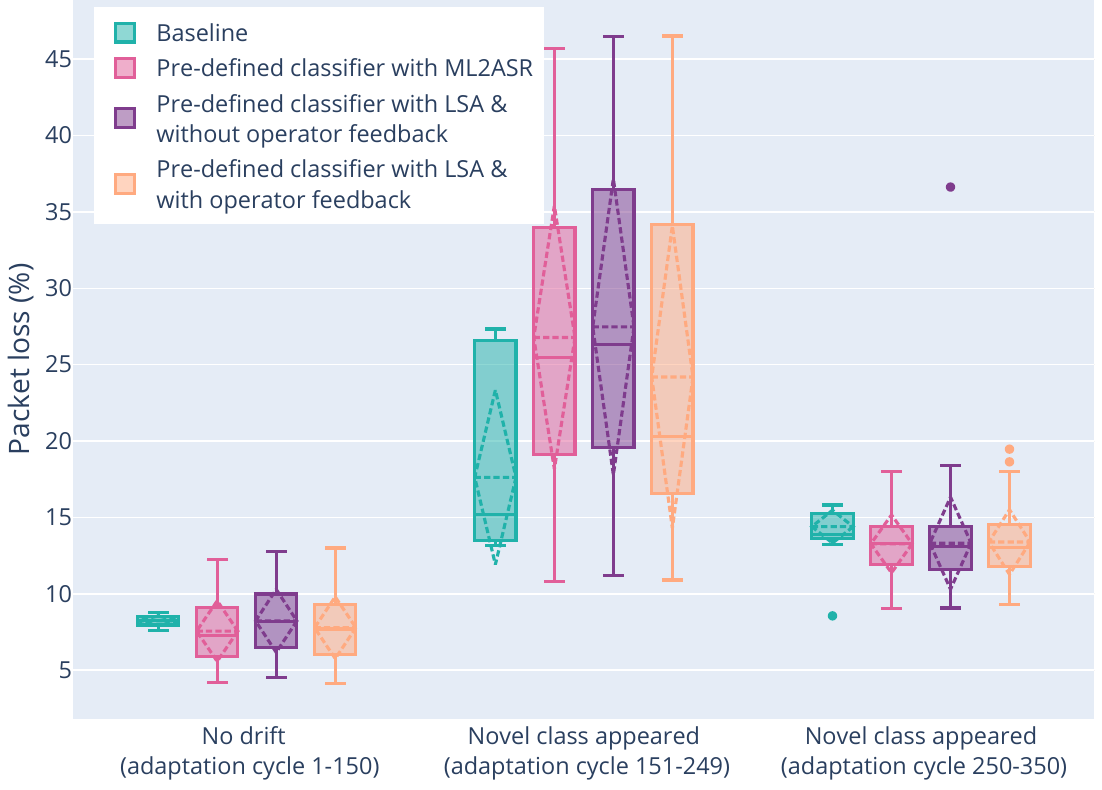}
		\caption{$\langle$(B), G, R$\rangle$}
		\label{fig: pl blue green red 2}
	\end{subfigure}
	\begin{subfigure}[b]{0.32\textwidth}
		\includegraphics[width=\textwidth]{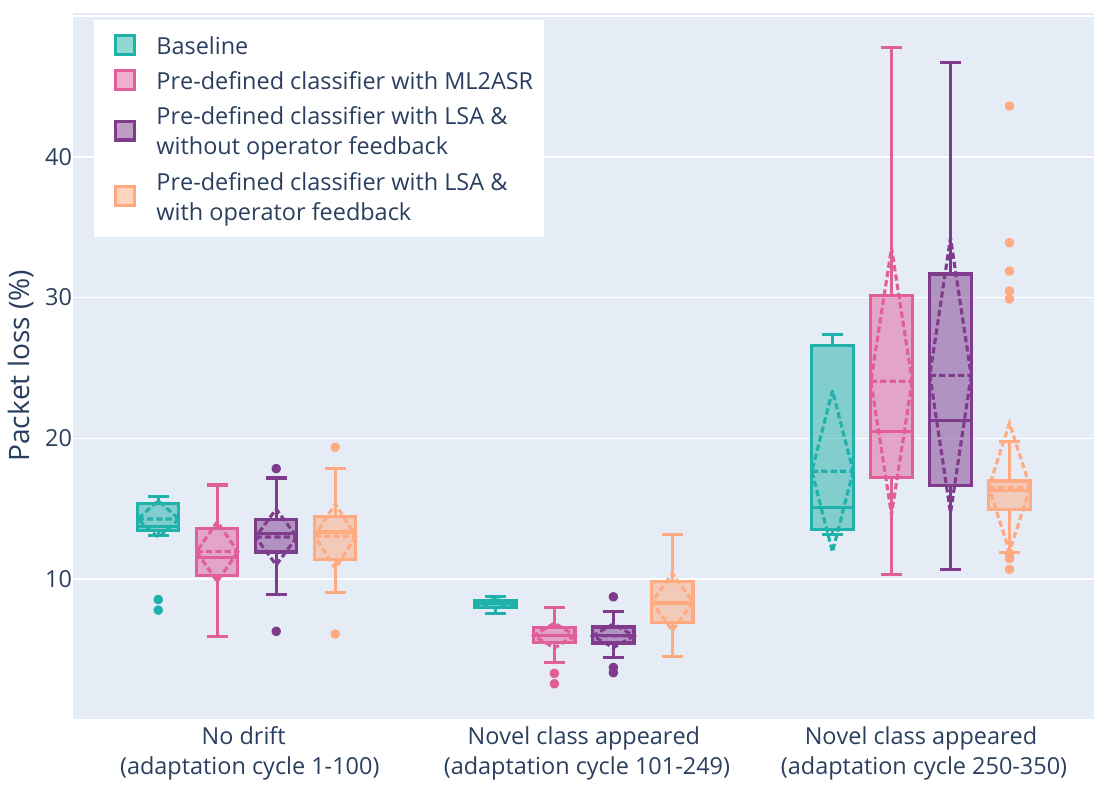}
		\caption{$\langle$(R), B, G$\rangle$}
		\label{fig: pl red blue green 2}
	\end{subfigure}
	\bigskip
	
	\begin{subfigure}[b]{0.32\textwidth}
		\includegraphics[width=\textwidth]{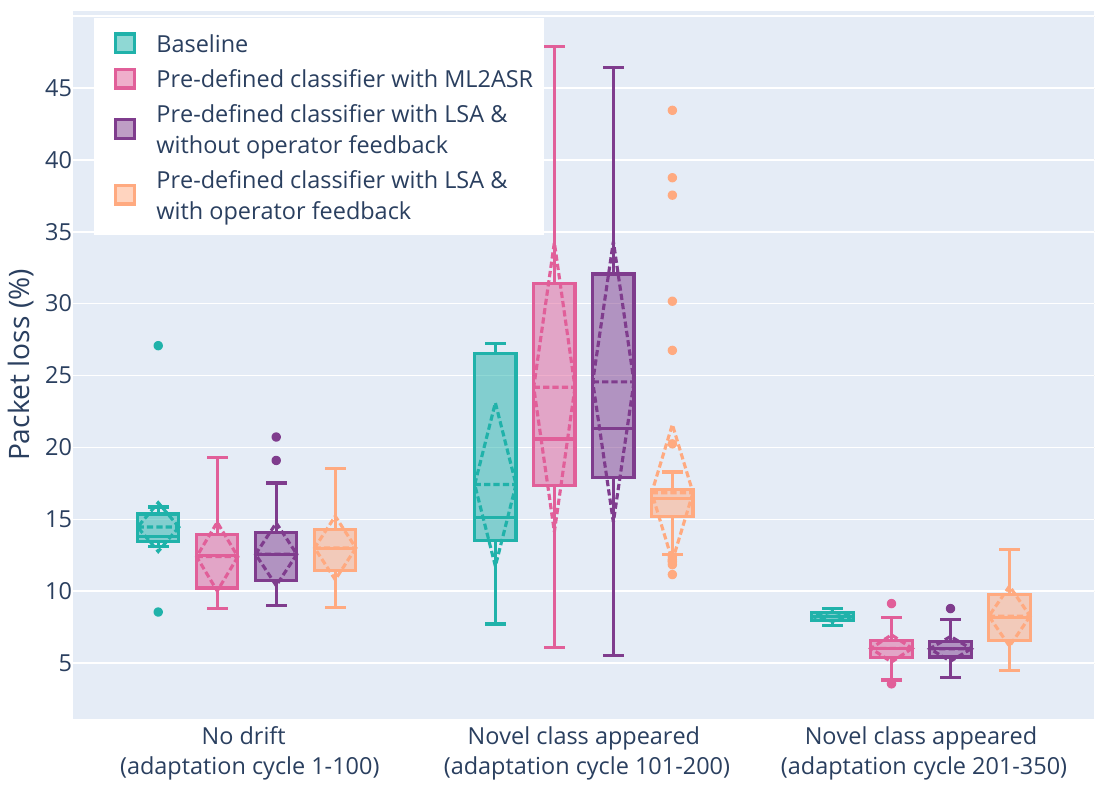}
		\caption{$\langle$(R), G, B$\rangle$}
		\label{fig: pl red green blue 2}
	\end{subfigure}
	\begin{subfigure}[b]{0.32\textwidth}
		\includegraphics[width=\textwidth]{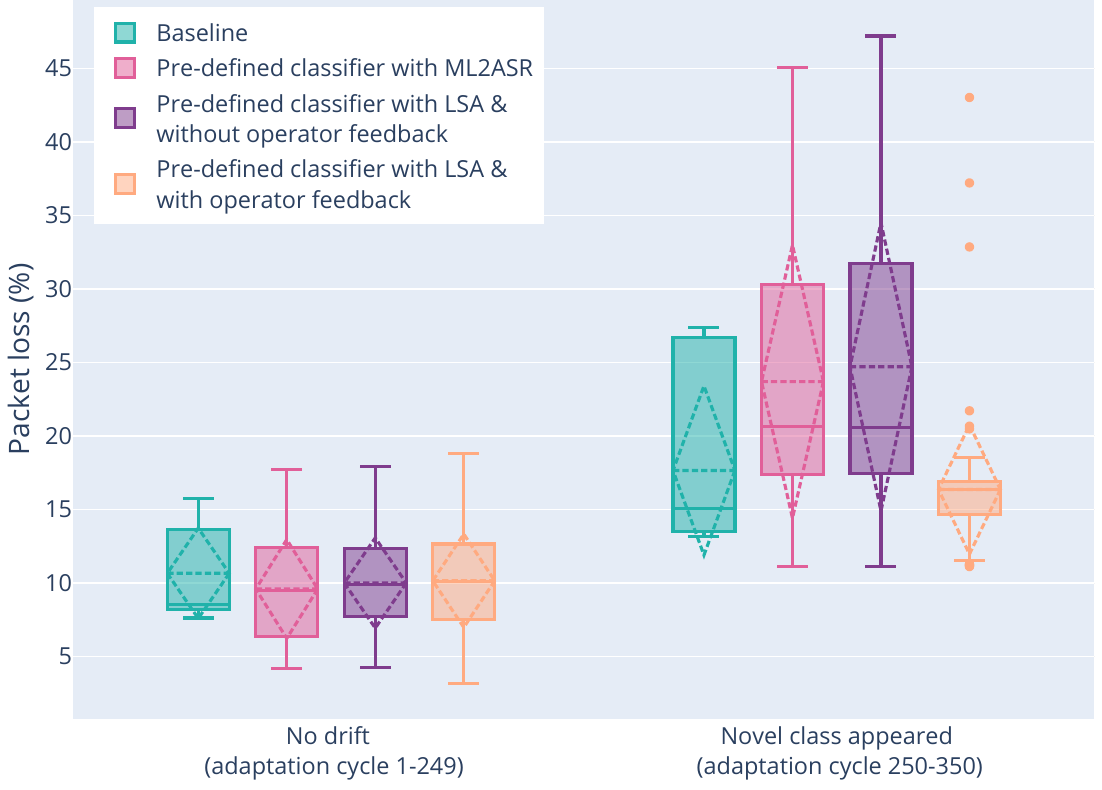}
		\caption{$\langle$(B, R), G$\rangle$}
		\label{fig: pl (blue red) green 2}
	\end{subfigure}
	\begin{subfigure}[b]{0.32\textwidth}
		\includegraphics[width=\textwidth]{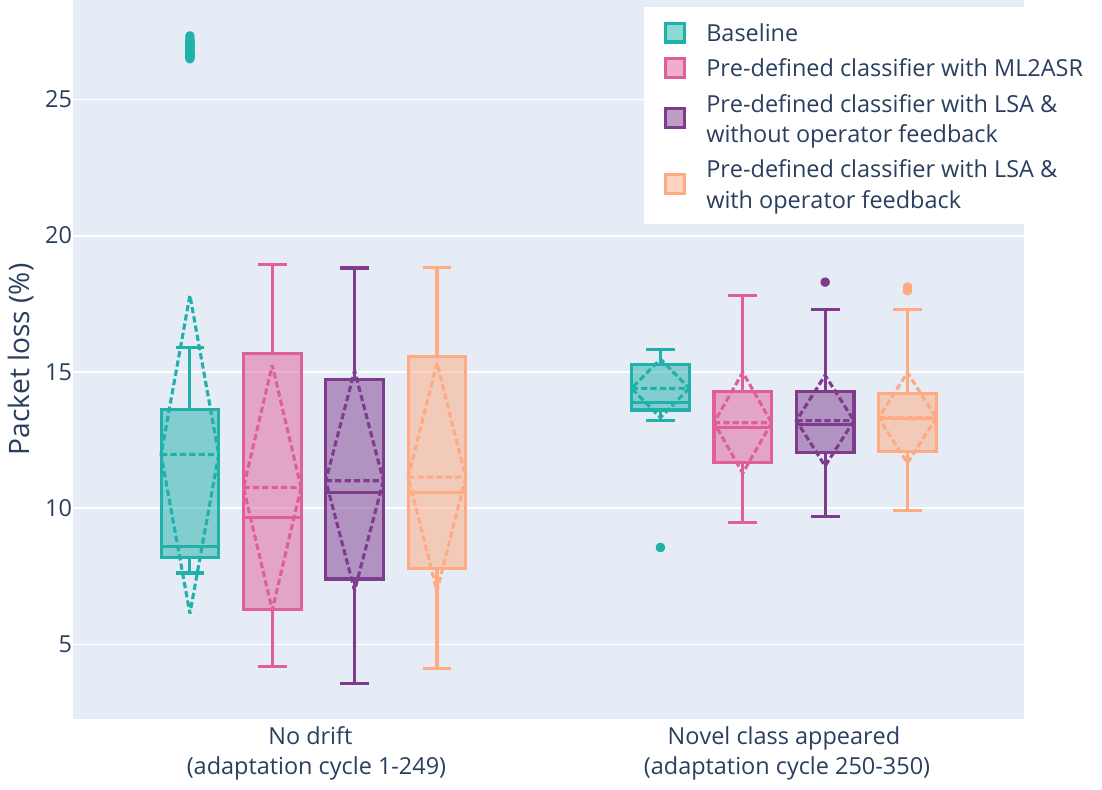}
		\caption{$\langle$(B, G), R$\rangle$}
		\label{fig: pl (blue green) red 2}
	\end{subfigure}
	\caption{In terms of packet loss, evaluation of the lifelong self-adaptation  with and without the feedback of the stakeholder by comparing with the state-of-the-art (the pre-defined classifier supported by ML2ASR), the pre-defined classifier, and the baseline. The preference of the stakeholder is here $\langle$``less energy consumption'', ``less packet loss''$\rangle$. The appearance order of classes corresponding to each plot is mentioned in its caption.}\vspace{-10pt}
	\label{fig:all compare performance on packet loss 2}
\end{figure}

\begin{figure}[htbp]
	\centering
	\begin{subfigure}[b]{0.32\textwidth}
		\includegraphics[width=\textwidth]{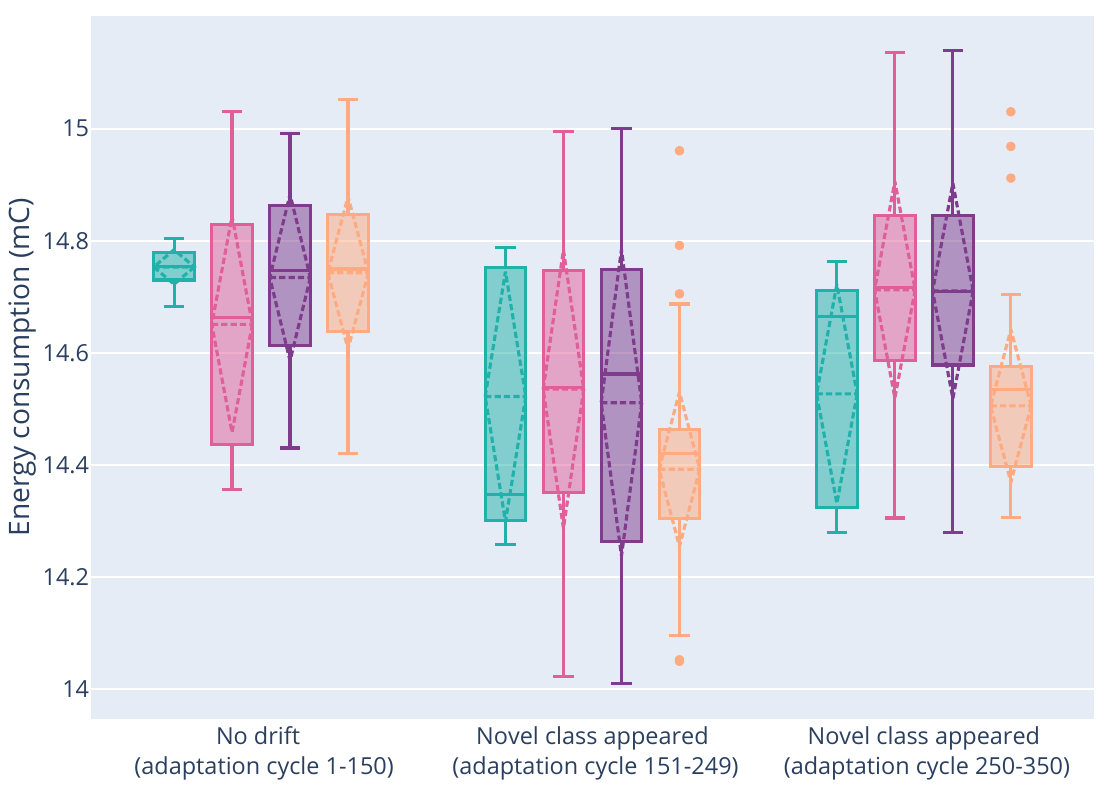}
		\caption{$\langle$(B), R, G$\rangle$}
		\label{fig: ec blue red green 2}
	\end{subfigure}
	\begin{subfigure}[b]{0.32\textwidth}
		\includegraphics[width=\textwidth]{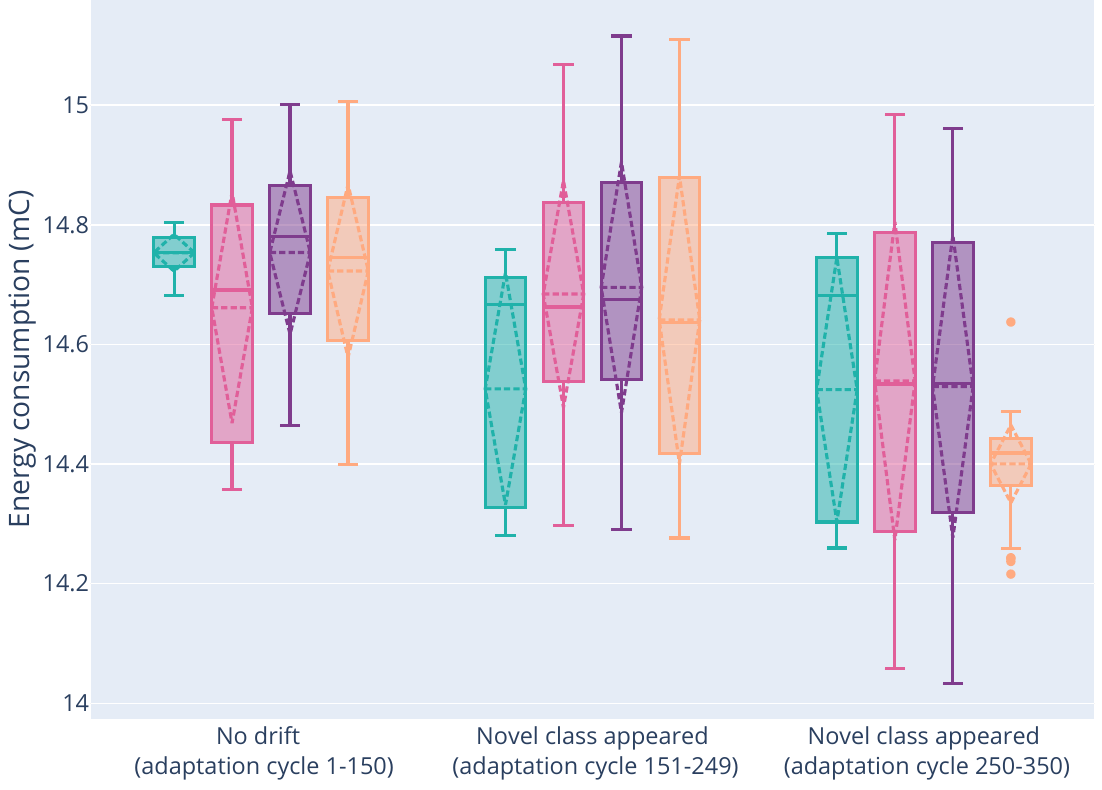}
		\caption{$\langle$(B), G, R$\rangle$}
		\label{fig: ec blue green red 2}
	\end{subfigure}
	\begin{subfigure}[b]{0.32\textwidth}
		\includegraphics[width=\textwidth]{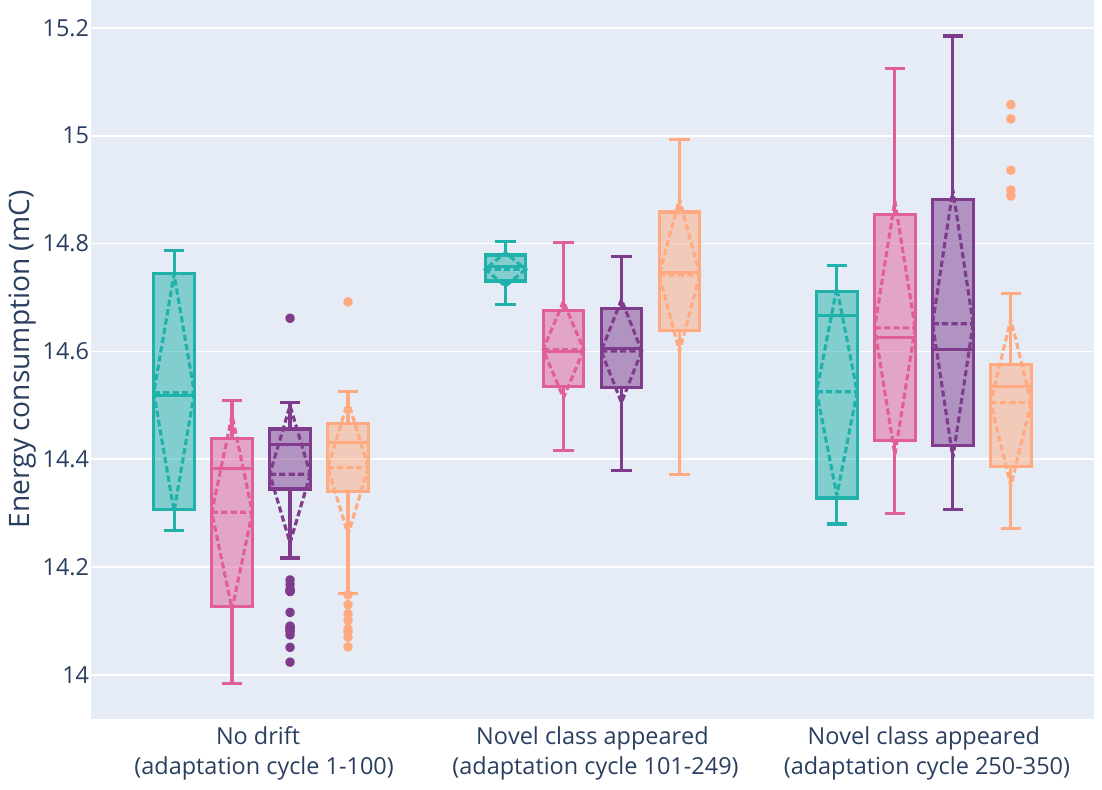}
		\caption{$\langle$(R), B, G$\rangle$}
		\label{fig: ec red blue green 2}
	\end{subfigure}
	\bigskip
	
	\begin{subfigure}[b]{0.32\textwidth}
		\includegraphics[width=\textwidth]{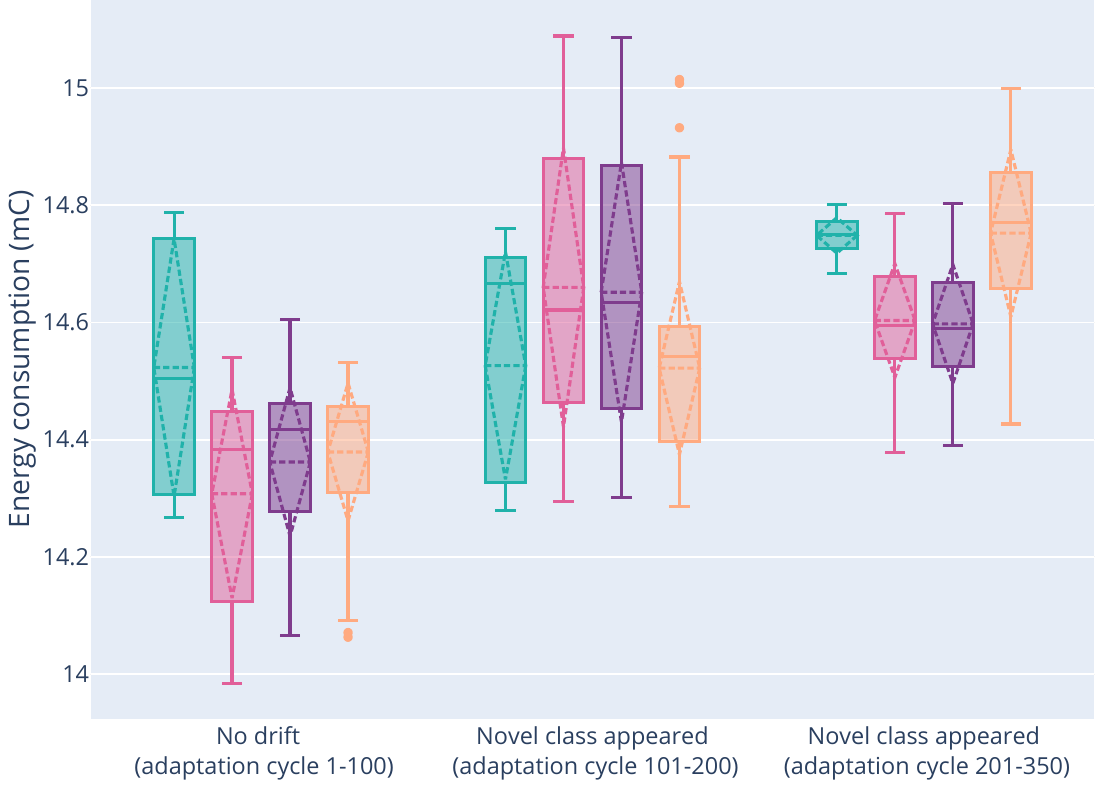}
		\caption{$\langle$(R), G, B$\rangle$}
		\label{fig: ec red green blue 2}
	\end{subfigure}
	\begin{subfigure}[b]{0.32\textwidth}
		\includegraphics[width=\textwidth]{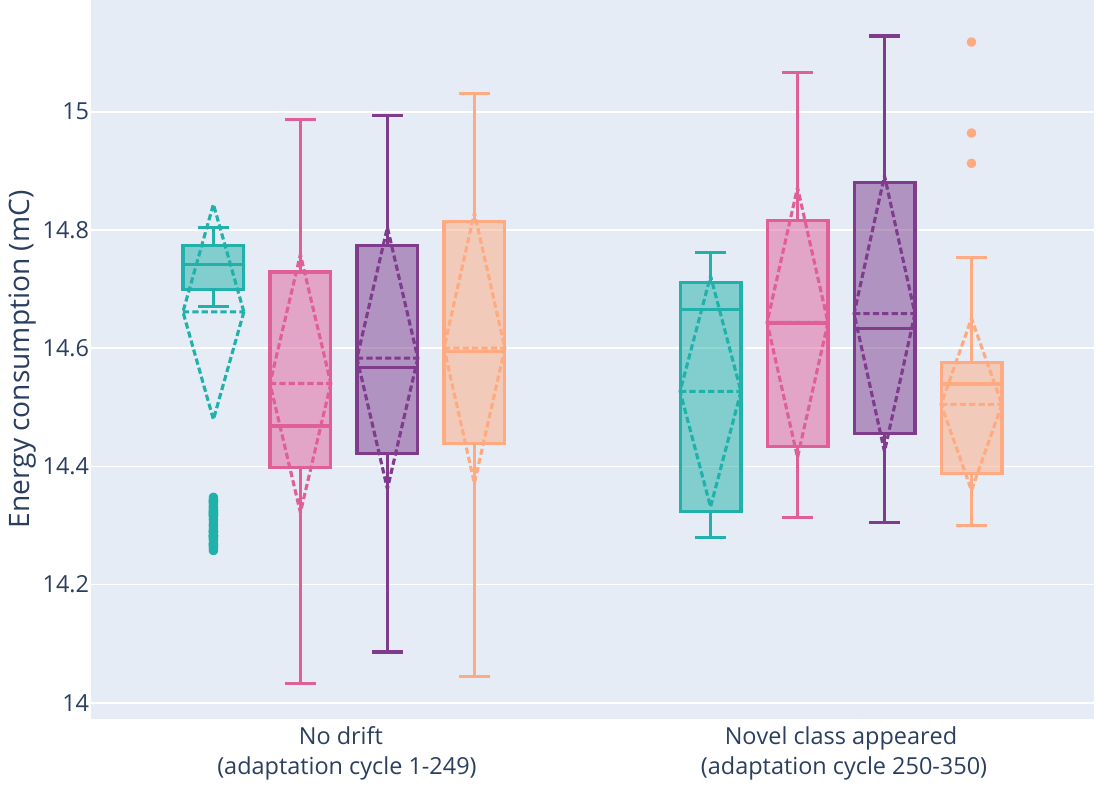}
		\caption{$\langle$(B, R), G$\rangle$}
		\label{fig: ec (blue red) green 2}
	\end{subfigure}
	\begin{subfigure}[b]{0.32\textwidth}
		\includegraphics[width=\textwidth]{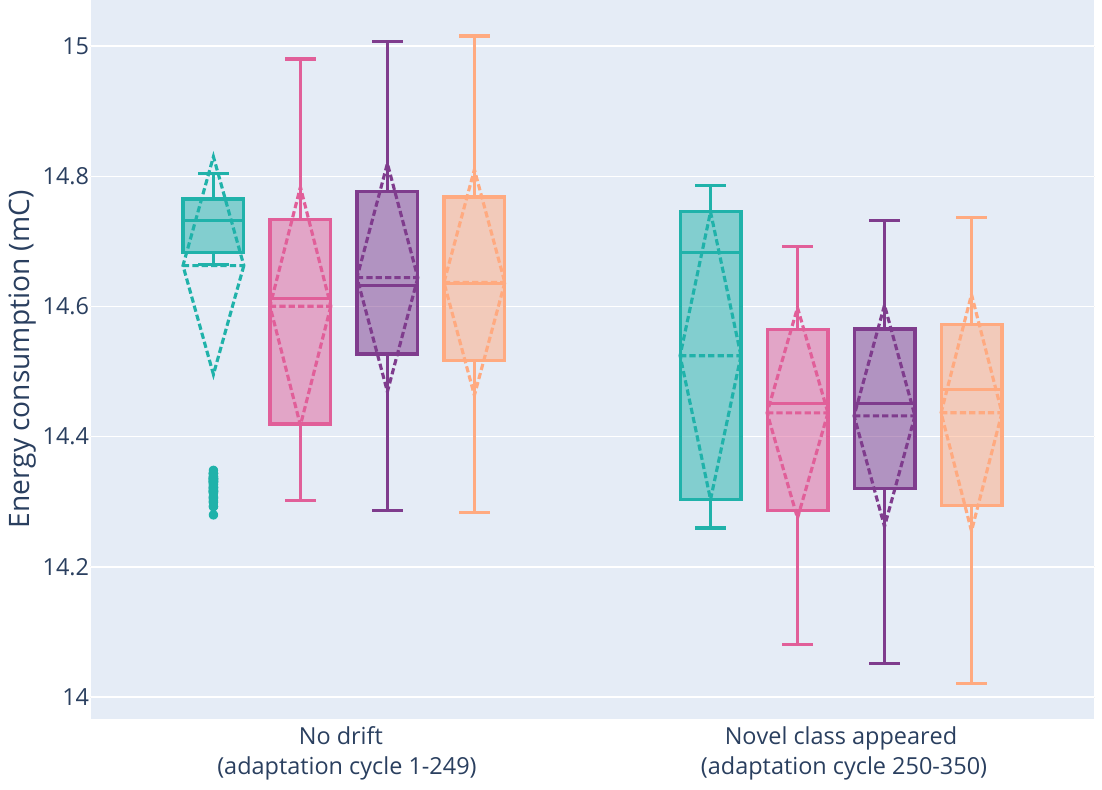}
		\caption{$\langle$(B, G), R$\rangle$}
		\label{fig: ec (blue green) red 2}
	\end{subfigure}
	\caption{In terms of energy consumption, evaluation of the lifelong self-adaptation  with and without the feedback of the stakeholder by comparing with the state-of-the-art (the pre-defined classifier supported by ML2ASR), the pre-defined classifier, and the baseline. The preference of the stakeholder is here $\langle$``less energy consumption'', ``less packet loss''$\rangle$.  The appearance order of classes corresponding to each plot is mentioned in its caption.}\vspace{-10pt}
	\label{fig:all compare performance on energy consumption 2}
\end{figure}

\begin{figure}[htbp]
	\centering
	\begin{subfigure}[b]{0.32\textwidth}
		\includegraphics[width=\textwidth]{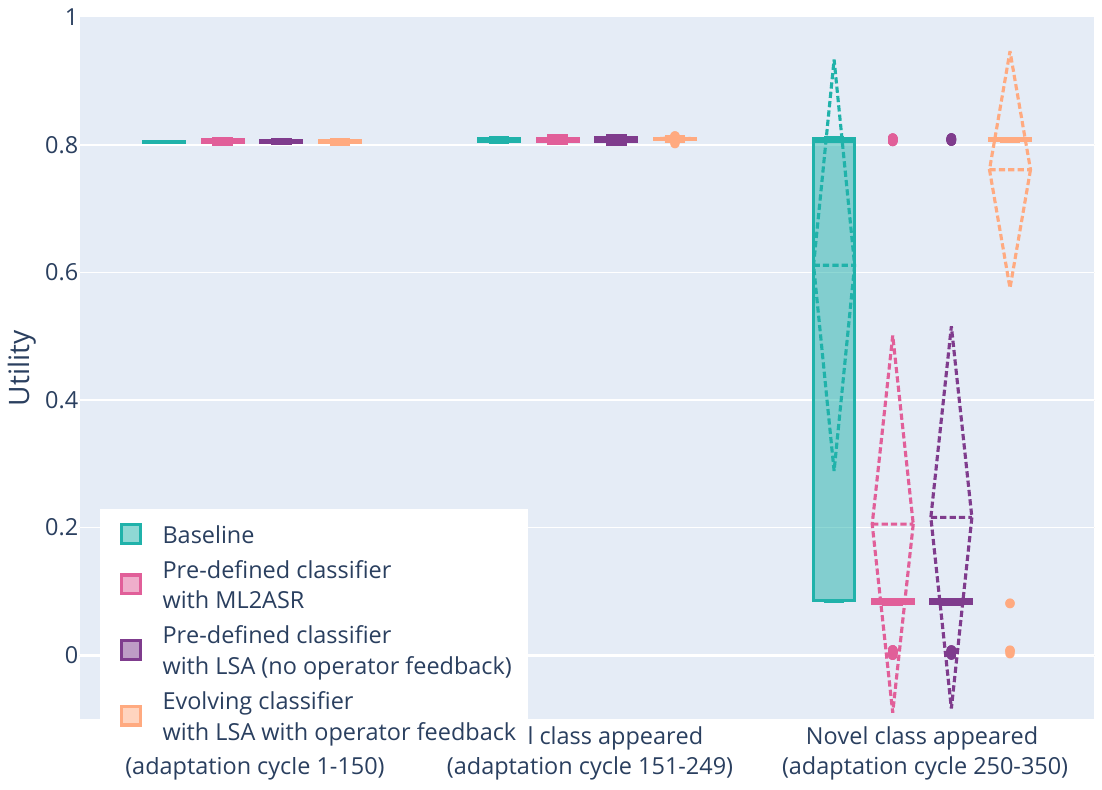}
		\caption{$\langle$(B), R, G$\rangle$}
		\label{fig: utility blue red green 2}
	\end{subfigure}
	\begin{subfigure}[b]{0.32\textwidth}
		\includegraphics[width=\textwidth]{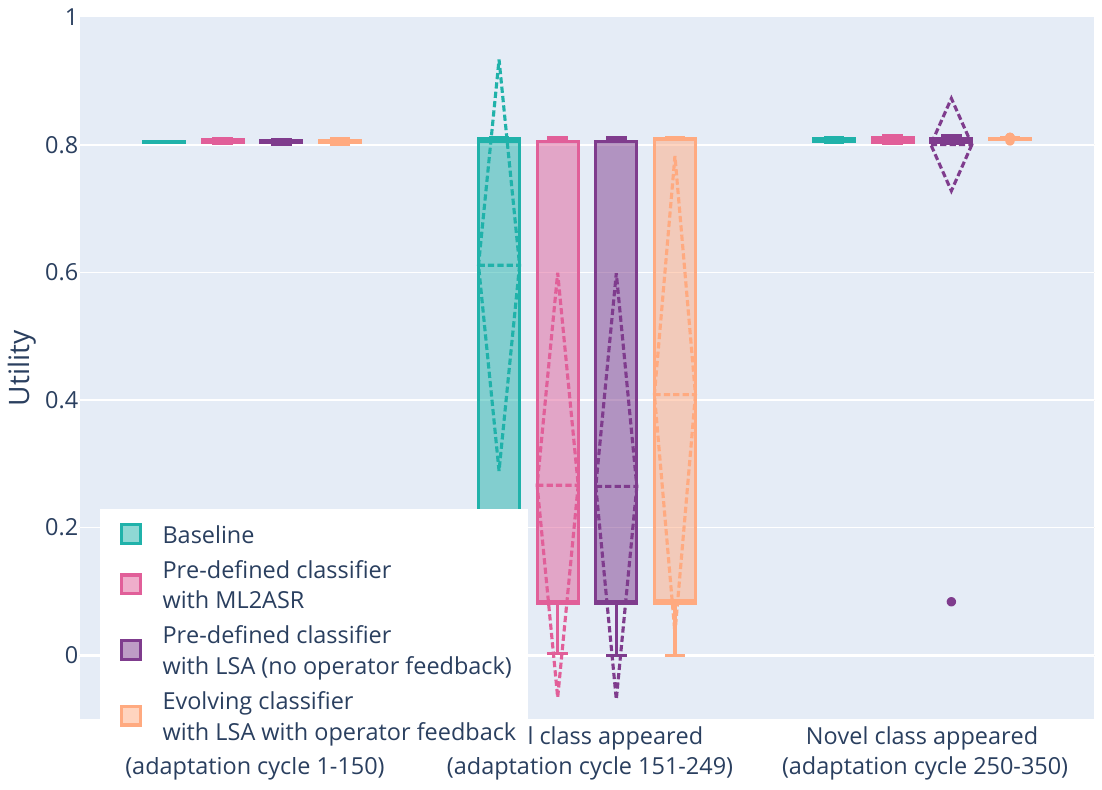}
		\caption{$\langle$(B), G, R$\rangle$}
		\label{fig: utility blue green red 2}
	\end{subfigure}
	\begin{subfigure}[b]{0.32\textwidth}
		\includegraphics[width=\textwidth]{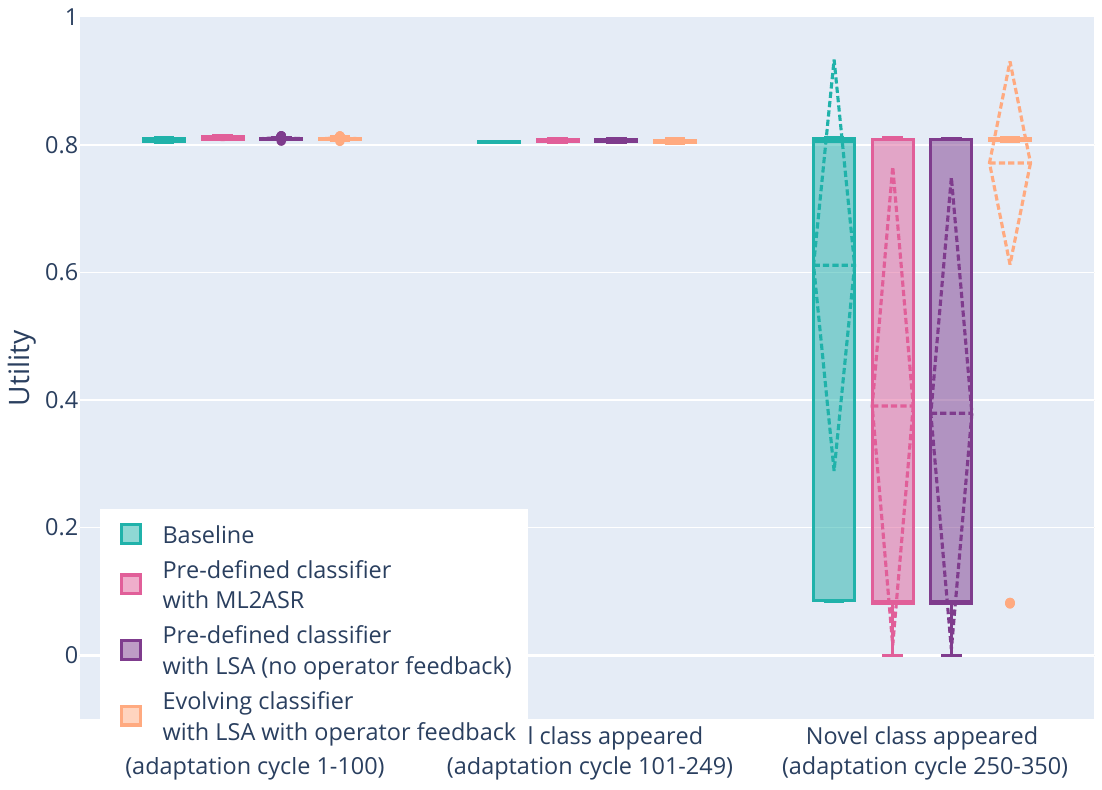}
		\caption{$\langle$(R), B, G$\rangle$}
		\label{fig: utility red blue green 2}
	\end{subfigure}
	\bigskip
	
	\begin{subfigure}[b]{0.32\textwidth}
		\includegraphics[width=\textwidth]{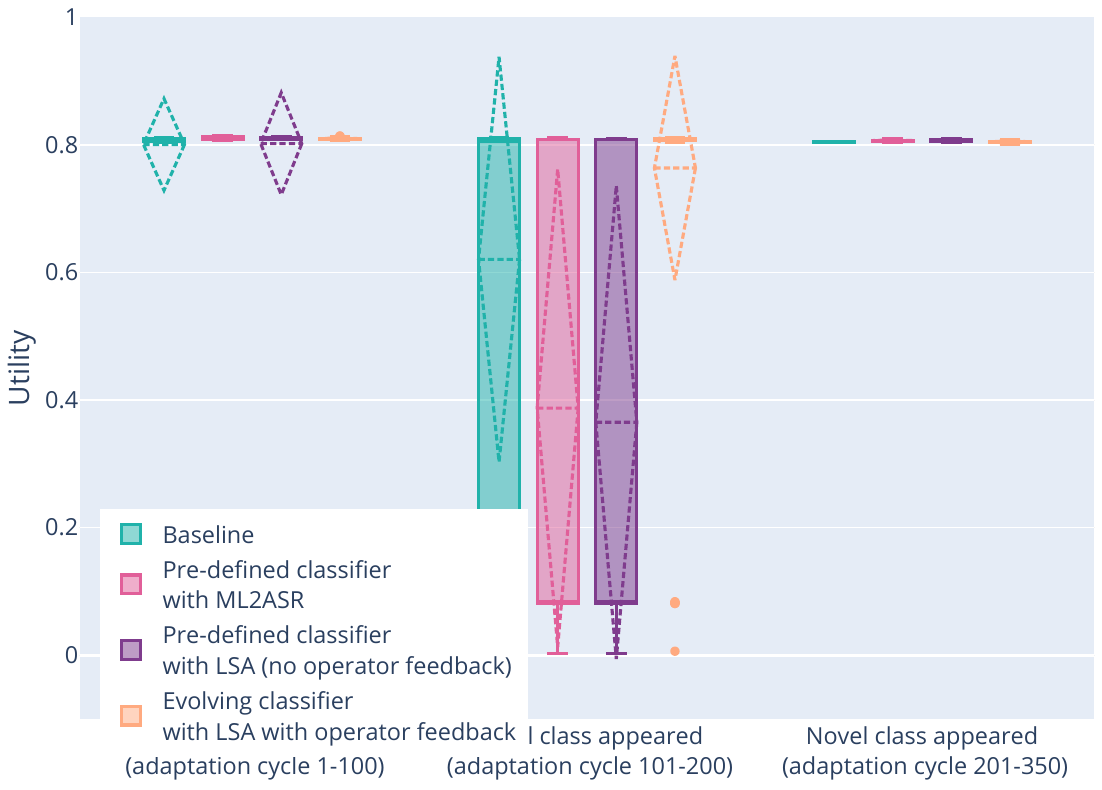}
		\caption{$\langle$(R), G, B$\rangle$}
		\label{fig: utility red green blue 2}
	\end{subfigure}
	\begin{subfigure}[b]{0.32\textwidth}
		\includegraphics[width=\textwidth]{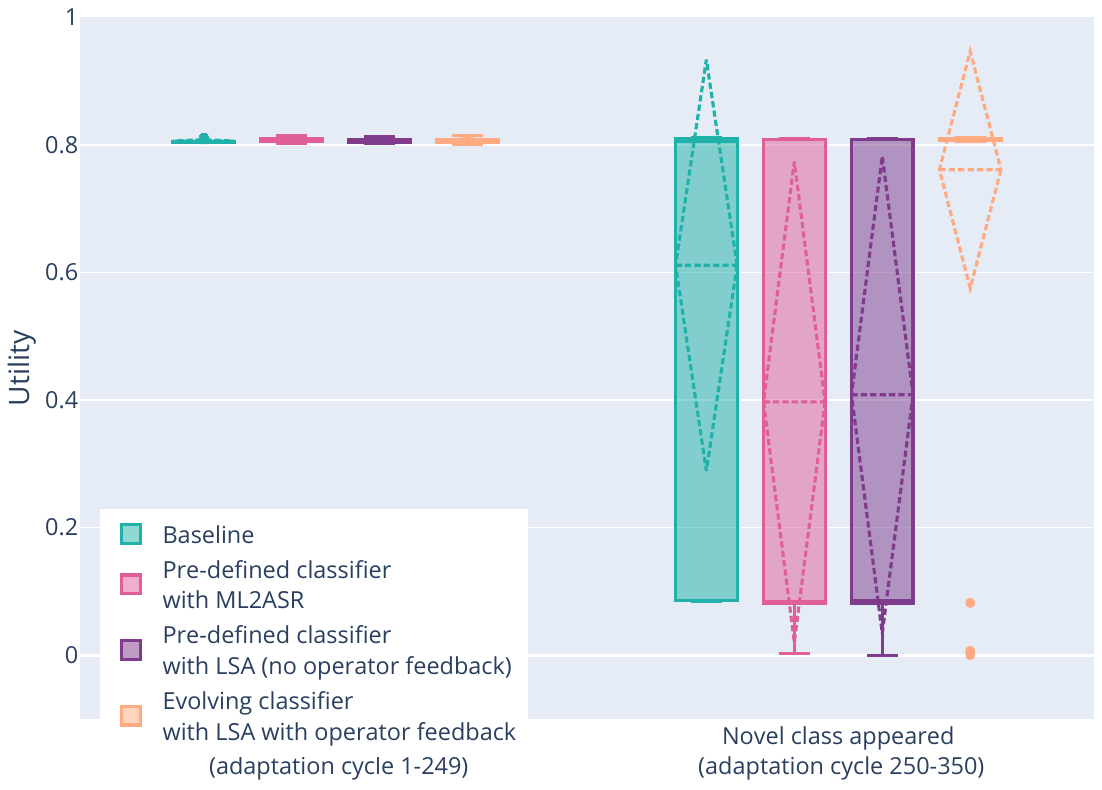}
		\caption{$\langle$(B, R), G$\rangle$}
		\label{fig: utility (blue red) green 2}
	\end{subfigure}
	\begin{subfigure}[b]{0.32\textwidth}
		\includegraphics[width=\textwidth]{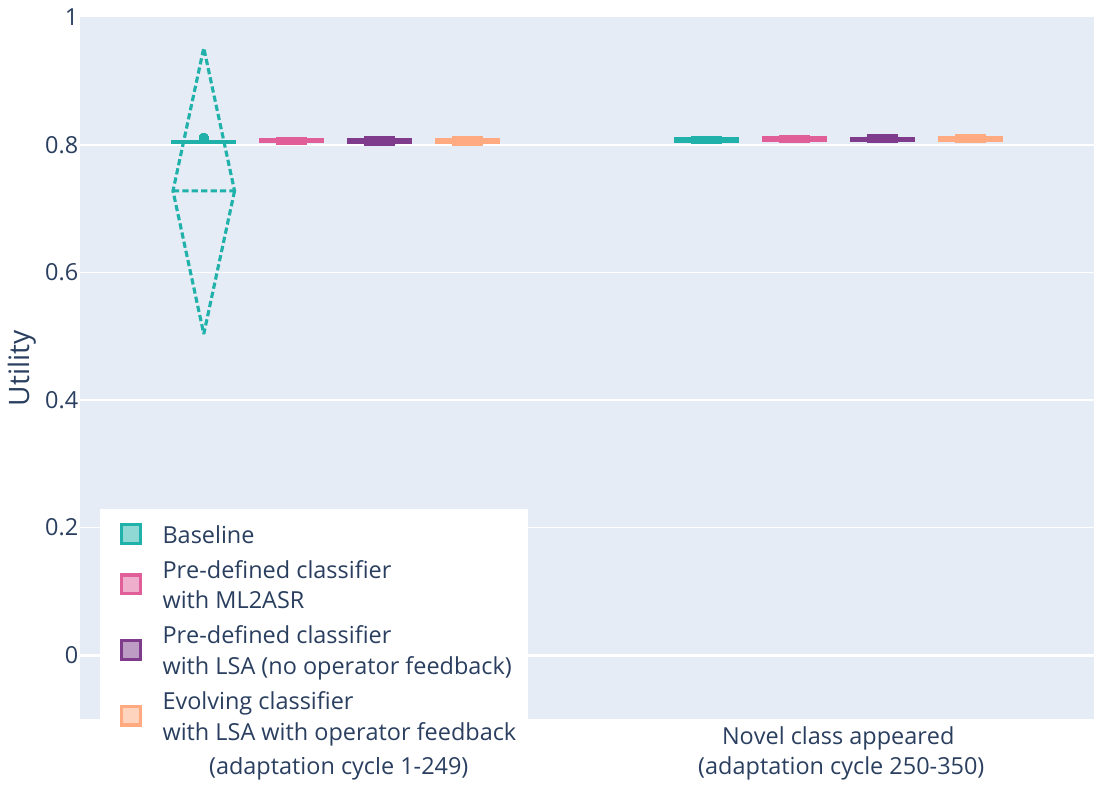}
		\caption{$\langle$(B, G), R$\rangle$}
		\label{fig: utility (blue green) red 2}
	\end{subfigure}
	\caption{In terms of utility, evaluation of the lifelong self-adaptation  with and without the feedback of the stakeholder by comparing with the state-of-the-art (the pre-defined classifier supported by ML2ASR), and the baseline. The preference of the stakeholder is here $\langle$``less energy consumption'', ``less packet loss''$\rangle$ (by weight of 0.8 and 0.2).  The appearance order of classes corresponding to each plot is mentioned in its caption.}\vspace{-10pt}
	\label{fig:all compare performance on utility 2}
\end{figure}

\begin{figure}[htbp]
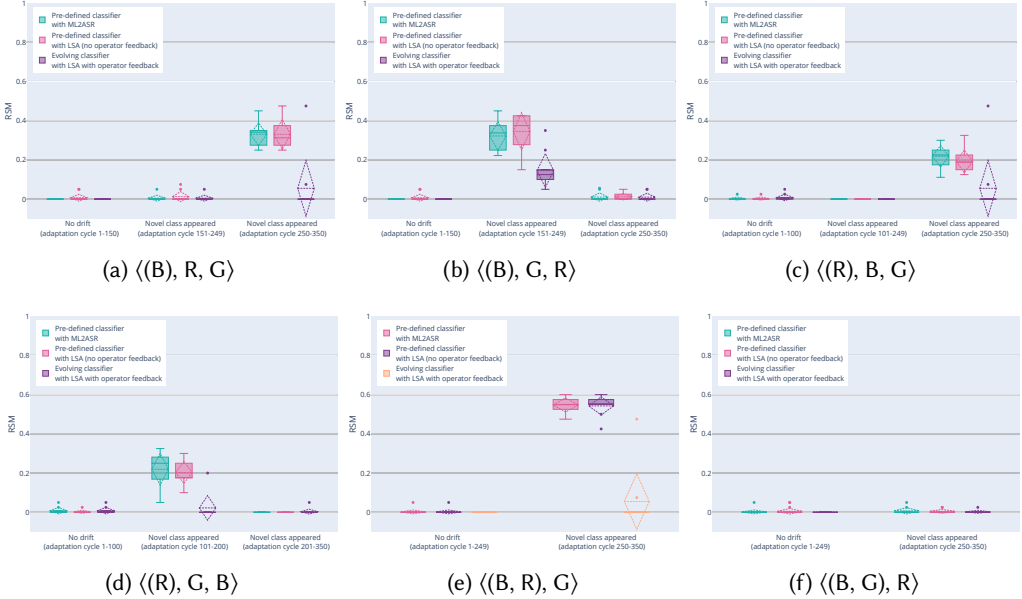

	\centering
	\begin{subfigure}[b]{0.32\textwidth}
		\includegraphics[width=\textwidth]{figures/scenarios_1/rsm/rsm_1.pdf}
		\caption{$\langle$(B), R, G$\rangle$}
		\label{fig: rsm blue red green 2}
	\end{subfigure}
	\begin{subfigure}[b]{0.32\textwidth}
		\includegraphics[width=\textwidth]{figures/scenarios_1/rsm/rsm_2.pdf}
		\caption{$\langle$(B), G, R$\rangle$}
		\label{fig: rsm blue green red 2}
	\end{subfigure}
	\begin{subfigure}[b]{0.32\textwidth}
		\includegraphics[width=\textwidth]{figures/scenarios_1/rsm/rsm_3.pdf}
		\caption{$\langle$(R), B, G$\rangle$}
		\label{fig: rsm red blue green 2}
	\end{subfigure}
	\bigskip
	
	\begin{subfigure}[b]{0.32\textwidth}
		\includegraphics[width=\textwidth]{figures/scenarios_1/rsm/rsm_4.pdf}
		\caption{$\langle$(R), G, B$\rangle$}
		\label{fig: rsm red green blue 2}
	\end{subfigure}
	\begin{subfigure}[b]{0.32\textwidth}
		\includegraphics[width=\textwidth]{figures/scenarios_1/rsm/rsm_7.pdf}
		\caption{$\langle$(B, R), G$\rangle$}
		\label{fig: rsm (blue red) green 2}
	\end{subfigure}
	\begin{subfigure}[b]{0.32\textwidth}
		\includegraphics[width=\textwidth]{figures/scenarios_1/rsm/rsm_8.pdf}
		\caption{$\langle$(B, G), R$\rangle$}
		\label{fig: rsm (blue green) red rsm 2}
	\end{subfigure}
	
	\caption{The impact of drift on the \rsm\ value. In each group, bars indicate the value of \rsm\ for: (i) the predefined classifier, (ii) the state-of-the-art (the pre-defined classifier supported by ML2ASR), (iii) lifelong self-adaptation without and (iv) with the feedback of the stakeholder, respectively from left to right. Also, the related class appearance order is determined in the caption of each figure. The preference of the stakeholder is here $\langle$``less energy consumption'', ``less packet loss''$\rangle$.}\vspace{-10pt}
	\label{fig:all compare performance on rsm 2}
\end{figure}

\re{
\section{Statistical Analysis of the Evaluation Results}
\label{sec: appendix statistical test results}
}
We present the statistical analysis of the evaluation results presented in the main text of the paper. The text is structured based on the figures of the evaluation as they appear in the main text. 

In the following, $\mu_{i,j,k}$ represents the value under test of an approach, with $i$ representing the related subfigure with possible values $\{a,b,c,d,e,f\}$ (his set can be reduced to a subset depending on the number of subfigures of the corresponding figure - e.g., $\{a,b\}$ - or $i$ can be omitted when there are no subfigures, we then write $\mu_{j,k}$), $j$ represents the metric value with possible values $\{r,u\}$, with $r$ representing \rsm\ and $u$ representing the utility, and $k$ represents the actual approach with possible values ``Baseline'', ``ML2ASR'', ``LSA'', and ``LSA with operator feedback''. 

\subsection{Quality Properties of Lifelong Self-Adaptation (Figure~\ref{fig:compare performance on quality attributes})}
Figure~\ref{fig:compare performance on quality attributes} shows the quality properties of lifelong self-adaptation compared to the other approaches for the base scenario. Since the different approaches optimize for utility we cannot define hypotheses for individual quality properties. Hence, we only present basic statistics in Table~\ref{tab:19a} and Table~\ref{tab:19b}.

\begin{table}[h]
\caption{Statistics corresponding to Figure~\ref{fig:packet loss distribution over adaptation cycles}: packet loss in the base scenario}\label{tab:19a}
\begin{adjustbox}{max width=\textwidth}
\begin{tabular}{cl|lll|lll}
\multicolumn{1}{l}{}                           &                   & \multicolumn{3}{c|}{No drift}                                                                   & \multicolumn{3}{c}{Novel class appeared}                                                      \\ \cline{3-8} 
\multicolumn{1}{l}{}                           &                   & \multicolumn{1}{c}{Median} & \multicolumn{1}{c}{Mean} & \multicolumn{1}{c|}{Standard deviation} & \multicolumn{1}{c}{Median} & \multicolumn{1}{c}{Mean} & \multicolumn{1}{c}{Standard deviation} \\ \hline
\multicolumn{1}{c|}{\multirow{4}{*}{Approaches}} & Baseline          &    \multicolumn{1}{c}{8.556}                  & 
           \multicolumn{1}{c}{10.665}                      
      &    \multicolumn{1}{c|}{3.072}                                     &
           \multicolumn{1}{c}{15.093}
      &    \multicolumn{1}{c}{17.654}                      & 
            \multicolumn{1}{c}{5.751}              \\
\multicolumn{1}{c|}{}                          & ML2ASR            & \multicolumn{1}{c}{9.593}                           &       \multicolumn{1}{c}{9.457}                   &        
        \multicolumn{1}{c|}{3.094}&                          
        \multicolumn{1}{c}{36.630}&
        \multicolumn{1}{c}{37.320}&
        \multicolumn{1}{c}{3.570}\\
\multicolumn{1}{c|}{}                          & LSA               & \multicolumn{1}{c}{9.765}                           &       \multicolumn{1}{c}{9.910}                   &        
        \multicolumn{1}{c|}{3.250}                            &\multicolumn{1}{c}{38.033}                           &\multicolumn{1}{c}{38.035}                          &\multicolumn{1}{c}{3.808}                                        \\
\multicolumn{1}{c|}{}                          & LSA with operator & \multicolumn{1}{c}{9.895}                           & \multicolumn{1}{c}{9.813}                         &     
    \multicolumn{1}{c|}{3.040}
&   \multicolumn{1}{c}{16.263}                        &      
    \multicolumn{1}{c}{17.962}                    &          \multicolumn{1}{c}{7.503}                             
\end{tabular}
\end{adjustbox}
\end{table}

\begin{table}[h]
\caption{Statistics corresponding to Figure~\ref{fig:energy consumption distribution over adaptation cycles}: energy consumption in the base scenario}\label{tab:19b}
\begin{adjustbox}{max width=\textwidth}
\begin{tabular}{cl|lll|lll}
\multicolumn{1}{l}{}                           &                   & \multicolumn{3}{c|}{No drift}                                                                   & \multicolumn{3}{c}{Novel class appeared}                                                      \\ \cline{3-8} 
\multicolumn{1}{l}{}                           &                   & \multicolumn{1}{c}{Median} & \multicolumn{1}{c}{Mean} & \multicolumn{1}{c|}{Standard deviation} & \multicolumn{1}{c}{Median} & \multicolumn{1}{c}{Mean} & \multicolumn{1}{c}{Standard deviation} \\ \hline
\multicolumn{1}{c|}{\multirow{4}{*}{Approaches}} & Baseline          &    \multicolumn{1}{c}{14.742}                  & 
           \multicolumn{1}{c}{14.661}                      
      &    \multicolumn{1}{c|}{0.182}                                     &
           \multicolumn{1}{c}{14.665}
      &    \multicolumn{1}{c}{14.527}                      & 
            \multicolumn{1}{c}{0.195}              \\
\multicolumn{1}{c|}{}                          & ML2ASR            & \multicolumn{1}{c}{14.616}                           &       \multicolumn{1}{c}{14.588}                   &        
        \multicolumn{1}{c|}{0.226}&                          
        \multicolumn{1}{c}{14.809}&
        \multicolumn{1}{c}{14.823}&
        \multicolumn{1}{c}{0.139}\\
\multicolumn{1}{c|}{}                          & LSA               & \multicolumn{1}{c}{14.683}                           &       \multicolumn{1}{c}{14.640}                   &        
        \multicolumn{1}{c|}{0.215}                            &\multicolumn{1}{c}{14.809}                           &\multicolumn{1}{c}{14.834}                          &\multicolumn{1}{c}{0.152}                                        \\
\multicolumn{1}{c|}{}                          & LSA with operator & \multicolumn{1}{c}{14.708}                           & \multicolumn{1}{c}{14.651}                         &     
    \multicolumn{1}{c|}{0.224}
&   \multicolumn{1}{c}{14.545}                        &      
    \multicolumn{1}{c}{14.529}                    &          \multicolumn{1}{c}{0.155}                             
\end{tabular}
\end{adjustbox}
\end{table}

\subsection{Impact of Drift of Adaptation Spaces on The Utility (Figure~\ref{fig:resuls-utilities})}
Figure~\ref{fig:resuls-utilities} shows the impact of the drift of adaptation spaces on the utility of the system. Table~\ref{tab:20} shows the basic statistics.

\label{tab: normality test fig 20}

\begin{table}[h]
\caption{Statistics corresponding to Figure~~\ref{fig:resuls-utilities}: utility in the base scenario}\label{tab:20}
\begin{adjustbox}{max width=\textwidth}
\begin{tabular}{cl|lll|lll}
\multicolumn{1}{l}{}                           &                   & \multicolumn{3}{c|}{No drift}                                                                   & \multicolumn{3}{c}{Novel class appeared}                                                      \\ \cline{3-8} 
\multicolumn{1}{l}{}                           &                   & \multicolumn{1}{c}{Median} & \multicolumn{1}{c}{Mean} & \multicolumn{1}{c|}{Standard deviation} & \multicolumn{1}{c}{Median} & \multicolumn{1}{c}{Mean} & \multicolumn{1}{c}{Standard deviation} \\ \hline
\multicolumn{1}{c|}{\multirow{4}{*}{Approaches}} & Baseline          &    \multicolumn{1}{c}{0.834}                  & 
           \multicolumn{1}{c}{0.835}                      
      &    \multicolumn{1}{c|}{0.036}                                     &
           \multicolumn{1}{c}{0.779}
      &    \multicolumn{1}{c}{0.806}                      & 
            \multicolumn{1}{c}{0.088}              \\
\multicolumn{1}{c|}{}                          & ML2ASR            & \multicolumn{1}{c}{0.843}                           &       \multicolumn{1}{c}{0.862}                   &        
        \multicolumn{1}{c|}{0.054}&                          
        \multicolumn{1}{c}{0.606}&
        \multicolumn{1}{c}{0.589}&
        \multicolumn{1}{c}{0.052}\\
\multicolumn{1}{c|}{}                          & LSA               & \multicolumn{1}{c}{0.837}                           &       \multicolumn{1}{c}{0.845}                   &        
        \multicolumn{1}{c|}{0.040}                            &\multicolumn{1}{c}{0.591}                           &\multicolumn{1}{c}{0.577}                          &\multicolumn{1}{c}{0.058}                                        \\
\multicolumn{1}{c|}{}                          & LSA with operator & \multicolumn{1}{c}{0.838}                           & \multicolumn{1}{c}{0.844}                         &     
    \multicolumn{1}{c|}{0.037}
&   \multicolumn{1}{c}{0.776}                        &      
    \multicolumn{1}{c}{0.798}                    &          \multicolumn{1}{c}{0.099}                             
\end{tabular}
\end{adjustbox}
\end{table}

For the baseline and LSA with operator feedback when the novel class appeared, denoted by $\mu_{u,\text{Baseline}}$, $\mu_{u,\text{LSA with operator}}$ respectively, we define the following alternative hypothesis:  

\begin{itemize}
    \item The utility of the ``Baseline'' is higher than the utility of the ``evolving classifier with LSA and operator feedback''  when a novel class appeared:
    \begin{align*}
        H_1: & \mu_{u,\text{Baseline}} > \mu_{u,\text{LSA with operator}} \\
        H_0: &  \mu_{u,\text{Baseline}} = \mu_{u,\text{LSA with operator}}
\end{align*}
\end{itemize}

We start with determining the appropriate test we can use to test the null hypothesis. First, we test whether the data of the utilities for the baseline and LSA with operator are normally distributed using the Anderson–Darling test~\cite{d2017goodness}. Based on the results of Table~\ref{tab: normality test fig 20}, with a significance level ($\alpha$) of $0.05$, all normality tests were rejected. 

\begin{table}[h]
\center
\caption{\re{Normality tests for distributions in Figure~\ref{fig:resuls-utilities}, by Anderson-Darling test method. }}
\begin{adjustbox}{max width=\textwidth}
\begin{tabular}{|c|c|}
\hline
 Approach & p-value\\ \hhline{==}
Baseline &  \reject{0.000} \\ \hline
LSA with operator &   \reject{0.000} \\ \hline
\end{tabular}
\end{adjustbox}
\label{tab: normality test fig 20}
\end{table}

Next, we use the Spearman correlation method~\cite{myers2010research} to test the dependency between the pair of data. With a significance level of 0.05, the test results shown in Table~\ref{tab: dependency test fig 20} indicate that the dependency test for the pair is rejected.

\begin{table}[h]
\center
\caption{\re{Dependency test for distributions in Figure~\ref{fig:resuls-utilities}, by Spearman correlation test method. }}
\begin{adjustbox}{max width=\textwidth}
\begin{tabular}{|c|c|}
\hline
 Approach & p-value\\ \hhline{==}
(Baseline, LSA with operator) &   \reject{0.000} \\ \hline
\end{tabular}
\end{adjustbox}
\label{tab: dependency test fig 20}
\end{table}

Since both the normality test and the dependency test are rejected, we apply the Mann-Whitney U test\footnote{The Mann-Whitney U test compares differences between the probability distribution of two independent groups when the dependent variable is either ordinal or continuous, but not normally distributed.}~\cite{MacFarland2016} to test the null hypothesis in Table~\ref{tab: hypotheses test fig 20} for Figure~\ref{fig:resuls-utilities}.\footnote{Mann-Whitney U Tests and Wilcoxon Rank Sum Tests reported in this paper have been carried out as one-tailed tests.}  

\begin{table}[!htbp]
\center
\caption{\re{Test results for the hypothesis for Figure~\ref{fig:resuls-utilities}.}}
\begin{adjustbox}{max width=\textwidth}
\begin{tabular}{|c|c|c|c|}
\hline
\multicolumn{1}{|c|}{Hypothesis}  & p-value & Test method \\ \hhline{====}
\multicolumn{1}{|l|}{\begin{tabular}[c]{@{}c@{}}
$H_0: \mu_{u,\text{Baseline}} = \mu_{u,\text{LSA with operator}}$
        \end{tabular}} 
        & 0.114 & Mann-Whitney U
          \\ \hline
\end{tabular}
\end{adjustbox}

\label{tab: hypotheses test fig 20}
\end{table}

With a significance level of 0.05, the test result does not support the alternative hypothesis. Hence, there is no evidence that the utility of the ``Baseline'' is higher than the utility of the ``evolving classifier with LSA and operator feedback''. 

\subsection{Impact of Drift of Adaptation Spaces on The \rsm\ (Figure~\ref{fig: rsm improvement})}
\label{sec: appendix fig 21}

Figure~\ref{fig: rsm improvement} shows the impact of drift of adaptation spaces on the \rsm\ for the different approaches.
As LSA with operator feedback clearly outperforms two other approaches, i.e., ML2ASR and LSA, in the base scenario after the appearance of a novel class, we only present basic statistics in Table\,\ref{tab:fig21}.

\begin{table}[h]
\caption{Statistics corresponding to Figure~\ref{fig: rsm improvement}: \rsm\ in the base scenario}\label{tab:fig21}
\begin{adjustbox}{max width=\textwidth}
\begin{tabular}{cl|lll|lll}
\multicolumn{1}{l}{}                           &                   & \multicolumn{3}{c|}{No drift}                                                                   & \multicolumn{3}{c}{Novel class appeared}                                                      \\ \cline{3-8} 
\multicolumn{1}{l}{}                           &                   & \multicolumn{1}{c}{Median} & \multicolumn{1}{c}{Mean} & \multicolumn{1}{c|}{Standard deviation} & \multicolumn{1}{c}{Median} & \multicolumn{1}{c}{Mean} & \multicolumn{1}{c}{Standard deviation} \\ \hline
\multicolumn{1}{c|}{\multirow{3}{*}{Approaches}} & ML2ASR          &    \multicolumn{1}{c}{0.000}                  & 
           \multicolumn{1}{c}{0.002}                      
      &    \multicolumn{1}{c|}{0.010}                                     &
           \multicolumn{1}{c}{0.550}
      &    \multicolumn{1}{c}{0.548}                      & 
            \multicolumn{1}{c}{0.036}              \\
\multicolumn{1}{c|}{}                          & LSA               & \multicolumn{1}{c}{0.000}                           &       \multicolumn{1}{c}{0.002}                   &        
        \multicolumn{1}{c|}{0.010}                            &\multicolumn{1}{c}{0.550}                           &\multicolumn{1}{c}{0.543}                          &\multicolumn{1}{c}{0.046}                                        \\
\multicolumn{1}{c|}{}                          & LSA with operator & \multicolumn{1}{c}{0.000}                           & \multicolumn{1}{c}{0.000}                         &     
    \multicolumn{1}{c|}{0.000}
&   \multicolumn{1}{c}{0.000}                        &      
    \multicolumn{1}{c}{0.055}                    &          \multicolumn{1}{c}{0.141}                             
\end{tabular}
\end{adjustbox}
\end{table}

\subsection{Utility for All Preference Orders of Stakeholders Split For Classes Appearance Orders (Figure~\ref{fig: utility total for class appearances})}
\label{sec: appendix fig 22}

Figure~\ref{fig: utility total for class appearances} shows the utility for all preference orders of stakeholders split for the appearance orders of classes. Table~\ref{tab:fig22} shows the basic statistics. We define the following set of alternative hypotheses:

\begin{table}[h]
\caption{Statistics corresponding to Figure~\ref{fig: utility total for class appearances}: utility for all preference orders of stakeholders}\label{tab:fig22}
\begin{adjustbox}{max width=\textwidth}
\begin{tabular}{cl|lll|lll}
\multicolumn{1}{l}{}                           &                   & \multicolumn{3}{c|}{(a)}                                                                   & \multicolumn{3}{c}{(b)}                                                      \\ \cline{3-8} 
\multicolumn{1}{l}{}                           &                   & \multicolumn{1}{c}{Median} & \multicolumn{1}{c}{Mean} & \multicolumn{1}{c|}{Standard deviation} & \multicolumn{1}{c}{Median} & \multicolumn{1}{c}{Mean} & \multicolumn{1}{c}{Standard deviation} \\ \hline
\multicolumn{1}{c|}{\multirow{3}{*}{Approaches}} & Baseline          &    \multicolumn{1}{c}{0.806}                  & 
           \multicolumn{1}{c}{0.765}                      
      &    \multicolumn{1}{c|}{0.192}                                     &
           \multicolumn{1}{c}{0.806}
      &    \multicolumn{1}{c}{0.767}                      & 
            \multicolumn{1}{c}{0.189}              \\
\multicolumn{1}{c|}{}                          & ML2ASR            & \multicolumn{1}{c}{0.801}                           &       \multicolumn{1}{c}{0.634}                   &        
        \multicolumn{1}{c|}{0.301}&                          
        \multicolumn{1}{c}{0.803}&
        \multicolumn{1}{c}{0.651}&
        \multicolumn{1}{c}{0.289}\\
\multicolumn{1}{c|}{}                          & LSA               & \multicolumn{1}{c}{0.802}                           &       \multicolumn{1}{c}{0.637}                   &        
        \multicolumn{1}{c|}{0.298}                            &\multicolumn{1}{c}{0.803}                           &\multicolumn{1}{c}{0.646}                          &\multicolumn{1}{c}{0.290}                                        \\
\multicolumn{1}{c|}{}                          & LSA with operator & \multicolumn{1}{c}{0.809}                           & \multicolumn{1}{c}{0.803}                         &     
    \multicolumn{1}{c|}{0.110}
&   \multicolumn{1}{c}{0.809}                        &      
    \multicolumn{1}{c}{0.703}                    &          \multicolumn{1}{c}{0.264}                             
\end{tabular}
\end{adjustbox}
\newline
\vspace*{0.25cm}
\newline
\begin{adjustbox}{max width=\textwidth}
\begin{tabular}{cl|lll|lll}
\multicolumn{1}{l}{}                           &                   & \multicolumn{3}{c|}{(c)}                                                                   & \multicolumn{3}{c}{(d)}                                                      \\ \cline{3-8} 
\multicolumn{1}{l}{}                           &                   & \multicolumn{1}{c}{Median} & \multicolumn{1}{c}{Mean} & \multicolumn{1}{c|}{Standard deviation} & \multicolumn{1}{c}{Median} & \multicolumn{1}{c}{Mean} & \multicolumn{1}{c}{Standard deviation} \\ \hline
\multicolumn{1}{c|}{\multirow{3}{*}{Approaches}} & Baseline          &    \multicolumn{1}{c}{0.806}                  & 
           \multicolumn{1}{c}{0.775}                      
      &    \multicolumn{1}{c|}{0.171}                                     &
           \multicolumn{1}{c}{0.806}
      &    \multicolumn{1}{c}{0.777}                      & 
            \multicolumn{1}{c}{0.168}              \\
\multicolumn{1}{c|}{}                          & ML2ASR            & \multicolumn{1}{c}{0.807}                           &       \multicolumn{1}{c}{0.728}                   &        
        \multicolumn{1}{c|}{0.251}&                          
        \multicolumn{1}{c}{0.808}&
        \multicolumn{1}{c}{0.728}&
        \multicolumn{1}{c}{0.250}\\
\multicolumn{1}{c|}{}                          & LSA               & \multicolumn{1}{c}{0.808}                           &       \multicolumn{1}{c}{0.756}                   &        
        \multicolumn{1}{c|}{0.253}                            &\multicolumn{1}{c}{0.808}                           &\multicolumn{1}{c}{0.721}                          &\multicolumn{1}{c}{0.256}                                        \\
\multicolumn{1}{c|}{}                          & LSA with operator & \multicolumn{1}{c}{0.807}                           & \multicolumn{1}{c}{0.806}                         &     
    \multicolumn{1}{c|}{0.089}
&   \multicolumn{1}{c}{0.808}                        &      
    \multicolumn{1}{c}{0.809}                    &          \multicolumn{1}{c}{0.088}                
\end{tabular}
\end{adjustbox}
\newline
\vspace*{0.25cm}
\newline
\begin{adjustbox}{max width=\textwidth}
\begin{tabular}{cl|lll|lll}
\multicolumn{1}{l}{}                           &                   & \multicolumn{3}{c|}{(e)}                                                                   & \multicolumn{3}{c}{(f)}                                                      \\ \cline{3-8} 
\multicolumn{1}{l}{}                           &                   & \multicolumn{1}{c}{Median} & \multicolumn{1}{c}{Mean} & \multicolumn{1}{c|}{Standard deviation} & \multicolumn{1}{c}{Median} & \multicolumn{1}{c}{Mean} & \multicolumn{1}{c}{Standard deviation} \\ \hline
\multicolumn{1}{c|}{\multirow{3}{*}{Approaches}} & Baseline          &    \multicolumn{1}{c}{0.806}                  & 
           \multicolumn{1}{c}{0.712}                      
      &    \multicolumn{1}{c|}{0.253}                                     &
           \multicolumn{1}{c}{0.810}
      &    \multicolumn{1}{c}{0.821}                      & 
            \multicolumn{1}{c}{0.042}              \\
\multicolumn{1}{c|}{}                          & ML2ASR            & \multicolumn{1}{c}{0.603}                           &       \multicolumn{1}{c}{0.495}                   &        
        \multicolumn{1}{c|}{0.285}&                          
        \multicolumn{1}{c}{0.809}&
        \multicolumn{1}{c}{0.825}&
        \multicolumn{1}{c}{0.037}\\
\multicolumn{1}{c|}{}                          & LSA               & \multicolumn{1}{c}{0.587}                           &       \multicolumn{1}{c}{0.494}                   &        
        \multicolumn{1}{c|}{0.280}                            &\multicolumn{1}{c}{0.809}                           &\multicolumn{1}{c}{0.827}                          &\multicolumn{1}{c}{0.038}                                        \\
\multicolumn{1}{c|}{}                          & LSA with operator & \multicolumn{1}{c}{0.807}                           & \multicolumn{1}{c}{0.780}                         &     
    \multicolumn{1}{c|}{0.149}
&   \multicolumn{1}{c}{0.809}                        &      
    \multicolumn{1}{c}{0.825}                    &          \multicolumn{1}{c}{0.039}                
\end{tabular}
\end{adjustbox}
\end{table}

\begin{itemize}

    \item For each scenario $i$, the utility of the ``Baseline'' is higher than the utility of the ``evolving classifier with LSA and operator feedback'' when a novel class appeared:

\begin{align*}
        H_1^{(i)}: & \mu_{i, u,\text{Baseline}} > \mu_{i, u,\text{LSA with operator}}\\
        H_0^{(i)}: & \mu_{i, u,\text{Baseline}} = \mu_{i, u,\text{LSA with operator}}
\end{align*}

    \item For each scenario $i$, the utility of the ``evolving classifier with LSA and operator feedback'' is higher than the utility of the ``pre-defined classifier with ML2ASR'':

\begin{align*}
    H_1^{(i)}: & \mu_{i, u,\text{LSA with operator}} > \mu_{i, u,\text{ML2ASR}}\\
    H_0^{(i)}: & \mu_{i, u,\text{LSA with operator}} = \mu_{i, u,\text{ML2ASR}}
\end{align*}

 \item For each scenario $i$, the utility of the ``evolving classifier with LSA and operator feedback'' is higher than the utility of the ``pre-defined classifier with LSA'':

\begin{align*}
        H_1^{(i)}: & \mu_{i, u,\text{LSA with operator}} > \mu_{i, u,\text{LSA}}\\
        H_0^{(i)}: & \mu_{i, u,\text{LSA with operator}} = \mu_{i, u,\text{LSA}}
\end{align*}
\end{itemize}
Note that the above generic hypotheses represent six distinct concrete hypotheses with the value of $i$ chosen from the set of scenarios $\{a,b,c,d,e,f\}$.

We start with determining the appropriate test we can use to test null hypotheses. First, we test whether the data of the utilities for each approach are normally distributed using the Anderson–Darling test. Based on the results of Table~\ref{tab: normality test fig 22}, with a significance level ($\alpha$) of $0.05$, all normality tests were rejected. 

\begin{table}[h]
\center
\caption{\re{Normality tests for distributions in subfigures of Figure~\ref{fig: utility total for class appearances}, by Anderson-Darling test method }}
\label{tab: normality test fig 22}
\begin{adjustbox}{max width=\textwidth}
\begin{tabular}{|c|c|c|c|c|c|c|}
\hline
 Approach & \begin{tabular}[c]{@{}c@{}}p-value (a)\end{tabular}
          & \begin{tabular}[c]{@{}c@{}}p-value (b)\end{tabular}
          & \begin{tabular}[c]{@{}c@{}}p-value (c)\end{tabular}
          & \begin{tabular}[c]{@{}c@{}}p-value (d)\end{tabular}
          & \begin{tabular}[c]{@{}c@{}}p-value (e)\end{tabular}
          & \begin{tabular}[c]{@{}c@{}}p-value (f)\end{tabular}\\ \hhline{=======}
Baseline &  \reject{0.000} & \reject{0.000} & \reject{0.000} & \reject{0.000} & \reject{0.000} & \reject{0.000}\\ \hline
ML2ASR &   \reject{0.000} & \reject{0.000} & \reject{0.000} & \reject{0.000} & \reject{0.000} & \reject{0.000}\\ \hline
LSA &   \reject{0.000} & \reject{0.000} & \reject{0.000} & \reject{0.000} & \reject{0.000} & \reject{0.000}\\ \hline
LSA with operator &   \reject{0.000} & \reject{0.000} & \reject{0.000} & \reject{0.000} & \reject{0.000} & \reject{0.000}\\ \hline
\end{tabular}
\end{adjustbox}
\end{table}

Next, we use the Spearman correlation method to test the dependency between related pairs of data. With a significance level of 0.05, the test results shown in Table~\ref{tab: dependency test fig 22} indicate that the dependency tests for all pairs are rejected.

\begin{table}[h]
\center
\caption{\re{Dependency tests for distributions in subfigures of Figure~\ref{fig: utility total for class appearances}, by Spearman correlation test method}}
\label{tab: dependency test fig 22}
\begin{adjustbox}{max width=\textwidth}
\begin{tabular}{|c|c|c|c|c|c|c|}
\hline
 Approach & \begin{tabular}[c]{@{}c@{}}p-value (a)\end{tabular}
          & \begin{tabular}[c]{@{}c@{}}p-value (b)\end{tabular}
          & \begin{tabular}[c]{@{}c@{}}p-value (c)\end{tabular}
          & \begin{tabular}[c]{@{}c@{}}p-value (d)\end{tabular}
          & \begin{tabular}[c]{@{}c@{}}p-value (e)\end{tabular}
          & \begin{tabular}[c]{@{}c@{}}p-value (f)\end{tabular}\\ \hhline{=======}
(Baseline, LSA with operator) &   \reject{0.000} & \reject{0.000} & \reject{0.000} & \reject{0.000} & \reject{0.000} & \reject{0.000}\\ \hline
(ML2ASR, LSA with operator) &   \reject{0.000} & \reject{0.000} & \reject{0.000} & \reject{0.000} & \reject{0.000} & \reject{0.000}\\ \hline
(LSA, LSA with operator) &   \reject{0.000} & \reject{0.000} & \reject{0.000} & \reject{0.000} & \reject{0.000} & \reject{0.000}\\ \hline
\end{tabular}
\end{adjustbox}
\end{table}

Since both the normality and dependency tests are rejected for all cases, we apply the Mann-Whitney U test to test all the null hypotheses of Table~\ref{tab: hypotheses test fig 22} for Figure~\ref{fig: utility total for class appearances}.

\begin{table}[!htbp]
\center
\caption{\re{Test results for all hypotheses for Figure~\ref{fig: utility total for class appearances}, related hypotheses to each subfigure are separated by a double horizontal line from the others.}}
\begin{adjustbox}{max width=\textwidth}
\begin{tabular}{|c|c|c|c|}
\hline
\multicolumn{2}{|c|}{Hypothesis}  & p-value & Test method \\ \hhline{====}
\multicolumn{2}{|l|}{\begin{tabular}[c]{@{}c@{}}
$H_0^{(a)}: \mu_{a, u,\text{Baseline}} = \mu_{a,u,\text{LSA with operator}}$ 
        \end{tabular}}& 0.953 & Mann-Whitney U  \\ \hline
\multicolumn{2}{|l|}{\begin{tabular}[c]{@{}c@{}}
$H_0^{(a)}: \mu_{a, u,\text{LSA with operator}} = \mu_{a,u,\text{ML2ASR}}$ 
        \end{tabular}}& \reject{0.000} & Mann-Whitney U  \\ \hline
\multicolumn{2}{|l|}{\begin{tabular}[c]{@{}c@{}}
$H_0^{(a)}: \mu_{a, u,\text{LSA with operator}} = \mu_{a,u,\text{LSA}}$ 
        \end{tabular}}& \reject{0.000} & Mann-Whitney U  \\ \hhline{====}

\multicolumn{2}{|l|}{\begin{tabular}[c]{@{}c@{}}
$H_0^{(b)}: \mu_{b, u,\text{Baseline}} = \mu_{b,u,\text{LSA with operator}}$ 
        \end{tabular}}& 0.110 & Mann-Whitney U  \\ \hline
\multicolumn{2}{|l|}{\begin{tabular}[c]{@{}c@{}}
$H_0^{(b)}: \mu_{b, u,\text{LSA with operator}} = \mu_{b,u,\text{ML2ASR}}$ 
        \end{tabular}}& \reject{0.000} & Mann-Whitney U  \\ \hline
\multicolumn{2}{|l|}{\begin{tabular}[c]{@{}c@{}}
$H_0^{(b)}: \mu_{b, u,\text{LSA with operator}} = \mu_{b,u,\text{LSA}}$ 
        \end{tabular}}& \reject{0.000} & Mann-Whitney U  \\ \hhline{====}

\multicolumn{2}{|l|}{\begin{tabular}[c]{@{}c@{}}
$H_0^{(c)}: \mu_{c, u,\text{Baseline}} = \mu_{c,u,\text{LSA with operator}}$ 
        \end{tabular}}& 0.767 & Mann-Whitney U  \\ \hline
\multicolumn{2}{|l|}{\begin{tabular}[c]{@{}c@{}}
$H_0^{(c)}: \mu_{c, u,\text{LSA with operator}} = \mu_{c,u,\text{ML2ASR}}$ 
        \end{tabular}}& {0.747} & Mann-Whitney U  \\ \hline
\multicolumn{2}{|l|}{\begin{tabular}[c]{@{}c@{}}
$H_0^{(c)}: \mu_{c, u,\text{LSA with operator}} = \mu_{c,u,\text{LSA}}$ 
        \end{tabular}}& {0.837} & Mann-Whitney U  \\ \hhline{====}

\multicolumn{2}{|l|}{\begin{tabular}[c]{@{}c@{}}
$H_0^{(d)}: \mu_{d, u,\text{Baseline}} = \mu_{d,u,\text{LSA with operator}}$ 
        \end{tabular}}& 0.708 & Mann-Whitney U  \\ \hline
\multicolumn{2}{|l|}{\begin{tabular}[c]{@{}c@{}}
$H_0^{(d)}: \mu_{d, u,\text{LSA with operator}} = \mu_{d,u,\text{ML2ASR}}$ 
        \end{tabular}}& {0.539} & Mann-Whitney U  \\ \hline
\multicolumn{2}{|l|}{\begin{tabular}[c]{@{}c@{}}
$H_0^{(d)}: \mu_{d, u,\text{LSA with operator}} = \mu_{d,u,\text{LSA}}$ 
        \end{tabular}}& {0.598} & Mann-Whitney U  \\ \hhline{====}

\multicolumn{2}{|l|}{\begin{tabular}[c]{@{}c@{}}
$H_0^{(e)}: \mu_{e, u,\text{Baseline}} = \mu_{e,u,\text{LSA with operator}}$ 
        \end{tabular}}& 0.638 & Mann-Whitney U  \\ \hline
\multicolumn{2}{|l|}{\begin{tabular}[c]{@{}c@{}}
$H_0^{(e)}: \mu_{e, u,\text{LSA with operator}} = \mu_{e,u,\text{ML2ASR}}$ 
        \end{tabular}}& \reject{0.000} & Mann-Whitney U  \\ \hline
\multicolumn{2}{|l|}{\begin{tabular}[c]{@{}c@{}}
$H_0^{(e)}: \mu_{e, u,\text{LSA with operator}} = \mu_{e,u,\text{LSA}}$ 
        \end{tabular}}& \reject{0.000} & Mann-Whitney U  \\ \hhline{====}

\multicolumn{2}{|l|}{\begin{tabular}[c]{@{}c@{}}
$H_0^{(f)}: \mu_{f, u,\text{Baseline}} = \mu_{f,u,\text{LSA with operator}}$ 
        \end{tabular}}& 0.997 & Mann-Whitney U  \\ \hline
\multicolumn{2}{|l|}{\begin{tabular}[c]{@{}c@{}}
$H_0^{(f)}: \mu_{f, u,\text{LSA with operator}} = \mu_{f,u,\text{ML2ASR}}$ 
        \end{tabular}}& {0.652} & Mann-Whitney U  \\ \hline
\multicolumn{2}{|l|}{\begin{tabular}[c]{@{}c@{}}
$H_0^{(f)}: \mu_{f, u,\text{LSA with operator}} = \mu_{f,u,\text{LSA}}$ 
        \end{tabular}}& {0.809} & Mann-Whitney U  \\ \hline

\end{tabular}
\end{adjustbox}
\label{tab: hypotheses test fig 22}
\end{table}

With a significance level of 0.05, some test results support the corresponding alternative hypotheses, while other results do not. 
Concretely, we found evidence that: (i) the utility of the ``evolving classifier with LSA and operator feedback'' is higher than the utility of ``ML2ASR'' and utility of the ``pre-defined classifier with LSA'' in scenarios (a), (b), and (e). 
All the remaining alternative hypotheses are not supported by the test results. Concretely, we found no evidence that: (i) the utility of the ``Baseline'' is higher than the utility of the ``evolving classifier with LSA and operator feedback'' in all scenarios from (a) to (f), (ii)  the utility of the ``evolving classifier with LSA and operator feedback'' is higher than the utility of the ``ML2ASR'' in scenarios (c), (d), and (f), and (iii) the utility of the ``evolving classifier with LSA and operator feedback'' is higher than the utility of the ``pre-defined classifier with LSA'' in scenarios (c), (d), and (f), the same as the ``ML2ASR''.


\subsection{\rsm\ for All Preference Orders of Stakeholders Split for Classes Appearance Order (Figure~\ref{fig: rsm total for class appearances})}
\label{sec: appendix fig 23}

Figure~\ref{fig: rsm total for class appearances} illustrates the \rsm\ for all preference orders of stakeholders split for classes appearance orders. Table~\ref{tab:fig23} shows the basic statistics. 
We define the following set of alternative hypotheses: 

\begin{table}[h]
\caption{Statistics corresponding to Figure~\ref{fig: rsm total for class appearances}: \rsm\ for all preference orders of stakeholders}\label{tab:fig23}
\begin{adjustbox}{max width=\textwidth}
\begin{tabular}{cl|lll|lll}
\multicolumn{1}{l}{}                           &                   & \multicolumn{3}{c|}{(a)}                                                                   & \multicolumn{3}{c}{(b)}                                                      \\ \cline{3-8} 
\multicolumn{1}{l}{}                           &                   & \multicolumn{1}{c}{Median} & \multicolumn{1}{c}{Mean} & \multicolumn{1}{c|}{Standard deviation} & \multicolumn{1}{c}{Median} & \multicolumn{1}{c}{Mean} & \multicolumn{1}{c}{Standard deviation} \\ \hline

\multicolumn{1}{c|}{\multirow{3}{*}{Approaches}}                        & ML2ASR            & \multicolumn{1}{c}{0.200}                           &       \multicolumn{1}{c}{0.227}                   &        
        \multicolumn{1}{c|}{0.175}&                          
        \multicolumn{1}{c}{0.186}&
        \multicolumn{1}{c}{0.221}&
        \multicolumn{1}{c}{0.169}\\
\multicolumn{1}{c|}{}                          & LSA               & \multicolumn{1}{c}{0.200}                           &       \multicolumn{1}{c}{0.228}                   &        
        \multicolumn{1}{c|}{0.176}                            &\multicolumn{1}{c}{0.163}                           &\multicolumn{1}{c}{0.230}                          &\multicolumn{1}{c}{0.180}                                        \\
\multicolumn{1}{c|}{}                          & LSA with operator & \multicolumn{1}{c}{0.118}                           & \multicolumn{1}{c}{0.082}                         &     
    \multicolumn{1}{c|}{0.092}
&   \multicolumn{1}{c}{0.125}                        &      
    \multicolumn{1}{c}{0.145}                    &          \multicolumn{1}{c}{0.115}                             
\end{tabular}
\end{adjustbox}
\newline
\vspace*{0.25cm}
\newline
\begin{adjustbox}{max width=\textwidth}
\begin{tabular}{cl|lll|lll}
\multicolumn{1}{l}{}                           &                   & \multicolumn{3}{c|}{(c)}                                                                   & \multicolumn{3}{c}{(d)}                                                      \\ \cline{3-8} 
\multicolumn{1}{l}{}                           &                   & \multicolumn{1}{c}{Median} & \multicolumn{1}{c}{Mean} & \multicolumn{1}{c|}{Standard deviation} & \multicolumn{1}{c}{Median} & \multicolumn{1}{c}{Mean} & \multicolumn{1}{c}{Standard deviation} \\ \hline
\multicolumn{1}{c|}{\multirow{3}{*}{Approaches}}                          & ML2ASR            & \multicolumn{1}{c}{0.000}                           &       \multicolumn{1}{c}{0.107}                   &        
        \multicolumn{1}{c|}{0.139}&                          
        \multicolumn{1}{c}{0.000}&
        \multicolumn{1}{c}{0.102}&
        \multicolumn{1}{c}{0.141}\\
\multicolumn{1}{c|}{}                          & LSA               & \multicolumn{1}{c}{0.000}                           &       \multicolumn{1}{c}{0.102}                   &        
        \multicolumn{1}{c|}{0.138}                            &\multicolumn{1}{c}{0.000}                           &\multicolumn{1}{c}{0.105}                          &\multicolumn{1}{c}{0.151}                                        \\
\multicolumn{1}{c|}{}                          & LSA with operator & \multicolumn{1}{c}{0.000}                           & \multicolumn{1}{c}{0.038}                         &     
    \multicolumn{1}{c|}{0.085}
&   \multicolumn{1}{c}{0.000}                        &      
    \multicolumn{1}{c}{0.031}                    &          \multicolumn{1}{c}{0.062}                
\end{tabular}
\end{adjustbox}
\newline
\vspace*{0.25cm}
\newline
\begin{adjustbox}{max width=\textwidth}
\begin{tabular}{cl|lll|lll}
\multicolumn{1}{l}{}                           &                   & \multicolumn{3}{c|}{(e)}                                                                   & \multicolumn{3}{c}{(f)}                                                      \\ \cline{3-8} 
\multicolumn{1}{l}{}                           &                   & \multicolumn{1}{c}{Median} & \multicolumn{1}{c}{Mean} & \multicolumn{1}{c|}{Standard deviation} & \multicolumn{1}{c}{Median} & \multicolumn{1}{c}{Mean} & \multicolumn{1}{c}{Standard deviation} \\ \hline
\multicolumn{1}{c|}{\multirow{3}{*}{Approaches}}                          & ML2ASR            & \multicolumn{1}{c}{0.438}                           &       \multicolumn{1}{c}{0.495}                   &        
        \multicolumn{1}{c|}{0.426}&                          
        \multicolumn{1}{c}{0.081}&
        \multicolumn{1}{c}{0.068}&
        \multicolumn{1}{c}{0.062}\\
\multicolumn{1}{c|}{}                          & LSA               & \multicolumn{1}{c}{0.450}                           &       \multicolumn{1}{c}{0.436}                   &        
        \multicolumn{1}{c|}{0.124}                            &\multicolumn{1}{c}{0.068}                           &\multicolumn{1}{c}{0.064}                          &\multicolumn{1}{c}{0.060}                                        \\
\multicolumn{1}{c|}{}                          & LSA with operator & \multicolumn{1}{c}{0.118}                           & \multicolumn{1}{c}{0.094}                         &     
    \multicolumn{1}{c|}{0.110}
&   \multicolumn{1}{c}{0.068}                        &      
    \multicolumn{1}{c}{0.063}                    &          \multicolumn{1}{c}{0.061}                
\end{tabular}
\end{adjustbox}
\end{table}



\begin{itemize}

    \item For each scenario $i$, the \rsm\ of the ``evolving classifier with LSA and operator feedback'' is less than the \rsm\ of the ``pre-defined classifier with ML2ASR'':

\begin{align*}
        H_1^{(i)}: & \mu_{i, r,\text{LSA with operator}} < \mu_{i, r,\text{ML2ASR}} \\
        H_0^{(i)}: & \mu_{i, r,\text{LSA with operator}} = \mu_{i, r,\text{ML2ASR}}
\end{align*}

 \item For each scenario $i$, the \rsm\ of the ``evolving classifier with LSA and operator feedback'' is less than the \rsm\ of the ``pre-defined classifier with LSA'':

\begin{align*}
        H_1^{(i)}: & \mu_{i, r,\text{LSA with operator}} < \mu_{i, r,\text{LSA}} \\
        H_0^{(i)}: & \mu_{i, r,\text{LSA with operator}} = \mu_{i, r,\text{LSA}}
\end{align*}
\end{itemize}
Here too, these generic hypotheses represent six distinct concrete hypotheses dependent on the value of $i$ chosen from this set of scenarios $\{a,b,c,d,e,f\}$.

We start with determining the appropriate test we can use to test null hypotheses. First, we test whether the data of the \rsm\ for each approach are normally distributed using the Anderson–Darling test. Based on the results of Table~\ref{tab: normality test fig 23}, with a significance level of $0.05$, all normality tests are rejected.

\begin{table}[!htbp]
\center
\caption{\re{Normality tests for distributions in subfigures of Figure~\ref{fig: rsm total for class appearances}, by Anderson-Darling test method }}
\label{tab: normality test fig 23}
\begin{adjustbox}{max width=\textwidth}
\begin{tabular}{|c|c|c|c|c|c|c|}
\hline
 Approach & \begin{tabular}[c]{@{}c@{}}p-value (a)\end{tabular}
          & \begin{tabular}[c]{@{}c@{}}p-value (b)\end{tabular}
          & \begin{tabular}[c]{@{}c@{}}p-value (c)\end{tabular}
          & \begin{tabular}[c]{@{}c@{}}p-value (d)\end{tabular}
          & \begin{tabular}[c]{@{}c@{}}p-value (e)\end{tabular}
          & \begin{tabular}[c]{@{}c@{}}p-value (f)\end{tabular}\\ \hhline{=======}
ML2ASR &   \reject{0.001} & \reject{0.001} & \reject{0.000} & \reject{0.000} & \reject{0.028} & \reject{0.000}\\ \hline
LSA &   \reject{0.009} & \reject{0.000} & \reject{0.000} & \reject{0.000} & \reject{0.043} & \reject{0.000}\\ \hline
LSA with operator &   \reject{0.000} & \reject{0.003} & \reject{0.000} & \reject{0.000} & \reject{0.000} & \reject{0.000}\\ \hline
\end{tabular}
\end{adjustbox}
\end{table}

Next, we use the Spearman correlation method to test the dependency between related pairs of data. With a significance level of 0.05, the test results shown in Table~\ref{tab: dependency test fig 23} indicate that the dependency tests for all pairs are rejected.

\begin{table}[h]
\center
\caption{\re{Dependency tests for distributions in subfigures of Figure~\ref{fig: rsm total for class appearances}, by Spearman correlation test method}}
\label{tab: dependency test fig 23}
\begin{adjustbox}{max width=\textwidth}
\begin{tabular}{|c|c|c|c|c|c|c|}
\hline
 Approach & \begin{tabular}[c]{@{}c@{}}p-value (a)\end{tabular}
          & \begin{tabular}[c]{@{}c@{}}p-value (b)\end{tabular}
          & \begin{tabular}[c]{@{}c@{}}p-value (c)\end{tabular}
          & \begin{tabular}[c]{@{}c@{}}p-value (d)\end{tabular}
          & \begin{tabular}[c]{@{}c@{}}p-value (e)\end{tabular}
          & \begin{tabular}[c]{@{}c@{}}p-value (f)\end{tabular}\\ \hhline{=======}
(ML2ASR, LSA with operator) &   \reject{0.009} & \reject{0.000} & \reject{0.000} & \reject{0.000} & \reject{0.000} & \reject{0.000}\\ \hline
(LSA, LSA with operator) &   \reject{0.027} & \reject{0.000} & \reject{0.000} & \reject{0.000} & \reject{0.000} & \reject{0.000}\\ \hline
\end{tabular}
\end{adjustbox}
\end{table}

Since both the normality and dependency tests are rejected for all cases, we apply the Mann-Whitney U test to test the null hypotheses 
in Table~\ref{tab: hypotheses test fig 23} for Figure~\ref{fig: rsm total for class appearances}.

\begin{table}[h]
\center
\caption{\re{Test results for all hypotheses for Figure~\ref{fig: rsm total for class appearances}, related hypotheses to each subfigure are separated by a double horizontal line from the others.}}
\begin{adjustbox}{max width=\textwidth}
\begin{tabular}{|c|c|c|c|}
\hline
\multicolumn{2}{|c|}{Hypothesis}  & p-value & Test method \\ \hhline{====}
\multicolumn{2}{|l|}{\begin{tabular}[c]{@{}c@{}}
$H_0^{(a)}: \mu_{a, r,\text{LSA with operator}} = \mu_{a,r,\text{ML2ASR}}$ 
        \end{tabular}}& \reject{0.000} & Mann-Whitney U  \\ \hline
\multicolumn{2}{|l|}{\begin{tabular}[c]{@{}c@{}}
$H_0^{(a)}: \mu_{a, r,\text{LSA with operator}} = \mu_{a,r,\text{LSA}}$ 
        \end{tabular}}& \reject{0.000} & Mann-Whitney U  \\ \hhline{====}

\multicolumn{2}{|l|}{\begin{tabular}[c]{@{}c@{}}
$H_0^{(b)}: \mu_{b, r,\text{LSA with operator}} = \mu_{b,r,\text{ML2ASR}}$ 
        \end{tabular}}& \reject{0.028} & Mann-Whitney U  \\ \hline
\multicolumn{2}{|l|}{\begin{tabular}[c]{@{}c@{}}
$H_0^{(b)}: \mu_{b, r,\text{LSA with operator}} = \mu_{b,r,\text{LSA}}$ 
        \end{tabular}}& \reject{0.020} & Mann-Whitney U  \\ \hhline{====}

\multicolumn{2}{|l|}{\begin{tabular}[c]{@{}c@{}}
$H_0^{(c)}: \mu_{c, r,\text{LSA with operator}} = \mu_{c,r,\text{ML2ASR}}$ 
        \end{tabular}}& \reject{0.009} & Mann-Whitney U  \\ \hline
\multicolumn{2}{|l|}{\begin{tabular}[c]{@{}c@{}}
$H_0^{(c)}: \mu_{c, r,\text{LSA with operator}} = \mu_{c,r,\text{LSA}}$ 
        \end{tabular}}& \reject{0.010} & Mann-Whitney U  \\ \hhline{====}

\multicolumn{2}{|l|}{\begin{tabular}[c]{@{}c@{}}
$H_0^{(d)}: \mu_{d, r,\text{LSA with operator}} = \mu_{d,r,\text{ML2ASR}}$ 
        \end{tabular}}& \reject{0.012} & Mann-Whitney U  \\ \hline
\multicolumn{2}{|l|}{\begin{tabular}[c]{@{}c@{}}
$H_0^{(d)}: \mu_{d, r,\text{LSA with operator}} = \mu_{d,r,\text{LSA}}$ 
        \end{tabular}}& \reject{0.014} & Mann-Whitney U  \\ \hhline{====}

\multicolumn{2}{|l|}{\begin{tabular}[c]{@{}c@{}}
$H_0^{(e)}: \mu_{e, r,\text{LSA with operator}} = \mu_{e,r,\text{ML2ASR}}$ 
        \end{tabular}}& \reject{0.000} & Mann-Whitney U  \\ \hline
\multicolumn{2}{|l|}{\begin{tabular}[c]{@{}c@{}}
$H_0^{(e)}: \mu_{e, r,\text{LSA with operator}} = \mu_{e,r,\text{LSA}}$ 
        \end{tabular}}& \reject{0.000} & Mann-Whitney U  \\ \hhline{====}

\multicolumn{2}{|l|}{\begin{tabular}[c]{@{}c@{}}
$H_0^{(f)}: \mu_{f, r,\text{LSA with operator}} = \mu_{f,r,\text{ML2ASR}}$ 
        \end{tabular}}& {0.341} & Mann-Whitney U  \\ \hline
\multicolumn{2}{|l|}{\begin{tabular}[c]{@{}c@{}}
$H_0^{(f)}: \mu_{f, r,\text{LSA with operator}} = \mu_{f,r,\text{LSA}}$ 
        \end{tabular}}& {0.447} & Mann-Whitney U  \\ \hline
        
\end{tabular}
\end{adjustbox}
\label{tab: hypotheses test fig 23}
\end{table}

With a significance level of 0.05, the test results support the alternative hypotheses for scenarios (a) to (e), i.e., we found evidence that for these scenarios: (i) the \rsm\ of the ``evolving classifier with LSA and operator feedback'' is less than the \rsm\ of the ``ML2ASR'', and (ii) the \rsm\ of the ``evolving classifier with LSA and operator feedback'' is less than the \rsm\ of the ``pre-defined classifier with LSA''. For scenario (f) the alternative hypotheses are not supported by the test results.

\subsection{Utility for All Class Appearance Order Scenarios Split for Preference Order of Stakeholders (Figure~\ref{fig: utility total})}
\label{sec: appendix fig 24}

Figure~\ref{fig: utility total} shows the utility for all class appearance order scenarios split for preference order of stakeholders.
Table~\ref{tab:fig24} shows the basic statistics. We define the following set of alternative hypotheses:

\begin{table}[h]
\caption{Statistics corresponding to Figure~\ref{fig: utility total}: utility for all class appearance order scenarios}\label{tab:fig24}
\begin{adjustbox}{max width=\textwidth}
\begin{tabular}{cl|lll|lll}
\multicolumn{1}{l}{}                           &                   & \multicolumn{3}{c|}{(a)}                                                                   & \multicolumn{3}{c}{(b)}                                                      \\ \cline{3-8} 
\multicolumn{1}{l}{}                           &                   & \multicolumn{1}{c}{Median} & \multicolumn{1}{c}{Mean} & \multicolumn{1}{c|}{Standard deviation} & \multicolumn{1}{c}{Median} & \multicolumn{1}{c}{Mean} & \multicolumn{1}{c}{Standard deviation} \\ \hline

\multicolumn{1}{c|}{\multirow{4}{*}{Approaches}}                          & Baseline            & \multicolumn{1}{c}{0.834}                           &       \multicolumn{1}{c}{0.822}                   &        
        \multicolumn{1}{c|}{0.068}&                          
        \multicolumn{1}{c}{0.805}&
        \multicolumn{1}{c}{0.719}&
        \multicolumn{1}{c}{0.236}\\

\multicolumn{1}{c|}{}                          & ML2ASR            & \multicolumn{1}{c}{0.822}                           &       \multicolumn{1}{c}{0.778}                   &        
        \multicolumn{1}{c|}{0.124}&                          
        \multicolumn{1}{c}{0.806}&
        \multicolumn{1}{c}{0.591}&
        \multicolumn{1}{c}{0.340}\\
\multicolumn{1}{c|}{}                          & LSA               & \multicolumn{1}{c}{0.831}                           &       \multicolumn{1}{c}{0.776}                   &        
        \multicolumn{1}{c|}{0.126}                            &\multicolumn{1}{c}{0.806}                           &\multicolumn{1}{c}{0.589}                          &\multicolumn{1}{c}{0.340}                                        \\
\multicolumn{1}{c|}{}                          & LSA with operator & \multicolumn{1}{c}{0.829}                           & \multicolumn{1}{c}{0.817}                         &     
    \multicolumn{1}{c|}{0.082}
&   \multicolumn{1}{c}{0.807}                        &      
    \multicolumn{1}{c}{0.757}                    &          \multicolumn{1}{c}{0.189}                             
\end{tabular}
\end{adjustbox}
\end{table}


\begin{itemize}

    \item For each scenario $i$, the utility of the ``Baseline'' is higher than the utiltiy of the ``evolving classifier with LSA and operator feedback'':

\begin{align*}
        H_1^{(i)}: & \mu_{i, u,\text{Baseline}} > \mu_{i, u,\text{LSA with operator}}\\
        H_0^{(i)}: & \mu_{i, u,\text{Baseline}} = \mu_{i, u,\text{LSA with operator}}
\end{align*}

    \item For each scenario $i$, the utility of the ``evolving classifier with LSA and operator feedback'' is higher than the utility of the ``pre-defined classifier with ML2ASR'':

\begin{align*}
        H_1^{(i)}: & \mu_{i, u,\text{LSA with operator}} > \mu_{i, u,\text{ML2ASR}}\\
        H_0^{(i)}: & \mu_{i, u,\text{LSA with operator}} = \mu_{i, u,\text{ML2ASR}}        
\end{align*}

 \item For each scenario $i$, the utility of the ``evolving classifier with LSA and operator feedback'' is higher than the utility of the ``pre-defined classifier with LSA'':

\begin{align*}
        H_1^{(i)}: & \mu_{i, u,\text{LSA with operator}} > \mu_{i, u,\text{LSA}}\\
        H_0^{(i)}: & \mu_{i, u,\text{LSA with operator}} = \mu_{i, u,\text{LSA}} 
\end{align*}

\end{itemize}

Note that the above generic hypotheses represent distinct concrete hypotheses with the value of $i$ chosen from the set of $\{a,b\}$, i.e., the two scenarios for the preference order.

We start with determining the appropriate test we can use to test null hypotheses. First, we test whether the data of the utility for each scenario are normally distributed using the Anderson–Darling test. Based on the results of Table~\ref{tab: normality test fig 24}, with a significance level ($\alpha$) of $0.05$, all normality tests are rejected.

\begin{table}[h]
\center
\caption{\re{Normality tests for distributions in subfigures of Figure~\ref{fig: utility total}, by Anderson-Darling test method }}
\label{tab: normality test fig 24}
\begin{adjustbox}{max width=\textwidth}
\begin{tabular}{|c|c|c|}
\hline
 Approach & \begin{tabular}[c]{@{}c@{}}p-value (a)\end{tabular}
          & \begin{tabular}[c]{@{}c@{}}p-value (b)\end{tabular} \\ \hhline{===}
Baseline &  \reject{0.000} & \reject{0.000} \\ \hline
ML2ASR &   \reject{0.000} & \reject{0.000} \\ \hline
LSA &   \reject{0.000} & \reject{0.000} \\ \hline
LSA with operator &   \reject{0.000} & \reject{0.000} \\ \hline
\end{tabular}
\end{adjustbox}
\end{table}

Next, we use the Spearman correlation method to test the dependency between related pairs of data. With a significance level of 0.05, the test results in Table~\ref{tab: dependency test fig 24} for Figure~\ref{fig: utility total} indicate that we can reject the dependency test for the pairs in scenario (b), yet, we cannot reject the dependency test for the pairs in scenario (a). Therefore, we use the Mann-Whitney U test to test the null hypotheses for scenario (b). We use both the Mann-Whitney U test and the Wilcoxon signed-rank test\footnote{The Wilcoxon signed-rank test compares differences between two pairs of data when the dependent variable is either ordinal or continuous, but not normally distributed.}~\cite{conover1980practical} to test the null hypotheses for scenario (a).

\begin{table}[h]
\center
\caption{\re{Dependency tests for distributions in subfigures of Figure~\ref{fig: utility total}, by Spearman correlation test method}}
\label{tab: dependency test fig 24}
\begin{adjustbox}{max width=\textwidth}
\begin{tabular}{|c|c|c|}
\hline
 Approach & \begin{tabular}[c]{@{}c@{}}p-value (a)\end{tabular}
          & \begin{tabular}[c]{@{}c@{}}p-value (b)\end{tabular}\\ \hhline{===}
(Baseline, LSA with operator) &   0.107 & \reject{0.000} \\ \hline
(ML2ASR, LSA with operator) &   0.075 & \reject{0.000}\\ \hline
(LSA, LSA with operator) &   0.053 & \reject{0.001} \\ \hline
\end{tabular}
\end{adjustbox}
\end{table}

With a significance level of 0.05, the test results support the following alternative hypotheses in both scenarios: (i) the utility of ``evolving classifier with LSA and operator feedback'' is higher than the utility of the ``ML2ASR'' and (ii) the utility of ``evolving classifier with LSA and operator feedback'' is higher than the utility of the ``pre-defined classifier with LSA''.  The test results do not provide evidence to support the alternative hypothesis ``the utility of the ``Baseline'' is higher than the utility of ``evolving classifier with LSA and operator feedback'' in both scenarios (a) and (b).

\begin{table}[!htbp]
\center
\caption{\re{Test results for all mentioned hypotheses for Figure~\ref{fig: utility total}, related hypotheses to each subfigure are separated by a double horizontal line from the others.}}
\label{tab: hypotheses test fig 24}
\begin{adjustbox}{max width=\textwidth}
\begin{tabular}{|c|c|c|c|}
\hline
\multicolumn{1}{|c|}{Hypothesis}  & p-value & Test method \\ \hhline{====}
\multicolumn{1}{|l|}{\begin{tabular}[c]{@{}c@{}}
$H_0^{(a)}: \mu_{a, u,\text{Baseline}} = \mu_{a,u,\text{LSA with operator}}$ 
        \end{tabular}}& 
        \multicolumn{1}{|l|}{\begin{tabular}[c]{c@{}c@{}}
0.106 \\ 0.878\end{tabular}}& 
\multicolumn{1}{|l|}{\begin{tabular}[c]{c@{}c@{}}
Mann-Whitney U \\ Wilcoxon signed-rank U\end{tabular}}
          \\ \hline
\multicolumn{1}{|l|}{\begin{tabular}[c]{@{}c@{}}
$H_0^{(a)}: \mu_{a, u,\text{LSA with operator}} = \mu_{a,u,\text{ML2ASR}}$ 
        \end{tabular}}&         \multicolumn{1}{|l|}{\begin{tabular}[c]{c@{}c@{}}
\reject{0.000} \\ \reject{0.000}\end{tabular}}& 
\multicolumn{1}{|l|}{\begin{tabular}[c]{c@{}c@{}}
Mann-Whitney U \\ Wilcoxon signed-rank U\end{tabular}}
          \\ \hline
\multicolumn{1}{|l|}{\begin{tabular}[c]{@{}c@{}}
$H_0^{(a)}: \mu_{a, u,\text{LSA with operator}} = \mu_{a,u,\text{LSA}}$ 
        \end{tabular}}&
        \multicolumn{1}{|l|}{\begin{tabular}[c]{c@{}c@{}}
\reject{0.000} \\ \reject{0.000}\end{tabular}}& 
\multicolumn{1}{|l|}{\begin{tabular}[c]{c@{}c@{}}
Mann-Whitney U \\ Wilcoxon signed-rank U\end{tabular}}
          \\ \hhline{====}

\multicolumn{1}{|l|}{\begin{tabular}[c]{@{}c@{}}
$H_0^{(b)}: \mu_{b, u,\text{Baseline}} = \mu_{b,u,\text{LSA with operator}}$ 
        \end{tabular}}& 1.000 & Mann-Whitney U  \\ \hline
\multicolumn{1}{|l|}{\begin{tabular}[c]{@{}c@{}}
$H_0^{(b)}: \mu_{b, u,\text{LSA with operator}} = \mu_{b,u,\text{ML2ASR}}$ 
        \end{tabular}}& \reject{0.000} & Mann-Whitney U  \\ \hline
\multicolumn{1}{|l|}{\begin{tabular}[c]{@{}c@{}}
$H_0^{(b)}: \mu_{b, u,\text{LSA with operator}} = \mu_{b,u,\text{LSA}}$ 
        \end{tabular}}& \reject{0.000} & Mann-Whitney U  \\ \hline
\end{tabular}
\end{adjustbox}
\end{table}

\subsection{\rsm\ for All Class Appearance Order Scenarios Split for Preference Order of Stakeholders (Figure~\ref{fig: rsm total}) }
\label{sec: appendix fig 25}
Figure~\ref{fig: rsm total} shows the \rsm\ values for all class appearance order scenarios split for preference order of stakeholders.
Table~\ref{tab:fig25} shows the basic statistics. 

\begin{table}[h]
\caption{Statistics corresponding to Figure~\ref{fig: rsm total}: \rsm\ for all class appearance order scenarios}\label{tab:fig25}
\begin{adjustbox}{max width=\textwidth}
\begin{tabular}{cl|lll|lll}
\multicolumn{1}{l}{}                           &                   & \multicolumn{3}{c|}{(a)}                                                                   & \multicolumn{3}{c}{(b)}                                                      \\ \cline{3-8} 
\multicolumn{1}{l}{}                           &                   & \multicolumn{1}{c}{Median} & \multicolumn{1}{c}{Mean} & \multicolumn{1}{c|}{Standard deviation} & \multicolumn{1}{c}{Median} & \multicolumn{1}{c}{Mean} & \multicolumn{1}{c}{Standard deviation} \\ \hline

\multicolumn{1}{c|}{\multirow{3}{*}{Approaches}}  & ML2ASR            & \multicolumn{1}{c}{0.000}                           &       \multicolumn{1}{c}{0.150}                   &        
        \multicolumn{1}{c|}{0.186}&                          
        \multicolumn{1}{c}{0.139}&
        \multicolumn{1}{c}{0.199}&
        \multicolumn{1}{c}{0.166}\\
\multicolumn{1}{c|}{}                          & LSA               & \multicolumn{1}{c}{0.025}                           &       \multicolumn{1}{c}{0.149}                   &        
        \multicolumn{1}{c|}{0.187}                            &\multicolumn{1}{c}{0.125}                           &\multicolumn{1}{c}{0.204}                          &\multicolumn{1}{c}{0.175}                                        \\
\multicolumn{1}{c|}{}                          & LSA with operator & \multicolumn{1}{c}{0.000}                           & \multicolumn{1}{c}{0.032}                         &     
    \multicolumn{1}{c|}{0.092}
&   \multicolumn{1}{c}{0.125}                        &      
    \multicolumn{1}{c}{0.111}                    &          \multicolumn{1}{c}{0.088}                             
\end{tabular}
\end{adjustbox}
\end{table}

We define the following set of alternative hypotheses:

\begin{itemize}

    \item For each scenario $i$, the \rsm\ of the ``evolving classifier with LSA and operator feedback'' is less than the \rsm\ of the ``pre-defined classifier with ML2ASR'':

\begin{align*}
        H_1^{(i)}: & \mu_{i, r,\text{LSA with operator}} < \mu_{i, r,\text{ML2ASR}}\\
        H_0^{(i)}: & \mu_{i, r,\text{LSA with operator}} = \mu_{i, r,\text{ML2ASR}}
\end{align*}

 \item For each scenario $i$, the \rsm\ of the ``evolving classifier with LSA and operator feedback'' is less than the \rsm\ of the ``pre-defined classifier with LSA'':

\begin{align*}
        H_1^{(i)}: & \mu_{i, r,\text{LSA with operator}} < \mu_{i, r,\text{LSA}}\\
        H_0^{(i)}: & \mu_{i, r,\text{LSA with operator}} = \mu_{i, r,\text{LSA}}
\end{align*}
\end{itemize}

Note that the generic hypotheses represent distinct concrete hypotheses dependent on the value of $i$ chosen from the set of $\{a,b\}$, i.e., the two scenarios for the preference order.

To determine the appropriate test we can use to test the null hypotheses, we start with testing whether the data of the \rsm\ for each scenario are normally distributed using the Anderson–Darling test. Based on the results of Table~\ref{tab: normality test fig 25}, with a significance level of $0.05$, all normality tests are rejected.

\begin{table}[h]
\center
\caption{\re{Normality tests for distributions in subfigures of Figure~\ref{fig: rsm total}, by Anderson-Darling test method }}
\label{tab: normality test fig 25}
\begin{adjustbox}{max width=\textwidth}
\begin{tabular}{|c|c|c|}
\hline
 Approach & \begin{tabular}[c]{@{}c@{}}p-value (a)\end{tabular}
          & \begin{tabular}[c]{@{}c@{}}p-value (b)\end{tabular} \\ \hhline{===}
ML2ASR &   \reject{0.000} & \reject{0.000} \\ \hline
LSA &   \reject{0.000} & \reject{0.000} \\ \hline
LSA with operator &   \reject{0.000} & \reject{0.000} \\ \hline
\end{tabular}
\end{adjustbox}
\end{table}

Next, we use the Spearman correlation method to test the dependency between related pairs of data. With a significance level of 0.05, the test results shown in Table~\ref{tab: dependency test fig 25} for Figure~\ref{fig: rsm total} indicate that the dependency tests for all pairs are rejected.

\begin{table}[h]
\center
\caption{\re{Dependency tests for distributions in subfigures of Figure~\ref{fig: rsm total}, by Spearman correlation test method}}
\label{tab: dependency test fig 25}
\begin{adjustbox}{max width=\textwidth}
\begin{tabular}{|c|c|c|}
\hline
 Approach & \begin{tabular}[c]{@{}c@{}}p-value (a)\end{tabular}
          & \begin{tabular}[c]{@{}c@{}}p-value (b)\end{tabular}\\ \hhline{===}
(ML2ASR, LSA) &   \reject{0.000} & \reject{0.000} \\ \hline
(ML2ASR, LSA with operator) &   \reject{0.000} & \reject{0.000}\\ \hline
(LSA, LSA with operator) &   \reject{0.000} & \reject{0.000} \\ \hline
\end{tabular}
\end{adjustbox}
\end{table}

Since both normality and dependency tests are rejected for all cases, we apply the Mann-Whitney U test to test the hypotheses of Table~\ref{tab: hypotheses test fig 25}.


\begin{table}[h]
\center
\caption{\re{Test results for all mentioned hypotheses for Figure~\ref{fig: rsm total}, related hypotheses to each subfigure are separated by a double horizontal line from the others.}}
\label{tab: hypotheses test fig 25}
\begin{adjustbox}{max width=\textwidth}
\begin{tabular}{|c|c|c|c|}
\hline
\multicolumn{2}{|c|}{Hypothesis}  & p-value & Test method \\ \hhline{====}
\multicolumn{2}{|l|}{\begin{tabular}[c]{@{}c@{}}
$H_0^{(a)}: \mu_{a, r,\text{LSA with operator}} = \mu_{a,r,\text{ML2ASR}}$ 
        \end{tabular}}& \reject{0.000} & Mann-Whitney U  \\ \hline
\multicolumn{2}{|l|}{\begin{tabular}[c]{@{}c@{}}
$H_0^{(a)}: \mu_{a, r,\text{LSA with operator}} = \mu_{a,r,\text{LSA}}$ 
        \end{tabular}}& \reject{0.000} & Mann-Whitney U  \\ \hhline{====}

\multicolumn{2}{|l|}{\begin{tabular}[c]{@{}c@{}}
$H_0^{(b)}: \mu_{b, r,\text{LSA with operator}} = \mu_{b,r,\text{ML2ASR}}$ 
        \end{tabular}}& \reject{0.000} & Mann-Whitney U  \\ \hline
\multicolumn{2}{|l|}{\begin{tabular}[c]{@{}c@{}}
$H_0^{(b)}: \mu_{b, r,\text{LSA with operator}} = \mu_{b,r,\text{LSA}}$ 
        \end{tabular}}& \reject{0.000} & Mann-Whitney U  \\ \hline
        
\end{tabular}
\end{adjustbox}
\end{table}

With a significance level of 0.05, the test results support the alternative hypotheses, i.e., we found evidence for both scenarios that: (i) the \rsm\ of the ``evolving classifier with LSA and operator feedback'' is less than the \rsm\ of the ``pre-defined classifier with ML2ASR'', and (ii) the \rsm\ of the ``evolving classifier with LSA and operator feedback'' is less than the \rsm\ of the ``pre-defined classifier with LSA''.


\subsection{ Impact of The Operator Feedback on Utilities And \rsm\ in All Scenarios 
 (Figure~\ref{fig: utility rsm total total})}
\label{sec: appendix fig 26}
Figure~\ref{fig: utility rsm total total} illustrates the impact of the operator feedback on utilities and \rsm\ in all scenarios.
Table~\ref{tab:fig26} shows the basic statistics. 

\begin{table}[h]
\caption{Statistics corresponding to Figure~\ref{fig: utility rsm total total}: utility and \rsm\ for all scenarios}\label{tab:fig26}
\begin{adjustbox}{max width=\textwidth}
\begin{tabular}{cl|lll|lll}
\multicolumn{1}{l}{}                           &                   & \multicolumn{3}{c|}{(a) Utility}                                                                   & \multicolumn{3}{c}{(b) \rsm}                                                      \\ \cline{3-8} 
\multicolumn{1}{l}{}                           &                   & \multicolumn{1}{c}{Median} & \multicolumn{1}{c}{Mean} & \multicolumn{1}{c|}{Standard deviation} & \multicolumn{1}{c}{Median} & \multicolumn{1}{c}{Mean} & \multicolumn{1}{c}{Standard deviation} \\ \hline

\multicolumn{1}{c|}{\multirow{2}{*}{Approaches}}  & LSA               & \multicolumn{1}{c}{0.807}                           &       \multicolumn{1}{c}{0.682}                   &        
        \multicolumn{1}{c|}{0.273}                            &\multicolumn{1}{c}{0.125}                           &\multicolumn{1}{c}{0.176}                          &\multicolumn{1}{c}{0.183}                                        \\
\multicolumn{1}{c|}{}                          & LSA with operator & \multicolumn{1}{c}{0.809}                           & \multicolumn{1}{c}{0.787}                         &     
    \multicolumn{1}{c|}{0.149}
&   \multicolumn{1}{c}{0.000}                        &      
    \multicolumn{1}{c}{0.072}                    &          \multicolumn{1}{c}{0.098}                             
\end{tabular}
\end{adjustbox}
\end{table}


We define the following set of alternative hypotheses:

\begin{itemize}

 \item The utility of the ``evolving classifier with LSA and operator feedback'' is higher than the utility of the ``pre-defined classifier with LSA'':

\begin{align*}
        H_1^{(a)}: & \mu_{a, u,\text{LSA with operator}} > \mu_{a, u,\text{LSA}}\\
        H_0^{(a)}: & \mu_{a, u,\text{LSA with operator}} = \mu_{a, u,\text{LSA}}
\end{align*}

\item The \rsm\ of the ``evolving classifier with LSA and operator feedback'' is less than the \rsm\ of the ``pre-defined classifier with LSA'':

\begin{align*}
        H_1^{(b)}: & \mu_{b, r,\text{LSA with operator}} < \mu_{b, r,\text{LSA}}\\
        H_0^{(b)}: & \mu_{b, r,\text{LSA with operator}} = \mu_{b, r,\text{LSA}}
\end{align*}

\end{itemize}

We start with determining the appropriate test we can use to test null hypotheses. First, we test whether the data of the \rsm\ and the utility for each appraoch are normally distributed using the Anderson–Darling test. Based on the results of Table~\ref{tab: normality fig26}, with a significance level ($\alpha$) of $0.05$, all normality tests were rejected.

\begin{table}[h]
\center
\caption{\re{Normality tests for distributions in subfigures of Figure~\ref{fig: utility rsm total total}, by Anderson-Darling test method }} \label{tab: normality fig26}
\begin{adjustbox}{max width=\textwidth}
\begin{tabular}{|c|c|c|}
\hline
 Approach & \begin{tabular}[c]{@{}c@{}}p-value (a)\end{tabular}
          & \begin{tabular}[c]{@{}c@{}}p-value (b)\end{tabular} \\ \hhline{===}
LSA &   \reject{0.000} & \reject{0.000} \\ \hline
LSA with operator &   \reject{0.000} & \reject{0.000} \\ \hline
\end{tabular}
\end{adjustbox}
\end{table}

Next, we use the Spearman correlation method to test the dependency between related pairs of data. With a significance level of 0.05, the test results shown in Table~\ref{tab: dependency fig26} indicate that dependency tests for all pairs are rejected.

\begin{table}[h]
\center
\caption{\re{Dependency tests for distributions in subfigures of Figure~\ref{fig: utility rsm total total}, by Spearman correlation test method}}\label{tab: dependency fig26}
\begin{adjustbox}{max width=\textwidth}
\begin{tabular}{|c|c|c|}
\hline
 Approach & \begin{tabular}[c]{@{}c@{}}p-value (a)\end{tabular}
          & \begin{tabular}[c]{@{}c@{}}p-value (b)\end{tabular}\\ \hhline{===}
(LSA, LSA with operator) &   \reject{0.000} & \reject{0.000} \\ \hline
\end{tabular}
\end{adjustbox}
\end{table}

Since both normality and dependency tests are rejected for all cases, we apply the Mann-Whitney U test to test the hypotheses of Table~\ref{tab: hypotheses test fig 26} for Figure~\ref{fig: utility rsm total total}.

\begin{table}[h]
\center
\caption{\re{Test results for all mentioned hypotheses for Figure~\ref{fig: utility rsm total total}, related hypothesis to each subfigure is separated by a double horizontal line from the other.}} \label{tab: hypotheses test fig 26}
\begin{adjustbox}{max width=\textwidth}
\begin{tabular}{|c|c|c|c|}
\hline
\multicolumn{2}{|c|}{Hypothesis}  & p-value & Test method \\ \hhline{====}
\multicolumn{2}{|l|}{\begin{tabular}[c]{@{}c@{}}
$H_0^{(a)}: \mu_{a, u,\text{LSA with operator}} = \mu_{a,u,\text{LSA}}$ 
        \end{tabular}}& \reject{0.000} & Mann-Whitney U  \\ \hhline{====}

\multicolumn{2}{|l|}{\begin{tabular}[c]{@{}c@{}}
$H_0^{(b)}: \mu_{b, r,\text{LSA with operator}} = \mu_{b,r,\text{LSA}}$ 
        \end{tabular}}& \reject{0.000} & Mann-Whitney U  \\ \hline
        
\end{tabular}
\end{adjustbox}
\end{table}

With a significance level of 0.05, the test result support the alternative hypotheses. Hence, we found statistical evidence that: (i) the utility of the ``evolving classifier with LSA and operator feedback'' is higher than the utility of the ``pre-defined classifier with LSA'', and (ii) the \rsm\ of the ``evolving classifier with LSA and operator feedback'' is less than the \rsm\ of the ``pre-defined classifier with LSA''.

\end{document}